\documentclass[Afour,sageh,times]{sagej}
\usepackage{algorithm}
\usepackage{algorithmic}
\usepackage{adjustbox}
\usepackage{amsfonts}
\usepackage{amssymb}
\usepackage{amsthm}
\usepackage{booktabs}
\usepackage{array}
\usepackage{bm}
\usepackage{calc}
\usepackage{comment}
\usepackage{graphicx}
\usepackage[colorlinks,bookmarksopen,bookmarksnumbered,citecolor=red,urlcolor=red]{hyperref}
\usepackage{longtable}
\usepackage{mathrsfs}
\usepackage{mathtools}
\usepackage{moreverb}
\usepackage{multirow}
\usepackage{ragged2e}
\usepackage{stmaryrd}
\usepackage{tablefootnote}
\usepackage{textcomp}
\usepackage{threeparttable}
\usepackage{tikz}
\usepackage{url}

\usetikzlibrary{mindmap,backgrounds}

\def\volumeyear{2026} 

\graphicspath{{images/}}

\newcommand{\bbR}{\mathbb{R}}
\newcommand{\vect}[1]{\mathbf{#1}} 
\newcommand{\mtrx}[1]{\mathbf{#1}}
\newcommand{\argmin}[1]{\underset{#1}{\operatorname{arg}\,\operatorname{min}}\;}

\newcolumntype{M}[1]{>{\RaggedRight}m{#1}}

\begin{document}

\runninghead{Gutierrez, Cloud, and Beksi}

\title{Movement Primitives in Robotics: \\A Comprehensive Survey}

\author{Nolan B. Gutierrez, Joseph M. Cloud, and William J. Beksi}

\affiliation{Department of Computer Science and Engineering, The University of
Texas at Arlington, Arlington, USA}

\corrauth{William J. Beksi, Department of Computer Science and Engineering, The
University of Texas at Arlington, Arlington, USA.} 
\email{william.beksi@uta.edu}

\begin{abstract}
Biological systems exhibit a continuous stream of movements, consisting of
sequential segments, that allow them to perform complex tasks in a creative and
versatile fashion. This observation has led researchers towards identifying
elementary building blocks of motion known as \textit{movement primitives},
which are well-suited for generating motor commands in autonomous systems, such
as robots. In this survey, we provide an encyclopedic overview of movement
primitive approaches and applications in chronological order. Concretely, we
present movement primitive frameworks as a way of representing robotic control
trajectories acquired through human demonstrations. Within the area of
robotics, movement primitives can encode basic motions at the trajectory level,
such as how a robot would grasp a cup or the sequence of motions necessary to
toss a ball. Furthermore, movement primitives have been developed with the
desirable analytical properties of a spring-damper system, probabilistic
coupling of multiple demonstrations, using neural networks in high-dimensional
systems, and more, to address difficult challenges in robotics. Although
movement primitives have widespread application to a variety of fields, the
goal of this survey is to inform practitioners on the use of these frameworks
in the context of robotics. Specifically, we aim to (i) present a systematic
review of major movement primitive frameworks and examine their strengths and
weaknesses; (ii) highlight applications that have successfully made use of
movement primitives; and (iii) examine open questions and discuss practical
challenges when applying movement primitives in robotics. 
\end{abstract}

\keywords{
  Movement primitives, 
  robot learning,
  learning from demonstration
}

\maketitle

\section{Introduction}
\label{sec:introduction}
Robots are traditionally preprogrammed with the knowledge needed to perform
specific tasks. However, this introduces the following challenges: (i) it
necessitates expert understanding of control, motion planning, and sensor
integration; (ii) the robot could potentially overfit to the given task, thereby
diminishing its self-sufficiency. Moreover, due to the dynamic nature of
environments, hardcoding every possible scenario into the robot becomes
untenable. To surmount these issues, considerable effort has been invested in
creating robotic systems capable of harnessing human-task knowledge. This
allows robots to learn and perform jobs more intuitively, and non-experts to
instruct robots more easily. This concept is known as \textit{learning from
demonstration} (LfD) \cite{atkeson1997robot}, and it is sometimes referred to as
\textit{programming by demonstration}
\cite{billard2008survey,calinon2018learning}.

\begin{figure*}
\center
\includegraphics[width=0.19\textwidth]{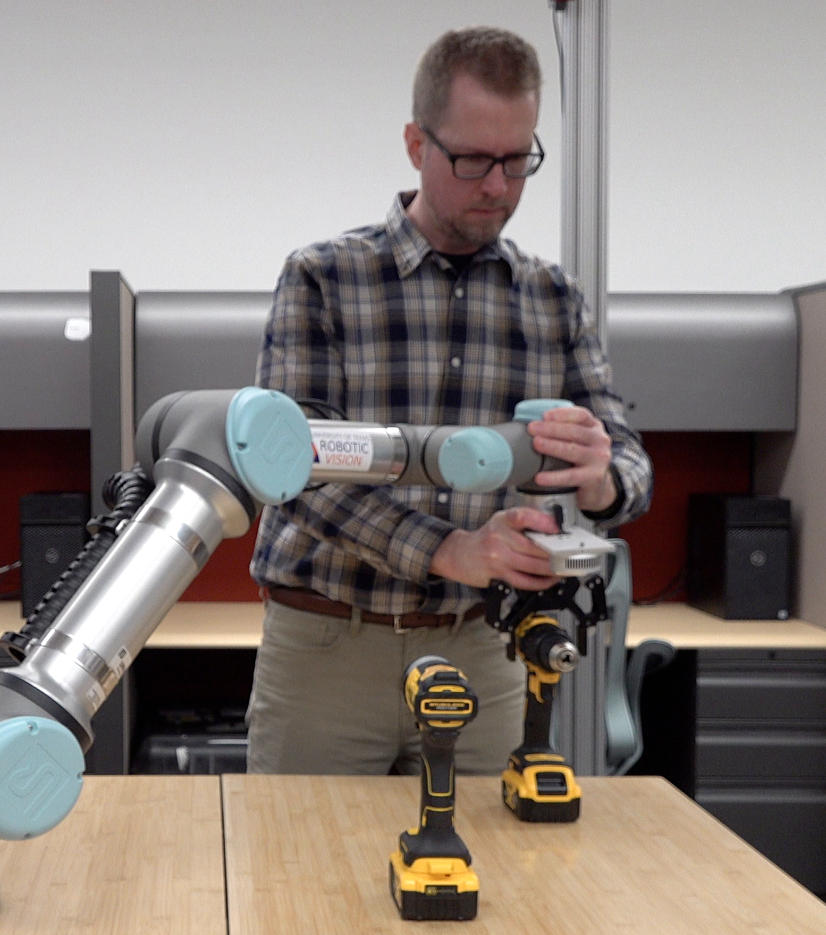} 
\includegraphics[width=0.19\textwidth]{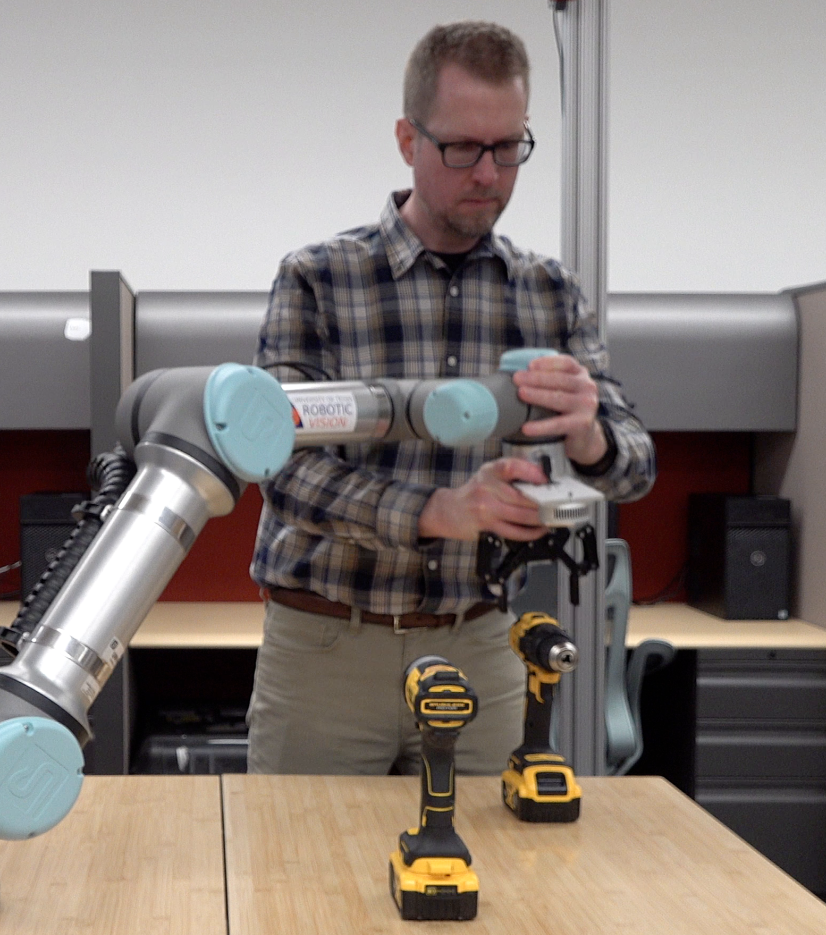}
\includegraphics[width=0.19\textwidth]{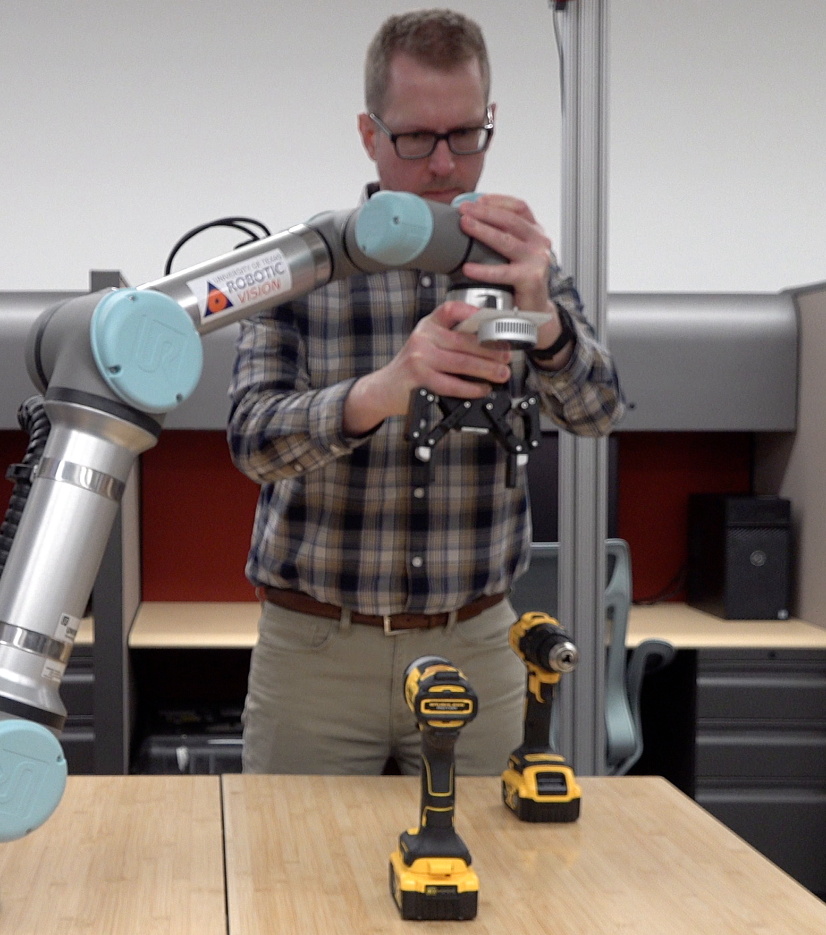}
\includegraphics[width=0.19\textwidth]{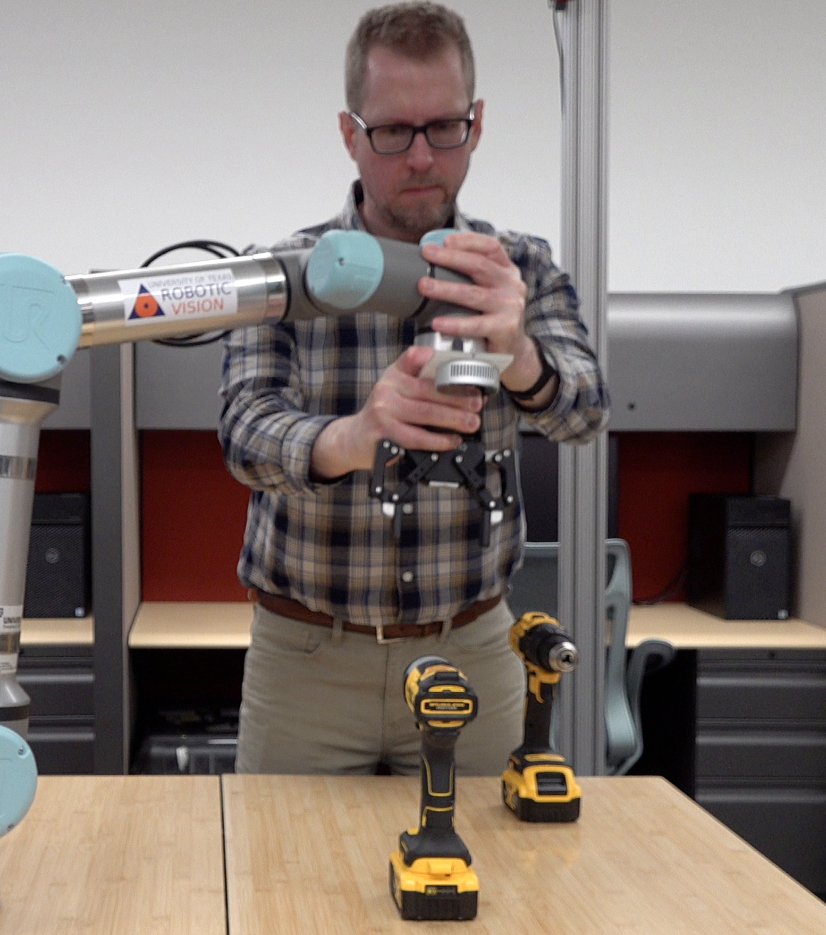}
\includegraphics[width=0.19\textwidth]{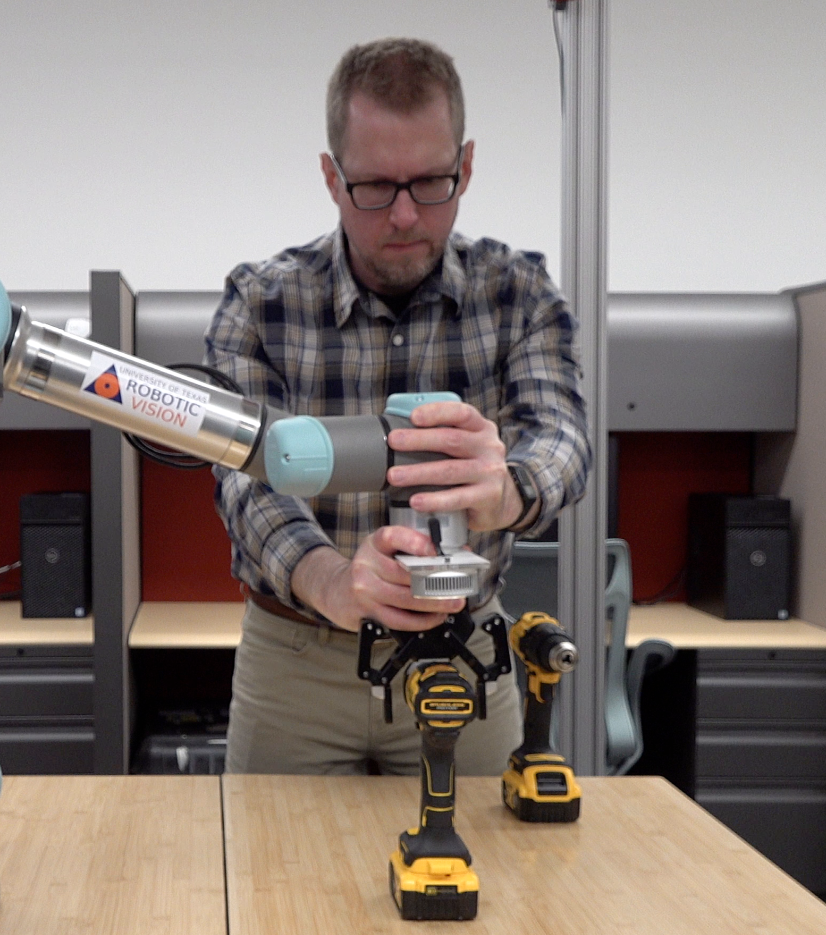}
\caption{An operator employing kinesthetic teaching to demonstrate a
pick-and-place task to a robot manipulator.}
\label{fig:learning_from_demonstration}
\end{figure*}

In the field of robotics, LfD is a methodology where skills are transferred from
a human teacher to a machine \cite{argall2009survey,ravichandar2020recent}
(Figure~\ref{fig:learning_from_demonstration}). This method is underpinned by
the need for intuitive human instruction. By learning behaviors through task
demonstrations, non-experts are able to adapt robots more readily to fit new
requirements or tasks. LfD frameworks form a branch of supervised learning with
the goal of approximating a state-action mapping function, known as a
\textit{policy}. These policies are initially provided by a teacher during a
demonstration. The process of LfD includes observing the demonstration,
encapsulating it, and finally reproducing it \cite{dindo2010adaptive}. Various
methods are used for capturing demonstrations, including recording the human
motion of the task \cite{kulic2008incremental,kulic2012incremental}, kinesthetic
teaching \cite{sauser2012iterative}, and immersive teleoperation
\cite{peternel2013learning}.

One prominent technique for learning these policies involves the use of movement
(or motor) primitives (MPs). These primitives encode basic movements at the
trajectory level, learned through demonstrations. Over time, MPs have been
developed to tackle various challenges while retaining desirable properties that
enable them to adapt to change with ease. There are many MP frameworks as we
will discuss in this survey, but all these methods should generate specific
movements to complete a task, adapt to environmental changes, combine primitives
to achieve complex tasks, and generalize primitives for various tasks or
environmental configurations \cite{tavassoli2023learning}. For example, this
includes how a robot might grasp an object or the sequence of motions a
manipulator needs to coordinate its joints for a throwing motion
\cite{suomalainen2022survey}. Additionally, MP frameworks can be modified to
account for obstacles or respond to external perturbations
\cite{rasines2024robots}.

\subsection{Motivation for Movement Primitives}
\label{subsec:motivation_for_movement_primitives}
In this survey, we identify several traits that are used to motivate the use of
MPs for various robotic applications. These ideas are distilled into the
following distinct criteria.
\begin{itemize}
  \item \textbf{Simplicity}: MPs provide a basic mechanism for generating
  repeatable patterns of motion.
  \item \textbf{Learning}: MPs learn from demonstration data, which can be
  acquired via a number of ways and actuate different types of robotic control
  systems.
  \item \textbf{Reusability}: A \textit{primitive} conceptually means that these
  ``building-block" movements can be reused across tasks and environments.
  \item \textbf{Adaptability}: MP methodologies are parameterized such that they
  can be modified in a deterministic fashion.
  \item \textbf{Human-like behavior}: The behavior exhibited by MPs has the
  capacity to appear human-like since primitives can be learned through human
  demonstrations.
\end{itemize}

\subsection{Taxonomy of Movement Primitives Research}
\label{subsec:taxonomy_of_movement_primitives_research}
\begin{figure*}
\centering
\includegraphics[width=\textwidth]{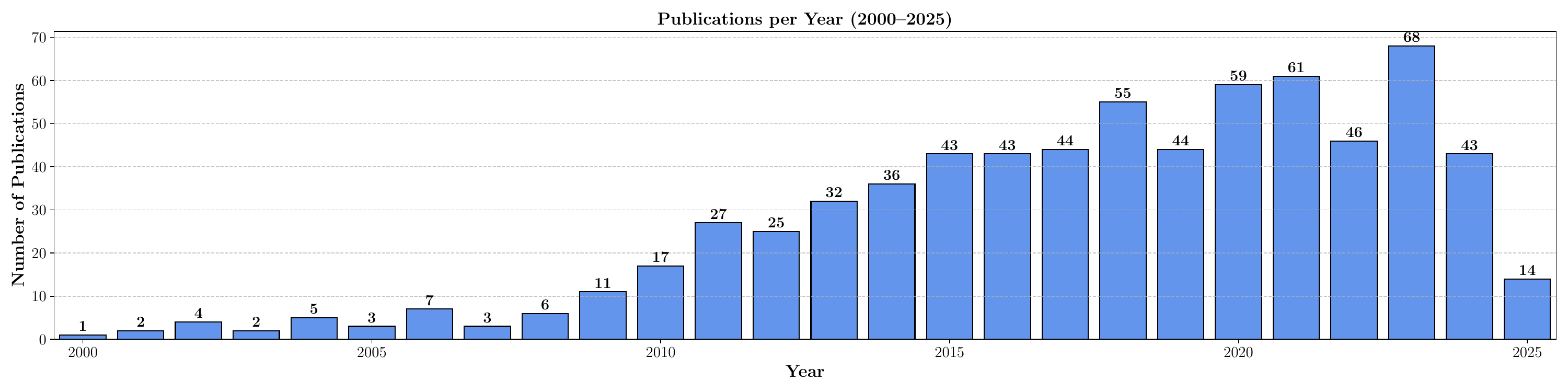}
\caption{The distribution of MP publications over time.}
\label{fig:publications_per_year}
\end{figure*}

The MP literature is broad and extensive, with contributions from numerous
research groups. Many of these works are intertwined, build upon one another, or
address a similar problem using a different approach. Not only does this make
the historical development of MPs difficult to follow, it also presents
challenges to those interested in getting started with MPs. In this survey, we
provide a structured and unified presentation of MP research. Concretely, we
disentangle prior works from the same research groups, and make it clear to
practitioners the connections and separations between various techniques. Our
chronological organization of the diverse related works preserves the historical
development of MPs and makes it easier to navigate this wide body of research.
Additionally, we provide a comprehensive and categorized survey of all major MP
applications, which can help inspire practitioners to apply MPs to miscellaneous
problems in robotics. In summary, this survey contains over 700 papers from
nearly three decades of MP research (Figure~\ref{fig:publications_per_year}).

\subsection{Key Contributions}
\label{subsec:key_contributions}
The aim of this survey is to assist practitioners in navigating the extensive
amount of information available on MPs. It strives to provide an overview of
the theory, applications, and practical challenges of working with MPs. We
summarize the contributions of this survey as follows.
\begin{itemize}
  \item We introduce and provide a comparison of established MP frameworks.
  \item We exhaustively organize applications of MPs over the years.
  \item We identify future research directions for MPs in the context of LfD.
  \item We provide a curated list of open-source software and papers on MPs.
\end{itemize}

\subsection{Note to Practitioners}
\label{subsec:note_to_practitioners}
Over the years, many research papers have been written and software programs
developed for MP frameworks. For researchers and developers interested in these
frameworks, we provide a list of publicly available papers and open-source
software \cite{awesome2026}. This repository, featuring various MP adaptations,
will be continuously updated with links to both relevant software and papers.
Contributions to the repository are welcome through pull requests consistent
with the markup language format.

\section{Introduction to Movement Primitives}
\label{sec:movement_primitives}
\begin{table*}
\caption{Acronyms and their definitions (alphabetically sorted).}
\label{tab:acronyms_and_definitions}
\centering
\scalebox{0.90}{
\begin{tabular}{llll}
  \toprule
  Acronym          & Definition & 
  Acronym          & Definition \\
  \midrule
  AL-DMP           & Arc-length DMP & 
  BC               & Behavior cloning \\ 
  BLF              & Barrier Lyapunov function & 
  CDMP             & Cartesian space DMP \\ 
  CBF              & Control barrier function & 
  CCDMP            & Coordinate changed DMP \\ 
  CNMP             & Conditional neural MP & 
  CNN              & Convolutional neural network \\ 
  CLF              & Control Lyapunov function & 
  CNP              & Conditional neural process \\ 
  CPG              & Central pattern generator & 
  DL               & Deep learning \\ 
  DNN              & Deep neural network & 
  DoF              & Degree of freedom \\ 
  DMP              & Dynamic movement primitive & 
  DTW              & Dynamic time warping \\ 
  EM               & Expectation maximization & 
  EMG              & Electromyography \\ 
  FMP              & Fourier MP & 
  GMM              & Gaussian mixture model \\ 
  GMR              & Gaussian mixture regression & 
  GP               & Gaussian process \\ 
  GPR              & Gaussian process regression & 
  HBM              & Hierarchical Bayesian model \\ 
  HMM              & Hidden Markov model & 
  HRI              & Human-robot interaction \\ 
  IL               & Imitation learning & 
  ILC              & Iterative learning control \\ 
  IMU              & Inertial measurement unit & 
  IP               & Interaction primitive \\ 
  KL               & Kullback-Leibler & 
  KMP              & Kernelized MP \\ 
  LfD              & Learning from demonstration & 
  LWR              & Locally weighted regression \\ 
  LWPR             & Locally weighted projection regression & 
  MoMP             & Mixture of MP \\
  MLE              & Maximum likelihood estimation & 
  MP               & Movement/motor primitive \\ 
  MPC              & Model predictive control & 
  OA               & Obstacle avoidance \\ 
  PCA              & Principal component analysis & 
  PI$^2$           & Policy improvement with path integrals \\ 
  PI$^{\text{BB}}$ & Policy improvement with black-box optimization & 
  PoWER            & Policy learning by weighting exploration with the returns \\ 
  PDMP             & Parametric DMP & 
  ProDMP           & Probabilistic DMP \\ 
  ProMP            & Probabilistic MP & 
  RBFNN            & Radial basis function neural network \\ 
  REPS             & Relative entropy policy search & 
  RL               & Reinforcement learning \\ 
  RRT              & Rapidly exploring random tree & 
  SDMP             & Stylistic DMP \\ 
  SPD              & Symmetric positive definite & 
  SVM              & Support vector machine \\ 
  TP-DMP           & Task-parameterized DMP & 
  VR               & Virtual reality
  \\
  \bottomrule
\end{tabular}
}
\end{table*}

MPs mark a paradigm shift in robotics, introducing a flexible and dynamic
approach to robot motion. By encoding movements into modular, reusable patterns,
they enable robots to adapt and learn new tasks across disparate environments.
This section delves into the historical development of MPs and examines their
role in facilitating task adaptation and learning. To enhance understanding
throughout this survey, Table~\ref{tab:acronyms_and_definitions} defines key terms
central to the discussion of MPs.

\subsection{Historical Overview}
\label{subsec:historical_overview}
Computational motor control has traditionally been approached from the following
viewpoints: \textit{optimal control} and \textit{dynamical systems}.  The
optimal control perspective views motor control as the evolutionary result of a
nervous system that attempts to optimize general organizational principles
(e.g., energy consumption, accurate task achievement, etc.) through the use of
optimal control theory \cite{schaal2007dynamics}. Conversely, the dynamical
systems point of view emphasizes motor control as a process of self-organization
between an animal and its environment via the use of nonlinear differential
equations for modeling behavior \cite{ijspeert2008central}.

\cite{schaal2000nonlinear,ijspeert2001trajectory,ijspeert2002movement,ijspeert2002learningb}
were the first to develop a computational approach to motor control that
provides a unifying framework from both the optimal control and dynamical
systems perspectives. They demonstrated how to allow for the representation of
self-organizing processes and the optimization of movement based on a reward
criteria. In addition, they emphasized that models of movement generation should
not have explicit time dependencies, similar to autonomous dynamical systems, in
order to accommodate coupling and perturbation effects in a straightforward way.

MPs may be categorized into state-based
\cite{calinon2010learning,khansari2011learning} and trajectory-based
representations \cite{schaal2005learning,neumann2009learning,rozo2013learning,
rueckert2015extracting}. Trajectory-based primitives typically use time as the
driving force of the movement. They require simple, usually linear controllers,
and scale well to a large number of degrees of freedom (DoF). On the contrary,
state-based primitives do not require knowledge of a time step, but often need
to use more complex nonlinear policies. The corresponding increase in complexity
has limited the application of state-based primitives to a rather small number
of dimensions, such as the Cartesian coordinates of the task space of a robot.

\section{Dynamic Movement Primitives}
\label{sec:dynamic_movement_primitives}
Dynamic movement primitives (DMPs) \cite{schaal2001real,
schaal2003control,schaal2006dynamica,schaal2006dynamicb} stem from the early
ideas of \cite{bullock1988neural}, and are used to encode both discrete and
rhythmic movements. They are represented by a second-order differential equation
based on the physics of a well-understood spring-damper system. An advantageous
feature of the DMP framework is that a policy can be learned from a single
demonstration such that generated motor skills are compliant with different
environmental parameters and spaces (e.g., the task can be learned in joint or
Cartesian space). 

\subsection{Background}
\label{subsec:dmp_background}
The goal in \textit{learning} DMPs is to find the forcing term with which we can
perturb the spring-damper system to replicate the demonstrated motion. More
formally stated,
\begin{equation}
  \tau\dot{u} = a_z(\beta_z(g - x) - u) + (g - x_0)f,
  \label{eq:dmp_eqn_1}
\end{equation}
\begin{equation}
  \tau\dot{x} = u,
  \label{eq:dmp_eqn_2}
\end{equation}
where $x$ and $u$ are the position and velocity of the robot's joints, $\tau$
is the duration of the demonstration, and $x_0$ and $g$ are the initial and
target positions in joint space, respectively. The terms $a_z$ and $\beta_z$
are control gains selected to render the system critically damped, and $f$ is a
forcing term that acts as the control input to drive the system towards the
goal. To achieve critical dampness, $\beta_z = a_z/4$ as derived in
\cite{ijspeert2013dynamical}. 

Note that we present the system in \eqref{eq:dmp_eqn_1}-\eqref{eq:dmp_eqn_2}
using first-order notation. The forcing term is further defined as
\begin{equation}
    \label{eq:dmp_eqn_3}
    f (s) = \frac{\sum_{i=0}^N \omega_i \, \psi_i(s)}{\sum_{i=0}^N \psi_i(s)} s,
\end{equation}
where $\psi_i(s)=exp(-h_i(s-c_i)^2)$ are Gaussian basis functions with width
$h_i$ and centers $c_i$. The $w_i$ are adjustable weights that change the shape
of the Gaussians over the $s$ domain. The forcing term is dependent on $s$,
\begin{equation}
  \tau\dot{s} = \alpha s. 
  \label{eq:dmp_eqn_4}
\end{equation}
In \eqref{eq:dmp_eqn_4}, $s$ represents a variable that starts from time $t=0$
and becomes zero at $t=\tau$. From this, $s$ is computed according to
\eqref{eq:dmp_eqn_4} by setting $\alpha$ such that $s$ becomes zero at the final
time. If the system learns motor policies in joint space, then
\eqref{eq:dmp_eqn_4} behaves the same for all joints.
\eqref{eq:dmp_eqn_1}-\eqref{eq:dmp_eqn_3} are computed independently for each
joint/DoF.

This formulation can be easily extended into vector form to handle multiple DoF.
However, one of the issues in formulating it this way is that starting and goal
locations that are physically near each other will limit the influence of the
forcing term. Later work by \cite{park2008movement} resolves this by separating
the constant $g -x_0$ from the forcing term and dissipating it using $s$, i.e.,
\begin{equation}
  \!\!\!  \tau \dot{\vect{v}} = \mtrx{K} (\vect{g} - \vect{x}) - \mtrx{D} \vect{v} - \mtrx{K} (\vect{g}- \vect{x}_0)s + \mtrx{K}\vect{f}(s).
  \label{eq:dmp_eqn_5}
\end{equation}
In \eqref{eq:dmp_eqn_5}, $\mtrx{K}$ and $\mtrx{D}$ are now $d \times d$
diagonal matrices for $d$-dimensional trajectories. The other notable change is
that the contribution of the constant $g - x_0$ is phased out with $s$, so the
forcing term is no longer dependent on it.

To learn the forcing term from demonstration, we first rearrange
\eqref{eq:dmp_eqn_1} to solve for the target forcing term given a demonstrated
trajectory,
\begin{equation}
f_{target} = \tau^2\ddot{x}_{demo} - a_z(\beta_z(g - x_{demo}) - \tau\dot{x}_{demo}),
\label{eq:dmp_learning_1}
\end{equation}
where $x_{demo}$, $\dot{x}_{demo}$, and $\ddot{x}_{demo}$ represent the
position, velocity, and acceleration from the demonstration. The weights $w_i$
for each basis function in \eqref{eq:dmp_eqn_3} are learned using locally
weighted regression (LWR) as described by
\cite{schaal1998constructive,schaal2002scalable}, by minimizing the locally
weighted quadratic error criterion
\begin{equation}
J_i = \sum_{t=1}^T \psi_i(t)(f_{target}(t) - w_is(t)(g - x_0))^2,
\label{eq:dmp_learning_2}
\end{equation}
where $T$ is the number of points in the demonstration. This weighted linear
regression problem has the following analytical solution:
\begin{equation}
w_i = \frac{\sum_{t=1}^T \psi_i(t)f_{target}(t)s(t)(g - x_0)}{\sum_{t=1}^T \psi_i(t)(s(t)(g - x_0))^2}.
\label{eq:dmp_learning_3}
\end{equation}
The learning can be performed either in batch mode using these equations
directly, or incrementally as the demonstration data arrives using recursive
least squares.

\cite{schaal2006dynamica} summarize the results leading to the formulation of
DMPs as a general approach to motor control in robotics and biology. Concretely,
to allow for the large flexibility of human limb control, the concept of motor
pattern generators is augmented by a component that can be adjusted to a
specific task. This leads to the concept of a DMP where the attractor landscape
is assumed to represent the desired kinematic state of a limb (e.g., positions,
velocities, and accelerations). A nascent DMP system was constructed and
deployed on a 30-DoF humanoid robot. The desired position, velocity, and
acceleration information was derived from the states of the DMPs and fed to a
compute-torque controller.

\subsection{Improvements and Extensions}
The DMP framework utilizes differential equations to adapt to and generate a
given movement. Yet, a weak point of DMPs is the generalization to new movement
targets. Motivated by neurophysiology, \cite{hoffmann2009biologically} present
an improved modification of the original DMP framework. The equations generalize
movements to novel targets while avoiding singularities and large accelerations.
Moreover, the equations can represent a movement in 3D task space without
dependence on the choice of coordinate system (i.e., invariance under invertible
affine transformations). The formalism is further extended to robot obstacle
avoidance by adding an empirical term to the differential equations that allows
for moving around an obstruction. 

Adapting DMPs to new task requirements is problematic when the demonstrated
trajectories are only available in joint space. This is because DMP parameters
generally do not correspond to variables meaningful for the task at hand.
Furthermore, the problem becomes more extreme with increasing DoF (e.g.,
humanoid movements). Nevertheless, DMP parameters can directly relate to task
variables when they are learned in latent spaces resulting from a dimensionality
reduction of the demonstrated trajectories. For example, \cite{bitzer2009latent}
learn discrete point-to-point movements and propose a modification of the
Gaussian process (GP) latent variable model, a powerful nonlinear dimensionality
reduction technique. The adaptation makes the method more ideal for the use of
DMPs by favoring latent spaces with highly-regular structure. 

DMPs consist of a simple nonlinear oscillator to generate the phase and
amplitude of periodic movements. A major drawback of this approach is that it
requires the frequency of the demonstration signal to be explicitly specified.
\cite{gams2009line} address this issue via a two-layered system that (i) learns
and encodes a periodic signal, and (ii) modulates the learned periodic
trajectory in response to external events. The first layer consists of a
dynamical system that extracts the fundamental frequency of the input signal and
is based on adaptive frequency oscillators. In the second layer, another
dynamical system is responsible for learning the waveform. The combination of
the two dynamical systems allows for rapidly teaching new trajectories without
any knowledge of the frequency of the demonstration signal. In proceeding work,
\cite{petrivc2011line} improve the first layer by removing the need for an
additional algorithm to determine the basic frequency of the input signal.
Extra properties and use cases of the two-layered system are presented in
\cite{gams2012performing}.

In observing that human motion has specific sequence features called
\textit{style}, \cite{matsubara2010learning} propose the concept of stylistic
DMPs for motor learning and control in humanoid robotics. Stylistic DMPs
compactly encode diverse motion styles in human behavior observed via multiple
demonstrations. This allows manipulation of various styles, even while a robot
is moving, by a control variable called a style parameter in the generated
motions. Compared to previous stylistic modeling techniques, such as motion
library–based approaches \cite{gams2009generalization}, stylistic DMPs offer
greater scalability and thus wider applicability in real-world environmental
tasks.

\cite{mulling2010learning} address the restriction of DMPs to representing only
single elementary actions via a mixture of motor primitives (MoMPs). Consisting
of a library of component DMPs, the framework employs a gating network activated
by external stimuli to initiate the relevant DMP, where the weighted sum of the
primitive defines the desired movement. Not only does the resulting policy
facilitate the selection of the right DMP, but it also supports its
generalization. This challenge, especially in predicting essential context
information such as the precise time and location to strike a ball, is
demonstrated in a table tennis scenario with an anthropomorphic robot arm. The
system learns smooth movements that properly distribute the forces over the
different DoF, producing a human-like pattern of movement. In
\cite{mulling2013learning}, MoMPs are expanded to gather a repertoire of DMPs
from human demonstrations, enhancing its adaptability to new situations.

Segmenting complex robot movements into a sequence of primitives is a difficult
problem for many applications. \cite{meier2011movement} solve this dilemma by
reformulating DMPs such that they are parameterized by the movement duration and
goal position of the encoded motion (i.e., the segmentation point if no
co-articulation were present). Since DMPs are designed to have spatial and
temporal generalization properties, having the duration and goal position as
open parameters allows for the representation of different versions of the same
MP. Using this reformulation, an expectation maximization (EM) algorithm is
developed to estimate the duration and goal position of a partially observed
trajectory. With the aid of the EM algorithm and the assumption that a library
of DMPs is available, the segmentation problem is reduced to sequential movement
recognition.

Learning new robot actions is a difficult problem due to the potentially massive
search space to be explored. To this end, \cite{nemec2011exploiting} speed up
the DMP learning process by combining ideas from imitation learning (IL) and
reinforcement learning (RL) with statistical generalization as follows. First, a
set of trajectories distributed across the task space are acquired using IL.
Then, the training data is used to compute a control policy suitable for the
current situation. Next, an initial approximation is improved via RL by learning
on a manifold of reduced dimensionality. Finally, fine tuning of the RL policy
is done by learning in a high-dimensional space. The method is improved in
\cite{nemec2012applying} by generating promising search space directions from
previous examples and in \cite{nemec2013efficient} by employing RL as an
error-learning algorithm within the constrained domain.
\cite{vuga2014speed,vuga2015enhanced} further enhanced the adaptation speed of
the system through a combination of iterative learning control (ILC) and RL.
Simulated and physical experiential evaluations are conducted on learning
pouring and ball-throwing tasks.  

\cite{stulp2011hierarchical} tackle the following challenges in hierarchical RL
via DMPs: compact representations and high-dimensional tasks. Specifically,
policy improvement with path integrals (PI$^2$)
\cite{theodorou2010reinforcement,theodorou2010generalized,stulp2011learningb} is
applied to sequences of DMPs (e.g.,
\cite{kalakrishnan2011learning,stulp2012path,stulp2012adaptive,stulp2012model,
stulp2013robot}). To do this, the shape of the DMP and the subgoals between two
DMPs are optimized up to both the costs of the next primitive in the sequence
and the end of the entire DMP sequence, respectively. This enables the DMP
parameters to be learned simultaneously at different levels of temporal
abstraction. For robots, this has desirable properties over discrete
representations in terms of compactness, control, and scalability. Robot
demonstrations show how simultaneously learning shapes and goals in sequences
leads to lower costs when compared to only accounting for local costs, and
behavior that mimics humans in the case of a pick-and-place task. 

Although complex robot motor skills can be represented using DMPs, in many cases
a robot may need to learn a new elementary movement even if an existing DMP
covers a similar scenario. To address this problem,
\cite{kober2011reinforcement} describe how to learn mappings from representation
to meta-parameters using RL. In particular, a kernelized version of
reward-weighted regression is utilized to modulate elementary movements through
the meta-parameters of the situation at hand. By extending DMPs with a learned
meta-parameter function, the problem is reframed in an episodic RL setting. Two
robot applications, dart throwing and ball hitting, highlight the generalization
of the approach.

Trajectories generated by a control policy can be modulated by changing the
parameters of the dynamical system. For instance, \cite{matsubara2011learning}
present parametric DMPs (PDMPs), a way to learn highly-scalable control policies
of basis movement skills using LfD. Unlike style variable models that directly
model joint trajectories, PDMPs encode multiple demonstrations through the
shaping of a parametric-attractor landscape in a set of differential equations.
This allows the DMP extension to learn control policies that synthesize
movements with new motion styles, without losing the useful properties of DMPs
(e.g., stability and robustness against perturbations), by specifying the linear
coefficients of the bases as parameter vectors. The feasibility and extended
scalability of the PDMP framework is demonstrated on a dual-arm robot.

Traditional periodic DMPs do not encode the transient behavior needed to start a
rhythmic motion and thus require ad hoc procedures to initiate periodic
movements. To address this issue, \cite{ernesti2012encoding} introduce a
formulation that encodes both periodic patterns and their transient behaviors
within a single DMP. The approach uses a 2D canonical system where a limit cycle
represents the periodic pattern, while trajectories converging to this limit
cycle encode transient behaviors. Multiple transients from different initial
conditions can be encoded while sharing the same periodic pattern. The system is
validated on a humanoid robot learning wiping motions from human demonstrations,
where different starting positions and wiping patterns (circular and
figure-eight) are successfully reproduced.

DMPs typically consist of a transformation system and a canonical system, where
the latter dictates the decay of the former over time. Prior extensions have
explored the use of multiple DMPs (e.g., library of primitives) to complete
tasks. \cite{kulvicius2011modified,kulvicius2012joining} specifically focus on
the dynamic joining of trajectories. In this approach, overlapping kernels are
utilized across adjacent DMPs instead of employing a single set of parameters
for each individual trajectory. This allows for smoother transitions between the
executions of different DMPs by ensuring continuous velocity and acceleration
profiles at the junction points. The adaptation of goal functions to a piecewise
linear form is shown to enhance the accuracy and flow between movements on a
simulated handwriting task. 

Demonstrations are typically performed slowly to achieve high accuracy, but
robots may need to execute these motions faster while maintaining task
constraints. To address this dilemma, \cite{nemec2013velocity} propose a
speed-scaled DMP formulation that incorporates a phase-dependent temporal
scaling factor for velocity adaptation. Three approaches are compared to learn
this scaling: feedback error adaptation, ILC, and RL to handle intermediate
rewards during episodic learning. In bimanual tasks using dual robot
manipulators, feedback control proved inadequate due to its inability to
anticipate contacts. However, both ILC and RL successfully accelerated the
motion while maintaining force constraints. Additional results on speed
adaption for DMPs are presented in \cite{vuga2016speed}.

A large number of Gaussian approximations needs to be performed when learning a
movement via DMPs. Furthermore, adding them up for all joints yields too many
parameters to be explored when using RL, which requires a prohibitive number of
simulations/experiments to converge to a solution with an optimal reward. To
handle this problem, \cite{colome2014dimensionalitya} propose two alternatives:
(i) explore only along the most relevant directions, (ii) split the trajectory
fitting problem during exploration using a second layer of Gaussians. In
proceeding work, \cite{colome2018dimensionality} improve linear dimensionality
reduction by providing a more intuitive algebraic description of such motions.
This leads to an effective way of dealing with the exploration and exploitation
trade-off in an RL parameter space, and allows for faster algorithm convergence
to a potentially better solution. Experimental results are performed with a
dual-arm robot on a cloth folding task.

The DMP framework in its original form is constrained to kinematic movement.
\cite{gams2014couplinga} were the first to extend DMPs to enable dynamic
behavior via force/torque feedback as follows. First, sensory feedback is
recorded as a robot moves along a trajectory. Next, an iterative control
algorithm is used to learn a coupling term, which is then applied to the
original trajectory in a feed-forward manner. The coupling term can either be
the real force between two manipulators, or it can represent a virtual external
force arising from an interaction. This allows the trajectory to be modified in
accordance to the desired positions or external forces. Both the coupling and
the learning algorithms are shown to be stable (i.e., robust to noise and
systematic errors), and that the method can be applied to use real force
feedback. In follow-up work, \cite{gams2014adapting} provide a comprehensive
evaluation of three approaches for adapting DMPs to external force feedback and
\cite{kramberger2018passivity} extend the coupled DMP framework to
admittance-coupled DMPs.

The original DMP model can be used to build a library of different primitives
and sequences as basic movements to perform complete tasks. However, due to the
generalization constraint of DMPs, not all task variations can be covered.
\cite{yin2014learning} improve the generalization of DMPs by modeling the
movements of the same skills with multiple demonstrations. To do this, the
radial basis function traditionally used in DMPs is replaced with a Gaussian
mixture model (GMM). The GMM is then utilized to learn the joint probability of
the nonlinear dynamical system from multiple demonstrations. This modification
allows for capturing the movement characteristics from several demonstrations of
the same skill, which improves the generalization ability of DMPs. The method is
evaluated on mini-jerk trajectories of static and moving targets via simulation.

An issue that occurs when DMPs define control policies in Cartesian space is
that there exists no minimal, singularity-free representation of orientation.
This creates problems when integrating differential equations on $SO(3)$ since
general methods do not have any information about the structure of the group. As
a result, it can cause the parameters to leave the constraint manifold. DMP
formulations capable of encoding Cartesian orientation were originally
introduced by \cite{pastor2011online} and expanded upon by
\cite{ude2014orientation}. In particular, \cite{ude2014orientation} show how
Cartesian space DMPs (CDMPs) can be defined for non-minimal, singularity-free
representations of orientation (e.g., rotation matrices and quaternions).
Standard extensions (e.g., phase stopping, goal switching, scaling) are
formulated for the case of quaternion-based DMPs via a logarithmic map. This
also includes a combination of phase stopping and standard DMPs to enable a
robot to resume its movement if unexpected perturbations occur. To allow for
more smooth and steady twisting, \cite{liu2019modified} add a dynamical
quaternions goal subsystem to CDMPs.

Additional work that considers the orientation of the system's motion includes a
modified formulation of unit quaternion DMPs by \cite{koutras2020correct}.
Specifically, undesired oscillatory behavior that can produce highly-deviant
motion patterns in \cite{ude2014orientation} is fixed, while guaranteeing the
generation of orientation parameters that lie in $SO(3)$. Although formulations
of DMPs capable of encoding Cartesian orientation have been proposed (e.g.,
\cite{pastor2011online,ude2014orientation}), these methods do not account for
the problem of merging multiple movements. On the other hand,
\cite{saveriano2019merging} describe how DMPs and unit quaternions can be used
to encode orientation trajectories. Three approaches are presented to combine a
set of DMPs and generate a smooth trajectory for a robot via the use of unit
quaternions. 

Compliant MPs \cite{denivsa2015learning,denivsa2016review} extend DMPs to
include torque profiles alongside position trajectories for robot control. The
methodology seeks to improve accuracy and compliance in task execution without
explicit task dynamics models. This is achieved by integrating both position and
torque data, and using statistical learning for task variation adaptation. The
approach ensures safety and compliance in human-robot shared spaces, while
maintaining precise trajectory tracking. Statistical learning is also used to
adapt compliant MPs for new task variations based on a database of existing MPs.
A method to expand this database of motions is initially presented in
\cite{petrivc2015bio} and further developed in
\cite{petrivc2017effect,petrivc2018accelerated}. In
\cite{batinica2017compliant}, the compliant MP framework is broadened to
bimanual tasks.

\cite{dragan2015movement} provide an understanding of DMPs that relates them to
trajectory optimization. To do this, the problem of adapting a trajectory to a
new start and end configuration is formalized as an optimization problem over a
Hilbert space of trajectories. Specifically, the adapted trajectory is defined
as the one that remains closest to the original demonstration while satisfying
the new endpoint constraints. This formalism provides choices for the inner
product, which leads to different adaptation processes. Furthermore, it is shown
that by updating the moving target tracked by the dynamical system, DMPs
implicitly update a moving target to perform this optimization for a particular
norm. Compared to the default norm used in DMPs, an evaluation using synthetic
data as well as a robot arm show that a learned norm enables more accurate
reconstruction of unseen demonstrations.

Despite allowing quick adaptation to novel start and end positions, DMPs may
generate undesired motion profiles that are not under full control. To mitigate
this issue, \cite{cardoso2015novel} propose a solution whereby imitation is
formulated as a global optimization problem. This can allow a much higher degree
of control on the trajectory that imitates the demonstration. More specifically,
the enhanced DMP framework aims to find the trajectory that best imitates the
acceleration profile of the demonstration given the different conditions in
which the movement has to be executed. By formulating the problem in this
manner, a number of additional constraints can be added. For example, these
constraints may allow compliance with joint limits (e.g., positions, velocities,
accelerations, etc.) over the whole trajectory, or to the ability to add
waypoints that were not in the original demonstration.

Early work on DMPs did not investigate the quality of the learned trajectories.
To address the problems of trajectory reproduction quality and efficient
adaptation, \cite{wang2016dynamic} derive a modified version of the DMP
framework called DMP+. The DMP+ formulation utilizes truncated kernels and the
addition of local biases. Truncated kernels provide a mechanism for the forcing
term to converge to zero and ensures the stability of the dynamical system.
Additionally, it limits the number of kernels affected by trajectory
modifications. Local biases can be added to achieve improved modeling accuracy
and to enable efficient updates of learned trajectories. Not only does this
allow for modifying learned trajectories by updating only a subset of kernel
weights, but it also achieves lower reproduction error by better approximating
local gradients of the forcing term. The DMP+ method preserves stability and
convergence, and it can be used with existing DMP techniques.

The original DMP formulation does not consider multi-robot interaction. Past
researchers have targeted the development of a coupling term for the underlying
dynamical system and its associated learning strategies. Yet, the results are
dependent on the quality of the learning methods. Differently,
\cite{zhou2016coordinate} formulate an interactive DMP in a leader-follower
configuration where the relationship between leader and follower is explicitly
represented by a coordinate changed DMP (CCDMP). Instead of using an external
force, CCDMPs define a mapping from the leader to the follower with regard to
position and orientation. This is done via a coordinate transformation matrix
$R^G_L$, where $G$ is the global coordinate and $L$ is the leader's coordinate,
to transform the follower's state from the leader's coordinate to the global
one. The coordinate transformation matrix entries are updated according to the
leader's state. CCDMPs simplify the learning process and are shown to meet the
requirements of several applications (e.g., handover and wiping tasks). 

\cite{calinon2012statistical} and \cite{meier2016probabilistic} both introduce
methods for learning a probabilistic representation of DMPs. In the study by
\cite{calinon2012statistical}, an IL approach is developed that merges
probabilistic machine learning techniques and dynamical systems. The method
relies on the superposition of virtual spring-damper systems to control a
robot's movement, and it extends DMP models by formulating the estimation
problem of dynamical systems parameters as a Gaussian mixture regression (GMR)
problem with projection into various coordinate systems. On the other hand,
\cite{meier2016probabilistic} present a probabilistic representation of DMPs by
reformulating them as a linear dynamical system that incorporates control
inputs. This allows the direct application of algorithms such as Kalman
filtering and smoothing to perform inference on sensor measurements during
movement. Performing inference in this model leads to an automatic feedback term
for online modulation of DMP execution. Moreover, it is shown how inference
enables the measurement of the likelihood of successful execution of a given
motion primitive. 

A standard DMP will continue its time evolution regardless of any significant
perturbation, which is likely to be both undesirable and unintuitive.
\cite{karlsson2017two} address this issue via the design of a 2-DoF controller
for reference trajectory tracking and perturbation recovery. The feed-forward
part of the controller tracks the DMP trajectory in the absence of large
perturbations. This allows for mitigating a slow trajectory evolution due to
temporal coupling acting on small tracking errors. In addition, feedback control
with moderate gains is used to suppress deviations. Nonetheless, when there are
no major disturbances the position error must be small enough such that the
dynamical system does not unnecessarily slow down due to the temporal coupling.
A stability analysis for the proposed temporally coupled DMPs is provided in
\cite{karlsson2018convergence} and the control algorithm is extended to
incorporate orientations in \cite{karlsson2020temporally}.

Despite the popularity of DMPs, an examination of the couplings between
coordinated primitive modules has not been thoroughly investigated.
\cite{wensing2017sparse} provide an analysis of DMPs within the framework of
contraction and introduce a functional tool for DMPs through spatially-sparse
inhibition. The conducted contraction analysis of DMPs provides results related
to scaling primitives in space through general diffeomorphisms, on the stability
of rhythmic DMPs in general networked combinations, and robustness to parameter
heterogeneity in coupled oscillators. In addition to using low-dimensional
inputs to shape rhythmic high-dimensional behavior, DMPs are shown to have the
ability to be globally shaped through a spatially-sparse modification to the DMP
vector fields. This capability is used to manage start/stop transitions for
phase oscillators in robot locomotion experiments.

The goal of task-parameterized skill learning is to provide adaptive motion
encoding for new situations. Motivated by the idea of adapting encoded motions,
\cite{pervez2018learning} introduce the concept of a task-parameterized DMP
(TP-DMP). To tackle the extrapolation issue, a GMM is employed to capture the
local behavior of each demonstration and learning is formulated as a density
estimation problem. By mixing the GMM, task-specific generalization and
supplementary incomplete data to address data sparsity among task parameters is
achieved. This allows for the ability to maintain the local behavior of each GMM
by keeping the means and mixing weights fixed, while adapting the covariances
and mixing coefficients. TP-DMPs can learn in both the task and joint space, and
they can manage the learning of DMP meta parameters.

In addition to encoding movement trajectories, determining how to adapt a DMP
model to stiffness/force is an important issue. \cite{yang2018dmps} address this
problem by extending the DMP framework to generalize movement trajectories and
stiffness profiles learned from human muscle activities. This is done via a
scheme that combines electromyography (EMG) signal-based variable impedance
skill transfer with DMP-based motion sequence planning. Specifically, human limb
muscle movements are monitored for variable stiffness estimation. The extension
can simultaneously encode both trajectories and stiffness profiles in a unified
manner, which allows for generalization among the two representations.
Additionally, a dual-arm control strategy with a haptic feedback mechanism is
designed for skill transfer and evaluations are conducted on dual-arm robot.

The number of DMP basis functions is typically chosen empirically by trial and
error, or by iteratively increasing the number of basis functions and retraining
until the reproduction error drops below a predefined threshold. To determine
the minimum required number of basis functions, \cite{papageorgiou2018sinc}
utilize sinc functions as kernels of DMP models for encoding point-to-point
kinematic behaviors. Concretely, the number of basis functions, which represents
the complexity of the DMP, is computed by the Nyquist-Shannon sampling frequency
and the learning requires only the acquisition of the samples. The following use
cases are examined: (i) when a given frequency band is needed and (ii) when a
given reproduction accuracy is desired. The minimum required kernels to achieve
a given accuracy is shown to be much less than an LWR Gaussian-based DMP.

In LfD there can be a large variability in the speed of execution across
demonstrations. This can lead to problems when multiple demonstrations are
compared to extract relevant information for learning. \cite{ude2016trajectory}
handle this issue by proposing a DMP extension, called an arc-length DMP
(AL-DMP), where the spatial and temporal components of motion are separated.
This is done by formulating the DMP equations as derivatives of the arc-length
instead of time. The speed of movement, which contains information about timing,
is separately encoded. Compared to prior work where the issue of variability in
the speed of execution of the demonstrations is addressed by estimating the
optimal time alignment, this is not necessary for AL-DMPs since the learning of
spatial movement is independent of the timing issues. Additional results along
with the limitations of AL-DMPs are presented in \cite{gavspar2018skill}.

Allowing online modifications of a DMP trajectory can result in a loss of
predictability, which is undesirable in scenarios where the velocity needs to be
limited. \cite{dahlin2019adaptive} propose temporal coupling for DMPs to tackle
the challenge of generating adaptive trajectories under velocity constraints,
while preserving the intended path shape. DMPs are utilized for trajectory
modeling, which offers features for online trajectory adjustment through spatial
and temporal parameter adaptation. Relying on a repulsive potential that
maintains the velocity within specified limits, the approach intrinsically
guarantees path shape invariance. The stability of the method is proved for the
continuous time case and verified through simulations. 

DMPs are extensively utilized for trajectory planning, yet the original
framework can not generate trajectories for the following situations: (i) the
goal point is the same as the start point, (ii) the goal point is close to the start
point, and (iii) the goal point intersects with the start point. Additionally,
when the original trajectory is on a curved surface, it is difficult to ensure
that the generalized trajectory remains on the surface. To solve these
problems, \cite{han2019trajectory} develop a modified DMP method with a scaling
factor and acceleration coupling term as follows. First, an adjusted cosine
similarity is used as the performance index of the generalized curve. Then, the
cosine similarity is optimized to obtain the scaling factor. Lastly, trajectory
optimization and force control with surface constraints are achieved by adding
the acceleration coupling term to the original DMP. More results on the
modified DMP framework are discussed in \cite{han2022modified}.

For many tasks (e.g., part assembly/disassembly) reverse execution would allow
automatic derivation of certain required operations from their forward
counterparts. However, the original DMP framework does not support
reversibility. The first approach towards reversing DMPs was proposed by
\cite{nemec2018efficient}, which was followed by \cite{iturrate2019towards} and
\cite{nemec2019incremental,simonivc2021analysis}. The later methodologies
require the use of two separate DMPs to achieve reversibility.
\cite{sidiropoulos2021reversible} propose a formulation that enables
reversibility via a linear system with a globally asymptotically stable origin
using a single DMP. This not only permits replaying a trajectory back from its
goal, but also moving back and forth along the learned path anywhere during the
execution. Additionally, training requires only position measurements and is
decoupled from the DMP's stiffness and damping gains.

\cite{ugur2020compliant} extend DMPs to parameterize trajectories based on
specific properties of the environment. The extension, called a compliant
parametric DMP, can encode and generate complex trajectories, learn sensory
feedback models observed during action executions, and automatically switch to a
mode that enables external help during high-variance portions of a demonstrated
skill. For instance, the method can allow a robot to change to a mode that
follows an average trajectory, hence complying with external forces during
execution of the corresponding part of the skill. Compliance is realized by
exploiting the difference between the actual and expected sensory feedback, and
is computed from learned sensory feedback models. The base system extends DMPs
to encode and reproduce the required actions, while parametric hidden Markov
models (HMMs) are used to encode the nonlinear part of the trajectories.

DMP encoded movements are designed to be executed towards a stationary goal
known before the start of the trajectory. Although a DMP can be perturbed to
another static goal while in motion, it cannot adapt to a non-stationary goal.
\cite{koutras2020dynamic} present an enhancement to DMPs that provides
adaptation to moving goals without depending on any known or estimated model for
the goal's motion except for the current position and velocity. The proposed
method is designed to maintain the demonstrated velocity levels throughout a
motion. By leveraging an adaptive temporal scaling parameter that is typically a
constant, reaching the moving goal is achieved while retaining the learned
kinematic pattern. The methodology is theoretically proved and validated by both
simulations and experiments.

DMPs need just one demonstration to learn a desired trajectory, as well as
spatial and temporal scaling characteristics. The spatial scaling of the DMP
formulation is performed separately for each coordinate, thus it is frame
dependent. Nonetheless, this can be undesirable in tasks that require
preservation of the 3D path pattern. To remedy the problem,
\cite{koutras2020novel} propose a formulation that has global scaling abilities.
The concept is based on the observation that DMPs perform different motion
generalizations by selecting distinct coordinate systems. More exactly, the
spatial scaling problems present in the other formulations stem from the frame
coordinate dependence of the scaling. The approach fixes this issue by
appropriately rotating the executed trajectory and scaling its magnitude to
alleviate frame dependency.

For many robot control problems, aspects such as stiffness matrices, damping
matrices, and manipulability ellipsoids, are naturally described as symmetric
positive definite (SPD) matrices to capture the geometric characteristics of
these factors. However, DMPs cannot be directly used with quantities expressed
as SPD matrices since they are limited to Euclidean space. To correct this,
\cite{abu2020geometry} introduce the use of Riemannian metrics to reformulate
DMPs such that they can operate with SPD data on an SPD manifold. The extension
of DMPs to Riemannian manifolds allows for the generation of smooth trajectories
using data that does not belong to the Euclidean space. In follow-up work,
\cite{abu2024unified} use differential geometry to extend the original DMP
framework to other Riemannian manifolds. This allows discrete DMPs to
effectively represent data evolving on different Riemannian manifolds, which
permits the generation of smooth trajectories with data that does not belong to
Euclidean space. \cite{xu2024imitating} extend this geometry-aware DMP approach
to not only adapt learned trajectories to SPD via points, but also establish a
correlation between SPD-based trajectories and position trajectories across
different nonlinear motion velocity scales.

Even though DMPs have been correctly formulated to learn point-to-point
movements for both orientation and translation, periodic movements are missing
from the initial framework. To close this gap, \cite{abu2021periodic} present a
DMP formulation that enables the encoding of periodic orientation trajectories.
Within this framework the following approaches are developed: (i) Riemannian
metric-based projection DMPs, and (ii) unit quaternion-based periodic DMPs.
Riemannian metric-based projection DMPs exploit the fact that the space of unit
quaternions is a Riemannian manifold that locally behaves as a Euclidean space
(i.e., tangent space). Hence, unit quaternion trajectories can be projected onto
the tangent space, then a periodic DMP can be fitted, and finally the output of
the DMP can be projected onto the unit quaternion manifold. Unit
quaternion-based periodic DMPs directly encode the unit quaternion trajectory
and ensure the unitary norm of the integrated quaternions.

Modifications to the DMP framework that guarantee additional robustness are
proposed by \cite{ginesi2021overcoming}. To do this, a set of compactly
supported basis functions is introduced and used during the learning process.
These basis functions not only provide an accurate approximation, but they also
have analytical and numerical advantages with respect to Gaussian basis
functions. Specifically, the invariance of DMPs to affine transformations can be
utilized to improve trajectory generalization against the choice of
hyperparameters and goal positions. In addition, an algorithm that learns a
unique DMP from multiple observations without the need to rely on probabilistic
approaches or additional parameters is developed.

\cite{cohen2021motion} introduce a hierarchical method for the adaptation of DMP
parameters. Specifically, the parameters are updated during runtime based on an
a priori learned map of the task and DMP meta-parameter manifold. To learn the
mapping, two approaches are compared: kernel estimation and deep neural networks
(DNNs). For efficient learning, the parameter space is analyzed using principal
component analysis (PCA) and locally linear embedding. PCA aids in finding the
regression equations, the number of required kernels, and for determining values
of the grid search for the hyperparameters of the DNN. Locally linear embedding
helps in obtaining a deeper understanding of the model complexity. Low runtime
estimation errors are obtained for both learning methods with an advantage
towards using kernel estimation when datasets are small.

Placing constraints in the joint and task spaces due to force interactions is
common in LfD. Ideally, DMPs should have the flexibility to incorporate
additional terms according to the constraints of the task. To provide this
capability, \cite{lu2021constrained} extend the DMP framework to handle
constrained trajectory planning via barrier Lyapunov functions (BLFs). This
allows for analyzing the stability of the system and restricting it to a set of
constraints. Concretely, BLFs are adopted to compute an additional acceleration
term and convergence of the generalized trajectories is proved. This ensures
trajectory stability and allows DMPs to compensate for dynamic changes in the
environment (e.g., obstacle avoidance, cooperative manipulation, etc.).

A DMP is a second-order dynamics system that generates position and velocity
trajectories by encoding motion in joint space. Nonetheless, feed-forward
control of robots with flexible joints is known to require reference
trajectories up to the fourth derivative of the position. To bridge this gap,
\cite{wahrburg2021flexdmp} develop an extension of the DMP framework towards
robots with flexible joints. The method retains the properties of inherent
stability, temporal and spatial invariance, and linearity in the parameters
describing the motion. Furthermore, the canonical system and forcing function
are structurally the same as a DMP. The key difference is in using a
fourth-order system for generating trajectories and in applying the forcing term
to the derivative of the jerk. The extension is demonstrated via simulation and
on a collaborative manipulator.

Due to the irreversible nature of traditional DMPs, a robot employed for an
assembly task cannot deal with part jamming situations. \cite{zhao2023robotic}
alleviate this issue through the development of reversible DMPs and trajectory
optimization as follows. First, the original DMP framework is made reversible by
extending it to Cartesian space with the following requirements: (i) global
asymptotic stability with convergence to the target point in the forward
execution and to the initial point in the backward execution; (ii) the
trajectory of the forward execution and backward execution must be the same.
Second, an optimization strategy based on the trajectory reproduced by the DMP
is designed to improve the assembly compliance. Three robotic peg-in-hole
assembly experiments are carried out on a robot platform to show the
effectiveness of the method.

Skills learned by DMPs are difficult to deploy when the use cases are
significantly different from the demonstrations. To address this limitation,
\cite{lu2023trajectory} extend DMPs for robot trajectory learning via an
adaptive neural network control method. The approach uses multi-mapping feature
vectors to rebuild the forcing function of the DMP. After creating the feature
vector for a new task, the adaptive neural network controller compensates for
dynamics errors and changed trajectories. The controller is incrementally
updated, and it can accumulate and reuse learned knowledge to improve learning
efficiency. A set of experiments highlights incremental skill learning with
higher tracking accuracies. 

In some cases the spatial generalization of DMPs can be problematic and lead to
excessive/unnatural spatial scaling. \cite{sidiropoulos2024dynamic} tackle this
issue by incorporating dynamic via points to adjust the DMP trajectory. To do
this, an online adaptation scheme is derived for the DMP weights, which is based
on minimizing the distance from the learned acceleration profile under the
equality constraints of the via points. Changes to the via points occur if they
are defined with respect to the position of the target/obstacle that is
perturbed during execution. Inequality constraints can also be imposed to
generate feasible trajectories in the presence of kinematic bounds related to
the robot and task environment. A real-world validation of the method is carried
out via experimental packing scenarios.

Current DMP formulations heavily rely on nonlinear forcing terms to fit
demonstrated trajectories, as their linear dynamical systems provide limited
expressiveness. \cite{stulp2024fitting} handle this restriction by using a
generalized logistic function as a delayed goal system. This enables the
dynamical system to inherently generate bell-shaped velocity profiles typical of
human movement. By optimizing the dynamical system to better fit demonstrations
before training the forcing term, the magnitude and variance of the forcing term
coefficients are significantly reduced. Using three human demonstration
datasets, the approach is shown to improve trajectory fitting compared to
previous DMP methods. Additionally, a robotic coat-hanging task highlights the
interpolation accuracy between demonstrated trajectories.

\subsection{Limitations}
DMPs are well understood and widely used. Yet, there exist several shortcomings
that limit their usage in robotic applications. These drawbacks include a
general reliance on Gaussian basis functions to perform function approximation,
the dependence on the choice of hyperparameters and desired goal position, and
the fact that DMPs are constrained to learn from a single unique demonstration. 

\section{Probabilistic Movement Primitives}
\label{sec:probabilistic_movement_primitives}
DMPs possess several advantages given their ability to efficiently represent an
MP using a single demonstration. However, what if there are multiple
demonstrations that share the neighborhood of a single type of movement?
Although multiple DMPs could be used to separately learn each demonstration,
they lack an effective way to blend these movements together. The ability to
combine and adapt the learned primitives by considering the distribution of the
demonstrations motivates the idea of a probabilistic movement primitive (ProMP)
\cite{paraschos2013probabilistica,paraschos2013probabilisticb}.

\subsection{Background}
ProMPs can be used to encode a generalized representation of multiple
demonstrations by learning a basis-function parameterized distribution of the
trajectories. These trajectories, similar to DMPs, may be acquired from
different spaces (e.g., joint or Cartesian space), and are also temporally
scalable by abstracting time to a phase variable. We retain the DMP notation for
trajectories, namely $q_{i,j}$ for each joint $j$ and time sample $i$. Let
$w_j\in\bbR^{1 \times L}$ be a weight matrix with $L$ terms. A linear basis
function model is then given by
\begin{equation}
  {y}_{i, j} = \begin{bmatrix} q_{i, j}\\ \dot{q}_{i, j} \end{bmatrix}
             = {{\Phi}(i) {w_j} + {\xi }_{x_j}},
\end{equation}
where ${\Phi}(i) = \begin{bmatrix}\phi(i) & \dot\phi(i) \end{bmatrix}^\top\in
\bbR^{2 \times {L}}$ is the time-dependent basis function matrix, and $L$ is the
number of basis functions. 

A ProMP encodes a Gaussian distribution over the weight vector $w_j$ and the
parameter vector $\theta_j = \{\mu_{w_j},\Sigma_{w_j}\}$ to reduce the number of
estimation parameters. Marginalizing the weights out renders the trajectory
distribution to 
\begin{equation}
  p({Q_j},{\theta_j}) = \int p{({Q_j}\,|\,w_j)p(w_j;{\theta_j})dw_j}.
\end{equation}
The distribution $ p({Q_j},{\theta_j})$ is defined as a hierarchical Bayesian
model (HBM) over the trajectories $Q_j$ \cite{paraschos2018using} and
{$p(w_j\,|\,\theta_j) = \mathcal{N}(w_j\,|\,\mu_{w_j},\Sigma_{w_j})$}. The
distribution of the state $p(x_{i,j}\,;\,{\theta_{j}})$ is provided by
\begin{equation}
  p(x_{i, j};{\theta_{j}}) \mbox{=} \mathcal{N}{(x_{i, j}\,|\,{\Phi}(i){\mu}_{w_j},{\Phi}(i){\Sigma}_{w_j}{\Phi}(i)^\top\mbox{+}{\Sigma}_{x_j})}.
  \label{eq:promp_eqn_1}
\end{equation}

A trajectory can be generated from a ProMP distribution using $w_j$, the basis
function {${\Phi}(i)$}, and \eqref{eq:promp_eqn_1}. The basis function is chosen
based on the type of robot movement, which can be either discrete or rhythmic
(analogous to DMPs), as well as the complexity and length of the demonstrations.
From \eqref{eq:promp_eqn_1}, the mean $\tilde{\mu}_{i, j}\in \bbR^{2}$ of the
ProMP trajectory at $i$ is ${\Phi}(i){\mu}_{w_j}$ and the covariance $\Sigma_{i,
j}$ is ${\Phi}(i){\Sigma}_{w_j}{\Phi}(i)^\top+{\Sigma}_{x_j}$. 

Multiple demonstrations are performed to learn a distribution over $w_j$, and
more demonstrations are desirable as the DoF increase. A combination of radial
and polynomial basis functions may be used for training a ProMP. From the
demonstrations, the parameters ${\theta_j}$ can be estimated using maximum
likelihood estimation (MLE) \cite{lazaric2010bayesian}. However, this may result
in unstable estimates of the ProMP parameters when there are insufficient
demonstrations. Thus, regularization is used to estimate the ProMP distribution
and it is also common to use a normal-inverse-Wishart distribution as a prior
distribution $p(\theta_j)$ to increase stability when training the ProMP
parameters. To obtain the posterior distribution over a ProMP, $\theta_j$ is
maximized via EM:
\begin{equation}
  p({\theta_j}\,|\,x_{i, j}) \propto p(\theta_j) p(x_{i, j}\,|\,\theta_j).
\end{equation}

The probabilistic framework employed by ProMPs allows one to model the coupling
between the DoF of a robot by estimating the covariance between different DoF.
For instance, working with trajectory distributions enables the modulation of
movements to a novel target by conditioning on the desired target's positions or
velocities. Hence, a ProMP can directly encode optimal behavior in systems with
linear dynamics, quadratic costs, and Gaussian noise \cite{todorov2002optimal}.
Moreover, the distributions can be defined in either joint space, task space, or
any other space that accommodates the application. 

Serving as a bridge between traditional deterministic MPs and probabilistic
modeling, ProMPs demonstrate versatility in capturing and generalizing complex
robotic movements across a wide variety of tasks. In formulating ProMPs,
\cite{paraschos2013probabilistica} and \cite{paraschos2013probabilisticb}
evaluated stroke and rhythmic movements across different robot manipulation
scenarios. For instance, a simulated task involves a 7-link planar robot that
must reach a target position in end-effector space while also reaching a via
point at a certain point in time. Other physical tasks include a robot playing
table tennis, Astrojax, hitting a ball, shooting a hockey puck in various
directions and distances, and a shaking task that must use the appropriate speed
to generate the desired sound of a musical instrument (e.g., maracas).

\subsection{Improvements and Extensions}
It is well known that the intrinsic dimensionality of many human movements is
small when compared to the number of employed DoF. Therefore, the movements may
be represented by a small number of synergies encoding the couplings between
DoF. In light of this observation, \cite{colome2014dimensionalityb} apply
dimensionality reduction to ProMPs to address the challenges posed by
high-dimensional robotic systems. By extracting data from a set of
demonstrations using probabilistic dimensionality reduction techniques,
trajectories are encoded in a low-dimensional space. The approach demonstrates
efficiency in both encoding trajectories and applying RL with relative entropy
policy search (REPS), which learns the parameters that maximize the expected
reward and can be used for arbitrary policy representations.

Force-control typically requires an accurate dynamics model of the robot and its
environment, which may not be easy to obtain. Although a dynamics model can be
learned, the process can be time-consuming and error-prone. Alternatively, the
desired movement of the robot and the contact forces can be jointly learned
through human demonstrations without relying on a forward or inverse model. For
example, \cite{paraschos2015model} present an approach that first extends the
ProMP framework to encode a state-action distribution, and then derives a
stochastic feedback controller without the use of a system dynamics model. They
show that model-free ProMPs can generalize to different grasping locations by
exploiting the correlations between motor commands and force feedback.
Additional details on sensory integration, a mixture model of primitives, and
how the method can be used for adapting the interaction forces to a user's input
are provided in \cite{paraschos2018probabilistic}.

Since ProMP parameters are high dimensional, a common practice for generalizing
to new tasks is to adapt only a small set of control variables known as meta
parameters. However, the encoding of these meta parameters is precoded in the
representation and thus cannot be adapted to novel tasks. Instead of relying on
fixed meta-parameter representations, \cite{rueckert2015extracting} introduce a
method for learning the encoding of task-specific control variables from data.
This is done by using HBMs to estimate a low-dimensional latent variable model
for ProMPs. The approach outperforms standard ProMPs in terms of learning and
generalization from a small amount of data. The HBMs are also extended by a
mixture model, which allows for the modeling of different movement types in the
same dataset. Experimental results on two kinesthetic teaching datasets show
that the control variables can be used to analyze the data or to generate new
robot trajectories. 

\cite{paraschos2017probabilistic} combine Bayesian task prioritization
(\cite{toussaint2010bayesian}), which allows for the computation of combined
motor control signals from multiple tasks at different operational spaces with
ProMPs. Concretely, they extend Bayesian task prioritization to provide a more
general derivation for torque control and show that existing prioritization
techniques are a special case of the Bayesian approach. This representation is
then used to solve a variety tasks with different end effectors. The LfD method
not only minimizes the amount of expert knowledge required, but it also avoids
the problem of specifying a cost function for a given task. It can be used to
adapt task-space movements without solving an inverse kinematics problem, while
staying close to the demonstrated data. Lastly, a ProMP controller extension is
proposed to handle deviations that occur from the demonstrated movements due to
the prioritization.

Appropriate robot responses to human coworkers during the execution of learned
movements are critical for fluent task execution, safety, and convenience. To
facilitate such responsive behaviors with ProMPs in human-robot interaction
(HRI) scenarios, a robot needs to be able to react to a human collaborator
during the execution of the ProMP. \cite{koert2019learning} enhance ProMPs by
integrating a goal-based intention prediction model with the purpose of enabling
robots to respond in real-time to their human counterparts during the execution
of learned movements. Two methods for intention-aware online adaptation are
introduced: (i) online spatial deformation to dynamically modify the shape of
the ProMP trajectories to prevent collisions; (ii) online temporal scaling to
adjust the velocity profile of a ProMP to evade time-dependent collisions.
Notably, the research reveals that these adaptations are not merely theoretical
improvements. Instead, the results show a tangible increase in perceived safety
and comfort for non-expert users, particularly during spatial deformation.

Learning a new robot skill, without prior knowledge, necessitates the
exploration of a large space of motor configurations. Nonetheless, time can be
saved by restricting the parameter search space and initializing it with the
solution of a similar task. Along this vein, \cite{stark2019experience} develop
a learning framework to facilitate the transfer of knowledge from existing
learned movement skills to a new task to reduce the need for learning from
scratch. By fusing ProMPs with descriptions of their effects, new skills are
initialized with parameters extrapolated from related ProMPs. This approach
results in a more efficient learning process. The real-world impact of this
improvement is demonstrated in an object pushing task involving a simulated
3-DoF robot, where the iterations required for learning a new task were reduced
by over 60\%, signifying a substantial improvement in learning efficiency and
skill quality.

In high-dimensional robotic systems, the number of ProMP parameters scales
quadratically with the dimensionality. \cite{colome2020dimensionality} mitigate
this issue by utilizing dimensionality reduction techniques via EM to extract
the unknown synergies from a given set of demonstrations while maximizing the
log-likelihood function with respect to the demonstrated motions. This provides
a reduced and intuitive algebraic description of the motion, addressing the
challenge of a large number of Gaussian approximations needed when using ProMPs.
Furthermore, using EM instead of PCA is beneficial since ProMPs provide a
time-dependent variance profile. Time points with a low variance are important
for movement, but the PCA approach is not permitted to distort these points.
These dimensionality reduction techniques are shown to be more efficient in
encoding a trajectory from data and for applying RL.

ProMPs capture the variability of an operator's demonstrations as a probability
distribution over trajectories. Therefore, they provide a sensible region of
exploration along with the ability to adapt to changes in the robot's
environment. Nonetheless, in comparison to their deterministic counterparts,
ProMPs require the estimation of a greater number of parameters to capture
variability and correlations between different robot joints.
\cite{gomez2020adaptation} tackle this issue by making use of prior
distributions over the parameters of a ProMP. This allows for robust estimates
of the parameters using fewer training instances. Furthermore, the influence of
the prior distribution decreases as more training data becomes available with
convergence to the MLE. Additionally, general-purpose operators to adapt ProMPs
in joint and task space are presented. The proposed training method and
adaptation operators are tested on a coffee preparation scenario and playing
robot table tennis. 

Due to the high dimensionality of the MP parameter space, policy optimization is
expensive in terms of both samples and computation. Nonetheless, motions in
highly-redundant kinematic structures exhibit a high correlation within the
configuration space. Based on this observation, \cite{tosatto2020dimensionality}
build upon the general framework of ProMPs and apply dimensionality reduction in
the parameter space using PCA. The key idea is to find a distribution of linear
combinations of the principal movements. The approach is tested both on a
robotic task (pouring) as well as the reconstruction of human movements.
Compared to dimensionality reduction in the configuration space, the methodology
exhibits a significant reduction in the parameter space with a modest loss of
accuracy.

The original ProMP formulation only provides solutions to specific movement
adaptation problems (e.g., obstacle avoidance). To address this problem,
\cite{frank2021constrained} propose a generic probabilistic framework for
adapting ProMPs. Specifically, previous adaptation techniques (e.g., various
types of obstacle avoidance, mutual avoidance, waypoints, etc.) are unified into
a single framework. This is done by formulating adaptation as a constrained
optimization problem. Concretely, the Kullback-Leibler (KL) divergence between
the adapted distribution and the distribution of the original primitive is
minimized while the probability mass associated with undesired trajectories is
constrained to be low. This allows for the construction of libraries of
probabilistic primitives and enables further downstream adaptation or
co-activation of primitives. Using this framework, adaptation techniques such as
temporally unbounded via points (i.e., additional points the system must pass
through to carry out a task) and dual-arm obstacle avoidance are derived. 

Within the ProMP framework, providing or requesting useful demonstrations may
not be easy, and quantifying what constitutes a good demonstration in terms of
generalization capabilities is nontrivial. \cite{kulak2021active} provide a
solution to this problem via an active learning method for contextual ProMPs
with the aim of improving the generalization capabilities by relying on fewer
demonstrations. Bayesian inference is utilized to quantify both aleatoric and
epistemic uncertainties in the ProMPs as follows. First, a ProMP is learned with
a Bayesian Gaussian mixture model. Then, the epistemic uncertainties captured by
the Bayesian Gaussian mixture model are used as an information gain metric to
actively learn the ProMPs. The effectiveness of the approach is demonstrated
both in simulation and on a 7-DoF robot arm.

ProMPs can be used to learn Cartesian movements, however the framework does not
handle quaternion trajectories. The use of quaternion trajectories involves the
following difficulties: (i) there is no closed-form solution for learning the
model parameters, and (ii) trajectory retrieval is constrained to lie on the
sphere $\mathcal{S}^3$. \cite{rozo2022orientation} solve these problems by
developing the first Riemannian manifold formulation of a ProMP framework that
makes the encoding and reproduction of full-pose end-effector trajectories
possible. The method provides via point modulation and blending capabilities,
which are naturally inherited from the original ProMP framework. In addition,
the Riemannian ProMPs are not susceptible to inaccuracies caused by
geometry-unaware operations. The approach is tested on synthetic data examples
along with LfD robot manipulation experiments.

In the ProMP framework, a sequential online learning method is employed whereby
only one data point is considered at each time and then the model parameters are
correspondingly updated. However, as the number of new data points to be fitted
increases, old points are no longer fitted accurately by the model. Not only do
\cite{fu2022probabilistic} demonstrate that the degree of uncertainty in the
prediction distribution gradually decreases as the number of observed data
points increases, but a solution to the problem is proposed using a weight
combination algorithm as follows. First, every point to be fitted is processed
one by one and the basis functions that fall within a highly-correlated range
with the point to be fitted are involved in the regression operation. Second,
the weight vector components corresponding to these basis functions are
concatenated and combined to obtain the complete weight vector. The approach is
mathematically proven to be better than the traditional online algorithm.

As described in Section~\ref{subsec:dmp_background}, DMPs denote a trajectory
using a forcing term instead of a direct representation of the trajectory
position. Therefore, numerical integration from acceleration to position has to
be applied to formulate the trajectory. This constitutes an additional workload
and makes the estimation of trajectory statistics difficult. On the other hand,
ProMPs are able to amass such statistics. This makes them the fundamental
enablers for acquiring variable-stiffness controllers along with a trajectory's
temporal and DoF correlation. Based on these observations, \cite{li2023prodmp}
introduce ProDMPs, a unified perspective of both DMPs and ProMPs. They show that
the trajectory of a DMP can be expressed by a linear basis function model that
depends on the parameters of the DMP (i.e., the weights of the forcing function
and the goal attractor). Akin to ProMPs, the linear basis functions can be used
to represent trajectory distributions while maintaining all the properties of
dynamical systems.

By maintaining the first-order differential relationship between velocity and
position via basis functions, ProMPs exhibit strong generalization abilities.
However, ProMPs are difficult to express in the state equation form.
\cite{wu2024probabilized} address this limitation by formulating a
\textit{probabilized} DMP. Specifically, a form of the state equation is
constructed that combines an HBM with the DMP equation.  This enables the
expression of the trajectory distribution for multi-trajectory learning while
retaining the convenience of ProMPs. Moreover, it provides the convergence
characteristics of the state equation representation. A set of IL experiments
show the method obtains conformability of the demonstrated trajectory under
strict constraint conditions.

ProMPs can enable robots to learn complex tasks from demonstrations where motion
trajectories are represented as stochastic processes with Gaussian assumptions.
Nevertheless, despite their computational efficiency, ProMPs have limited
expressiveness in capturing multimodal motion. To address this factor,
\cite{yin2024stein} combine variational inference and ProMPs to tackle the
challenge of learning skills from a set of multimodal demonstrations with
nonparametric distributions. Concretely, ProMP adaptation is formulated as
nonparametric probabilistic inference using Stein variational gradient descent.
As a result, generated trajectories are not limited to a Gaussian distribution.
They can be adapted via probabilistic inference while taking into account prior
demonstrations, even when the demonstrations are multimodal. The framework is
evaluated in environments with obstacles and new targets, using both simulation
and a real robot manipulator.

\cite{zhang2025wavelet} develop wavelet MPs, a framework that integrates ProMPs
with a discrete wavelet transform to model and learn discrete and rhythmic
trajectories from demonstrations. This eliminates the need for task-specific
basis function selection and simplifies the hyperparameter tuning process
commonly required in ProMPs. Furthermore, the creation of a local frame enables
the learning of discrete motion skills in one or more local coordinate systems
and generalizes them to any position in global space, ensuring stable trajectory
generation. Wavelet MPs leverage a time-frequency domain representation, which
provides a broader range of trajectory modeling capabilities while maintaining
lower computational complexity than Fourier-based methods (e.g.,
\cite{kulak2020fourier}). The method is validated through both virtual
environment tasks and robot stirring experiments. 

\subsection{Limitations}
The necessary additional preprocessing of demonstrations can be seen as a
limitation of ProMPs. Another drawback is that ProMPs assume temporal alignment
of the data by both aligning the starting points of a demonstration execution
and their durations. In practice, ProMP performance is sensitive to the
selection of basis parameters and thus can be difficult to tune. ProMPs are also
not suitable for high-dimensional input spaces since selecting the correct basis
functions becomes more challenging and computationally complex. Lastly, ProMPs
treat the distribution space of the demonstrations as a global frame of
reference, i.e., they cannot be readily extrapolated outside that space. 

\section{Kernelized Movement Primitives}
\label{sec:kernelized_movement_primitives}
Kernelized movement primitives (KMPs) \cite{huang2019kernelized} seek to
preserve the probabilistic strengths of ProMPs while better generalizing to
high-dimensional systems and the extrapolation of states that DMPs are capable
of. KMPs represent a nonparametric solution for LfD using kernels. They were
chiefly formulated to limit the use of hyperparameters and model
high-dimensional robot input spaces.

\subsection{Background}
In the KMP framework, demonstration data is represented by $\{\{
\vec{s}_{n,h},{\vec{\xi}}_{n,h}\}_{n=1}^{N}\}_{h=1}^{H}$, where $\vec{s}_{n,h}
\in \mathbb{R}^{\mathcal{I}}$ is the input and ${\vec{\xi}}_{n,h} \in
\mathbb{R}^{\mathcal{O}}$ is the output. $\mathcal{I}$, $\mathcal{O}$, $H$, and
$N$ represent the dimensionality. This representation allows KMPs to model
complex relationships between input and output variables. It makes them
well-suited for a wide range of applications in robotics and control systems.

To model the relationship between inputs and outputs, a reference database,
$\{\vec{s}_n,\hat{\vec{\mu}}_{n},\hat{\vec{\Sigma}}_{n}\}_{n=1}^{N}$, is
extracted using a GMM \cite{calinon2007learning}: 
\begin{equation}
  \left[\begin{matrix}
    \vec{s}\\\vec{\xi}
  \end{matrix}\right] \sim \sum_{c=1}^{C} \pi_c 
  \mathcal{N}(\vec{\mu}_c,\vec{\Sigma}_c).
  \label{eq:kmp_gmm}
\end{equation}
Here, $\pi_c$ represents the prior probability, and $\vec{\mu}_c$ and
$\vec{\Sigma}_c$ are the mean and covariance of the $c$-th Gaussian component in
the GMM. \eqref{eq:kmp_gmm} illustrates how the mixture of Gaussian
distributions are used to approximate the underlying distribution of the
demonstration data. This provides a probabilistic way to represent the data and
allows for better generalization when predicting new outputs given input
queries.

Given an input query $\vec{s}^*$, the KMP prediction can be computed by
calculating $\Sigma, \mu, \vect{K},$ and $\vec{k}^*$. Concretely, 
\begin{equation}
  {\vec{\mu}}_w^{*}=\vec{\Phi} ( \vec{\Phi}^{\top} \vec{\Phi} +\lambda \vec{\Sigma} )^{-1} {\vec{\mu}},
  \label{eq:kmp_muw}
\end{equation}
where
\begin{equation}
\begin{aligned}
  \vec{\Phi}   &= [\vec{\Theta}(\vec{s}_1) \ \vec{\Theta}(\vec{s}_2) \ \cdots \ \vec{\Theta}(\vec{s}_N)],\\
  \vec{\Sigma} &= \mathrm{blockdiag}(\hat{\vec{\Sigma}}_1, \ \hat{\vec{\Sigma}}_2, \ \ldots, \ \hat{\vec{\Sigma}}_N), \quad \\
  {\vec{\mu}}  &= [\hat{\vec{\mu}}_1^{\top} \ \hat{\vec{\mu}}_2^{\top} \ \cdots \ \hat{\vec{\mu}}_N^{\top}]^{\top}.
\end{aligned}
\label{eq:notations:define}
\end{equation}
In \eqref{eq:kmp_muw}, the weights ${\vec{\mu}}_w^{*}$ are computed using a
kernel ridge regression framework. $\lambda$ is a regularization parameter in
kernel ridge regression that balances fitting the training data and avoiding
overfitting. $\vec{\Phi}$ is a matrix of transformed input data, with each
column $\vec{\Theta}(\vec{s}_n)$ representing a feature mapping or kernel
function transformation of the input data. The KMP prediction as a function of
the kernel matrix and the learned weights is computed as
\begin{equation}
  \mathbb{E}(\vec{\xi}(\vec{s}^{*})) =\ \vec{{k}}^{*} (\vec{{K}}+\lambda \vec{\Sigma})^{-1} {\vec{\mu}}.
  \label{eq:kmp_mean}
\end{equation}
The kernel function $\vec{k}$, crucial in capturing the relationship between
inputs and outputs, is transformed into a kernel matrix $\vec{K}$, which is used
in the computation of the expected output. This prediction effectively estimates
the expected output value given a new input query.

The covariances, as derived in \cite{huang2019kernelized}, are given by 
\begin{equation}
  \mathbb{D}(\vec{\xi}(\vec{s}^{*})) = \frac{N}{\lambda}\left(\vec{k}(\vec{s}^{*}, \vec{s}^{*}) -\vec{k}^{*}(\vec{K}+\lambda \vec{\Sigma})^{-1} \vec{k}^{*\top}\right),
  \label{eq:kmp_var}
\end{equation}
where
\begin{equation}
  \vec{k}^{*} = [\vec{k}(\vec{s}^{*}, \vec{s}_{1}) \; \vec{k}(\vec{s}^{*}, \vec{s}_{2}) \; \cdots \; \vec{k}(\vec{s}^{*}, \vec{s}_{N})].
\label{eq:kmp_kstar}
\end{equation}
These covariances represent the uncertainty associated with the KMP prediction.
This is a useful measure when considering the reliability of the predictions and
it can be used to guide decision-making in robotics and control systems.
\eqref{eq:kmp_kstar} defines the kernel function $\vec{k}^{*}$, which is
utilized to compute $\vec{{K}}$, and is a key component of the KMP framework.
By leveraging the kernel-based approach, KMPs are able to handle
high-dimensional inputs and complex relationships between inputs and outputs,
hence making them a powerful and flexible solution for IL for robotic systems.

\cite{huang2019kernelized} demonstrate the utility of KMPs for robot skill
learning through the generation of trajectories for the following scenarios. In
the first scenario, KMPs are utilized to define a force-based adaptation problem
using new via points that are a function of the sensed forces. This force is
then used to determine a desired via point by the KMP, which the robot needs to
pass through in order to avoid an obstacle. The second scenario consists
employing KMPs to learn time-driven trajectories for a robot-assisted task
(e.g., soldering). In this situation the following properties are demonstrated:
(i) the task is accomplished using a single KMP without any trajectory
segmentation for different sequential subtasks; (ii) KMPs make the adaptation of
learned skills associated with high-dimensional inputs feasible; (iii) KMPs are
driven by the user hand positions, which allows for slower/faster hand movements
since the prediction of the KMP does not depend on time.

\subsection{Improvements and Extensions}
Although KMP-based approaches are effective for robot trajectory generation at
the level of Cartesian coordinates and joint angles, the learning of orientation
in the task space is still challenging. In contrast to position operations in
Euclidean space, orientation is accompanied by extra constraints such as the
unit norm of the quaternion representation and the orthogonal constraint of
rotation matrices. To handle these orientation-induced constraints,
\cite{huang2020toward} employ a kernelized approach via KMPs to learn
quaternions. This allows for orientation adaptation towards arbitrary desired
points that consist of both unit quaternions and angular velocities, while
taking into account high-dimensional inputs and smoothness constraints.

KMPs focus on IL while ignoring internal and external constraints. To deal with
the problem of linear constraints for IL, \cite{huang2020linearly} propose
linearly constrained KMPs. Specifically, the probabilistic properties of
multiple demonstrations are exploited and incorporated into a linearly
constrained optimization problem thus leading to a nonparametric solution. Not
only do  linearly constrained KMPs inherit key features of IL (e.g., learning
from multiple demonstrations, reproduction and adaptation of via/end points in
terms of position and velocity), but they also take into account arbitrary
linear equality (e.g., a planar constraint) and inequality constraints (e.g., a
linear combination of action components should be less than or greater than a
predefined value).

\cite{huang2021ekmp} introduce extended KMPs as a more generic and constrained
IL framework. An extended KMP is capable of the following: (i) learning
probabilistic features from multiple demonstrations; (ii) adapting learned
skills towards arbitrary desired points in terms of joint position and velocity;
(iii) avoiding obstacles at the robot link level; and (iv) satisfying arbitrary
linear and nonlinear equality and inequality hard constraints. Extended KMPs
learn both joint position and velocity simultaneously, while maintaining the
corresponding derivative relationship. This permits fast convergence and smooth
trajectories even under nonlinear hard constraints and obstacle avoidance
requirements.

KMPs excel at generalizing and handling high-dimensional inputs, yet they can be
slightly inadequate in reproduction accuracy. To address this deficiency,
\cite{liu2023variable} make two improvements to the KMP algorithm. First,
multivariate Gaussian process regression (GPR) is utilized to model the
reference trajectory, which improves the reproduction accuracy of KMP. Second,
an optimization problem is formulated to learn the hyperparameters of the KMP
kernel function, which reduces the dependence on experience. In addition, a
variable impedance control approach that provides a trade-off between contact
compliance and tracking accuracy by using the probabilistic properties of the
KMP is developed. The proposed enhancements are validated through simulations
and various experiments including a Chinese character writing task.

\subsection{Limitations}
KMPs are a kernel-based nonparametric approach that allow for learning complex
and high-dimensional trajectories, yet they have a few limitations. First,
similar to most regression algorithms, the computational complexity of KMPs
increases in proportion to the size of the training data. Second, although KMPs
have the capability to perform trajectory adaptation, the choice of desired
points is empirical. Third, choosing a kernel is crucial for KMP performance. A
Gaussian kernel is frequently used for KMPs since it can be nontrivial to
determine the optimal kernel.

\section{Conditional Neural Movement Primitives}
\label{sec:conditional_neural_movement_primitives}
Similar to ProMPs, conditional neural movement primitives (CNMPs)
\cite{seker2019conditional} use multiple demonstrations to encode primitive
movements. However, instead of LfD using likelihood estimation, CNMPs utilize
conditional neural processes (CNPs) \cite{garnelo2018conditional} to achieve a
prior via neural networks. CNPs are a type of probabilistic modeling framework
that can be used to learn and predict complex, nonlinear relationships between
input and output variables. They use neural networks to model the distribution
of the output variables given the input variables. Furthermore, CNPs may be
conditioned on context or environment variables to enable adaptability and
generalization.

\subsection{Background}
A CNMP consists of the following components: (i) an encoder network $E$, which
is used to encode observations into a latent representation; (ii) latent
representations that are aggregated into prior knowledge and used for
conditioning on the observations; and (iii) a query network $Q$ that generates
distribution parameters $(\mu_q,\sigma_q)$ based on input $x_q$. Given
observation pairs $\{(x_i,y_i)\}_{i=0}^n$, formulating a CNMP architecture with
CNPs is done using
\begin{equation}
  \mu_q, \sigma_q = Q_\theta (x_q \oplus \cfrac{\Sigma_i^n E_\phi ((x_i,u_i))}{n}).
\end{equation}
The optimization of $E$ and $Q$ is performed via the following loss function
\begin{equation}
  L(\theta,\phi) = -\log P(y_q\,|\,\mu_q, \text{softmax}(\sigma_q)).
\end{equation}

In the context of CNMPs, CNPs are used to extract a prior over the training data
for the primitive and associated sensor data. Doing so enables the CNMP
framework to adapt to different contexts or environments, which can enable
highly flexible and adaptable control. They can also be used to generate
movements with a variety of characteristics, such as variable speed or
acceleration.

\cite{seker2019conditional} utilize CNMPs within an LfD framework that learns
complex temporal sensorimotor relations to external parameters and goals.
Concretely, CNMPs extract prior knowledge directly from training data by
sampling observations to predict a conditional distribution over other target
points. CNMPs are demonstrated by conducting experiments that require a robot
arm to adapt to unexpected events during execution by conditioning the system
with sensor readings. Specifically, the following tasks are evaluated: (i) 2D
obstacle avoidance where the system learns a movement trajectory to avoid
obstacles; (ii) robotic manipulation to determine the generalization capability
to environment configurations inside and outside the range of demonstrations;
(iii) pick-and-place to analyze the capability of the CNMP model to react to
external perturbations. 

\subsection{Improvements and Extensions}
\cite{akbulut2021acnmp} extend CNMPs with an RL component. The LfD and RL
framework, named adaptive CNMPs, learns the distribution of the input movement
trajectories conditioned on the task parameters. Adaptive CNMPs employ an
encoder-decoder network to represent the relations between the task parameters
and movement trajectories from a few demonstrations. If the generated
trajectories fail to achieve the desired goals (e.g., change in environment,
out-of-range queries, etc.), then the framework begins to adapt its skill
representation by simultaneously performing RL and LfD. To do this, internal
system parameters are updated via RL-guided actions and then interspersed with
error-based supervised learning based on the demonstration set. 

Adaptive CNMPs can also facilitate skill transfer between two morphologically
different robots. For example, a robot can automatically learn a new task by
observing the execution of the task by another robot. This is done by training
two adaptive CNMP models to create a common representation in their latent
layers by a proxy skill demonstration that is available to both models. A
refined version of the adaptive CNMP framework uses a novel reward CNMP to
generate trajectories by sampling from the latent space conditioned on rewards
\cite{akbulut2021reward}. This is accomplished with an encoder-decoder
architecture that learns the distribution of trajectory points and corresponding
rewards to output a trajectory as a function of time.

CNMPs utilize a deterministic latent variable whose stochasticity is only
provided by a Gaussian output. Although this is well suited to capture unimodal
aleatoric uncertainty, it cannot represent multimodal distributions. In
addition, how to perform motion blending, temporal modulation, and rhythmic
movements with CNMPs is not clear. CNMPs may also produce trajectories that are
far from the desired via points. To mitigate these weaknesses,
\cite{przystupa2023deep} propose a framework that provides all of the
functionality of ProMPs while enabling the ability to work with high-dimensional
variables provided by CNMPs. Specifically, the CNMP deterministic mean
aggregator is replaced with a Bayesian aggregator for improved predictive
performance. The formulation provides a natural way to perform motion blending
directly in the latent space. 

\subsection{Limitations}
A major limitation of CNMPs is that they require large amounts of data and
computational resources for neural network training. In addition, they can be
sensitive to initialization along with the choice of training algorithm and
hyperparameters. As a generative model, CNMPs can only predict isotropic
trajectory variance at each time step. Consequently, they do not model
correlations across time steps or dimensions. Finally, different skills require
the training of different models, which decreases both the teaching efficiency
and reusability of each skill.

\section{Fourier Movement Primitives}
\label{sec:fourier_movement_primitives}
Most MP frameworks primarily focus on discrete motions for point-to-point tasks.
In contrast, Fourier movement primitives (FMPs) \cite{kulak2020fourier} are a
type of MP that represents movements as a combination of sinusoidal basis
functions through a Fourier transform. Similar to ProMPs, FMPs can be used to
generate smooth, continuous trajectories for a variety of different types of
movements, including oscillatory movements such as swinging or walking. 

\subsection{Background}
In an FMP, the MP is represented as a sum of sinusoidal basis functions, each
with a specific frequency and phase. The FMP algorithm is based on the notion
that many movements can be represented as a combination of simple sinusoidal
oscillations, and that these oscillations can be captured using a Fourier
transform. Analogous to ProMPs, FMPs use multiple demonstrations to learn a
primitive. However, since motion is represented in the frequency domain, FMPs
can automatically extract the demonstrations and no manual alignment is
necessary.

FMPs follow the ProMP learning process. The difference is the use of complex
weights,
\begin{equation}
  \forall i \in \llbracket 1; N \rrbracket : \bm{y}_i = \tilde{\bm{\Phi}}\tilde{\bm{w}}_i,
  \label{eq:fmp_1}
\end{equation}
where $\tilde{\bm{w}}_i$ consists of the complex weights, $(\tilde{\bm{w}}_i)_{i
= 1,\ldots,N}$, that a distribution is learned over. To maintain unified
magnitude and phase statistics, the real and imaginary weights are concatenated,
\begin{equation}
 \tilde{\bm{w}}_i = [\Re(\tilde{\bm{w}}_i)^\top, \Im(\tilde{\bm{w}}_i)^\top].
  \label{eq:fmp_2}
\end{equation}

The weights are learned by fitting a Gaussian mixture with an EM algorithm as
follows. First, the associated weights, means, and covariances (i.e.,
$\bm{\theta} = (\pi_j,\bm{\mu}_j, \bm{\Sigma}_j)_{j=1,...,M}$) are calculated, 
\begin{equation}
  p(\bm{w}\,|\,\theta) = \sum_{j=1}^{M} \pi_j \mathcal{N}(\bm{w}\,|\,\bm{\mu}_j,\bm{\Sigma}).
  \label{fmp_eq_3}
\end{equation}
Then, to determine a starting location, a partial trajectory must be mapped into
the Fourier domain. Based on \cite{kulak2020fourier}, $K$ time steps of the
demonstration data $\bm{y}_{1:K}$ are provided. Next, the weights are determined
by $M$ least squares problems,
\begin{equation}
  \bm{w}^j = (\bm{\Phi}_{1:K}^H\bm{\Phi}_{1:K} + \lambda\bm{\Sigma}_j^{-1})^{-1}(\bm{\Phi}_{1:K}^H\bm{y}_{1:K} + \lambda\bm{\Sigma}_j^{-1}\mu_j). 
  \label{eq:fmp_4}
\end{equation}
\eqref{eq:fmp_4} is solved by minimizing
\begin{equation}
\begin{split}
  j^* = \argmin{j \in \llbracket 1; M \rrbracket}(\parallel \bm{y}_{1:K} - \bm{\Phi}_{1:K}\bm{w}^j \parallel^2 - \\ 
        \lambda \log(\pi_j) + \lambda \parallel \bm{w} - \bm{\mu}_j \parallel^2_{\Sigma_j^-1})
\end{split}
\label{eq:fmp_5}
\end{equation}
such that the trajectory can be mapped into the Fourier domain with $\bm{w}_K =
\bm{w}^{j^*}$. To achieve tracking in the Fourier domain, FMPs use a
proportional controller for a trajectory of length $T$ with time step $t$, i.e.,
\begin{equation}
  \bm{w}_{t+1} = \bm{w}_t + \text{dt} \beta \text{diag}(\Sigma_{j^*}^{-1})(\mu_{j^*}-\bm{w}_t).
  \label{eq:fmp_6}
\end{equation}
Finally, the tracked trajectory is generated by 
\begin{align}
  \bm{y}_{t+1}^{\text{des}} &= \bm{\Phi}_{t+1}\bm{w}_{t+1}, \\
  \Sigma_{\bm{y}_{t+1}^{\text{des}}} &= \bm{\Phi}_{t+1}\Sigma_{j^*}\bm{\Phi}_{t+1}^H.
\end{align}

Compared to other periodic MP methods, FMPs leverage Fourier series to extract
multiple frequencies underlying the demonstrations without requiring a manual
choice or tuning of the basis functions. Doing so eliminates a major
preprocessing step required for other methods. In FMPs, the basis functions are
derived from a Fourier series and are automatically determined based on the
frequency content of the demonstrations. This makes FMPs more flexible and
adaptive. Another advantage of FMPs is that they require minimal preprocessing
of the demonstration data. The input to the FMP algorithm is a set of motion
trajectories. The algorithm automatically extracts the underlying frequencies
and generates a Fourier series representation of the motion. This makes FMPs
more robust to noise and other imperfections in the demonstration data. 

FMPs capture both the magnitude and phase statistics of the demonstration data.
This enables them to model not just the shape of the motion trajectory, but also
its dynamics and temporal properties. This is important for jobs that require
precise timing and coordination, such as robot manipulation tasks. Lastly, FMPs
do not require manual alignment of the demonstration data. The algorithm can
learn the underlying frequency content of the data and generate motion
trajectories that match the frequency content of the input data. This makes FMPs
useful for learning and generating motion trajectories from a variety of
sources, including demonstrations from human experts or motion capture data.

FMPs were designed to tackle tasks that involve rhythmic movements (e.g.,
wiping, polishing, etc.). The advantages of FMPs for contact manipulation
applications are the following: (i) extraction of multiple frequencies hidden in
the demonstrations; (ii) no need to choose or tune basis functions; (iii) very
little preprocessing is required; (iv) unified magnitude and phase statistics
are provided. In \cite{kulak2020fourier}, FMPs are demonstrated on polishing and
other wiping tasks to illustrate real-world use cases of a periodic MP method. 

\subsection{Improvements and Extensions}
Research focused on learning periodic motion to complete tasks such as wiping,
stirring, winding, etc., has evolved from extensions of oscillatory MP
frameworks. The majority of this work involves the encoding of periodic motions
in DMPs. To date, there have been no direct improvements/extensions to the
original FMP framework.

\subsection{Limitations}
FMPs are not suitable for discrete motions. This limits the use of FMPs to
robotic tasks comprised of rhythmic movements. In addition, although FMPs do not
require choosing hyperparameters, the length of the signal $T$ needs to be
manually selected and must also be sufficiently long in duration to include at
least one period of the demonstration. FMPs also assume that multiple
demonstrations are provided. Similar to DMPs and ProMPs, FMPs assume expert
demonstrations with low noise \cite{kalinowska2021ergodic}.

\section{Comparison of Movement Primitives}
\label{sec:comparison_of_movement_primitives}
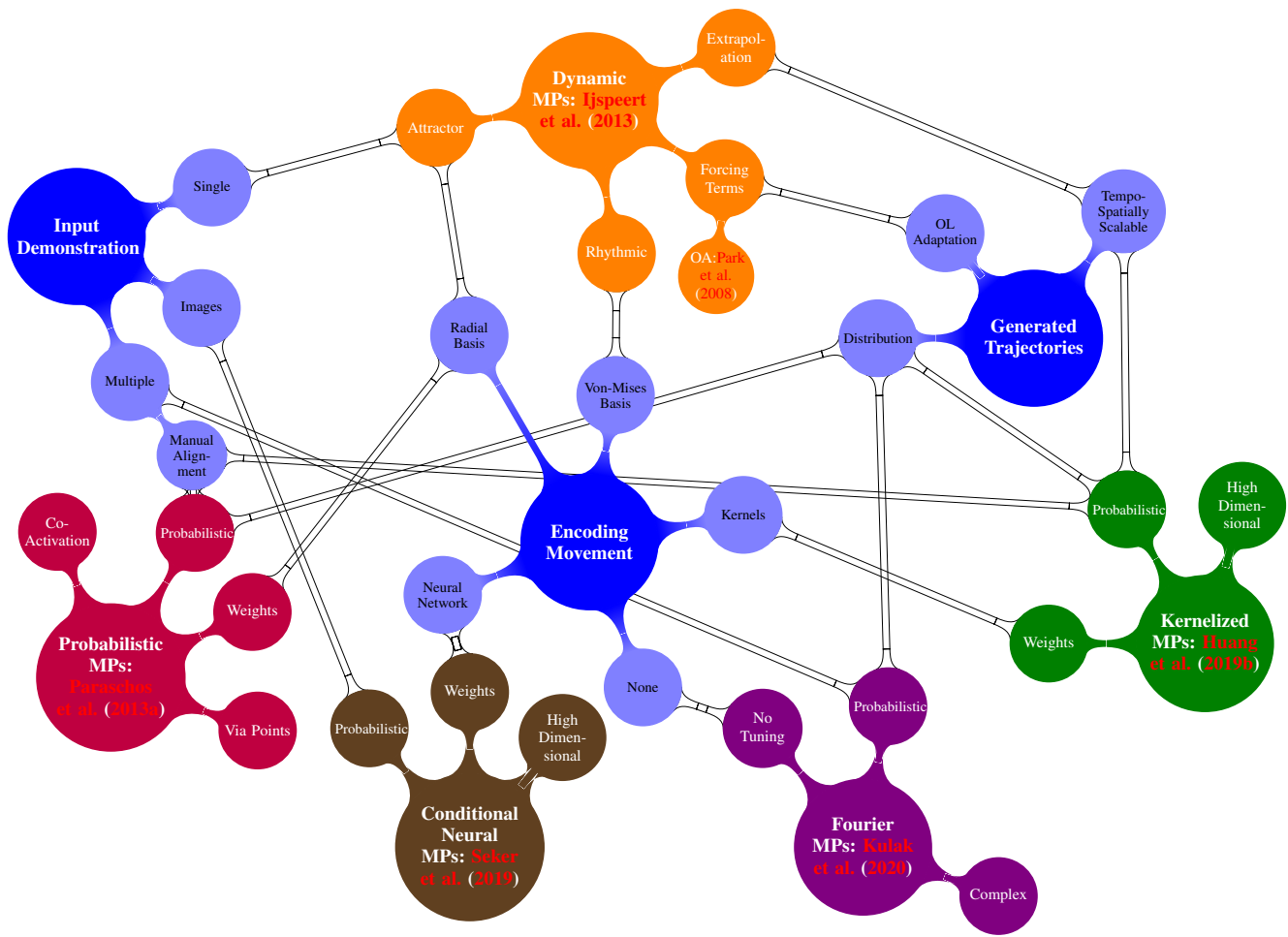
\begin{figure*}
\centering
\resizebox{\linewidth}{!}{%
\begin{tikzpicture}[mindmap,
 level 1 concept/.append style={level distance=130,sibling angle=30},
 extra concept/.append style={color=blue!50,text=black}]
 
\begin{scope}[mindmap, concept color=orange, text=white]
  \node [concept, font=\fontsize{16pt}{18pt}\selectfont\bfseries] {Dynamic MPs: \cite{ijspeert2013dynamical}}[clockwise from=-5]
    child [grow=190]
      {node [concept, font=\fontsize{13pt}{15pt}\selectfont] (att) {Attractor}}
    child [grow=-80]
      {node [concept, font=\fontsize{13pt}{15pt}\selectfont] (rhy) {Rhythmic}}
    child [grow=20]
      {node [concept, font=\fontsize{13pt}{15pt}\selectfont] (ext) {Extrapol- \\ ation}}
    child [grow=-30]
      {node [concept, font=\fontsize{13pt}{15pt}\selectfont] (for) {Forcing Terms}
    child [grow=-95] {node [concept, font=\fontsize{13pt}{15pt}\selectfont] (obs) {OA:\cite{park2008movement}}}};
   \end{scope}
 
\begin{scope}[mindmap, concept color=purple,text=white]
  \node [concept, font=\fontsize{16pt}{18pt}\selectfont\bfseries] at (-14,-17) {Probabilistic MPs: \cite{paraschos2013probabilistica}}
    child [grow=60, level distance=140]
      {node [concept, font=\fontsize{12pt}{13pt}\selectfont] (pro) {Probabilistic}}
    child [grow=-20]
      {node [concept, font=\fontsize{13pt}{15pt}\selectfont] (via) {Via Points}}
    child [grow=25]
      {node [concept, font=\fontsize{13pt}{15pt}\selectfont] (weip) {Weights}}
    child [grow=110]
      {node [concept, font=\fontsize{13pt}{15pt}\selectfont] (coa) {Co-Activation}};
   \end{scope}
 
\begin{scope}[mindmap, concept color=green!50!black,text=white]
  \node [concept, font=\fontsize{16pt}{18pt}\selectfont\bfseries] at (18,-16) {Kernelized MPs: \cite{huang2019kernelized}}
    child [grow=75, level distance=120]
      {node [concept, font=\fontsize{13pt}{15pt}\selectfont] (hid) {High Dimensional}}
    child [grow=120, level distance=130]
      {node [concept, font=\fontsize{12pt}{13pt}\selectfont] (probk) {Probabilistic}}
    child [grow=180]
      {node [concept, font=\fontsize{13pt}{15pt}\selectfont] (weik) {Weights}};
   \end{scope}
 
\begin{scope}[mindmap, concept color=brown!50!black,text=white]
  \node [concept, font=\fontsize{16pt}{18pt}\selectfont\bfseries] at (-3.5,-22) {Conditional Neural MPs: \cite{seker2019conditional}}
    child [grow=50, level distance=120]
      {node [concept, font=\fontsize{13pt}{15pt}\selectfont] (hidcn) {High Dimensional}}
    child [grow=130, level distance=130]
      {node [concept, font=\fontsize{12pt}{13pt}\selectfont] (probcn) {Probabilistic}}
    child [grow=90]
      {node [concept, font=\fontsize{13pt}{15pt}\selectfont] (weicn) {Weights}};
\end{scope}
   
\begin{scope}[mindmap, concept color=violet, text=white]
   \node [concept, font=\fontsize{16pt}{18pt}\selectfont\bfseries] at (8,-22) {{Fourier MPs: \cite{kulak2020fourier}}}
       child [grow=-20, level distance=120]
         {node [concept, font=\fontsize{13pt}{15pt}\selectfont] (com) {Complex}}
       child [grow=130, level distance=130]
         {node [concept, font=\fontsize{13pt}{15pt}\selectfont] (not) {No Tuning}}
       child [grow=80, level distance=120]
         {node [concept, font=\fontsize{12pt}{13pt}\selectfont] (probf) {Probabilistic}};
   \end{scope}
 
\begin{scope}[mindmap, concept color=blue]
 
  \node [concept, font=\fontsize{16pt}{18pt}\selectfont\bfseries, text=white] at (-15,-4) {Input \\ Demonstration}
    child [concept color=blue!50, grow=20, level distance=120]
      {node [concept, font=\fontsize{12pt}{13pt}\selectfont] (sin) {Single}}
    child [concept color=blue!50, grow=330, level distance=120]
      {node [concept, font=\fontsize{12pt}{13pt}\selectfont] (img) {Images}}
    child [concept color=blue!50, grow=-70] {node [concept, font=\fontsize{12pt}{13pt}\selectfont] (mul){Multiple}
    child [concept color=blue!50, grow=-50, level distance=80]
      {node [concept, font=\fontsize{12pt}{13pt}\selectfont] (man) {Manual Alignment}}};
 
  \node [concept, font=\fontsize{16pt}{18pt}\selectfont\bfseries, text=white] at (0,-13) {Encoding Movement} [clockwise from=150]
    child [concept color=blue!50, grow=120, level distance=200] {node [concept, font=\fontsize{12pt}{13pt}\selectfont] (rad) {Radial Basis}}
    child [concept color=blue!50, grow=80, level distance=125]
      {node [concept, font=\fontsize{12pt}{13pt}\selectfont] (von) {Von-Mises Basis}}
    child [concept color=blue!50, grow=-70]
      {node [concept, font=\fontsize{12pt}{13pt}\selectfont] (non) {None}}
    child [concept color=blue!50, grow=-160]
      {node [concept, font=\fontsize{12pt}{13pt}\selectfont] (nn) {Neural Network}}
    child [concept color=blue!50, grow=10]
      {node [concept, font=\fontsize{12pt}{13pt}\selectfont] (ker) {Kernels}};
 
   \node [concept, font=\fontsize{16pt}{18pt}\selectfont\bfseries, text=white] at (13, -7) {Generated Trajectories}
     child [concept color=blue!50, grow=130, level distance=115]
       {node [concept, font=\fontsize{12pt}{13pt}\selectfont] (ola) {OL Adaptation}}
     child [concept color=blue!50, grow=180] {node [concept, font=\fontsize{12pt}{13pt}\selectfont] (dis) {Distribution}}
     child [concept color=blue!50, grow=55] {node [concept, font=\fontsize{12pt}{13pt}\selectfont] (tem) {Tempo-Spatially Scalable}};
\end{scope}
  
\begin{pgfonlayer}{background}
  \draw [circle connection bar]
    (weik) edge (ker)
    (att) edge (rad)
    (rhy) edge (von)
    (weip) edge (rad)
    (not) edge (non)
    (probf) edge (mul)
    (ext) edge (tem)
    (for) edge (ola)
    (probf) edge (dis)
    (pro) edge (dis)
    (att) edge (sin)
    (pro) edge (man)
    (probk) edge (tem)
    (probk) edge (man)
    (probk) edge (dis)
    (probcn) edge (img)
    (weicn) edge (nn);
  \end{pgfonlayer}
\end{tikzpicture}
}
\caption{An overview of MP frameworks in terms of their similarities and
differences.}
\label{fig:mp_mindmap}
\end{figure*}

This section compares and contrasts MP methods in chronological order. First, we
provide a high-level overview of the pros and cons of each individual MP
framework. Then, a detailed comparison of each MP method against the other MP
frameworks is given. We also lay out a visual (Figure~\ref{fig:mp_mindmap}) and
tabular (Table~\ref{tab:mp_framework_overview}) comparison of the approaches
described.

DMPs are capable of learning from human demonstrations, adapting to new tasks or
environments, and handling non-stationary movements. They may be used for both
open-loop and closed-loop control. However, DMPs can be sensitive to the choice
of parameters and they require a good initialization. In addition, DMPs may not
handle highly-stochastic environments well since they only use a single
demonstration.

ProMPs were designed to deal with uncertainty and stochasticity in a robot's
environment. This makes it possible to generate movements that are robust to
disturbances and unexpected events. Nonetheless, ProMPs can be computationally
expensive and may not handle nonlinear movements as well as other frameworks.
Preprocessing demonstrations to ensure manual alignment is also required.

KMPs can adapt to high-dimensional data and they may be used to generate
periodic patterns. A kernel function can also be selected to create rhythmic
movement. Nevertheless, KMPs may be computationally expensive and sensitive to
the choice of kernel functions. 

CNMPs can learn from human demonstrations, adapt to new tasks or environments,
and used to generate smooth primitives. Yet, they can be computationally
expensive and subject to initialization sensitivity. CNMPs also require a large
amount of training data to learn effectively.

FMPs were created for periodic movements and can be used to generate smooth,
natural-looking movements for robots. They do not require manual alignment of
trajectories. However, FMPs may not work well with highly nonlinear movements or
discrete movements. They require a window parameter, which may indirectly define
the trajectory length despite limiting the amount of preprocessing required.

\subsection{Dynamic Movement Primitives}
DMPs are formed from an analytically well-known spring-damper system where an
appended forcing term is learned to generate behavior based on a single
demonstration. Compared to other MPs, the advantage of DMPs is their
flexibility: adding additional forcing terms is straightforward and the system
has been demonstrated in a wide array of robotic applications. A potential
limitation of a DMP is that it only uses a single demonstration and it assumes
that the demonstration is perfect. In many real-world tasks, human experts may
need to provide several demonstrations. Hence, DMPs may not be able to properly
represent the movement. Another restriction of the spring-damper system is that
it assumes convergence in point-to-point tasks with a terminal velocity of zero.
Yet, in some tasks this may not be the case (e.g., swinging at a ball).
Finally, in practice DMPs are sensitive to the basis parameters. Therefore, not
only is manual tuning required, but it will also differ between tasks.

\subsection{Probabilistic Movement Primitives}
ProMPs are a probabilistic formulation of MPs that are motivated by the need to
blend MPs simultaneously instead of just successive executions. ProMPs learn
from several demonstrations instead of just one as seen with DMPs. From this,
they learn the mean and standard deviations given the basis weight parameters.
Consequently, they tend to be sensitive to the selection of these basis
parameters. Using multiple demonstrations has the advantage of widening the
space of a task. However, this requires additional preprocessing as ProMPs
assume that the trajectories are properly aligned and that they are similar in
length. Varying either one of these properties will skew the distribution of the
trajectories. Another limitation is that the distribution of trajectories is
global whereas DMPs can vary the start and goal locations arbitrarily in space.
Similar to DMPs, ProMPs also assume a final velocity of zero.

\subsection{Kernelized Movement Primitives}
KMPs seek to address some of the restrictions found in both DMPs and ProMPs
through kernel learning. Instead of explicitly representing MPs with basis
function parameters, KMPs use a kernel treatment making it suitable for
high-dimensional input spaces. One of the advantages of KMPs is that they can
extrapolate trajectories similar to DMPs and unlike ProMPs, despite also using a
probabilistic representation. Additionally, in contrast to ProMPs, KMPs treat
the IL process as an optimization problem with the goal of minimizing the
information loss between the KMP trajectory and the demonstration set. On the
other hand, since a KMP requires multiple demonstrations, it cannot be used in a
``one-shot'' learning approach like a DMP. Another downside is that in longer
trajectories computing the inverse of a kernel matrix is more computationally
demanding when compared to other MP frameworks.

\begin{table*}
\caption {A comparison of MP methods.}
\label{tab:mp_framework_overview}
\centering
\scalebox{0.85}{	 
\begin{tabular}{lcccccc}
  \toprule %
  & DMPs & ProMPs  & KMPs & CNMPs & FMPs \\ \toprule %
  Discrete 
    & \cite{schaal2006dynamica} 
    & \cite{paraschos2013probabilistica} 
    & \cite{huang2019kernelized} 
    & \cite{seker2019conditional} 
    & - \\
    \midrule
   Rhythmic 
     & \cite{schaal2006dynamica} 
     & - 
     & \cite{huang2019kernelized} 
     & - 
     & \cite{kulak2020fourier} \\
     \midrule
   Multiple Demonstrations 
     & \cite{li2023prodmp} 
     & \cite{paraschos2013probabilistica} 
     & \cite{huang2019kernelized} 
     & \cite{seker2019conditional} 
     & \cite{kulak2020fourier} \\
     \midrule
   Obstacle Avoidance 
     & \cite{park2008movement} 
     & \cite{frank2021constrained}
     & \cite{huang2021ekmp} 
     & \cite{seker2019conditional} 
     & - \\
     \midrule
   No Tuning of Parameters 
     & - 
     & - 
     & - 
     & - 
     & \cite{kulak2020fourier} \\
     \midrule
   Time Complexity 
     & $\mathcal{O}(n)$ 
     & $\mathcal{O}(\log(n))$ 
     & $\mathcal{O}(n^3)$ 
     & $\mathcal{O}(n^3)$ 
     & $\mathcal{O}(n^3)$ \\
     \bottomrule
\end{tabular}  
}
\end{table*}

\subsection{Conditional Neural Movement Primitive}
CNMPs extend MPs to adapt to changes from external stimuli. For example, a CNMP
concerned with the concept of proprioception could use the force sensing of a
robotic gripper to adapt the behavior of the MP in the joints of a robot arm.
The drawback of this approach is that it requires many more learning parameters
than other MP formulations such as DMPs. Additionally, the primitives learned do
not generalize in the space of the sensor (e.g., magnitudes of force), which
does not align with the typical feature set of MP frameworks.

\subsection{Fourier Movement Primitives}
Although FMPs are similar to rhythmic ProMPs in formulation, they also possess a
unique method of encoding MPs. Specifically, FMPs are probabilistic and thus
expect multiple demonstrations. They use a Fourier series to extract the
frequencies underlying multiple demonstrations, which makes them suited for
rhythmic tasks. Due to this, FMPs do not require hyperparameters unlike other MP
methods. In addition, since they are able to resolve rhythmic trajectories
within the frequency domain, no manual alignment of trajectories is necessary
(as opposed to ProMPs and KMPs). Nonetheless, a window must be specified.
Therefore, at minimum, the period of the trajectory must be known.

\section{Other Movement Primitive Approaches}
\label{sec:other_movement_primitive_approaches}
In this survey, the term ``movement primitives'' refers to frameworks that
simplify the learning and execution of complex motor tasks to aid robotic
movement adaptation. Nevertheless, not only has there been many modifications
and extensions to DMPs over the years \cite{saveriano2023dynamic}, but the same
has also been done for the other MP frameworks. In some of these works, the
authors give their method a new name (e.g., ``contextual movement primitives''
\cite{jing2018learning}, etc.). Yet, these techniques largely retain the form
and function of their base MPs. 

We avoid enumerating these other frameworks in this survey unless they are
appreciably distinct in the following ways: (i) input demonstrations, (ii)
encoded movement, or (iii) output trajectories. This criteria allows us to
systemically categorize the work contributed to MPs and better highlight the
overall impact. However, we believe that it is important to highlight the
influence of these other methods on MP frameworks, and vice versa. These works
are generally based on dynamical systems, probabilistic, or RL techniques. In
this section, we briefly describe some of these approaches.

\subsection{Dynamical Systems Approaches}
With the goal of efficiently generating natural human-like movements,
\cite{lim2005movement} adopt a dynamical systems approach to MPs. Different from
DMPs, a framework for merging movement storage, dynamic models, and optimization
is proposed. Each MP is represented as a set of joint trajectory basis
functions, extracted via PCA of human motion capture data.  Utilizing basis
functions constructed from a set of reference motions, suboptimal motions
closely resembling the reference motions are rapidly generated. The technique is
demonstrated on case studies using a set of human arm movements. 

\cite{degallier2006movement}, \cite{degallier2007hand},
\cite{degallier2008modular}, \cite{gay2010integration}, and
\cite{degallier2011toward} explore systems of differential equations to address
the problem of real-time online trajectory generation. The use of nonlinear
dynamical systems, especially their attractor properties, makes them ideal for
creating simple motor primitives. These attractor properties include the
following: (i) robustness against small perturbations, (ii) low computational
cost, (iii) on-the-fly change of parameters, (iv) synchronization with external
signals, and (v) integration of sensory feedback terms. To show seamless
switching between discrete and rhythmic movements, these frameworks are
evaluated on a humanoid robot performing crawling, reaching, and a drumming
task.

Dynamical systems can encode trajectories through time-independent functions
that define the temporal evolution of the motions. This permits the
generalization of motions to unobserved parts of the space directly from the
application of the functions to a new set of input variables. Based on this
observation, \cite{gribovskaya2009learning} present a framework that allows
learning nonlinear dynamics of motion in manipulation tasks and the generation
of dynamical laws for the control of position and orientation.
\cite{gribovskaya2011learning} extend the theoretical aspects of the approach
and measure its robustness against perturbations and noise. Furthermore, an
in-depth analysis is made against DMPs with experimental results conducted using
a humanoid platform. 

\cite{nakanishi2011stiffness} develop a dynamical systems representation tuned
to the requirements of rhythmic movements via a methodology that optimizes for
control commands, the temporal aspect of movements, and time-varying stiffness
profiles. In particular, a phase oscillator is formulated to represent periodic
movements. Instead of Gaussian basis functions, the approach maintains all the
key benefits (e.g., ease of frequency, amplitude, and offset modulation) while
providing robustness to perturbations during control using Fourier basis
functions. Similar to DMPs, the formulation permits the ability to easily scale
the frequency, amplitude, and offset of the reference trajectory and potentially
incorporate additional coupling with external signals.

\cite{khansari2012dynamical} present an obstacle avoidance framework based on
dynamical systems that ensures impenetrability of multiple convex shaped
objects. In particular, the method can be used to perform obstacle avoidance in
Cartesian and joint spaces using both autonomous and non-autonomous controllers.
The approach requires a global model of the environment and analytical modeling
of the obstacle boundaries. Obstacle avoidance is done by modulating the
original dynamics of the controller. Furthermore, the modulation is
parameterizable and allows for determining a safety margin in order to increase
a robot's reactiveness under uncertainty. The system is evaluated on a set of
2D motions described by dynamical systems that were inferred from human
demonstrations using DMPs.

How to create adaptive smooth sequences of actions and decide among skill
options in a continuous manner, without the need for recurrent planning, is
addressed by \cite{luksch2012adaptive}. Concretely, a hierarchical dynamical
systems architecture for movement generation is proposed. The approach views MPs
as dynamical systems, blended either serially or in parallel, that facilitate
continuous skill transitions by employing continuous-time recurrent neural
networks for sequencing. Unlike conventional DMPs, this method employs a neural
dynamics-based coordination of MPs, continuous transitioning among skills, and
predictive decision-making to minimize anticipated action failures. The
framework is demonstrated on simulated reach-and-grasp experiments. Building
upon this work, \cite{manschitz2014learning} represent demonstrations as a
sequence graph where finding the transitions between consecutive movements is
treated as a classification problem. The approach is utilized by
\cite{manschitz2015learning} to learn sequential robot skills via kinesthetic
teaching. 

\cite{kober2015learning} propose to learn skills that do not rely only on
kinematics, but also take into account interaction forces represented by a
sequence of MPs. These primitives are obtained from a small number of continuous
kinesthetic demonstrations. Specifically, they consist of a set of control
variables and contain kinematic and contact force modalities. The control
variables and attractor goals are simultaneously determined in order to
reconstruct and explain the observation data. The methodology is evaluated on a
box pulling and flipping task using objects with different object geometries and
arrangements.

DMP stability is enforced by suppressing the nonlinear perturbation at the end
of the motion where the smooth switch from nonlinear to stable linear dynamics
is controlled by a phase variable. The phase variable acts as external
stabilizer, which distorts the temporal pattern of the dynamics. As a result,
DMPs lack the ability to generalize outside of the demonstrated trajectory.
\cite{neumann2015learning} address this problem through an approach that serves
as a framework generalizing the class of demonstrations that are learnable by
means of provably stable dynamical systems. To do this, the framework extends
the application domain of the stable estimator of dynamical systems by
generalizing the class of learnable demonstrations via diffeomorphic
transformations.

Modeling MPs via dynamical systems can be convenient for closed-loop
implementations. For instance, \cite{perrin2016fast} consider the problem of
learning dynamical systems from demonstrations. Specifically, a diffeomorphic
matching algorithm is developed and used to learn nonlinear dynamical systems
with the guarantee that the learned systems have global asymptotic stability.
For a given set of demonstration trajectories, a diffeomorphism is computed that
maps the forward orbits of a reference stable time-invariant system onto the
demonstrations. This deforms the whole reference system into one that reproduces
the demonstrations while still being stable. The method can be applied to
efficiently learn a variety of stable autonomous systems with applications to
DMPs.

MPs provide a compact and modular representation of continuous trajectories,
while autonomous systems yield control policies that are time independent. In
light of this observation, \cite{li2025movement} design an approach that
combines the advantages of both representations by transforming MPs into
autonomous systems. The main idea is to convert the explicit representation of a
trajectory as an implicit shape encoded as a distance field. This conversion
enables the definition of an autonomous dynamical system with modular reactions
to perturbation. Moreover, asymptotic stability guarantees are provided by using
Bernstein basis functions in the MPs, which represent trajectories as
concatenated quadratic B\'ezier curves and allow for an analytical solution to
computing distance fields. Experimental results on a set of tasks using a 7-DoF
arm show that the method offers benefits in terms of model capability and
computational efficiency when compared against a DMP baseline. 

\subsection{Probabilistic Approaches}
Instead of encoding trajectories directly, \cite{ruckert2013graphical} introduce
a framework called planning MPs that endows MPs with an intrinsic probabilistic
planning system. The approach learns the parameters of a cost function through
RL, which is then used by a stochastic optimal control planner to generate
movements. Concretely, the system combines model-based and model-free RL by
learning both the system dynamics model for planning, followed by policy search
to learn the cost function parameters. The methodology is demonstrated on both a
simple via-point task and a humanoid balancing task, showing better
generalization to new situations by maintaining learned task constraints rather
than using trajectory scaling heuristics. Although planning MPs share the
``probabilistic'' keyword with ProMPs, they represent a fundamentally different
approach. Instead, the focus is on integrating planning and optimal control
rather than learning distributions over trajectories directly from
demonstrations as in ProMPs. 

Robots must have basic interactive capabilities to engage with human partners.
Yet, programming these skills is challenging since each partner can have
distinct ways of executing movements. To solve this problem,
\cite{amor2014interaction} develop an interaction primitive (IP) representation.
The primary objective of the representation is to estimate a predictive DMP
distribution that relates MP weights to a partial trajectory. This allows for
the estimation of full weights based on a partial movement, subsequently
generating the remainder of the movement. Demonstrations recorded from two
interacting humans show how to learn IPs. \cite{ewerton2015learning} extend IPs
to learn multiple interaction patterns via a GMM by creating a mixture of
interaction ProMPs. A fully Bayesian reformulation of IPs is proposed by
\cite{campbell2017bayesian} for HRI tasks.

\cite{koch2015learning} explore humanoid gait generation based on MPs learned
from optimal and dynamically feasible motion trajectories. The key idea is to
first learn MPs in a GP framework from a precomputed set of trajectories as
training data. Then, instead of finding a new optimal solution from scratch for
every motion task, the goal is to use these MPs to construct a motion that is as
close as possible to the criterion used to compute the training data. During the
learning process, the joint angle trajectories of all the actuated joints are
utilized to learn morphable MPs based on GPs and PCA. Simulations show that five
MPs generated via the proposed approach are sufficient to generate steps with 24
different lengths for a humanoid robot.

Task-parameterized models can be used to automatically adapt robot movements to
novel situations. Based on this insight, \cite{calinon2016tutorial} introduce
the idea of a task-parameterized GMM for movement adaptation, focusing on
integrating task parameters such as frames of reference and coordinate systems
to dynamically adjust robot actions. This covers regularization to prevent
overfitting, subspace clustering for high-dimensional data management, and
continuous movement generation through GMR and dynamic trajectory models.
Task-parameterized GMMs are evaluated via the incorporation of a control
strategy based on linear quadratic tracking to optimize robot efficiency.  

In LfD, learning a nontrivial skill requires finding a mapping between a
potentially large input to output space. This can demand more demonstrations
than a user would be willing to provide. To address this problem,
\cite{manschitz2018mixture} introduce a mixture of attractors approach for
learning complex object-relative movements from demonstrations. This MP
representation inherently supports multiple coordinate frames, enabling skill
generalization to unseen object positions and orientations without relying on
initial parameter estimates. Skills are learned through convex optimization,
ensuring smooth, arbitrary-shaped movements.

Besides temporal adaptation, MPs also have the capability to be adapted through
via-point modulation. For example, via-point MPs devised by
\cite{zhou2019learning} are designed to enhance the adaptability of MPs.
Concretely, via-point MPs achieve adjustability to arbitrary via points through
a simple structured model that can handle both interpolation and extrapolation
of motion targets. By incorporating via points directly into the motion
generation process, the methodology can efficiently adapt to new task
constraints without sacrificing the simplicity and compactness expected of other
MP formulations. In proceeding work, \cite{daab2024incremental} introduce a set
of seven fundamental operations to incrementally learn libraries of via-point
MPs.

Multiple modes and models exist in IL, which can be taken into account when
learning mappings and generalized MP representations. Yet, dealing with mode or
model collapse is a challenging problem. To tackle this issue,
\cite{zhou2020movement} introduce a mixture density network that takes task
parameters as input and provides a GMM of MP parameters as output. An entropy
cost is developed to achieve a more balanced association of demonstrations to
GMM components and avoid mode and model collapse during training. Additional
performance is achieved through a failure cost function that reduces the
occurrence of the same failures for a given task parameter query.

To address complexities in learning geometric constraints for robot
manipulation, \cite{abu2021probabilistic} propose a probabilistic framework that
leverages KMPs to adapt demonstrations encapsulated in SPD matrices. This
allows for the adjustment of learned skills to novel situations with new start-,
via-, and end-SPD points, addressing a gap in LfD techniques that struggle to
update motion skills that rely on SPD matrices. The method leverages both
Cholesky decomposition and Riemannian metrics for transforming SPD data between
Euclidean and manifold spaces, facilitating the probabilistic modeling and
adaptation of SPD profiles.

The goal of IL is to not only to reproduce a robot's trajectory, but to also
allow for dynamic regulation of the trajectory in unforeseen situations. In
contrast to previous approaches (e.g., Gaussian kernels, radial basis functions,
etc.), \cite{sun2022type} integrate fuzzy theory into IL via a type-2 fuzzy
model-based MP. The type-2 fuzzy model-based MP is a regression-based IL method
that utilizes fuzzy logic. It describes the nonlinear properties of human
demonstrations using a group of type-2 fuzzy models where each model is composed
of a set of linear polynomials. Demonstrations of the method show trajectory
reproduction with modulation and superposition for real-time adaptation.

\cite{franzese2023interactive} present a safe interactive MP learning for
teaching bimanual manipulation tasks through kinesthetic demonstrations and
corrections. The approach introduces a graph-based encoding of MPs using GPR
that ensures trajectory convergence while maintaining task-relevant features.
This is achieved by combining the pose and time-belief dependent policies that
enable disambiguation of complex trajectories while remaining reactive to
perturbations. The framework allows for teaching of single-arm trajectories
independently, and then synchronizing them through interactive corrections using
variable stiffness based on the policy's epistemic uncertainty to facilitate
human feedback. The method is demonstrated using a robot manipulator on crate
picking and object handover tasks. 

Learning MPs under a probabilistic framework is essential for capturing better
demonstration information. For instance, \cite{jin2023gaussian} leverage this
understanding to develop a nonparametric GP MP. Unlike KMPs, the approach is not
only free of manual feature design, but it also allows for natural kernel
learning under a GP framework. Furthermore, instead of just a high probability
of hitting all via points as ensured by ProMPs/KMPs, if the covariance matrix of
the via points is positive definite, then the generated trajectory is guaranteed
to pass through the desired set of via points in a deterministic manner.
Analogous to ProMPs, different GP MPs can also be analytically blended to allow
for smooth switching between trajectories.

\subsection{Reinforcement Learning Approaches}
RL has a synergistic relationship with MPs \cite{kober2013reinforcement}.
However, early RL algorithms could not scale beyond systems with more than 3-
or 4-DoFs and/or deal with parameterized policies. As a result, using RL with
MPs was not applicable high-dimensional systems such as humanoid robots.
\cite{peters2006policy,peters2008reinforcement} recognized that policy gradient
methods were a notable exception to this limitation. In policy gradient
techniques, the main problem is to obtain a good estimator of the gradient,
which can be realized directly from data generated during the execution of a
task. Three major ways of estimating first order gradients (finite-difference
gradients, vanilla policy gradients, and natural policy gradients) along with
practical algorithms are detailed. This includes the use of gradient policy
methods to learn the DMP parameters for a baseball bat swinging task.

Motor skills in humanoid robotics can be learned with MPs, yet most interesting
motor learning problems are high-dimensional RL problems. To this end,
\cite{kober2008policy} further investigate policy search for high-dimensional
domains with continuous states and actions. In particular, the immediate reward
case is extended and generalized to episodic RL. Moreover, the relation to
policy gradient methods is highlighted. This results in the derivation of the
EM-inspired policy learning by weighting exploration with the returns (PoWER)
algorithm, which is especially well-suited for learning trial-based tasks in
motor control. The approach is used to encode DMP control policies via IL for
underactuated swing-up and ball-in-cup tasks using a robot arm. Additional
evaluations of the proposed methodology are presented by
\cite{kober2011policy,kober2014policy}. 

Robot skill learning must be able to scale and fulfill crucial real-time
requirements. Recognizing this dilemma, \cite{peters2013towards} divide the
skill learning problem into parts that can be well-understood from a robotics
point of view. In particular, three major questions are presented: (i) how can
we develop efficient motor learning methods? (ii) how can anthropomorphic robots
learn basic skills similar to humans? (iii) can complex skills be composed with
these elements? These questions are addressed via model-free RL methods, which
do not maintain an internal behavior simulator (i.e., a forward model), but
operate directly on the data. Policy features, based on DMPs, are chosen to
apply these RL techniques to robotic tasks. The tasks include ball-paddling,
ball-in-a-cup, dart games, and playing robot table tennis.

A popular paradigm in sensorimotor control is to train policies directly in raw
action spaces (e.g., using torque, joint angle, or end-effector position).
Nonetheless, this forces a robot to make a decision at each point in training,
thus preventing scalability to continuous, high-dimensional, and long-horizon
tasks. \cite{bahl2020neural} tackle this issue by embedding a dynamics structure
into DNN-based policies. To do this, neural dynamic policies are developed to
make predictions in trajectory distribution space as opposed to prior policy
learning methods where actions represent the raw control space. The embedded
structure allows for performing end-to-end policy learning under both RL and IL
setups. DMPs are employed as the structure for the dynamical system, yet they
may be swapped for a different differentiable dynamical structure. Simulation
experiments are conducted on picking and throwing tasks with a 6-DoF arm.

Advances in deep learning have had an impact on developing MP approaches. For
example, different learning-based methodologies have been investigated to
determine the necessary DMP parameters for a desired trajectory and task. Many
of these neural network and RL approaches are described in a survey by
\cite{rovzanec2022neural}. This includes a description of the fusion of
traditional DMPs with neural networks for learning and generating complex
motions. The survey provides an overview of the applications of GMMs, neural
networks, and autoencoders to DMPs. Methods that use RL to optimize and learn
new behaviors from a reference DMP are also reported.

Deep RL has incorporated various robot tasks through DNNs. However, this
approach is difficult to apply to robot contact tasks due to the exertion of
excessive force from the random RL search process. \cite{kim2020reinforcement}
address this problem using neural networks and an IL algorithm that optimizes
force exertion in random search processes. Unlike traditional methods that
struggle with precise contact tasks due to position errors, the proposed neural
network-based MP generates continuous trajectories for force controllers,
learned through a deep deterministic policy gradient algorithm. The IL algorithm
enables a stable generation of trajectories similar to the demonstrated
trajectories.

Episode-based RL can be used for efficiently exploring the behavior space since
exploration is implemented in the parameter space of the MP. Using this
knowledge, \cite{otto2023deep} introduce a deep RL approach for MP-based
planning policies. This is done by combining an episode-based RL algorithm with
differentiable trust region layers to perform policy updates within
higher-dimensional DMP spaces. Being episodic, the method allows for smoother
control trajectories and more effective exploration of the parameter space such
as those for complex movement tasks where the dynamics and temporal aspects of
the movement can impact the outcome of a task. \cite{otto2023mp3} extend the
approach to incorporate replanning strategies. This permits adaptation of the
MP parameters throughout motion execution, addressing the lack of online motion
modification in stochastic domains requiring feedback.

Incorporating hard constraints into RL may expedite the learning of manipulation
tasks, enhance safety, and reduce the complexity of the reward function. To
realize these benefits, \cite{padalkar2025towards} propose a representation that
can learn MPs from demonstrations by allowing modifications through null-space
actions while respecting the linear inequality constraints. Concretely, a
linearly constrained null-space kernelized MP representation is developed where
a nonparametric LfD framework generates motions that adhere to linear inequality
constraints on the robot state. This provides a null-space projector that allows
actions generated by RL policies to modify the mean behavior of the LfD policy.
The system is evaluated via an insertion task involving a torque-controlled
7-DoF manipulator.

\section{Applications of Movement Primitives}
\label{sec:applications_of_movement_primitives}
During the last few decades, many methods have demonstrated robotic systems
utilizing MPs. In this section, we provide an encyclopedic review of these
approaches and categorize them based on their application area.
Tables~\ref{tab:dmp_applications}-\ref{tab:cnmp_applications} provide a concise
taxonomy of these MP applications.

\subsection{Applications of Dynamic Movement Primitives}

\subsubsection{Contact Manipulation}
\paragraph{Assembly.}
A primary application of DMPs is robotic assembly. For example,
\cite{abu2014solving,abu2015adaptation} adapt robot trajectories in automated
assembly tasks to ensure better disturbance rejection, improve stability, and
increase overall performance. Discrete trajectories and forces, encoded as DMPs,
are obtained by demonstration and iteratively updated to a specific environment
configuration as follows. First, the system adapts Cartesian space trajectories
to match the applied forces. When a new workpiece is acquired, the demonstrated
skill is then transferred to the new configuration, and the movement is adapted
to the trained forces and torques. Lastly, a set of exception strategies are
applied when the object (e.g., peg) fails to enter a hole due to pose estimation
errors. \cite{kramberger2016learning} build on these results by developing
algorithms for learning assembly sequences and constraints using a robot arm.

Cooperative manipulation (e.g., when a group of robots jointly manipulate an
object from an initial to a final configuration while preserving the robot
formation) is a significant problem in robotics. \cite{umlauft2014dynamic}
address this challenge by designing DMPs that allow robots to move in formation and
tolerate external disturbances. Concretely, DMPs are utilized to generate
individual manipulator trajectories to a desired final configuration that is in
agreement with the initial configuration of a multi-robot team. A
formation-preserving feedback approach is used for each DMP to handle individual
DMPs that violate formation rigidity or interference that may occur on a single
manipulator. The effectiveness of the proposed controller with respect to
disturbances is demonstrated via dual-robot arm experiments.

While a single demonstration can encode basic motions, adapting to variations in
the environment often requires multiple demonstrations or additional learning.
\cite{gams2015learning} combine trajectory adaptation through DMP coupling terms
with statistical generalization to handle these novel situations. The approach
first learns force-based adaptations via ILC to modify a demonstrated trajectory
for specific environmental conditions. These coupling terms are then collected
into a database that can be used with GPR to generate appropriate adjustments
for new scenarios. The system is validated on a window frame assembly task where
a robot manipulator learns to adapt its trajectories based on physical human
guidance. A key feature is that both successful adaptations and intermediate
learning attempts can be stored in the database, which reduces the
demonstrations needed while enabling rapid adaptation to new situations.

Dual robot arms can increase the performance of a variety of assembly tasks.
For instance, they are essential for tasks involving the manipulation of heavy
and long objects, and they can enable the transfer of diverse human skills to
robots. Nevertheless, this additional flexibility requires more complex control
algorithms. \cite{likar2015adaptation} tackle this challenge via a force
adaptation scheme for bimanual systems as follows. First, a desired policy is
shown by human demonstration using kinesthetic teaching, where both trajectories
and interaction forces are recorded and represented as DMPs. Then, the captured
trajectories are divided into absolute and relative coordinates and iteratively
modified to increase overall performance. The theoretical results are evaluated
through an implementation of a peg-in-a-hole task with dual robot arms.

Industrial robots typically operate at a high impedance, which complicates
control during interactions in unstructured environments that involve
unpredictable events and external perturbations. To address this issue for the
task of part assembly, \cite{peternel2015human} propose the following LfD
approach. First, a human operator controls the robot via a haptic and handheld
impedance control interface. Then, based on a linear spring-return
potentiometer, the impedance control interface maps the button position to the
robot arm stiffness. This allows the operator to modulate the robot compliance
for a given task. Finally, the motion and stiffness trajectories are encoded via
DMPs and learned using LWR. The system is demonstrated on an assembly task that
involves putting two parts together.

Force feedback at the teleoperator's end effector can provide additional
information about a robot's state. However, it may also introduce instability
via undesired commanded end-effector motions. Building on the work of
\cite{peternel2015human}, \cite{peternel2018robotic} create a human-in-the-loop
approach that combines direct control of the end-effector stiffness with 3-DoF
end-effector force feedback. The method integrates a human sensorimotor system
into the robot control loop using a teleoperation setup. In addition, a handheld
stiffness control interface controlled by the motion of a human finger is
developed. A teaching approach is then performed to collect and encode
trajectories by DMPs. The system is used to solve the following assembly tasks:
sliding a bolt into a groove and screwing in a self-tapping screw.

DMPs have been demonstrated on a multitude of manipulation tasks within a
manufacturing setting. Nonetheless, how to best represent complex, multistep
robotic tasks is still an open problem. A solution provided by
\cite{wu2019incremental} introduces an introspective MP framework based on DMPs.
Introspective MPs grant robots the capability to assimilate information from
unstructured demonstrations. Concretely, introspective MPs are learned using a
nonparametric Bayesian model such that they can assess the quality of multimodal
sensory data. This enables incremental construction of manipulation tasks in
the presence of faults. An experimental evaluation of the robustness towards
perturbations and adaptive exploration is conducted on a human-robot
collaborative packaging task.

In practical multistep applications that require a set of distinct skills, it
can be desirable to combine sub-controllers built with different
representations. For example, \cite{angelov2020composing} introduce a
hierarchical policy that allows for the sequencing of a diverse set of
sub-controllers. Concretely, expert demonstrations are used to construct a DMP
model and convert a gradient-based learning process into a planning process
without explicitly specifying pre- and post-conditions. A trained goal score
estimator, which can be used across sub-controllers, selects a controller within
the operation domain for the current state by evaluating future states within
the plan. The approach is applied on both simulated and real-world gear assembly
tasks using a mobile manipulator.

DMPs, which require no human tuning, enable \cite{sloth2020towards} to develop a
reliable kinesthetic teaching approach for enabling robotic insertion tasks.
Specifically, a spiral search algorithm with configurable parameters ensures
robust object manipulation under pose uncertainty. To learn key points in the
desired trajectory, a sequence of four sub-demonstrations is first collected.
Subsequently, the data is encoded and executed using DMPs. During task
execution, a reactive DMP then executes the trajectory using a linear DMP until
forces are present. Lastly, the spiral search procedure is acti-

\onecolumn
\renewcommand{\arraystretch}{1.26}
\begin{longtable}{M{0.14\textwidth-2\tabcolsep - 1.25\arrayrulewidth} M{0.86\textwidth-2\tabcolsep - 1.25\arrayrulewidth}}
\caption{Applications of DMPs organized by topic area.}\\
  \toprule 
  \textbf{Area} & \textbf{References} \\\midrule
  Contact Manipulation & 
  Assembly: \cite{abu2014solving,umlauft2014dynamic,abu2015adaptation,gams2015learning,likar2015adaptation,peternel2015human,kramberger2016learning,peternel2018robotic,wu2019incremental,angelov2020composing,sloth2020towards,spector2020deep,iturrate2021quick,zeng2021generalization,dong2023novel}.
  \newline
  Brain-Machine Interfaces: \cite{hotson2016high}.
  \newline
  Deformable Objects: \cite{joshi2017robotic,nemec2018adaptive,joshi2019framework,cui2022coupled,si2022adaptive,chandra2023dual}.
  \newline
  Grasping: \cite{kroemer2010combining,kroemer2011grasping,pastor2012towards,pastor2013dynamic,rombokas2012tendon,rombokas2012reinforcement,li2015teaching,de2018reinforcement,solak2019learning,beik2020model}.
  \newline
  Mobile Manipulation: \cite{beetz2010generality,niekum2012learning,niekum2015learning,chi2017learning,li2017reinforcement,zhao2018cooperative}. 
  \newline
  Insertion: \cite{kramberger2017generalization,pervez2017novel,pervez2019motion,iturrate2019improving,ti2019human,nemec2020learning,spector2021learning,chang2022impedance,davchev2022residual,wang2022adaptive,li2023enhanced,nguyen2025quasi}. 
  \newline
  Pick-and-Place: \cite{pastor2009learning,stulp2011learninga,stulp2012reinforcement,aein2013toward,stein2014convexity,lioutikov2016learning,hung2017programming,kim2018learning,kramberger2020adapting,li2020generalization,noohian2022framework,chen2023object,ning2023novel,chen2024vision,chen2024robust,ning2024mt,zhang2024using}. 
  \newline
  Pouring/Pumping/Turning: \cite{nemec2009task,tamosiunaite2011learning,lee2013skilla,lee2013skillb,petrivc2014onlinea,worgotter2015structural,yang2018learning,yang2018robot,agostini2020manipulation,noseworthy2020task,bai2022robot,liao2022dynamic,yu2022human,amadio2022target,kong2023dynamic,hanks2024comparing,liu2024robot,lu2025dynamic}. 
  \newline
  Grinding/Polishing/Sawing/Scraping/Smoothing/Sweeping/Wiping: \cite{gams2010line,do2014learn,chebotar2014learning,gams2014learning,peternel2014teaching,montebelli2015handing,gams2016adaptation,peternel2016adaptation,zhou2016learning,peternel2017method,pervez2017learning,peternel2017robots,shahriari2017adapting,dometios2018vision,peternel2018robot,sutanto2018learning,conkey2019learning,li2020pattern,yang2022learning,zhou2022combination,liao2024simultaneously}.
  \newline
  Flipping/Hitting/Throwing: \cite{peters2006reinforcement,kober2008learning,kober2009learning,kober2010imitation,kober2010movement,kormushev2010robot,nemec2010learning,pastor2011skill,kober2012reinforcement,daniel2012learning,daniel2012hierarchical,daniel2013learning,daniel2016hierarchical,kupcsik2013data,kupcsik2017model,parisi2015reinforcement,huang2016jointly,lundell2017generalizing,pahic2018user,lonvcarevic2019learning,lonvcarevic2020reduction,lonvcarevic2020generalization,lonvcarevic2021generalization,lonvcarevic2021accelerating,hazara2017model,hazara2018speeding,hazara2019transferring,hazara2019active,prakash2019dynamic,lonvcarcvic2021accelerated,pahic2021robot,lonvcarevic2022combining,lonvcarevic2022fitting,huang2023toward,sun2023integrating}.
  \newline
  Surgery: \cite{ghalamzan2015incremental,chi2018trajectory,ginesi2019knowledge,straivzys2020surfing,su2020reinforcement,denivsa2021semi,su2021toward,schwaner2021autonomous,iturrate2023handheld,straivzys2023learning,zhang2023step,scheikl2024movement}.
  \newline
  Painting/Writing: \cite{xu2004multiple,tan2012robots,steinmetz2015simultaneous,li2018enhanced,pahic2018deep,ridge2019learning,luo2020generalized,pahic2020training,pahic2020reconstructing,song2020robot,papageorgiou2022dirichlet,dong2023dynamic,dong2023robot,xue2023robotic}.
  \\\cline{1-2}
  Field Robotics & 
  Agricultural Robotics: \cite{la2021study,lauretti2023robot,la2024combining,lauretti2024new}.
  \newline
  Space Robotics: \cite{cloud2023lunar,cloud2025vision}.
  \newline
  Underwater Robotics: \cite{carrera2015cognitive,yang2024learning}.
  \newline
  Unmanned Aerial Vehicles: \cite{perk2006motion,fang2014control,tomic2014learning,kim2018cooperation,lee2018integrated,kim2019incorporating,lee2020trajectory,zhang2021novel,rothomphiwat2024robust}.
  \\\cline{1-2}
  Humanoid Robotics &
  Bipedal Motion: \cite{nakanishi2003learning,nakanishi2004framework,nakanishi2004learning,pongas2005rapid,gopalan2013feedback,ruckert2013learned,gams2014rich,andre2015adapting,andre2016skill}.
  \newline
  Joint/Limb Coordination: \cite{ijspeert2002learninga,stulp2009compacta,kim2010learning,tan2011computational,stulp2013learning,li2014novel,mao2014learning,reinhart2014efficient,gams2015accelerating,luo2015learning,pfeiffer2015gesture,reinhart2015efficient,silverio2015learning,mandery2016using,queisser2016incremental,bockmann2017kick,mukovskiy2017adaptive,duan2018constrained,queisser2018bootstrapping,amatya2020human,lin2020arm,liu2020workspace,nah2020dynamic}.
  \\\cline{1-2}
  Human-Robot Interaction & 
  Active Compliance: \cite{basa2015learning,gao2018projected,bian2020extended,dong2021dmp,dou2022robot,escarabajal2023combined,hong2023human,xing2023dynamic,zhang2024innovative,zhou2024dynamic}.
  \newline
  Active Learning: \cite{fabisch2014active,maeda2017active}.
  \newline
  Cooperative Control: \cite{guerin2014adjutant,cui2016environment,petrivc2016cooperative,gutzeit2018besman,nemec2018human,cui2019environment,anand2021real,lu2022dmps,wang2024cooperative}.
  \newline
  Imitation Learning: \cite{kormushev2011imitation,chang2013motion,lauretti2017learning,wu2018multi,vollmer2018user,eiband2019learning,abu2020passive,wang2020framework,kramberger2022robotic,escarabajal2023imitation,li2023human,seleem2023imitation,liu2024enhancing}.
  \newline
  Interaction Learning: \cite{strachan2004dynamic,akgun2010action,parlaktuna2012closed,kulvicius2013interaction,liu2014visual,wang2015study,duminy2016strategic,schroecker2016directing,nielsen2017individualised,caccavale2019kinesthetic,baumkircher2022collaborative,ding2022intelligent,lai2022user,lu2022novel,luo2023vision,odesanmi2023skill,shi2025exploring}.
  \newline
  Object Handover: \cite{prada2012dynamic,prada2013dynamic,prada2014implementation,koene2014experimental,gams2014couplingb,nemec2014human,widmann2018human,sidiropoulos2019human,abdelrahman2020context,sidiropoulos2021human,wu2022adaptive,zeng2021learning,cai2023probabilistic,iori2023dmp,perovic2023adaptive,mavsar2024simulation,franceschi2025human}.
  \newline
  Virtual Reality: \cite{lentini2020robot,tram2023intuitive}. 
  \\\cline{1-2}
  Motion Modeling & 
  Control Policy Generation: \cite{thota2016learning,travers2016shape,travers2018shape,kim2022learning,li2024efficient}. 
  \newline
  Human-Motion Reproduction: \cite{lantz2004rhythmic,xu2006internal,stulp2009compactb,gams2011constraining,meier2012movement,chen2015efficient,pehlivan2015dynamic,chen2016dynamic,fanger2016gaussian,haddadin2016optimal,hiratsuka2016trajectory,umlauft2017bayesian,zhang2017robust,cho2018relationship,pan2018realtime,wang2020robot,liang2021dynamic,lu2021incremental}. 
  \newline
  Trajectory Generation: \cite{ude2010task,forte2011real,ning2011accurate,forte2012line,nemec2012action,ning2012novel,ajallooeian2013general,denivsa2013discovering,denivsa2013new,denivsa2013synthesizing,krug2013representing,nemec2013transfer,vuga2013motion,weitschat2013dynamic,lemme2014self,zhao2014generating,colome2015friction,denivsa2015synthesis,forte2015exploration,krug2015model,lioutikov2015probabilistic,alizadeh2016learning,dewolf2016spiking,kramberger2016generalization,nemec2016bimanual,samant2016adaptive,tan2016applying,chen2017robot,denivsa2017cooperative,lioutikov2017learning,zhou2017task,dimeas2018towards,dimeas2019progressive,kastritsi2018phri,kastritsi2018progressive,papageorgiou2020passive,papageorgiou2020kinesthetic,yang2018biologically,vlachos2020control,liu2020learning,rotithor2020combining,dahlin2021temporal,jaques2021newtonianvae,jiang2021multiple,lonvcarevic2021specifying,rotithor2022stitching,rouse2021visualization,wang2021learning,zhang2021robot,liendo2022dmp,wang2022temporal,anarossi2023deep,boas2023dmps,liu2023demonstration,lonvcarevic2023encoder,si2023composite,sidiropoulos2023from,teng2023fuzzy,zhang2023human,coelho2024dmps,abu2024learning,lonvcarevic2024effects,shen2024research,xu2025generalizing}.
  \newline
  Trajectory Modification: \cite{petrivc2014onlineb,hangl2015reactive,karlsson2017autonomous,hu2018evolution,weitschat2018safe,kordia2021movement,pacheck2023physically}.
  \\\cline{1-2}
  Motion Planning & 
  Driving Assistance/Behavior: \cite{wang2018learning,wang2019regeneration,wang2019motion,guan2023coordinated}.
  \newline
  Obstacle Avoidance: \cite{park2008movement,tan2011potential,tan2011conceptual,gams2013modulation,rai2014learning,kim2015adaptability,lee2016planning,jiang2016mobile,kim2017motion,mei2017mobile,ossenkopf2017reinforcement,rai2017learning,chi2019learning,ginesi2019dynamic,lauretti2019hybrid,pairet2019learning,sharma2019dmp,ginesi2021dynamic,li2021reinforcement,lu2021dynamic,maeda2022blending,tu2022whole,zhai2022motion,liendo2023improved,liu2023superquadrics,shaw2023constrained,sidiropoulos2023novel,niu2023generalized,yuan2023hierarchical,jia2024trajectory,lu2024dynamic,theofanidis2024safe,li2025model,liu2025novel,singh2025collaborative} 
  \\\cline{1-2}
  Robotic Prosthetics & 
  Exoskeletons: \cite{kamali2016trajectory,peternel2016adaptive,huang2016hierarchical,huang2016learning,huang2018hierarchical,chen2017step,deng2018learning,lauretti2018learning,ma2018gait,huang2019learning,hwang2019intuitive,hwang2021gait,yuan2019dmp,zou2019adaptive,li2020enhanced,li2020skill,qiu2020exoskeleton,xu2020stair,zou2020slope,lanotte2021adaptive,qiu2021exoskeleton,luo2022trajectory,zhang2022motion,eken2023continuous,lu2023visual,xu2023dmp,xu2023learning,eken2024continuous,yu2024modified,zhou2024trajectory,hao2025hierarchical,huang2025lower}.
  \newline
  Powered Prosthetics: \cite{zhang2021dynamical,eken2023locomotion,hong2023vision}. 
  \\
  \bottomrule
\label{tab:dmp_applications}
\end{longtable}
\twocolumn

\hspace{-1em}vated. An experimental comparison shows that insertion tasks can
be successfully executed with and without pose uncertainty.

Although robots have been installed in various industries to handle
high-precision tasks, they are still limited in adaptability, flexibility, and
decision-making skills. This is especially true for sensitive, contact-rich
assembly tasks. To tackle this dilemma, \cite{spector2020deep} combine deep RL
with DMPs to handle assembly tasks such as insertion. High-dimensional
continuous state-action spaces are addressed by dividing a task into subtasks.
This permits the subtasks to perform control in Cartesian, rather than joint
space, and thus parameterize the control policy. In addition, DMPs are extended
with a coupling term to support compliant control. The simulated system is shown
to learn insertion skills that are invariant to size, shape, and space while
handling large uncertainties, and it can achieve similar performance when
transferred to a robot arm.

\cite{iturrate2021quick} present an LfD framework for gluing tasks. In the
encoding stage, the task is segmented based on the initial contact into an
approach phase and a process phase using DMPs. Additionally, surface normal
estimates are encoded as radial basis functions and synchronized with the
process phase. During execution, a parallel position/force controller is
utilized to meet both a position target and a force target, which is calculated
based on a user-specified force magnitude applied to the encoded normal
direction. The encoded normal is again used to vary the eigenstructure of the
force controller gain matrices in order to apply force control in the contact
direction. When transitioning, the impact forces are reduced by slowly
modulating the goal of the DMP. The system is evaluated on an industrial gluing
task for printed circuit board assembly.

To execute manipulation behaviors involving physical interaction, robots require
the acquisition and generalization of force-relevant skills (e.g., cutting,
insertion, plugging, screwing, opening doors or shelves, etc.). However,
generalizing such skills to new force conditions remains a challenge.
\cite{zeng2021generalization} address this problem through a framework that
enables skill generalization by iteratively adjusting compliant profiles based
on a biomimetic control strategy. In particular, trajectories are encoded using
DMPs, while stiffness, damping, and feed-forward force parameters are adapted
online to accommodate new force fields. A simulated catching-a-ball task, and
real-world experiments with a dexterous hand involving moving a plank and
plugging into a socket, demonstrate human-like compliance adaptation.

Reteaching robots when tasks change is a difficult endeavor. For these
scenarios, \cite{dong2023novel} propose a robot skill learning framework that
combines LfD with behavior trees. To do this, a human operator constructs a
behavior tree for specific tasks, which is then used to build a robot motion
planner. In detail, the motion skills and controller outcomes are encapsulated
into the action nodes of a behavior tree. A DMP model that can scale and rotate
the shape of the trajectory is employed to improve overall reusability and
applicability (\cite{koutras2020novel}). During deployment, the behavior tree
selects the corresponding DMP parameters according to the system state and
generates the trajectory. A set of experiments, including sweeping and assembly
tasks, are conducted on different robot arm platforms.

\paragraph{Brain-machine interfaces.} 
Brain-machine interfaces can provide a basic level of prosthetic control by
tapping into a human's cortical signals.  Nonetheless, many simple tasks (e.g.,
object manipulation) require precise trajectories, which has not yet been
obtained from direct neural control. To make progress towards this goal,
\cite{hotson2016high} present a framework for arbitrating between direct neural
control and nonlinear predictions of user intent as follows. First, the type of
task the user is engaged in is predicted through a hidden Markov model (HMM).
Then, the inferred task is used to inform a nonlinear prediction of the user's
desired kinematics via DMPs. Finally, to localize the user's intended position,
the prediction is fused with neural sensor measurements using an unscented
Kalman filter. The utility of system is demonstrated by reconstructing
trajectories taken by a non-human primate performing four different actions on
objects placed in various locations.

\paragraph{Deformable objects.} 
There is a growing need for robotic clothing assistance since it is one of the
most basic and essential tasks in the daily life of elderly people. Yet,
automated clothing assistance is challenging due to the following: (i) the robot
needs to perform cooperative manipulation by holding a non-rigid and highly
deformable piece of clothing utilizing both arms; (ii) the robot must maintain
safe HRI with the person whose pose can vary during assistance. To this end,
\cite{joshi2017robotic} provide an LfD approach where DMPs are used as a task
parameterization model for clothing assistance tasks. The methodology is broken
up into the following phases: teaching, trajectory learning, and testing. Six
DMP systems are used to control a dual-arm robot and the experimental evaluation
is designed to highlight both clothing tasks and failure cases. Subsequently,
\cite{joshi2019framework} fine-tune clothing assistance using a humanoid robot,
modeling the task in three phases: reaching, arm dressing, and body dressing.
DMPs model the arm dressing phase as a global trajectory modification. The body
dressing phase is modeled as a local trajectory modification via a Bayesian GP
latent model. The results show the generalizability to various people,
successfully performing a sleeveless shirt dressing task. In addition,
experiments with a humanoid robot demonstrate full dressing of a sleeveless
shirt on a human subject.

To enable safe physical HRI, robot joint controllers must operate with low
gains, which can lead to poor trajectory tracking. \cite{nemec2018adaptive},
building upon \cite{likar2015adaptation}, address this inherent trade-off by
introducing a modified relative coordinate formulation for bimanual control that
better decouples absolute and relative motion. The key contribution is combining
this improved coordinate system with an ILC scheme that progressively improves
tracking accuracy across multiple executions while maintaining low controller
gains. The approach is validated on a dual robot manipulator system performing a
tablecloth placement task with a human partner. In this task, a reduction in
tracking errors within five iterations is shown.

\cite{cui2022coupled} develop a dynamically coupled multiple DMP generalization
method to manipulate deformable objects. The technique addresses model parameter
uncertainty caused by soft-body objects by using a reference model to adaptively
introduce a coupling term into the multiple DMP generalization. This permits the
method to adaptively estimate the undetermined model parameters. The DMP
generalization, pattern preserving configuration, and collision avoidance are
compactly modeled in one integrated second-order system. The framework is
demonstrated on rope manipulation tasks via simulations, along with a dual-arm
robot, where deformable objects are modeled as a mass-spring-damper system. 

Many methods have been investigated to enhance the adaptability and
generalization of robot manipulation, yet it is still difficult to perform
complex contact-rich tasks on deformable objects without the assistance of
humans. \cite{si2022adaptive} address this problem via an LfD framework that
employs DMPs to enhance robot safety and adaptability as follows. First, a
hybrid control architecture for a bilateral teleoperation system that includes
hybrid force and position control as well as impedance control to achieve
human-robot compliant skill transfer is developed. Then, the teleoperation
system is employed to allow operators to correct the behavior of the robot.
Lastly, the new behavior is used to update the pre-learned compliant DMP model.
The approach is evaluated on the task of rolling dough of various hardness.

\cite{chandra2023dual} improve the effectiveness of CDMPs via a compact and
efficient representation of dual quaternions. The LfD methodology employs a
screw-based interpretation of a rigid transformation and exponential mapping of
the screw parameters to obtain the corresponding unit dual quaternion. This
ensures that the resulting dual quaternion always satisfies the unity norm
constraint. The framework is utilized for interpolating and filtering Cartesian
trajectories online. It can be used for safe teleoperation since the screw
interpolation is capable of limiting velocity twist to a desired value.
Additionally, the approach can deal with irregular demonstration data tracking
and corresponding noise. The framework is demonstrated on the task of
manipulating deformable objects using dual robot arms. 

\paragraph{Grasping.} 
Taking inspiration from human movements, \cite{kroemer2011grasping} introduce
techniques for robotic grasping in environments with clutter using potential
fields and only a small amount of visual information. This is done by
supplementing DMPs with information derived from early cognitive vision
descriptors as follows. First, an attractor field is defined as a DMP to encode
the desired movements with no obstacles. Then, the goals and IL trajectories are
combined to specify the attractor fields to handle obstacles. These methods
provide tools that a robot needs to reactively execute grasps of an object in
cluttered scenarios without relying on a complex planner. To demonstrate the
framework, a set of grasping experiments are conducted using a robot hand, arm,
and a stereo camera mounted on a pan-tilt unit. In proceeding work, the method
is integrated into a hierarchical RL/DMP-based controller that can determine
good grasps of objects and execute them \cite{kroemer2010combining}.

An executed DMP trajectory will associate sensory variables specific to that
manipulation skill. Not only can these associations be used to increase
robustness towards perturbations, but they can also allow failure detection and
switching towards other behaviors. For instance, \cite{pastor2012towards}
describe how associative skill memories can be acquired by IL, and then used to
create manipulation skills by determining subsequent skill memories online for a
particular manipulation goal. This is done via a method that generates motion
using knowledge from previous task executions.  These past sensor experiences
resemble the nominal behavior. They are stored with the corresponding executed
movement and enable the system to predict future sensor measurements. An
arm/hand robot is utilized to evaluate the approach on manipulation tasks. The
notion that all sensory events connected to DMPs form associative skill memories
is further described by \cite{pastor2013dynamic}.

Tendon-driven systems provide distinct advantages for space-constrained
mechanisms such as anthropomorphic hands. However, compared to torque-driven
systems, the inverse dynamics solution for a tendon-driven system is much more
complicated. To address this problem, \cite{rombokas2012tendon} express the
tendon space in a lower-dimensional space and provide a first-order model that
captures the dynamics of actual sensor recordings of the robot in the
environment. PI$^2$ is used to learn a control policy on tasks that involve two
fingers (e.g., sliding a switch). The trajectories, represented as DMPs, are
learned in the tendon space of the index finger and the thumb. In
\cite{rombokas2012reinforcement}, the approach is extended to include a less
informative cost feedback function, incorporate tendon synergies, and
demonstrated on a knob-turning task.  

Robots equipped with mechanical grippers are much more affordable and widely
deployed in practice. Taking advantage of this situation, \cite{li2015teaching}
develop a strategy to replace dexterous hands with dual grippers for
manipulating tools. The system encodes interaction patterns between tools and
human hands, and then transfers them to a robot's end effector. To do this, a
hierarchical architecture is designed to embed tool use into an LfD framework.
At the high-level, the method learns temporal orders for dual-arm coordination.
DMPs are learned at the lower-level as the manipulation primitives for multistep
execution. Experimental results demonstrate the learning and operation of three
human tools including an electric tacker, an electric drill, and a hot-glue pen.

\cite{de2018reinforcement} investigate the task of spinning a ball optimally
(i.e., with minimum contact forces and no translation or rotation in undesirable
directions) around the axis perpendicular to the palm of the robotic hand.
Since the optimal finger gaiting is not known a priori, a hierarchical RL
planning framework is developed to obtain an appropriate parameter
initialization. At the highest level, states and actions are abstract
representations that correspond to a subset of the states and actions at the
lower level. The key idea is to solve the levels in a decoupled way. This is
first done at the upper level using Q-learning. When a solution is found, it is
then used to initialize the lower-level search for a solution via PI$^2$ and
rhythmic DMPs. Experimental evaluations show that the hierarchical approach with
PI$^2$ is able to steadily improve the ball spinning performance. 

Human-like robot hands can accomplish a variety of tasks that would otherwise
require special-purpose end effectors. Nevertheless, these tasks are hard to
manually program due to complex trajectories (e.g., writing) or they have
significant acceleration profiles (e.g., cutting). \cite{solak2019learning}
solve these problems by learning in-hand robotic manipulation tasks from human
demonstrations using DMPs. To do this, the tasks are reproduced by employing
real-time feedback of the contact forces measured on the robot fingertips using
a compliant controller based on a virtual springs framework. This allows the
generalization capabilities of DMPs to be successfully transferred to the
dexterous in-hand manipulation problem. Using a collaborative robot arm, an
evaluation demonstrates that the method can achieve dexterous manipulation of
objects without having access to the object model. 

Long-distance teleoperation of hand-arm exoskeleton systems can experience high
latency and thus high uncertainties. \cite{beik2020model} propose a
model-mediated teleoperation framework that overcomes these issues by
integrating RL and DMPs within a simulated environment. Concretely, an
environment model is used to obtain instant feedback on trajectories by adapting
the trajectories using RL techniques. The methodology employs a two-layer
system. In the first layer, DMPs are utilized to reconstruct and adapt to
teleoperated trajectories in a virtual environment, providing instant feedback
to the teleoperator. The second layer fine-tunes unsuccessful trajectories using
policy search. Experimental results show that the double-layer framework rapidly
adapts to the high uncertainties found in long-distance robot teleoperation.

\paragraph{Mobile manipulation.} 
\cite{beetz2010generality} highlight mobile robot manipulation capabilities by
integrating obstacle avoidance with DMPs using vector fields. This is done by
first training one DMP for each principal trajectory with a regression learning
algorithm. Then, a vector field takes the current task position given by the
forward kinematics and utilizes information about obstacles and goals to
calculate a task space velocity vector. Next the velocity vector is translated
to joint speeds by an inverse kinematics module. If the robot's arm is pushed
too far away from the trajectory dictated by the DMPs, then it is controlled by
the vector field to provide a smooth movement to the goal. This permits the
system to avoid obstacles that move, appear, or disappear from the robot's
workspace. The combination of these mechanisms into a single low-level control
system is demonstrated on an autonomous manipulation platform.

Structured demonstrations typically consist of a sequence of subtasks, which are
easier to learn and generalize than an entire continuous task. Nonetheless,
effective task segmentation may require knowledge of the robot's kinematic
properties, internal representations, and existing skill competencies. Based on
this observation, \cite{niekum2012learning} develop a method that integrates a
Bayesian nonparametric approach for task segmentation via LfD. Segmentation and
recognition are performed using a beta process autoregressive HMM, while DMPs
are employed to address LfD, policy representation, and generalization. This can
enable the accumulation of a library of skills over time. Various skill
acquisition tasks are conducted to demonstrate the system both in simulation and
on a mobile manipulator. Follow-up work highlights the learning of tasks from
unstructured demonstrations \cite{niekum2015learning}. 

To learn complex motion sequences in human-robot environments,
\cite{li2017reinforcement} describe an RL strategy for grasping using a mobile
manipulator that handles varying manipulation dynamics in the presence of
uncertain external perturbations. This is done through employing vision sensor
feedback to record the position of the grasped object and by exploiting position
information to formulate an operational space target point as follows. First, a
redundancy resolution is computed by optimizing a primal dual neural network.
Next, optimal solutions obtained by the network are treated as the target points
of DMPs and used to model and learn joint trajectories. Finally, RL is utilized
to learn the trajectories with uncertainties and reduce the complexity of the
vision-based feedback. Experimental results show that the method can suppress
uncertain external perturbations.

\cite{zhao2018cooperative} present a two-phase system for cooperative mobile
manipulation. Specifically, the framework is composed of high-level
neural-dynamic optimization-based redundancy resolution in the operational
space, and low-level DMP-based RL in the joint space.  During the first phase, a
vision system acquires the target position of the reaching point in the task
space. In the learning phase, a sequence of DMPs (\cite{stulp2012reinforcement})
are used to encode the joint trajectories for two subgoals. These subgoals
consist of the reaching point and the final lifted-up point recorded in the
first phase. PI$^2$ is then applied to learn and regulate the joint trajectories
by adjusting the shape and goal parameters such that a desired optimal control
objective is attained. Using a mobile dual-arm robot, experimental results
highlight the ability to suppress uncertain perturbations and achieve successful
manipulation.

\paragraph{Insertion.} 
Robotic assembly operations not only require knowledge of the position and
orientation of trajectories, but also the accompanying force-torque profiles for
successful performance. To this end, \cite{kramberger2017generalization} propose
a methodology for generalizing orientation trajectories and force-torque
profiles for assembly tasks. Concretely, DMPs are employed to operate on
rotational differences between trajectories rather than directly on quaternions
to maintain the unit norm. The force-torque profiles are generalized using LWR
to adapt to new task conditions. Validation of the framework is performed on a
dual-arm robot manipulator via valve turning and peg-in-hole assembly tasks. The
approach allows robots to adapt assembly skills to varying geometries without
requiring new demonstrations, effectively bridging the gap between LfD and
autonomous adaptation to novel task conditions.

Although remote teleoperation can permit robot operations in extreme conditions,
it still demands a heavy mental workload from the human operator. To mitigate
this problem, \cite{pervez2017novel} develop a DMP learning scheme that can
handle less consistent, asynchronous, and incomplete demonstrations. To do this,
an EM algorithm is first used for estimating GMM parameters and phase variables.
Then, the DMP forcing term is encoded using the GMM. Lastly, after learning, the
forcing term is synthesized via GMR. This allows for synchronizing and encoding
demonstrations with temporal and spatial variances, different initial and final
conditions, and partial executions. The approach is evaluated on a peg-in-hole
task with a master-slave teleoperation system. Further experimental results are
presented in \cite{pervez2019motion}.

DMP encoded tasks can be reused with a different goal position, albeit with a
resulting distortion in the approach trajectory.  Although this may be
sufficient for some applications, the accuracy requirements for assembly tasks,
where tolerances are tight and workpieces are small, is much higher. In these
scenarios, \cite{iturrate2019improving} improve the accuracy of DMP
generalization via classification of temporal data as follows. The task is first
taught by guiding the robot using admittance control. Next, a convolutional
neural network (CNN) is used to segment the task based on contact forces
captured by a force-torque sensor, which are then encoded as DMPs. Experimental
results show increased generalization when compared to unsegmented DMPs, however
task replays may still fail when generalizing to new positions and orientations
due to imperfect demonstrations.

DMPs can generalize trajectories imitating a demonstration, yet they cannot
integrate the features of multiple trajectories for different targets. To
address this problem, \cite{ti2019human} employ a GMM and GMR to extract the
common characteristics and eliminate the uncertainty of multiple demonstrations.
Prior to GMM encoding, dynamic time warping (DTW) is utilized to temporally
align the signals and the DMP trajectory parameters are modeled via an LWR
method. In addition, multivariate GPR is used to construct the regression model
of the parameters to reflect human intentions, with respect to the target
position, which enhances the ability of the DMPs to learn multiple trajectories.
The feasibility of the method is evaluated on peg-in-hole experiments, with
obstacles, using a 6-DoF robot arm.

Despite careful policy optimization, assembly operations can fail due to
unforeseen situations. \cite{nemec2020learning} handle this problem with a
framework where, when failures occur, an operator demonstrates how to recover by
physically guiding the robot through the corrective motion. Using PCA to
generate a low-dimensional context descriptor, the system analyzes force/torque
data and then generalizes from past experiences using LWR to generate an
appropriate recovery policy, or requests a new demonstration if insufficient
similar experiences exist. The demonstration process uses incremental
kinesthetic teaching within a refinement tube. This enables operators to modify
only the relevant parts of the trajectory while preserving the rest. Over time,
the system becomes increasingly autonomous in handling assembly failures without
human intervention on peg-in-hole tasks with a robot manipulator.

During assembly tasks, unpredictable forces and torques may result when
uncertainty exists in the locations of the contacting parts. Therefore,
correcting the trajectory in response to haptic feedback despite location
uncertainties is a crucial skill. \cite{spector2021learning} address this issue
by learning a task-specific residual admittance policy that enables general
position and orientation corrections in response to force/torque feedback. The
policy is learned to correct movements generated by a baseline policy within the
framework of DMPs. Given the reference trajectories created by the baseline
policy, the action space of the policy is limited to the admittance parameters.
Deep RL is used to train a DNN that maps task specifications to proper
admittance parameters. Using a collaborative robot, empirical results show the
successful insertion of pegs of different shapes and sizes despite localization
and grasping errors.

DMPs have been extended to include force trajectories for contact-rich tasks
where position trajectories alone may not be robust over variations in the
contact geometry. Nevertheless, different task phases or DoF may require the
tracking of either position or force. For example, once contact is made it may
be more important to track the force demonstration trajectory in the contact
direction. To this end, \cite{chang2022impedance} utilize DMPs to learn position
and force trajectories from demonstrations, and then adapt the impedance
parameters online with a higher-level control policy via RL. This permits a
one-shot demonstration of the task with DMPs along with improved robustness and
performance from the impedance adaptation. The methodology is evaluated on a
peg-in-hole and adhesive strip application task using a robot arm.

Although DMPs are a popular way of extracting policies for a variety of tasks,
they can struggle on jobs that consist of inserting objects. To tackle this
problem, \cite{davchev2022residual} consider several possible ways to adapt DMPs
for peg-in-hole tasks. Specifically, an LfD framework that combines DMPs with RL
to learn a residual correction policy is proposed. Evaluations of the system
show that the overall performance of DMPs can be improved by applying residual
learning directly in task space and by operating on the full pose of the robot.
Moreover, the framework has the following benefits: (i) gentle on the robot's
joints, (ii) improves both task success and generalization of DMPs, and (iii)
enables transfer to different geometries and frictions through few-shot task
adaptation. 

To reduce the repetitive effort of teaching new tasks, \cite{wang2022adaptive}
present an imitation framework for robot manipulation. In particular, DMPs are
introduced into a hybrid trajectory and force-learning system in a modular
manner. The framework learns a specific class of contact-rich insertion tasks
based on the trajectory profile of a single task instance belonging to the task
class. This is done by learning the nominal trajectory via a combination of IL
and RL, and learning the force control policy by an RL-based controller.
Adaptive DMPs, where the coupling terms and the weights of the forcing terms are
learned through RL, adapt the trajectory profile of a single task to new tasks
with topologically similar trajectories. Experimental evaluations show that the
framework is sample efficient, safer, and generalizes better at learning
insertion tasks.

\cite{li2023enhanced} develop an LfD methodology to improve the generalization
performance of a robot on axle hole assembly tasks. The proposed method combines
DMPs and task parameterized-LfD, which allows for learning features from
multiple trajectories by performing probabilistic statistics on the trajectory
data. Task-related parameters based on the target pose are used to adapt the
robot's movement (\cite{calinon2016tutorial}). DTW is utilized to preprocess the
recorded robot demonstration data to obtain a smoother motion trajectory model.
In addition, the Fr\'{e}chet distance is used to compute the similarity between
the reproduced trajectory and the demonstration data. An evaluation criterion
based on curve similarity is constructed to determine the model's interpolation
and extrapolation performance on insertion tasks.

A sense of touch can be necessary for robots to handle objects effectively.
Based on this consideration, \cite{nguyen2025quasi} design a method to perform
contact-rich manipulation using only force/torque measurements operating under
quasi-static conditions as follows. First, the interaction forces/torques
between the manipulated object and its environment are recorded. Then, a
potential function is constructed from the collected force/torque data using GPR
with derivatives. Next, haptic DMPs are developed to generate robot
trajectories. In contrast to conventional DMPs, which mainly focus on kinematic
aspects, haptic DMPs incorporate force-based interactions by integrating the
constructed potential energy. The effectiveness of the methodology is
demonstrated via numerical tasks, including the peg-in-hole insertion problem.

\paragraph{Pick-and-place.} 
\cite{pastor2009learning} introduce a framework for learning and generalizing
manipulation skills via human demonstrations. The approach consists of three key
contributions: (i) goal generalization through a modified DMP formulation, (ii)
semantic labeling of movements based on task context, and (iii) the integration
of orientation and gripper control in end-effector space. The framework is
validated on a robot manipulator performing water-serving and pick-and-place
tasks with online adaptation to moving goals and obstacles. Movement transitions
are achieved by initializing successive DMPs with the velocity and position
states of their predecessors to avoid abrupt changes. Notably, the method uses
quaternions to represent end-effector orientation trajectories alongside the end
effector in Cartesian space and the finger position within joint space, all in
one unified framework.

Grasping under object uncertainty has seen innovations such as the adoption of
robust MPs to account for state estimation uncertainties. For example,
\cite{stulp2011learninga} combine DMPs with a probabilistic model-free RL
algorithm to generate grasping trajectories. To obtain DMPs that can handle
uncertainty, the object pose is randomly sampled from a probability
distribution. This enables a robot to optimize the probability of grasping the
object in any of the possible orientations provided by the distribution. In
\cite{stulp2012reinforcement}, the RL policy is extended and additional
experiments on the task of pick-and-place are conducted. Later work on this
topic by \cite{zhang2024using} introduces an implicit behavior cloning (BC)
framework that further improves the learning performance and generalizability of
RL-based robot motion planners using demonstrations and DMPs.

In robotic manipulation, traditional action representation methods often
struggle to generalize actions across different objects, positions, and
orientations. \cite{aein2013toward} address this issue through a layered
architecture for defining and executing human-like robotic manipulation actions.
The system utilizes a semantic event chain framework integrated with a finite
state machine to represent and execute actions procedurally. Unlike symbolic
representations, the approach encodes actions as sequences of object-relation
changes, enabling systematic generalization across different objects and
scenarios. Specifically, the method encodes a \textit{move} primitive as a
Gaussian kernel DMP. The architecture operates across three levels: high-level
symbolic abstraction using semantic event chains to define actions, a mid-level
finite state machine for sequencing action primitives, and low-level execution
primitives that leverage real-world sensor data to carry out actions.
Experiments are conducted on pushing and pick-and-place tasks using a robot arm.

Connected convex surfaces, separated by concave boundaries, play an important
role in the perception of objects along with their decomposition into parts.
Based on this observation, \cite{stein2014convexity} present a bottom-up
approach to segment 3D point clouds into object parts for robotic applications.
The method first approximates a scene using an adjacency-graph of spatially
connected surface patches. Next, edges in the graph are classified as either
convex or concave via a strictly local criterion. Finally, region growing is
used to locate locally convex connected subgraphs that represent object parts.
Action execution is carried out by a library of manipulation actions based on
modified DMPs (\cite{kulvicius2012joining}). A pick-and-place evaluation is
performed where the pose of an object handle is calculated from its 3D shape
using the proposed algorithm.

\cite{lioutikov2016learning} address the challenge of sequencing multiple DMPs
to execute complex multistep tasks that cannot be represented by a single
primitive. The approach enables independent learning of individual movement
components. Not only does this allow demonstrations to be simpler and more
robust, but it also preserves the generalization capabilities of DMPs across the
entire sequence. Furthermore, the framework permits temporal scaling of each
primitive and generalization of goals to adapt to different object positions and
execution speeds. A bimanual vegetable manipulation task is performed to
validate the system where each arm executes its own sequence of DMPs for
grasping, placing, and cutting. The evaluation shows successful generalization
to new object positions with doubled execution speed while maintaining task
performance.

A method for learning manipulation tasks from human demonstrations via DMPs is
developed by \cite{hung2017programming} as follows. First, demonstrations of a
task are recorded by a hand-motion tracker using an RGBD camera and color-marker
glove. Then, the recorded movements are segmented into sub-actions. The
segmentation is based on the velocity of the hand motion and distance of the
fingers. Next, each sub-action is mapped to a primitive skill. This results in a
library of common robot skills that can be automatically translated to a
specific robot movement. Lastly, a DMP model is learned to generate trajectories
that follow the demonstrated movements. An evaluation is performed on a water
dispensing task with a robot arm on the following actions: pick up a cup, place
the cup under the spout, press the button, and pick up the cup with some water.

DMPs are effective at reproducing demonstrated motions, however they struggle to
generalize to new behaviors. To mitigate this issue \cite{kim2018learning}
propose a hierarchical deep RL methodology, where DMPs serve as temporal
abstractions for learning complex skills from demonstration. In detail, the
framework combines a deterministic actor-critic algorithm, a hierarchical
strategy that decomposes tasks, and a two-level controller structure consisting
of a meta-controller and a sub-controller. Experimental validation on a 6-DoF
robot arm with a 1-DoF gripper shows that the method is able to refine
suboptimal demonstrations while generalizing to new goal positions during
pick-and-place tasks.

Part feeding is a critical but troublesome task, especially for flexible
manufacturing systems where classical methods are less applicable due to
constraints from small batch sizes and high part variance.
\cite{kramberger2020adapting} tackle this problem via an LfD framework that
exploits DMPs as the underlying trajectory representation. During the execution
phase, admittance control (defined during the demonstration) is used with
switching force control parameters for reliable execution of the part feeding
skill. This allows for lower contact forces to be achieved and consequently more
parts can be scooped, aligned, and fed to the process. The approach is evaluated
on several part bin feeding cases using a robot manipulator equipped with a
scooping tool.

\cite{li2020generalization} develop an LfD system based on DMPs for object
recognition. The visual recognition component consists of the following steps.
First, an RGBD camera captures images of a scene before and after the placement
of a target object. Then, an image mask separates the target object from the
background.  Lastly, the centroid of the object is calculated using the image
mask, and the centroid's coordinates in the camera coordinate system are
transformed into the world coordinate system. Skill generalization is summarized
as follows: (i) obtain the demonstration trajectories and set the DMP
parameters; (ii) learn the target forcing term and solve the weighting for the
basis functions; (iii) set the start/target positions and generate a new
trajectory. Simulations are run on a pick-and-place task using a robot
manipulator.

The benefit of combining LfD and RL is that a robot can learn a skill from a
demonstration and then improve it using RL. To be combined with RL, the LfD
method must be policy-based. DMPs are a suitable option since they model the
forcing term using basic functions and convergence to the goal position is
guaranteed. \cite{noohian2022framework} further explore this idea by modeling
DMPs within the framework of RL. This first requires modeling the forcing term
of the DMP as a neural network. Then, a suitable reward function is defined and
the DMPs are learned using deep RL. In particular, a twin-delayed deep
deterministic policy gradient algorithm is employed. Nonetheless, the framework
can be applied to any policy optimization or actor-critic deep RL algorithm.
Simulation results show that the system is able to learn trajectories with high
accuracy on pick-and-place tasks.

\cite{chen2023object} build a framework for 3D object recognition, localization,
and manipulation using an object detection algorithm, two low-cost cameras, and
a 7-DoF robot manipulator equipped with an anthropomorphic hand. Autonomous
pick-and-place tasks are executed using impedance control and difference-based
DMP trajectory planning. Specifically, the difference-based DMPs are utilized to
learn unique trajectories for different objects based on human demonstrations
and the object's coordinates, which improves the viability and safety of the
regenerated trajectories. A feedback impedance controller is designed in
Cartesian space to achieve compliant physical interaction with users.
Experimental evaluations are conducted on pick-and-place tasks for two different
objects.

Learning multi-step manipulation skills is crucial for improving the autonomy
and capability of redundant manipulators. For example, \cite{ning2023novel}
create a multi-task DMP framework that enables robots to learn and execute
sequential tasks. The architecture consists of four components: a sub-task
segmentation module based on finite-state machines, a parameter setting module,
a DMP-based robot skill learning module, and a velocity directional
manipulability-based pose optimization module. An exponential decay function is
introduced to smooth accelerations, while adaptive particle swarm optimization
is used to improve pose efficiency. Experiments with a 7-DoF manipulator
performing pick-and-place tasks show that the system improves execution time by
up to 36\%, enhances motion manipulability, and achieves a 100\% success rate.

Motivated by the need to develop a user-friendly LfD framework that can collect
demonstration data, \cite{chen2024vision,chen2024robust} propose the use of DTW
with GMM/GMR to capture and regress multiple dexterous and markerless hand
motions. DMPs with a variance-based force coupling term are developed to
adaptively assimilate human actions into trajectories. By considering the
estimated variance from the demonstration data, DMP parameters are automatically
fine-tuned and associated with the nonlinear terms for trajectory adaptation.
Furthermore, to compensate for unknown external disturbances, non-singular
terminal sliding mode control is applied for precise trajectory tracking. The
approach results in a more comprehensive motion representation, simplifies
multiple demonstrations, and mitigates the non-smoothness inherent in single
demonstrations. An experimental study is conducted using a robot arm on a
pick-and-place task.

\cite{ning2024mt} formulate a framework to enhance the efficiency and robustness
of multitask robot skill learning for contact manipulation (e.g., pick-and-place
tasks). The system consists of a coarse-to-fine sub-task segmentation module
that analyzes robot behaviors to obtain a sub-task sequence. To improve the
efficiency of the robot, DMPs are employed within a skill learning module that
performs actions by coordinating sub-parts. Lastly, a configuration optimization
module is designed by incorporating a velocity directional manipulability
measure as the evaluation index of robot kinematic performance. Experimental
results show that the methodology improves multitask robot skill learning,
especially in terms of learning efficiency, velocity directional manipulability,
and success rate.

\paragraph{Pouring/pumping/turning.} 
When solving a robotic task under circumstances that were not present during the
initial learning, it is necessary to adapt the trajectories. To this end,
\cite{nemec2009task} investigate how RL can be utilized to modify the subgoals
of DMPs. Specifically, since a smooth transition between DMPs without coming to
a full stop is often needed to effectively sequence the primitives, a
formulation that ensures smoothness (up to second-order derivatives) of the
transition between two consecutive DMPs is proposed. Jumps in velocities and
accelerations when joining two trajectories is solved by applying a first order
low-pass filter at the output of the DMP generator. This modification ensures
continuous accelerations, whereas the accelerations in the original formulation
are discontinuous. The methodology is applied towards learning appropriate robot
movements for pouring a liquid.

When using RL with DMPs, goals and temporal scaling parameters are usually
predefined and only the weights for shaping a DMP are learned. Nevertheless, in
many tasks the best goal position is not known a priori and must be learned. To
solve this problem, \cite{tamosiunaite2011learning} investigate the simultaneous
combination of goal and shape parameter learning for the task of liquid pouring.
Concretely, function approximation RL is used for goal and shape learning. A
value function approximation technique for goal learning is utilized due to the
structure of the reward landscape for the goal point in pouring. In the reward
landscape, the objective is to find large areas with zero reward in which no
liquid can be successfully poured and only a limited patch where a reward can be
obtained. Experiments are carried out in simulation as well as on a robot arm.

Robots should be able to learn relationships between task-relevant MPs and
affordances for activating the MPs. Based on this observation,
\cite{lee2013skilla} build a skill learning and inference system consisting of
the following modules: a human demonstration process, an autonomous segmentation
process, a DMP learning process, a Bayesian network learning process, a
motivation graph construction process, and a skill-inferring process.
Segmentation points are estimated from a GMM learned from motion trajectories.
The segmented motion trajectories are then formalized as DMPs, which guarantee
goal convergence under uncertainties and perturbations. To handle these
uncertainties, \cite{lee2013skillb} represent probabilistic affordances as
Bayesian networks to learn meanings that can activate the DMPs. The framework is
evaluated on a tea service (pouring) task using a robot arm.

\cite{petrivc2014onlinea} address the problem of accurate trajectory tracking
while ensuring compliant robotic behavior for periodic tasks. The online
approach learns task-specific movement trajectories and corresponding
force/torque profiles by combining central pattern generators (CPGs) with
kinematic and dynamic profiles encoded using DMPs. The framework is built on a
two-stage motion imitation system (\cite{gams2009line}) augmented with dynamic
torque primitives as follows. First, an adaptive frequency oscillator
(\cite{petrivc2011line}) is utilized to extract robot trajectories captured via
LfD. Next, the trajectory is executed under supervision using high feedback
gains to ensure accurate tracking. Finally, the robot movement is carried out
with the learned feed-forward task-specific dynamic model, allowing for low
position feedback gains. The methodology is demonstrated on two robotic tasks:
object manipulation and turning a crank.

\cite{worgotter2015structural} devise a general framework for a generative
process that improves learning and can be used at different levels of a robot's
cognitive architecture. To do this, structural bootstrapping is employed to
define a probabilistic process that utilizes existing knowledge together with
new observations to supplement a robot's database with missing information. The
system consists of the following three-layer architecture: a planning level, a
mid-level, and a sensorimotor level. To perform a task, the robot first makes a
symbolic plan. The mid-level then acts as a symbol-to-signal mediator and
couples the planning information to the sensorimotor level where DMPs are used
to represent the motion. Lastly, the sensorimotor level performs the execution
and monitors the task progress along with any errors caused by the robot's
actions. The framework is evaluated using a humanoid robot on pouring, mixing,
and wiping tasks in a kitchen environment.

In complex multistep tasks, it can be very difficult to generalize learned
policies and adopt them to new task situations. \cite{yang2018learning} address
this complication via a demonstration-segmentation-alignment-generalization
process for human-to-robot skill transfer. The key idea is that learned skills
are first automatically segmented into a sequence of subskills. Then, each
individual subskill is encoded and regulated accordingly. More specifically,
each set of the segmented DMP encoded trajectories is individually adapted
instead of the whole movement profiles. Human limb stiffness estimates from
surface EMG signals are also utilized for human-to-robot variable impedance
control, as well as the generalization of both trajectories and stiffness
profiles. The effectiveness of the framework is tested on pouring tasks using a
dual-arm robot.

The ability to leverage multiple demonstrations to encode ideal trajectories,
along with probabilistic approaches, can improve the performance of LfD systems.
Due to their generalization capability, DMPs can be used as the basis of the
motion model to ameliorate learning proficiency. For instance,
\cite{yang2018robot} develop a robot learning system consisting of motion
generation and trajectory tracking as follows. First, GMMs and GMR are
integrated with DMPs to learn and generalize motion skills. Then, an adaptive
controller is designed to track the set of generated trajectories. To do this, a
radial basis function neural network (RBFNN) is employed to compensate for
uncertain dynamics. This design allows the robot to extract more motion features
from multiple demonstrations. The approach is evaluated on pouring tasks using a
robot arm.

The usual methods for a task and motion planning framework are intractable when
applied to problems with large configuration spaces (e.g., robot manipulation).
To this end, \cite{agostini2020manipulation} resolve the intractability problem
by first leveraging computationally efficient linear planners to obtain feasible
motions. Then, a task planner is integrated with a hierarchical decomposition of
contextual actions by combining LfD and action segmentation. Initial and goal
poses, along with robot trajectories, are automatically extracted from the
demonstrations and used to fit DMPs. Finally, action contexts associated with
symbolic schemas allow for defining geometric constraints with respect to the
current task and action intentions. The system is evaluated on different pouring
and stacking tasks using a 7-DoF robot arm.

Motivated by the diversity of liquid pouring tasks, \cite{noseworthy2020task}
also use DMPs for this problem. Specifically, the task parameters that determine
what task to perform (e.g., the location of the container) and the unspecified,
learned manner parameters that describe how that task should be performed (e.g.,
the length of the pour) are made distinct. To do this, a task-conditioned
variational autoencoder is designed to ensure that learned manner parameters are
independent of the specified task parameters via adversarial training. The
adversarial training is then utilized to disentangle the parameter sets.
Evaluations using a robot arm show that the model generalizes to new task
instances while maintaining the interpretability of the learned parameters. 

In industrial environments valve turning is important for production safety, yet
it is a daunting manipulation task that requires robots to learn complex motor
skills. \cite{bai2022robot} take on this problem and create a method for
automatic valve turning by enabling a robot to acquire motor skills from
demonstrations through the following steps. First, recorded teaching motions are
temporally aligned using DTW. Then, GMMs and GMR are employed to capture the
statistical structure of the demonstrated human motions. Finally, DMPs are used
to learn and generate the motion trajectories. An experimental evaluation using
a 6-DoF robot arm performing a butterfly valve closure task shows that the
system can successfully complete the operation by effectively replicating
reaching and turning motions. 

Research on DMPs has tended to focus more on skill learning at the kinematic
level rather than at the dynamic level (e.g., stiffness, contact force). Based
on this observation, \cite{liao2022dynamic} propose an LfD framework that
simultaneously learns motion, stiffness, and force variations from one-shot
demonstrations as follows. First, a Riemannian-based DMP method is developed to
model motion, stiffness, and force in Cartesian space and on a 2D sphere
manifold. This provides a way to learn multi-space skills simultaneously,
including position, orientation, and 3D endpoint stiffness. Second, human-like
variable impedance control of a robot is realized by learning the extracted 3D
endpoint stiffness via a simplified geometric configuration of a human arm
skeleton. The system is evaluated on a water pouring task using a robot
manipulator. In proceeding work, \cite{liao2024simultaneously} incorporate
quadratic programming optimization to fine-tune the desired position of the
controller and evaluate on button pressing and polishing tasks.

One way for robots to perform collaborative tasks is to learn human stiffness
scheduling strategies. For example, to execute in-contact tasks (e.g., pumping)
while being as compliant as possible, \cite{yu2022human} introduce a framework
for transferring human-variable impedance skills to robots as follows. First,
EMG is employed to record human arm stiffness. Then, the stiffness is encoded
alongside movement trajectories into a DMP. The approach aims to enable robots
to mimic human-like adaptive impedance in task execution, particularly for
in-contact tasks that require a balance between force application and
compliance. To highlight the effectiveness of the learning method, an
application involving a water pumping task using a robot manipulator is
demonstrated.  

LfD systems for bimanual telemanipulation may struggle with generalizing learned
trajectories to different target orientations, which may be overcome by
modifying the reference frame in which movements are encoded. For instance,
\cite{amadio2022target} propose a target-referred DMP framework that expresses
learned trajectories relative to the target object rather than the world frame.
This allows robots to adapt manipulation skills to different object orientations
without distorting the demonstrated motion. The approach integrates
shared-autonomy teleoperation to collect demonstrations, encoding end-effector
trajectories in a target-relative frame. During execution, the target-referred
DMPs adapt these movements to new object poses, ensuring correct alignment for
bimanual tasks. Experimental validation on a robotic system demonstrates that
the methodology outperforms standard DMPs, particularly on tasks requiring
precise alignment (e.g., picking and placing parcels, turning a valve).

In robot motion planning, learning a complex task with long sequences of motion
is a major problem. There is also a need to handle uncertainty in the motion
model. \cite{kong2023dynamic} tackle these issues by modifying DMPs and applying
them to complex trajectories. Concretely, the solution consists of dynamically
joining trajectories using overlapping kernels. This is achieved via a method
that can produce a smooth target trajectory with high accuracy in both the
position and the velocity profiles. Moreover, a neural network-based controller
is designed to approximate uncertainties and ensure both the tracking accuracy
and imitation performance of the closed-loop system. The joining of multiple
DMPs and the reliability of the robot skill learning framework are evaluated on
pouring tasks.

DMPs can be expanded using neural networks to learn longer, more complex skills.
Yet, the complexity and length of a skill may have trade-offs in terms of
accuracy, task flexibility, and demonstration time. Based on this observation,
\cite{hanks2024comparing} compare the ability of neural DMPs to learn a skill
from full demonstrations to those that learn from sub-tasks. Specifically, the
accuracy of neural DMPs is analyzed and the key trade-offs between each approach
are examined. The findings show that both techniques are successful in
completing a pouring task. Only 24 demonstrations are required to learn and
generalize to new configurations of the task not seen during training.
Furthermore, the models trained using sub-tasks are more accurate and have more
task flexibility. Nonetheless, models trained on sub-tasks require a larger
investment from the human expert.

\cite{liu2024robot} design a robot skill learning system that incorporates
multi-space fusion by simultaneously considering motion/stiffness generation
and trajectory tracking. To do this, surface EMG signals from a human's arm are
captured to estimate endpoint stiffness. Then, GPR is combined with DMPs to
extract skill features from multiple demonstrations. The DMP formulation is
based on the $S^3$ Riemannian metric to allow learning, reproducing, and
generalizing the orientation properties of the robot represented by
quaternions.  Additionally, a controller is designed to track the trajectory
generated by the motion model and to reflect the learned stiffness
characteristics via an RBFNN that compensates for the uncertainty of the robot
dynamics. Experimental results with a 7-DoF robot executing a water pouring
task show the effectiveness of the method in performing contact tasks. 

Learning transferable tool-use skills is essential for improving flexibility in
fine robot manipulation tasks. In light of this importance,
\cite{lu2025dynamic} create a framework that enables generalized manipulation
skills across different tools and scenarios. To do this, the approach
introduces two DMP-based skills: an object operating skill that handles
tool-object interactions under environmental constraints, and a tool flipping
skill that permits changing contact positions between tools and objects while
maintaining motion continuity. A constraint-irrelevant skill technique
separates the transferable core of a demonstration from tool-specific
limitations. Using a 7-DoF robot, experiments on pushing, cutting, pouring, and
obstacle avoidance demonstrate smooth, conflict-free, and generalizable
tool-use trajectories, outperforming conventional DMP and rapidly exploring
random tree (RRT$^\ast$) methods in continuity, efficiency, and adaptability.

\paragraph{Grinding/polishing/sawing/smoothing/wiping.} 
Robotic tasks that require contact with surfaces (e.g., wiping a table) need
ways to handle contact forces appropriately. \cite{gams2010line} propose a
two-layered learning framework for such tasks. The first layer extracts the
frequency and waveform of periodic motions from visual demonstrations using
DMPs, while the second layer modifies these trajectories using LWR based on
force feedback. The adapted trajectories enable a robot to maintain both proper
contact forces and desired periodic movements when interacting with surfaces.
The approach uses active compliance control combined with force feedback to
safely adapt motions. To validate the system, a humanoid robot is tasked with
performing wiping motions across both flat and curved surfaces. The results show
that the robot learns the appropriate height adjustments to maintain consistent
sponge contact despite surface variations.

While robots can perform wiping tasks with predefined objects, adapting these
motions to wipe with new cleaning tools requires extensive manual parameter
tuning. \cite{do2014learn} tackle this issue via an iterative system that
explores and models wiping motions encoded as DMPs based on object properties as
follows. First, the system measures each object's softness through controlled
deformation tests and height through kinematic sensing. Then, support vector
regression is used to learn mappings from these properties to appropriate DMP
amplitude parameters. Lastly, the learned models enable prediction of suitable
amplitudes for new objects, which are validated through physical wiping
experiments using force-based adaptation rules. The method is validated on a
humanoid robot with 12 different objects where successful adaptation is
displayed with cubic objects, but limitations are revealed with cylindrical and
spherical shapes.

\cite{chebotar2014learning} present an approach for incorporating tactile
feedback into robot manipulation tasks through two key components. First,
computer vision algorithms are adapted for in-hand object localization using
tactile sensors and compared against probabilistic hierarchical object
representations, iterative closest point, and voting schemes. Second, DMPs are
modified to incorporate tactile feedback by learning desired trajectories from
demonstrations and RL to optimize the parameters. The framework reduces the
high-dimensional tactile feedback space through spectral clustering and PCA
before performing policy optimization. Experiments are conducted on a robot
manipulator carrying out a scraping task, which successfully adapts to both
elevated surfaces and ramps while maintaining the desired contact forces.

Physical human intervention provides a natural way to correct a robot's behavior
during task execution. To this end, \cite{gams2014learning} present two
complementary methods for adapting periodic DMPs: one maintains desired contact
forces with surfaces while the other enables trajectory refinement through human
coaching. Rather than using admittance control to modify reference trajectories,
the method directly updates DMP weights using force errors in the regression.
When a human physically pushes on the robot, these interaction forces are
treated as error signals to update specific portions of the learned trajectory.
The system is validated on wiping tasks where a robot maintains surface contact
while allowing trajectory modifications through both physical interaction and
visual gesture coaching. The direct weight modification approach eliminates the
delays inherent in traditional admittance control methods such as those
introduced via \cite{gams2010line}. Additional results on DMP adaptation methods
are discussed by \cite{gams2016adaptation}.

For robots to cooperate and physically interact with humans in a wide range of
settings they must be equipped with force-control skills. To enable this
functionality, \cite{peternel2014teaching} develop a multimodal LfD framework
based on autonomous controllers for highly-dynamic tasks involving cooperation
between humans and robots. The approach uses selected EMG muscle activation
signals as a human interface for controlling a robot's end-effector stiffness.
DMPs are utilized to encode the motion of the end effector and stiffness
parameters. Force feedback allows for modulation of the robot's compliance in
order to successfully cooperate with a human. The methodology is demonstrated on
a sawing task in which compliance needs to be carefully modulated according to a
human partner's actions.

In human manufacturing, the manipulation of materials is accomplished through
the use of handheld tools. However, the remapping of tool-mediated human skills
onto robots involves two main challenges: (i) encoding of human procedural
knowledge, (ii) handling the distribution of force requirements in space and
time across a physical interface. \cite{montebelli2015handing} address these
issues through an articulated coordination of the distribution of forces
necessary for wood planning. This is achieved by capturing the interaction
between the human teacher and a robot arm equipped with a force/torque sensor
using DMPs. At reenactment time, the system can mimic and discern the
demonstrated spatial requirements. Two experimental conditions are compared: (i)
reenactment where the tasks relying on the input are calculated by the DMP
model; (ii) reenactment with a fixed orientation whereby the orientation of the
tool is parallel to the plank and made maximally stiff.

An essential part of human-robot collaboration is the ability of the robot to
change its behavior according to the intention and states of the human partner.
To this end, \cite{peternel2016adaptation} present an approach for robot
adaptation to human fatigue that employs DMPs, LWR, and adaptive frequency
oscillators. Initially, the robot imitates the human to perform the task in a
leader-follower setting via human motor behavior feedback. At the same time, the
robot learns the skill online. When a predetermined level of human
fatigue is detected, the robot then uses the learned skill to take over the
physically demanding aspect of the task while the human controls and supervises
the high-level interaction. The method is demonstrated on co-manipulation tasks
(e.g., sawing). In \cite{peternel2018robot}, the human fatigue model is extended
to include recovery when the effort is reduced.

Standard DMPs cannot effectively handle interactions between multiple agents in
collaborative tasks. \cite{zhou2016learning} address this issue by introducing
coordinate change DMPs to represent relationships between leaders and followers
directly in task space. To do this, local movements are learned relative to the
leader's coordinate system. Specifically, the framework adds a coupling term
that enables learning of actions in complex environments and allows gradual
refinement based on sensory feedback. Through a force-based controller, the
coupling term gradually adjusts to maintain desired contact forces during
execution. The approach is demonstrated on a wiping task where a humanoid robot
adapts both its wiping pattern and force profile to moving surfaces.
Experimental results show successful adaptation to different surface
orientations while maintaining specified contact forces.

\cite{peternel2017method} develop a method that enables a robot to devise an
appropriate control strategy from demonstrations without prior knowledge of the
task as follows. First, data is collected using a motion capture system, force
sensor, and muscle activity measurements. The variables (position and force) in
the captured dataset are then segmented and analyzed for each axis of the
observed task frame separately where DMPs are used to encode the nominal
trajectory. Next, the appropriate controller (i.e., position or force) is
delegated to each axis of the task frame. In the final stage, checks are made
for a correlation between the variables and muscle activity patterns to
determine the desired stiffness behavior. The robot then uses the control
strategies for autonomous operations (e.g., wiping, sawing, drilling) via a
hybrid force/impedance controller.

Neural-based MP approaches integrate neural network architectures to learn MP
parameters from high-dimensional data (e.g., images). For instance,
\cite{pervez2017learning} utilize CNNs to learn deep MPs from visual input for
real-time human movement imitation. This is done by combining the generalization
capabilities of TP-DMPs with the feature learning ability of CNNs, eliminating
the need for manually extracting task-specific parameters by directly utilizing
images from cameras during motion reproduction. By using the proposed approach,
not only is there no need to formulate a task as an optimization problem, but
also only a few demonstrations are required. The system is evaluated with a
trash cleaning/sweeping task on a physical robot.

\cite{peternel2017robots} explore the concept of unskilled robots
collaboratively learning from skilled robots. The key motivation is that human
involvement can be reduced and the skill can be propagated faster among robots
performing similar collaborative tasks. The methodology is based on a
multi-stage learning process that can gradually learn the appropriate motion and
impedance behavior under the given task conditions. Trajectories are encoded as
DMPs, which are learned using LWR, while the phase is estimated by adaptive
oscillators. The learned trajectories are replicated via a hybrid
force/impedance controller. The framework is evaluated by performing a
collaborative sawing task (brick and wood) using two robots.

Although past work has provided solutions for surface wiping tasks, only a few
have dealt with the energy transfer between the robot and environment while
maintaining system stability. In this scenario, ensuring stability is closely
related to the concept of passivity, which is when the system does not produce
more energy than it receives. To this end, \cite{shahriari2017adapting}
introduce a methodology to encode trajectories as periodic DMPs in
impedance-controlled robots. This is accomplished by first designing a passivity
observer based on the input power ports and a demonstrated reference power.
Then, a DMP phase altering law is created to adjust the phase according to the
passivity criterion. Experimental results are carried out using an
impedance-controlled robot arm on surface polishing tasks. 

Robotic applications that involve direct human-robot physical contact (e.g.,
showering) are much more demanding in terms of safety than other assistive
robotic systems. In these scenarios, unexpected human motion may occur during
the operation of the robot. Therefore, the safety standards may include
operation of the robot on deformable human body parts in dynamic environments.
\cite{dometios2018vision} address these safety issues with a robotic
vision-based washing system. To do this, CCDMPs are combined with a
perception-based controller that adapts the motion of the robot end effector on
moving and deformable surfaces. Furthermore, the controller guarantees uniformly
asymptotic convergence to the leader MP, while ensuring avoidance of sensitive
regions. Experiments are conducted with a humanoid robot that applies washing
actions on the back region of a subject.

Many tasks (e.g., wiping, scrubbing, mixing) require motion along a
surface while maintaining a desired force. To this end,
\cite{conkey2019learning} design a method for learning hybrid force/position
control from demonstrations. Concretely, a dynamic constraint frame aligned to
the direction of the desired force is learned using CDMPs. In contrast to a
fixed constraint frame, the approach accommodates tasks with changing task
constraints over time. Furthermore, only one DoF for force control is activated
at any given time, ensuring that motion orthogonal to the direction of desired
force is possible. Additionally, the CDMP framework is extended to encourage
robust transition from free-space to constrained motion. Experiments using a
robot arm show active compensation for frictional forces between the end
effector and contact surface when performing sliding motions.

To enrich a repertoire of movements and make it easier for nontechnical users to
program a robot by demonstration, \cite{li2020pattern} develop a methodology
to obtain optimal DMP parameters and pattern information (i.e., discrete or
rhythmic) for unknown trajectories. Specifically, the DMP parameters of both
discrete and rhythmic dynamical systems are found using Bayesian optimization
where the correct pattern repeats the goal trajectory with the least amount of
deviation. Furthermore, virtual generalized trajectory behaviors are computed to
ensure reliability of the learned model in novel situations (e.g., new goal
positions in the discrete model or amplitudes in the rhythmic model). The
framework is demonstrated on whiteboard wiping and box stacking tasks using a
robot manipulator.

A robot could efficiently learn the underlying motion of a periodic task, and
repeat it as many times as necessary, if it is able to decompose the
demonstration of the task into periods. Based on this observation,
\cite{yang2022learning} explore learning periodic tasks from human visual
demonstrations. To do this, active learning is employed to optimize the
parameters of rhythmic DMPs. An objective function maximizes the similarity
between the motion of objects manipulated by a robot and the desired motion in
the video demonstrations. Bayesian optimization is used to develop a
sample-efficient learning algorithm and unsupervised keypoint model for
computing scores. The system is applied to deformable objects including winding
cables, stirring granular matter with a spoon, and wiping surfaces with a cloth.

To obtain high trajectory accuracy and shape invariance for robot machining
applications, \cite{zhou2022combination} propose a framework based on
combinations of DMPs. This is accomplished by first demonstrating predesigned
machining trajectories. After the trajectories are analyzed and segmented, the
features of the segmentation points are summarized to build a node mapping
policy. The policy helps with accurately restoring the geometric shape of the
demonstration trajectories when changes to the machining area occur. Lastly,
low-order DMPs are connected end-to-end via time modulation and a shared
canonical system. Experiments are conducted to validate the performance of the
robotic system on metal grinding jobs.

\paragraph{Flipping/hitting/throwing.} 
Traditional representations of motor behaviors in robotics are based on desired
trajectories generated from spline interpolations between points. However,
control policies generated from a spline trajectory tracking controller have a
variety of disadvantages such as a lack of robustness and generalization.
\cite{peters2006reinforcement} explore how to mitigate these concerns via
different RL approaches for improving the performance of DMPs. Specifically,
various methods for policy gradient estimation are implemented along with their
applicability to RL for DMPs. In particular, three major ways of estimating
first-order gradients (finite-difference gradients, vanilla-policy gradients,
and natural-policy gradients) are investigated. The methods are applied to motor
skill learning using a 7-DoF anthropomorphic robot arm on bat swinging tasks. 

The need to learn perceptual coupling for MPs has long been recognized as an
important endeavor. Nonetheless, learning perceptual coupling to an external
variable is not straightforward. In fact, it requires numerous trials in order
to properly determine the connections from external to internal focus. To solve
this issue, \cite{kober2008learning} develop an augmented version of a DMP that
incorporates perceptual coupling to an external variable using RL. Concretely,
it relies on an initialization through IL and subsequent self-improvement by RL
via the PoWER method. The resulting framework is demonstrated on learning a
ball-in-a-cup task with a simulated anthropomorphic robot arm. The work is
continued by \cite{kober2009learning} where the standard DMP formulation is used
and results are presented on a physical robot arm.

Learning the ball-in-a-cup game presents a challenging control problem due to
the continuous interaction between ball and cup dynamics.
\cite{kober2010imitation} propose a solution to the game by augmenting DMPs with
a perceptual coupling term that enables adaptation to external variables (e.g.,
ball position and velocity). Policies from human demonstrations are initialized
via RL using the PoWER algorithm to optimize both movement and coupling
parameters. The method successfully learns to catch the ball despite large
variations in initial conditions and it matches the learning speed of a human
child by converging after 600--800 trials. Furthermore, the results show that
perceptually-coupled DMPs can handle more perturbations than both uncoupled and
hand-tuned coupled versions, while maintaining good performance even when
perception and action occur in different spaces.

Despite the versatility of the standard DMP framework, it struggles to modify
learned movements for achieving specific target velocities at predetermined
times. To bridge this gap, \cite{kober2010movement} reformulate DMPs to allow
for a designated hitting point and velocity without compromising the framework's
inherent strengths. This enhancement lets a robot change the target velocity of
the movement while maintaining the overall duration and shape. Additionally, a
modification to overcome the issue of an initial acceleration step, which is
important for the safe generalization of learned movements, is presented. Using
a robot arm and four-camera system, the movement template is evaluated on two
table tennis situations: (i) obtaining a desired velocity by hitting a ball on a
string with a racket, and (ii) executing precise forehand returns.

A method based on a mixture of dynamical systems for learning the couplings
across multiple motor control variables is developed by
\cite{kormushev2010robot}. To encapsulate correlation information, a
demonstrated skill is first encoded in a compact form through a modified DMP
system. Next, an EM RL algorithm is used to modulate the mixture of dynamical
systems initialized from the user's demonstration. This process refines the
coordination matrices and attractor vectors associated with the set of
primitives. Not only does this highlight the advantages of considering
probabilistic approaches in RL, but it also shows the significance of applying
importance sampling to reduce the number of rollouts required when learning a
new skill. Two skill-learning experiments are conducted with a 7-DoF robot arm:
a reaching task where the robot needs to adapt the learned movement to avoid an
obstacle, and a dynamic pancake-flipping task. 

\cite{nemec2010learning} also explore the application of DMPs to the ball-in-cup
game. To teach this task to a robot manipulator, they evaluate two approaches:
(i) IL using DMPs, and (ii) RL using state-action-reward-state-action with
eligibility traces. The RL approach requires no prior knowledge and learns
successful swing-up motions in 220-300 simulation rollouts plus 40-90 real robot
rollouts, using cup position, velocity, and ball angle as state variables. Both
methods successfully enable the robot to swing up and catch the ball, although
they generate notably different motion patterns. For instance, the DMP-based
approach mirrors the human's single-swing strategy while the RL method discovers
a novel two-swing motion within the robot's acceleration constraints.

\cite{pastor2011skill} present a combination of LfD with RL to enable robust
manipulation skills while incorporating sensor feedback for failure prediction.
The methodology bootstraps skills through demonstration, refines them using
PI$^2$, and builds predictive models from sensor statistics to anticipate task
outcomes. Two tasks with a robot manipulator are used to validate the approach:
(i) a pool stroke that maximizes ball speed while maintaining accuracy, and (ii)
a dexterous manipulation task involving chopstick-based box flipping. For the
box-flipping task, the system learns to predict failures by monitoring
deviations in sensor signals to enable early detection of unsuccessful
executions with high accuracy. It requires minimal task-specific engineering,
primarily in defining appropriate coordinate frames and cost functions that
capture the success criteria.

Adapting DMPs to new situations typically requires relearning the entire
movement rather than adjusting its parameters. To mitigate this overhead,
\cite{kober2012reinforcement} propose to learn a mapping from situations to DMP
parameters via cost-regularized kernel regression. Concretely, an RL approach to
incorporate costs as uncertainty weights is used to guide exploration. This
enables meta-parameters (e.g., goal positions, movement durations, and
velocities) to be adjusted, while preserving the underlying movement shape. The
methodology is extensively validated on the following tasks with various robot
manipulators: (i) dart throwing, (ii) table tennis, and (iii) ball-throwing
games. An experimental evaluation shows that the technique can efficiently adapt
movements to new situations with significantly fewer trials compared to previous
policy-search methods.

Many robot skill learning methods focus on a single solution for a given task,
which limits their autonomy. Based on this observation,
\cite{daniel2012learning,daniel2012hierarchical} develop a framework for
learning multiple solutions for one task. Moreover, the system can choose from
these solutions for a given situation. The approach learns two levels of a
hierarchical policy representation. A high-level gating policy selects the
specific solution and an action-policy subsequently defines the movement to be
executed by the robot. Specifically, a gating network that selects between DMPs
based on the current context and the policies of the primitives that specify the
robot's actual actions, is simultaneously learned. Experimental results show
that the system can learn multiple solutions for a game of tetherball. In
\cite{daniel2013learning}, the policy search method is updated to learn
sequencing of multiple primitives while concurrently improving the individual
primitives. The framework is broadened by \cite{daniel2016hierarchical} to take
advantage of structured environments by identifying multiple subpolicies for a
given task.

Model-based RL policy search has limitations that can impede learning
generalized robot skills. For example, it assumes a specific structure of the
reward function as well as the lower-level policy. \cite{kupcsik2013data} relax
these assumptions and develop GP relative entropy policy search, a model-based
policy search algorithm that generalizes the lower-level policy parameters over
multiple contexts. Concretely, it exploits learned probabilistic forward models
of the robot and its environment to predict expected rewards of artificially
generated data points. Simulated and real robot arm experiments (hockey and
table tennis) highlight a reduction in the required amount of measurement data
to learn high-quality policies compared to model-free contextual policy search
approaches. The framework is further extended by \cite{kupcsik2017model}.

Acquiring motor skills via RL can provide robots with the ability to solve a
wide range of complex tasks. Nevertheless, comparing the effectiveness of RL
against human programming is not straightforward. To address this challenge,
\cite{parisi2015reinforcement} create a motor skill learning framework and
compare it to a manually designed program on the task of playing tetherball.
DMPs are used to represent the robot's trajectories and relative entropy policy
search is employed to train the motor skill learning system and improve its
behavior by trial and error. An experimental evaluation shows that although the
learning modules require parameter tuning, doing so is less time intensive than
building a mathematical model of the given task. Furthermore, the learned player
is able to outperform the handcrafted opponent.

\cite{huang2016jointly} develop a combined learning framework for a table tennis
robot. Instead of predicting a single striking point, a trajectory prediction
map is constructed to forecast the ball's entire rebound trajectory using its
initial state as follows. First, a robot trajectory generation map is created to
predict the joint movement pattern and the movement duration using the Cartesian
racket trajectories without the need for inverse kinematics. Then, a correlation
function is used to adapt the joint movement parameters according to the ball
flight trajectory. After obtaining the joint movement parameters, the joint
trajectories are generated using DMPs. Lastly, RL is employed to modify the
joint trajectories such that the robot can return balls well. The framework is
tested in both simulation and on a real system using 7-DoF robot arm.

\cite{lundell2017generalizing} design a parametric LfD approach that employs a
linear basis function model with global nonlinear basis functions to generalize
an imitated task to unseen situations as follows. First, the kinematics of an
initial demonstration are encoded as a DMP. Next, the shape parameters of the
DMP are optimized using the PoWER method to adapt the skill to new
circumstances. Then, the training data is utilized to build a global parametric
model of the skill. Finally, the global model is used for generalizing the model
parameters (e.g., DMP shape parameters) to new task parameters without
relearning the generalized model. A ball-in-a-cup task is performed to assess
how effective the method is in generalizing from an initial demonstration with a
fixed string length to variable lengths.

Efficient RL requires the determination of an appropriate reward function, which
is a difficult problem even for domain experts. To make progress on this issue,
\cite{pahic2018user,lonvcarevic2019learning} investigate if the PoWER algorithm
can be effectively utilized with a simple, qualitatively determined reward,
instead of a complex reward function. The baseline for the RL comparison is
policy search on DMPs using an expert-defined reward. Concretely, policy search
is applied using different reward functions in two spaces: the configuration
space of the robot and the latent space of a deep autoencoder network.
Experimental results show that RL, with discrete user feedback, can be
effectively applied for robot learning. In follow-up work,
\cite{lonvcarevic2020reduction} explore how much data is needed for the faithful
representation of an action. To reduce dimensionality,
\cite{lonvcarevic2020generalization,lonvcarevic2021generalization} compare
methods for obtaining a database of trajectories to train deep autoencoder
networks. \cite{lonvcarevic2021accelerating} combine the autoencoder-reduced
learning space and non-continual training of DNNs to provide a priori knowledge
with target-to-action mapping. 

\cite{hazara2017model} demonstrate that global parametric DMPs can be learned
incrementally, improve sample efficiency, and generalize to new variations of
the ball-in-a-cup task. Furthermore, a method for choosing optimal model
complexity on fewer samples is introduced. Experimental evaluations validate the
use of online incremental learning, showing that traditional selection criteria
need more training samples. \cite{hazara2018speeding} expand upon this research
via an empirical Bayes method to extract the uncertainty of predicted global
parametric DMP motions, guiding the incremental learner's exploration and
leading to improved learning speeds. In follow-up work,
\cite{hazara2019transferring} propose a dynamics-agnostic sample-efficient
approach to transfer global parametric DMPs to the real world. Global parametric
DMPs are employed by \cite{hazara2019active} to target multiple variations
(i.e., string lengths) of the ball-in-a-cup task. Specifically, a contextual
skill model utilizing active incremental learning is used to learn a DMP
database where task variations are ordered according to reward improvement.  

Table tennis requires complex interactive contact manipulation skills that can
handle uncertainty in the ball position and velocity. To mitigate this
uncertainty, \cite{prakash2019dynamic} propose a DMP-based framework for the
stable generation of dynamic trajectories as follows. First, DMP parameters are
learned from expert demonstrations. Next, multiple DMPs are combined using a
piecewise linear system to achieve stability under  uncertainty. The tracking
performance is also enhanced using a learning technique based on Lyapunov
stability. Lastly, a fuzzy fractional order sliding mode control is designed to
enhance the DMPs and enable interception of a moving ball. An evaluation on a
4-DoF robot arm displays the system's ability to smoothly adapt to a ball's
position in real-time.

Robot learning is often performed in simulation to expedite real-world skill
learning via simulation-to-reality transfer learning. For example,
\cite{lonvcarcvic2021accelerated} highlight the possibility of improving the
performance of transfer learning through RL. A DNN is utilized to extract the
coordinates of the target, which are then fed as input into a second DNN that
maps the target coordinates to policy parameters. The motion policy is encoded
as a DMP, i.e., an end-to-end solution takes as input an image and provides the
DMP parameters as the output. An experimental evaluation is conducted using a
7-DoF robot arm equipped with a ball-throwing spoon. In
\cite{lonvcarevic2022combining}, the performance of the method is improved by
adapting only one layer of the DNN model and demonstrated using a more
complicated (humanoid) platform.

A big problem when applying RL to robotics is that many of the generated
trajectories are not executable because of either the physical limitations of
the robot or due to issues with calculating inverse kinematics. To address this
limitation, \cite{lonvcarevic2022fitting} develop a method for throwing actions
that positions a trajectory in the constrained space of the robot in a way that
allows for the highest number of executable trajectories that are similar to the
demonstrated one. Trajectories are encoded in the form of CDMPs and stored in a
database. With a database of trajectories, the approximate best angle for
performing the throw is computed and compared using three different approaches:
an evolutionary algorithm, GPR, and a neural network. The technique is evaluated
with a humanoid robot.

\cite{huang2023toward} tackle the challenge of dual-arm cooperative flipping
manipulation, a contact-rich industrial task requiring synchronized motion and
orientation coupling between robot arms. The proposed framework builds upon
conventional DMPs by maintaining inter-arm motion dependencies at the object
level, ensuring consistent coordination during learning and generalization. A
dual-leader-dual-follower teleoperation system is used to collect demonstration
data, and orientation generalization is achieved through an additional
formulation for end effector pose control. Experiments conducted on a 7-DoF
robot demonstrate that the approach preserves contact, minimizes squeezing
forces, and generalizes flipping trajectories to objects of different sizes and
shapes, outperforming standard DMPs in smoothness and constraint preservation.

Nonprehensile manipulation consists of manipulating an object without a
form/force closure grasp, and it is common in many applications (e.g., shooting
a basketball, flipping a pancake, etc.). To enable this manipulation capability,
\cite{sun2023integrating} combine LfD and RL to learn a controller for complex
motor skills. Specifically, DMPs are first employed for encapsulating force
information by compactly encoding the demonstrated skills. Then, RL is utilized
for self-improvement to master the motor skills. In particular, an optimal
replay buffer technique is applied to reduce the number of rollouts, thus
ensuring an improvement for each trial. Using a 7-DoF robot, the methodology is
evaluated on a hockey task where the goal is to hit the puck into the highest
reward region.

\paragraph{Surgery.} 
Standard DMPs face challenges in generalizing demonstrations to new environments
and handling noisy human inputs. To address these limitations,
\cite{ghalamzan2015incremental} propose a three-tiered learning approach
inspired by human observational learning: mimicking, imitation, and emulation.
Concretely, the method uses (i) GMM/GMR to compute a noise-free average path
from demonstrations (mimicking), then (ii) employs DMPs to generalize this path
to new start/goal positions (imitation), and finally (iii) applies inverse
optimal control to learn a reward function that combines following the nominal
path with appropriate environmental responses (emulation). The methodology is
validated on surgical pick-and-place and sweeping tasks. Experimental results
show the robot's ability to adapt to new goal positions and static obstacle
configurations despite suboptimal demonstrations.

Addressing operator fatigue and cognitive workloads remains an open problem in
the field of robot-assisted endovascular intervention. To this end,
\cite{chi2018trajectory} propose an LfD RL-based DMP framework to enhance
catheterization tasks by minimizing unwanted contacts between the catheter tip
and the vessel wall. The learning-based methodology leverages a customized
robotic manipulator while adapting to different flow simulations, vascular
models, and catheterization tasks. In particular, the method directly addresses
the complex nonlinear flow dynamics inside the vasculature by incorporating the
PI$^2$ algorithm. The experiments demonstrate shorter trajectory lengths and
fewer contacts with the vessel wall, which can potentially reduce risks in
endothelial wall damage, embolization, and stroke.

Autonomous robotic surgery involves challenging and complex tasks such as a
real-time understanding of the surgical workflow, the execution of safe and
precise movements, and reactive decision making during unexpected and faulty
events. Within this context, \cite{ginesi2019knowledge} design a modular
framework with hierarchical reasoning for the automation of structured tasks. At
the high level, multiple actions are coordinated via an ontology that encodes
prior knowledge as rules and verifies preconditions for the execution of
actions. DMPs are implemented at the lower level for motion planning.
Furthermore, DMPs are used during replanning to compensate for failure events in
surgical scenarios both at the task and motion planning levels. The system is
tested on a peg-and-ring task using a robot manipulator.

Precision cutting tasks are regularly encountered in surgery, yet the design of
effective learning and adaptive control strategies for these applications
remains an open problem in robotics. \cite{straivzys2020surfing} address the
challenge of cutting along the boundary between two soft mediums, a problem that
is very difficult due to visibility constraints. To do this, a strategy is
developed using a binary medium classifier trained on joint-torque measurements.
In addition, a closed loop control law that relies on an error signal compactly
encoded in the decision boundary of the classifier is formulated. Using a mobile
manipulator, the system is demonstrated on a grapefruit cutting task by
modulating a nominal trajectory fitted using DMPs to follow the boundary between
the grapefruit pulp and peel. 

Developing autonomous robots that can perform specific surgical operations
(e.g., knotting, suturing, etc.) can not only decrease the length of these
procedures, but also reduce surgeon fatigue and increase accuracy. To this end,
\cite{su2020reinforcement} facilitate robot-assisted minimally invasive surgery
by transferring manipulation skills through RL. A GMM and GMR are used to model
high-dimensional manipulation skills obtained by human demonstrations using
DMPs. Further building upon this work, \cite{su2021toward} propose a method for
surgical teaching via demonstration by integrating cognitive learning and
control techniques. This allows a robot to learn a senior surgeon's skills. The
methodologies are demonstrated in a surgical setting with a 3D-printed human
abdomen and robot arm.

Multi-arm surgical robots require operators to coordinate several tools
simultaneously through complex motion sequences, imposing significant cognitive
demands. To alleviate this burden, \cite{denivsa2021semi} propose a
semi-autonomous framework in which the operator controls a single leader arm,
while a follower arm is automatically coordinated. The system integrates a
hierarchical database of demonstrated leader trajectories that encode the
corresponding follower motions (\cite{denivsa2015synthesis}). During execution,
a sliding-window search through the database identifies the leader's intended
motion in real time, and synchronized follower trajectories are generated via
DMPs with velocity adaptation. The methodology is evaluated on a bimanual
peg-transfer task using a dual-arm surgical robot, demonstrating effective
coordination and reduced operator workload.

Needle manipulation actions can be learned from a single human demonstration and
then used to autonomously execute part of a surgical task. For instance,
\cite{schwaner2021autonomous} develop an LfD framework that exploits existing
expert knowledge from an action library of surgical actions. In particular,
demonstrations are used to learn needle manipulation skills that compose parts
of a surgical task. Specifically, each action is learned from a single
demonstration, where DMPs encode the Cartesian space trajectories. To validate
the method, experiments are conducted on a surgical suturing task. The results
achieve an 81\% success rate in previously unseen, non-clinical settings with a
mean needle insertion error of 3.8 mm.

Compared to traditional laparoscopic surgery, robot-assisted minimally invasive
surgery has advantages for both patients and doctors. Nonetheless, the workload
experienced by surgeons is very high and there is a lot of repetition in many
surgical procedures (e.g., suturing, dissection, etc.). To this end,
\cite{iturrate2023handheld} present an LfD system for surgical tasks based on a
handheld forceps tracker and DMP-encoded motions. The framework provides a way
to collect surgical skill demonstrations, without access to a surgical robot,
but easily transfers to one. The system is evaluated on a robot where recorded
teleoperated trajectories are compared against trajectories generated by DMPs to
show generalizable representations of surgical skills

Although DMPs are well suited to learning manipulation skills from
demonstration, their reactive nature restricts their applicability for tool use
and object manipulation tasks involving nonholonomic constraints such as scalpel
cutting or catheter steering. \cite{straivzys2023learning} address this
limitation by constraining DMPs through an additional coupling term derived from
an analytical solution to the Udwadia-Kalaba method. This is done by first
encoding the pose trajectory using two uncoupled DMPs: a position DMP and an
orientation DMP. Then, the position DMP is extended with an analytically-derived
coupling term that imposes a nonholonomic equality constraint. Finally, the
orientation DMP is optimized to minimize the constraint coupling term. The
experimental evaluation consists learning nonholonomic DMPs from demonstration
for elliptical tissue excision.

Autonomous robots can provide surgeons with numerous benefits including a
reduction in cognitive load when performing repetitive and complex surgical
tasks. For example, \cite{zhang2023step} focus on automating robotic
appendectomy, a minimally invasive surgical procedure that is relatively less
complicated compared to other operations. The robotic motions to automate the
appendectomy are generated via LfD and DMPs using two patient-side manipulators.
The surgeon first has to identify the head and root of the appendix in a
laparoscopic image as the target. Then, after hand-eye calibration, the image
coordinates are converted to 3D spatial coordinates. Finally, DMPs allow the
endpoint of the trajectories to be adapted to the desired position identified by
the surgeon. The method is evaluated on comparative experiments utilizing both a
simulator and a surgical robot platform.

Policy learning in robot-assisted surgery lacks data efficient techniques that
exhibit the desired motion quality for delicate surgical interventions. To
tackle this problem, \cite{scheikl2024movement} introduce an IL framework that
focuses on the manipulation of deformable objects (e.g., tissue) for surgical
applications. Concretely, the approach combines the versatility of
diffusion-based IL with the high-quality motion generation capabilities of
ProDMPs. This enables the gentle manipulation of deformable objects while
maintaining data efficiency. Furthermore, the use of ProDMPs allows for
generating smooth high-frequency action sequences with guaranteed initial
conditions, without an increase in inference time. The framework is evaluated on
simulated and real-world robotic tasks using both state and image observations.

\paragraph{Painting/writing.} 
Internal models are a set of policies that describe the action mechanisms
between external input, internal states, and output actions. They play a key
role in movement planning and execution. Furthermore, under the hypothesis that
any human motion can be decomposed into DMPs, a set of second-order differential
equations can be utilized as the internal model to describe primitive movements.
To this end, \cite{xu2004multiple} extend the results of
\cite{ijspeert2002learningb} to multiple internal models that learn and describe
complex motion behaviors. This allows all movements to be classified into
sequential and parallel DMPs based on the task decomposition. An experimental
evaluation employs a 2-DoF robot arm to mimic human limb behavior when
performing letter writing movements.

Without any prior knowledge, it is extremely hard for robots to generate
human-like behaviors in dynamic environments. Based on this observation,
\cite{tan2012robots} create an LfD methodology for robots to learn motions and
corresponding semantic knowledge simultaneously. The goal is to find features of
the motions in correspondence with the inner features of behavior that are
strongly related to the semantic models. To learn the required movements, a
modified nonlinear dimensionality reduction algorithm (Isomap) is first utilized
to convert sampled 6D vectors of joint angles into 2D trajectories. A GP method
then represents the learned motion and knowledge models in a 2D latent space.
The framework is demonstrated via a humanoid robot that learns how to write
numbers and automatically relate the motions of writing digits to their semantic
knowledge models.

The use of DMPs for simultaneous kinesthetic teaching of positional and force
requirements for in-contact tasks (e.g., writing) is highlighted by
\cite{steinmetz2015simultaneous}. This is done using a specific sensor
configuration, namely a force/torque sensor is mounted between the tool and the
ﬂange of a robot arm with integrated torque sensors at each joint. Next, human
demonstrations are modeled using DMPs. After the demonstrations, a robot arm is
provided with the capacity to perform sequential in-contact tasks. During
reenactment of the task, the system can imitate and generalize from both the
demonstrated trajectories and the associated force profiles. The methodology is
evaluated on writing a sequence of demonstrated characters using a 7-DoF robot.

Having a natural and friendly way to interface with a robot is an important
aspect in social HRI situations. To this end, \cite{li2018enhanced} develop an
LfD interface that combines a GMM, GMR, and DTW with DMPs as follows. First,
trajectory data is collected by teleoperating a robot manipulator and clustering
the recorded trajectories via k-means. Next, DTW is employed to align the
trajectories. The generalization of the motions is achieved by using DMPs where
the GMM is used to estimate the parameters (e.g., prior probability,
expectations, and variance). Finally, GMR is applied to generate a synthesized
trajectory with reduced position errors. The methodology is evaluated using
robot arms on two tasks: obstacle avoidance and writing.

DNNs have the capability to perform highly nonlinear transformations between
input and output data, and learn direct perception-action couplings. Taking
advantage of this ability, \cite{pahic2018deep} propose a deep encoder-decoder
network to map raw image pixels to DMP parameters for robot manipulation tasks.
The approach leverages a dataset of images paired with an associated trajectory
to train a DNN that predicts the DMP parameters directly from an image.
Experiments are performed using synthetically generated datasets and MNIST to
reproduce handwriting movements from raw images. These movements are also
practically demonstrated through the execution of writing tasks by a robotic
arm. In proceeding work, \cite{ridge2019learning} improve the architecture by
including pretrained CNN layers in the image encoder. 

\cite{luo2020generalized} introduce a handwriting LfD system that enables a
robot to learn from handwritten examples to draw alphanumeric characters. The
approach is based on studying how Gaussian kernel shape and number influence
learning performance as follows. First, to determine how kernel width influences
learning performance, the learned trajectories of five letters (a, B, D, e, M)
are explored based on Gaussian kernels of different width. Then, the influence
of the kernel numbers on Euclidean errors is investigated by fixing the kernel
width and applying DMPs of different kernels to the five letters. Experimental
results using a mobile manipulator not only show the capability to rewrite
letters the way humans write, but also the ability to create new letters in a
similar writing style. 

\cite{pahic2020training} design an encoder-decoder framework for generating DMPs
that enables a robot to learn and reproduce complex motions. Instead of
minimizing the difference between DMP parameters, the proposed loss function
directly minimizes the physical distance between demonstrated and generated
trajectories. This formulation employs differentiable DMP equations for
efficient gradient computation, making it compatible with standard
backpropagation. The method is evaluated across two neural architectures,
showing improved trajectory reproduction compared to parameter-based losses, and
is further validated on handwritten digit datasets to demonstrate the capacity
to learn motor behaviors from visual input. In follow-up work,
\cite{pahic2020reconstructing} extend the model to output normalized AL-DMP
parameters, while \cite{pahic2021robot} leverage the trained autoencoder to
derive a latent action space from simulated data, enabling the synthesis of new
robot actions such as ball throwing.

To overcome the shortcomings of DMP and GMM/GMR methods, \cite{song2020robot}
develop a skill transfer system that encapsulates multiple demonstrations into
one motion model. Concretely, an unsupervised segmentation technique is proposed
to detect motion units in the demonstrated kinematic data using the concept of
key points. Consistent trajectory features are found using a hidden semi-Markov
model, which is then used to identify key points common to all the
demonstrations via a probabilistic approach. The DMP-modified framework can
generate trajectories of arbitrary shape and complexity from a small number of
demonstrations, and it does not rely on preset empirical parameters. The system
is evaluated on an imitation painting task using a 14-DoF dual-arm cooperative
robot.

For periodic motions, the amount of effort and time required for the explicit
programming of a robot is substantially more. To address this issue,
\cite{papageorgiou2022dirichlet} encode periodic motions based on a single
demonstration utilizing periodic sinc kernels (i.e., Dirichlet kernels)
introduced by \cite{papageorgiou2018sinc}. Concretely, the approach implements
Dirichlet base functions in DMPs to encode periodic motions. Not only does this
guarantee perfect reproduction of the periodic motions, but it also allows for
analytically computing the minimum number of required kernels based only on the
predefined accuracy. An experimental evaluation using two scenarios are
considered: the shake motion of a handshake, and the task of brush painting a
symbol on a planar surface.

Adapting the DMP framework by learning from both expert and non-expert
demonstrations can improve flexibility on complex manipulation tasks (e.g.,
writing). For instance, \cite{dong2023dynamic} design positive and negative
demonstration-based DMPs, which are obtained by utilizing DMP trajectories
generalized by GMR (\cite{calinon2007learning}) according to three aspects.
First, a new maximum log-likelihood balances the probabilities of positive and
negative demonstrations. Second, the maximum of this formulation is obtained by
iteratively calculating the lower bound of a Q-function. Finally, dataset
aggregation is utilized to allow the framework to handle unmodeled obstacles.
Experimental results show the method's success on several manipulation tasks
including letter writing, obstacle avoidance, and grasping in a grid box.

To enable robots to learn smooth and high-quality motion skills,
\cite{dong2023robot} propose an LfD framework based on generalized GMMs as
follows. First, generalized GMMs are used to model the demonstration data, which
improves performance for nonlinear data. Second, a GMR algorithm is adjusted to
make it applicable to generalized GMMs. The resulting GMR method is then
utilized to obtain smooth trajectories. Finally, the generated trajectories are
modeled by DMPs. In this way, a relatively perfect trajectory can be learned
from multiple demonstrations. Moreover, by using DMPs to model the trajectories
the corresponding motion skills not only have strong generalization performance,
but they also achieve real-time motion generation. The effectiveness of the
approach is verified by performing a Chinese-character writing task using a
robot manipulator.

Reproducing smooth dragging motions on industrial robots is important for
improving the ability to learn and generalize human-like curve drawing skills.
Recognizing this importance, \cite{xue2023robotic} propose a continuous drag
scheme that uses DMPs to enable compliant and effortless motion. The
architecture's key components consist of a modified DMP and a discrete
admittance model. Six DMPs enhanced by a scaling factor and force coupling term
are used to denote positions and $X$-$Y$-$Z$ Euler angles. These components are
integrated into an overall framework that includes modules for trajectory
acquisition, learning, reproduction, and generalization. Drawing experiments
using a 6-DoF collaborative robot demonstrate accurate and smooth reproduction
of demonstrated drag motions on both flat and curved surfaces, outperforming the
robot's built-in drag demonstration functionality.

\subsubsection{Field Robotics}
\paragraph{Agricultural robotics.} 
Boosting the productivity of harvesting operations is crucial to meet increasing
demand. For example, \cite{la2021study} explore how DMPs can be employed for
automating the motions of forestry cranes. Specifically, an LfD methodology is
developed where a forwarder crane is equipped with sensors to record motion data
while being controlled by expert operators. The goal of the motion planner is to
automatically retract logs back into a log bunk once the operator has manually
grabbed them via joystick. Simulated results show that the framework can
reliably reproduce the demonstrated motion after carefully preparing the
training dataset. In follow-up work, \cite{la2024combining} optimize the DMPs in
order to exploit the crane's redundancy to find alternative joint trajectories
that minimize energy costs. The objective is to mimic the Cartesian-space
motions performed by the operators while simultaneously identifying new joint
trajectories that improve energy performance.

Agricultural environments pose challenges for robotics, especially motion
planning, due to the complexity of the tasks to be performed. To address these
issues, \cite{lauretti2023robot} create an LfD framework based on DMPs that can
be applied to complex robotics activities such as those performed in the
agricultural domain. The approach employs Lie theory and integrates the
exponential and logarithmic maps into the DMP equations. This converts any
element of the Lie group into an element of the tangent space, and vice versa.
Furthermore, it includes a dynamic parameterization for the tangent space
elements in order to manage the discontinuity of the logarithmic map. The
proposed DMP motion planner is trained and tested on a mobile manipulator to
perform the following agricultural activities: digging, seeding, irrigation, and
harvesting. 

Traditional DMP scaling methods face a fundamental limitation: they either work
near demonstrated positions or require many demonstrations to generalize across
a robot's workspace. To tackle this restriction, \cite{lauretti2024new} develop
a DMP scaling method that uses demonstrations at workspace boundaries and linear
interpolation of DMP parameters. The approach requires only two demonstrations
rather than collecting large datasets of examples. The method is validated in
agricultural robotics, where a mobile manipulator performs complex tasks
including digging, seeding, irrigation, and harvesting. Notably, the proposed
scaling method prevents trajectory amplification issues that cause
single-demonstration techniques to generate motions exceeding the robot's
workspace limits. Experimental results show comparable accuracy to other methods
while maintaining a high success rate across the workspace. 

\paragraph{Space robotics.} 
To support sustainable infrastructure on the Moon, robots must be leveraged to
extract lunar resources for in-situ processing and construction. Within this
realm of space robotics, \cite{cloud2023lunar} utilize DMPs for navigating a
lunar rover in a circular trajectory around a lander while avoiding surface
hazards such as rocks. A steering angle approach is adapted by introducing an
obstacle avoidance parameter, which is configured to avoid rocks throughout a
set of testing exercises. This compact and generalized representation of the
motion allows a mobile robot to safely navigate in the presence of densely
populated obstacles while retaining the desired navigation pattern behavior. The
methodology is demonstrated within a simulation of the Moon against the
canonical obstacle avoidance approach for DMPs.  

Solutions for lunar rover autonomy require a multidisciplinary approach that
combines advances in computer vision, motion planning, and adaptive control
strategies. To this end, \cite{cloud2025vision} integrate DMPs with
state-of-the-art instance segmentation models to address hazard avoidance during
lunar excavator operations. The framework considers both visually detected
hazards and their characteristics, and is the first to combine vision-based
models with real-time obstacle avoidance for DMPs. Furthermore, a high-fidelity
simulation environment is designed to replicate the low solar angle lighting
effects of the lunar south pole. Simulation results highlight the system's
ability to avoid large obstacles, and most small ones, with minimal path
deviations and increases to drive length.

\paragraph{Underwater robotics.} 
DMPs have been shown to be robust to perturbations in complex underwater
environments caused by currents, low visibility, and sensor uncertainty. For
example, \cite{carrera2015cognitive} present an LfD approach that leverages DMPs
within a cognitive architecture for underwater tasks. The architecture consists
of three modules. The control layer handles the end-effector velocity, vehicle
velocity, and the end effector applied torque. The perception and localization
layer estimates the vehicle's position using an extended Kalman filter, and
template matching is used to identify a target object's position and
orientation. The LfD layer records, learns, and reproduces intervention tasks
using DMPs. Trained by humans to perform valve turning tasks, the proposed
method is evaluated on real-world experiments using a 4-DoF autonomous
underwater vehicle.  

Enhancing the autonomy of underwater robots can reduce operator cognitive burden
while also mitigating the impact of communication latency on operational
efficiency. In line with these observations, \cite{yang2024learning} introduce
an LfD approach that uses operator demonstrations as inputs and models the
trajectory characteristics of a task through DMPs. Different from existing DMP
applications, the method addresses the complexity and stochasticity of
underwater operational environments (e.g., current perturbations and floating
operations). A GMM and GMR are employed to extract features from multiple
demonstration trajectories to obtain trajectories as inputs to the DMPs.
Additionally, the commonly used homomorphic-based teleoperation mode is improved
to a heteromorphic mode. This allows the operator to focus more on the
end-operation task. The effectiveness of the system for underwater LfD is shown
via simulations.

\paragraph{Unmanned aerial vehicles.} 
Control algorithms for unmanned aerial vehicles must handle high-dimensional and
nonlinear environments, which makes the generation of aggressive flight
maneuvers difficult to accomplish. \cite{perk2006motion} address this problem by
taking an approach that adapts to environmental changes using a sequence of
target points irrespective of the initial conditions. To do this, DMPs are
coupled with nonlinear contraction theory. This enables the system to converge
to its coupled pair smoothly. DMP trajectories learned from human-piloted flight
data are first segmented into parts, and then combined at different initial
points to achieve maneuvers against various obstacles in different locations.
The methodology is tested using numerical simulations and physical experiments
with a 3-DoF benchtop model helicopter. 

DMPs have been applied to problems in field robotics including the control and
motion planning of mobile robots. For instance, \cite{fang2014control} use
higher-order statistics and DMPs for the control-oriented modeling of flight
demonstrations for quadrotors. This is done by first parsing the demonstration
data into segments in an unsupervised way. Then, DMPs are used to model the
nonlinear and rhythmic segments. Lastly, the methods are integrated into flight
demonstrations and evaluated via simulation by modeling a quadrotor's axial roll
maneuver. 

\cite{tomic2014learning} use an adapted DMP approach to learn and generalize
optimal control solutions for a quadrotor-type vehicle. Concretely, an algorithm
is developed to encode a grid of optimal solutions that are generated offline
into a second-order dynamical system. A DMP is then trained to handle unforeseen
goal states instantaneously, via weight interpolation, to enable real-time
generalization to new goals and in-flight adaptation. The method is demonstrated
via planar point-to-point and perching maneuvers in both simulation and physical
quadrotor experiments.

\cite{kim2018cooperation} explore the use of cooperation among multiple aerial
vehicles to overcome payload limitations and allow coordination in safely
carrying heavy objects. To do this, a learning-based motion-planning algorithm
is introduced. PDMPs are used to learn the relationship between known
environments and corresponding optimal motions, to produce trajectories in new
environments in real-time. By learning the relationship between motion
parameters and the environmental configuration from optimal (offline)
demonstrations, a safe trajectory corresponding to the actual environment with
unknown obstacles can then be quickly computed (online) without the need for
further optimization. In \cite{kim2019incorporating}, a process to ensure
safety-guaranteed demonstrations for PDMPs is formulated and evaluated on
cooperative tasks between two aerial manipulators.

An integrated framework that includes control, estimation of an unknown payload,
safety management, and obstacle avoidance for cooperative transportation via
aerial manipulators is proposed by \cite{lee2018integrated} as follows. An
online estimator is first designed to estimate the mass and inertial properties
of a payload. Then, an adaptive controller based on the estimated parameters is
developed. After joint space trajectories are generated to satisfy unilateral
constraints for safety, DMPs are employed to modify the trajectories in order to
avoid in-flight obstacles. Experiments are conducted using camera-equipped
aerial manipulators combined with a 2-DoF robot arm in unknown environments with
obstacles. 

A trajectory optimization and replanning algorithm for micro air vehicles in
cluttered environments is developed by \cite{lee2020trajectory}. The approach
first designs an offline global path optimization algorithm that generates a
safe trajectory. This trajectory enables the vehicle to avoid static obstacles,
denoted in a navigation map, and it allows the vehicle to satisfy its initial
and final velocities. DMPs along with a time-adjustment algorithm are used to
avoid obstacles and replan the vehicle's path. The approach is demonstrated via
simulations and outdoor autonomous flights using a custom-made micro air
vehicle.

The coupling of position and attitude is an often overlooked problem when
applying DMPs to the control of unmanned aerial vehicles. \cite{zhang2021novel}
solve this issue using IL as follows. First, DMPs are expanded to $SE(3)$ using
a dual quaternion approach. Since the dual quaternion uses only eight real
numbers to describe the motion of a general rigid body, it can concurrently
express rotation and translation as well as the coupling relationship between
position and attitude. Next, an online hardware-in-the-loop training system is
established. The dual quaternion DMP is trained via expert teaching data and
applied to describe the motion of a fixed-wing unmanned aerial vehicle in an
open-source simulation.

A major challenge of quadrotor precision landing is dealing with disturbances in
the environment (e.g., wind gusts), which can affect the trajectory and
stability. To overcome this problem, \cite{rothomphiwat2024robust} propose an
online adaptive trajectory planning method based on temporal scaling adaptation
of DMPs. Specifically, the modified DMPs allow the system to adapt the landing
trajectory to moving targets and handle disturbances by utilizing the current
position and velocity of the goal along with tracking errors as feedback. This
enables the desired trajectory to be dynamically adjusted in response to
tracking errors and the goal's state. The approach is evaluated under various
quadrotor disturbance scenarios using both simulations and real-world
environments. 

\subsubsection{Humanoid Robotics}
\paragraph{Bipedal motion.} 
Biped locomotion research is crucial for advancing the development of humanoid
robots. Compared to offline trajectory planning, biologically inspired control
approaches based on CPGs with neural oscillators have been utilized for rhythmic
motion generation. For example,
\cite{nakanishi2003learning,nakanishi2004framework,nakanishi2004learning} learn
biped locomotion from human demonstrations and adapt them through coupling
between a CPG and the mechanical system. DMPs are employed to rapidly learn
demonstrated trajectories and rescale rhythmic movements in terms of amplitude,
frequency, and offset of the patterns. An adaptation algorithm is also proposed
for the frequency of walking based on phase resetting and entrainment between
the phase oscillator and mechanical system using environmental feedback.
Numerical simulations along with a physical implementation of the framework are
conducted on a biped robot. \cite{pongas2005rapid} build upon this work by
developing a system that can not only phase lock with arbitrary rhythmic
external events, but can also accomplish phase locking at any phase of the DMP.

Since many tasks (e.g., wiping, paddling, drumming, walking/running, etc.) are
generically rhythmic, learning rhythmic behavior is a critical motor skill
acquisition problem. Rhythmic behaviors implemented with DMPs typically rely on
stabilizing controllers, such as balance controllers for walking gaits utilized
by bipedal robots. For instance, \cite{gopalan2013feedback} propose a feedback
error learning approach that adapts DMPs to minimize the controller's effort. In
particular, the method employs the stabilizing controller's output as an error
signal to modify the DMP parameters via gradient descent. The approach is
validated on a two-link robot arm maintaining vertical orientation while
oscillating, and a planar biped robot optimizing torso trajectories during
walking. The system effectively learns inherently stable trajectories that
minimize the required stabilizing control effort.

In biological motor control, it has been hypothesized that coherent activations
of groups of muscles can allow for exploiting shared knowledge. Inspired by this
concept, \cite{ruckert2013learned} propose a representation that employs
parametrized basis functions to combine the benefits of muscle synergies and
DMPs. For each task, a superposition of learned basis functions modulate the
stable attractor system of the DMPs. The approach leads to a compact
representation of multiple motor skills and enables efficient learning in
high-dimensional continuous systems. Furthermore, the movement representation
supports discrete and rhythmic movements, and includes DMPs as a special case.
Simulated evaluations are conducted on reaching tasks along with bipedal walking
using multiple patterns and different step heights.

DMPs were originally designed for discrete tasks, yet they can also be applied
to periodic motion. For example, \cite{gams2014rich} modulate periodic DMPs
based on force feedback. To do this, DMPs are coupled at the velocity level with
sensed forces where immediate corrections and feed-forward terms are learned
through repetitive control to minimize the force tracking error over multiple
periods. The approach is validated on a humanoid robot in two scenarios: a
bimanual task of rotating a fixed pedal racer while maintaining contact, and
operating a pedal racer by standing on it. In the bimanual task, the system
maintained constant contact despite noisy joint-torque sensors, though balance
support was needed. For the pedaling task, the system adapted approximate foot
trajectories to achieve pedaling motions, but balance assistance was required at
least once per period.

Periodic DMPs can provide a methodology for generating periodic biped
locomotions, particularly when combined with an established CPG framework. To
this end, \cite{andre2015adapting} integrate RL into a hybrid CPG/DMP system
that optimizes DMP parameters through the PoWER method and path integral policy
improvement with covariance matrix adaptation. This approach allows a bipedal
humanoid robot to adapt its walking strategy across inclined slopes, improving
frontal velocity and stability. Building on this methodology,
\cite{andre2016skill} refine associative skill memory with phase-indexed data
normalization and statistical failure detection. This results in a reduction of
storage requirements and enhances early failure prediction.   

\paragraph{Joint/limb coordination.} 
A lack of theories on motor learning in high-dimensional spaces led to the idea
of leveraging IL to speed up the learning process for complex motor systems.
\cite{schaal1999imitation} recognized that research into IL for humanoid robots
needed to include a theory of motor learning, compact state-action
representations (i.e., MPs), and of the interaction of action and perception.
Early results of this research consisted of approaches to generate rhythmic
movement patterns with nonlinear dynamical systems. For example, compared to
non-autonomous movement representations (e.g., splines),
\cite{ijspeert2002learninga} show that learned pattern generators can cope with
external perturbations. LfD evaluations of the approach are performed with a
30-DoF humanoid robot.

Many works have applied DMPs to bipedal motion. For instance,
\cite{stulp2009compacta} derive a set of standard solutions for reaching
behavior using human motion data. In addition, reaching trajectories are derived
for variations of a task where obstacles are present. The goals of the research
are the following: (i) demonstrate the advantages of explicitly representing
standard solutions to task variations; (ii) show how such variations are
compactly represented as DMPs; (iii) enable a robot to not only imitate reaching
motion, but also have the robot imitate a few ``principal trajectories" that
allow it to deal with task variations. Trajectories are considered as a whole
and relevant features are extracted via PCA. The approach is evaluated using a
humanoid robot.

Planning human-like robot trajectories for catching fast moving targets involves
two closely related problems: (i) accurately predicting the object's trajectory,
and (ii) rapid planning of precise trajectories for the robot's end effector.
Using DMPs to generate these trajectories is problematic since they are time
dependent. Building upon \cite{gribovskaya2009learning}, \cite{kim2010learning}
address this issue by imposing temporal constraints on motions encoded with DMPs
as follows. First, demonstrated trajectories and their velocities are encoded
with a GMM through an EM algorithm. Then, a timing controller is developed to
manage the duration of the DMP movement. This allows for gradually speeding up
or slowing down the learned DMP to adhere to precise temporal restrictions. The
approach is validated on experiments that involve a humanoid robot catching a
ball in motion.

\cite{tan2011computational} develop a computational framework for humanoid
robots to learn complex behaviors through the combination of robotic
self-exploration and human demonstrations. Concretely, an RRT-connect algorithm
is employed for exploration, a linear global model is utilized for recording
demonstrations, and a spatial-temporal extension of the Isomap method is used
for dimensionality reduction. This enables exploration to be conducted in a
low-dimensional latent space, which can allow a robot to learn complex
behaviors more readily. The log likelihood function of the distribution of
sampled data in the joint space is then utilized to project the data in the
latent space back to the joint space. A robot behavior imitation experiment is
carried out to demonstrate how the proposed framework works.

Learning parameterizable skills from multiple demonstrations can allow robots to
generalize their motions to new tasks. For example, \cite{stulp2013learning}
generalize prior approaches on learning parameterizable skills based on DMPs.
This is done by generalizing the DMP formulation such that the task parameters
are passed directly to the DMP's function approximator in an online fashion.
Instead of training DMPs from multiple demonstrations and requiring two separate
regressions, the proposed process can be done with a single regression. Not only
does learning the function approximator with one regression in the full space of
phase and tasks parameters allow for more compact models, but it also permits
the flexible use of different function approximator implementations such as
locally weighted projection regression (LWPR) and GPR. The method is evaluated
on via-point and object transport tasks using humanoid robots.

Adapting rhythmic motion patterns to different robot morphologies is challenging
without extensive parameter tuning. To handle this difficulty,
\cite{li2014novel} propose a bio-inspired architecture that combines CPGs with
DMPs in an actor-critic RL framework. The system separates locomotion modeling
into a baseline system using a four-cell CPG network for basic patterns and an
adaptation system using DMPs to reshape these patterns. The architecture is
validated through experiments on two different platforms: crawling on a humanoid
robot and galloping on a quadruped robot. The results show that the PoWER
algorithm achieves faster learning and better convergence than the episodic
natural actor-critic method. Not only does the system demonstrate successful
gait adaptation, but it also identifies the learning speed and fixed oscillation
frequencies as limitations.

Deploying humanoid robots to aid with everyday chores is a practical use case
for robotics. Yet, teaching via demonstration is difficult for tasks with
expansive solution spaces. With regard to manipulation actions,
\cite{mao2014learning} address this issue with a framework that learns hand
movement trajectories in new situations as follows. First, markerless
hand-tracking data is captured using a depth sensor. Next, the data is
postprocessed by segmenting the movements into manipulation and withdraw phases.
Finally, a DMP-based generative method learns the hand movements necessary to
control the robot's end effector. Experimental results on a dual-arm robot
validate the effectiveness of the approach in learning human hand movements from
demonstration without the use of motion capturing devices.  

Although DMPs allow for the generation of complex motions from simpler building
blocks, a less explored issue is at which level to encode and organize a library
of DMPs and how to retrieve them from the library to fit a particular task.
\cite{reinhart2014efficient} tackle this problem via a parameterized skill
memory that organizes a set of DMPs in a low-dimensional, topology-preserving
embedding space. Concretely, the skill memory acts as a mechanism that links
low-dimensional skill parametrizations to DMP parameters and complete motion
trajectories. The skill memory is implemented by means of a parameterized skill
library through a trainable neural dynamic memory based on associative radial
basis function networks. Experimental results using a humanoid robot on bimanual
manipulation tasks shows that the framework is beneficial for efficient,
reward-based retrieval of DMP parameters and simplifies the shaping of reward
functions. In proceeding work, \cite{reinhart2015efficient} leverage the
parameterized skill memory for efficient policy search in low-dimensional
spaces. 

Multi-arm manipulation tasks require precise coordination to maintain desired
relative positions and velocities between end effectors. To this end,
\cite{gams2015accelerating} accelerate DMP adaptation for dual-arm tasks by
incorporating both position and velocity error terms into the feedback learning
framework. Specifically, ILC with feedback error is employed to achieve faster
convergence of the coupling terms used for arm coordination. This enables
adaptation in joint space rather than just task space, which preserves the joint
configuration of redundant manipulators. The approach is demonstrated on a
discrete-periodic DMP formulation where one arm must adapt its trajectory to
maintain a constant grasp on a shared payload during both reaching and periodic
motions. The modified feedback structure reduces the number of learning
iterations from ten to four while maintaining the behavior.

\cite{luo2015learning} devise a framework to learn push recovery strategies for
bipedal humanoid robots via DMPs. The approach models bio-inspired strategies
(hip-ankle and stepping) through DMPs and refines them using the PoWER
algorithm. For hip-ankle strategies, DMP parameters are optimized to handle
specific push forces and GPR is employed to generalize across different force
magnitudes. For stepping strategies, IL is utilized to acquire basic gaits and
to exploit the DMP's attractor properties for online footstep adaptation. A
three-layer hierarchical structure (center of mass, body parts, joints) is used
to reduce learning complexity. The framework is validated on both simulated and
physical humanoid robots by demonstrating successful recoveries from pushes
during multiple standing and walking tasks. 

Robot gestures play a key role in human-robot communication, but are difficult
to design and modify for high-DoF systems. \cite{pfeiffer2015gesture} address
this issue through a system for learning and executing gestures on a humanoid
robot using DMPs to generalize the gestures. Three different learning modes are
supported: (i) recording predefined motions, (ii) kinesthetic teaching with
gravity compensation, and (iii) learning from external devices like motion
trackers. The system preprocesses recorded trajectories to reduce noise via
downsampling and cubic splines, performs collision checking, and then executes
the generalized gestures. Experimental results demonstrate successful learning
and reproduction of gestures from different starting conditions while
maintaining smooth transitions. Compared to traditional fixed trajectory
methods, the approach enables more natural and flexible gesture generation.

In the context of bimanual manipulation, \cite{silverio2015learning} present a
framework that encapsulates both position and orientation patterns by
emphasizing the full poses of end effectors for richer skill encoding. Complete
end-effector poses are learned by a robot through a task-parameterized GMM. This
permits the simultaneous encoding of the demonstrations in multiple frames, and
it can be extended to task-adaptive orientation control to encode and retrieve
coordination patterns between two end effectors. To select the desired impedance
of the orientation controller, a quaternion-based dynamical systems formulation
makes it possible to encode the dynamics of the task in $SO(3)$ via a virtual
attractor. The approach is experimentally validated using two 7-DoF robots
performing a bimanual sweeping task. 

Due to the complexity of kinematic chains, multi-dimensionality, and dynamic
constraints, the problem of multi-contact whole body motions is a difficult
problem in humanoid robotics. \cite{mandery2016using} make progress on this
topic via a probabilistic n-gram language model learned from human locomotion
data. The framework is comprised of three steps. First, the body is segmented
according to which parts make direct contact with the environment. These poses
are further divided with respect to the whole-body configuration. Second,
sequences of whole-body poses are discovered from motion capture data, where
DMPs directly learn transitions between consecutive poses. Third, a language
model is trained on the pose sequences. The method is evaluated on 140 motion
capture demonstrations where one or both hands are used for support. The results
show the ability to generate complex sequences of pose transitions, including
following a straight line.   

Repeated policy search for parameterizations of different skills is inefficient
if the structure of the skill variability is not exploited. Based on this
observation, instead of gathering demonstrations from a human teacher,
\cite{queisser2016incremental} apply policy optimization for parametrizing a new
task. Concretely, a bootstrapping algorithm is developed to successively improve
the initialization of the optimization process by determining the DMP parameters
based on previous experiences. Simulations are performed on a reaching scenario
with a humanoid robot and grid-shaped obstacle. The results indicate that not
only is the DMP space well suited for parameterized robot trajectory generation,
but also a smooth mapping between the DMP space and task parameterization is a
valid assumption. Additional experimental verifications are carried out by
\cite{queisser2018bootstrapping}. 

\cite{bockmann2017kick} utilize DMPs to generate kicking motions for
soccer-playing humanoid robots. To do this, a mathematical motor model is
developed to compensate for the robot's motor control delay and maintain balance
when kicking. Specifically, during a kick the robot is dynamically balanced
using a linear quadratic regulator with previews to keep the zero moment point
inside the support polygon. This involves establishing a way of estimating the
zero moment point based on a model of the robot's motor behavior. The approach
is evaluated on humanoid robots, showing its efficacy in maintaining balance and
performing the desired movements. Additionally, the kicking motion is
successfully used in the corner kick challenge at RoboCup 2015.

Planning human-like motions for humanoid robots requires simultaneously
coordinating multiple joints while ensuring that the robot maintains balance. To
this end, \cite{mukovskiy2017adaptive} present an architecture that combines
online planning through learned DMPs with MPC to
generate adaptive full-body motions as follows. First, primitives are learned
from human demonstrations of a drawer-opening task. Then, a dynamic filter with
MPC converts the demonstrations into physically feasible trajectories. Using
these trajectories as training data, new optimized primitives for robot control
are learned. Validated on both simulated and physical robots, the approach
successfully generates coordinated multistep sequences that adapt to varying
drawer positions. Furthermore, the framework achieves performance comparable to
optimal control methods, but with a much lower computational cost.

When RL is used to optimize DMP parameters, the exploration of noise may cause a
trajectory to exceed a robot's joint limits. To address this problem,
\cite{duan2018constrained} focus on robot skill learning with physical joint
limitations in humanoid robots via a framework based on constrained DMPs.
Inspired by previous joint limit control laws, constrained DMPs control a set of
transformed states called exogenous states. Specifically, the joint trajectories
are bound within defined safety limits as long as the policy respects the
restriction of the joint limits. This ensures that the resulting trajectory
stays within the given limits of the joint's movement. The constrained DMP
approach is evaluated in simulation against conventional DMPs using a humanoid
robot.

\cite{amatya2020human} present a cooperative DMP framework to model lower-limb
coordination during physical human-human interaction, specifically for
three-legged walking tasks. The methodology represents each participant's knee
trajectory using rhythmic DMPs. The DMPs are augmented with coupling terms that
modify forcing terms based on the measured forces of other limbs. These coupling
terms are then refined across gait cycles using an ILC approach, enabling the
system to adapt and converge based on past interaction errors. Experimental
results from human dyads demonstrate that the proposed framework more accurately
captures gait synchronization and reduces modeling error for at least one
participant in each pair when compared to a baseline DMP model.

Among all the motor skills that a robot must adapt to for various tasks, the
execution of the arm is particularly important. Based on this observation,
\cite{lin2020arm} propose a systematic framework for trajectory planning and
learning of robot arm motion. The key idea is to use DMPs to extract motion
characteristics from human demonstrations, and then realize the reproduction and
generalization of the human actions by adjusting the nonlinear terms online. To
do this, DMPs are first combined with LWR to learn point-to-point motions from a
single demonstration. Then, smoothing of the trajectory is performed by cubic
spline fitting. This culminates in the generalization of the DMPs based on
spatial and temporal scaling. Experimental results show point-to-point motions
can be achieved on a humanoid service robot.

Humanoid robots must be able to adapt their walking gaits to novel situations in
real-world scenarios. \cite{liu2020workspace} address this challenge of adaptive
locomotion by implementing DMPs in the workspace of a humanoid robot.
Concretely, two online DMP trajectory generators are employed to overcome the
limitations of traditional bipedal robot locomotion control methods. Not only
does this simplify the complexity of trajectory planning, but it also enables
learning directly from a human walking gait. In addition, the control strategy
mimics the vestibular reflexes of humans and it includes multiple feedback loops
to prevent a robot from overturning and slipping. The performance of the system
is evaluated through both simulations and practical experiments with a humanoid
robot walking over sloped terrain.

When trying to learn how to manipulate tools, attempting to match human
dexterity may lead to improved robot control. However, high-DoF tools can pose
significant challenges. As an example of this difficulty, \cite{nah2020dynamic}
take inspiration from the central nervous system of a human to learn how to
model a whip: a complex, exotic, and flexible tool with non-uniform mechanical
properties. To do this, the complexity of the proposed approach is managed using
DMPs. Specifically, five optimal parameters are discovered through optimization,
allowing for the successful execution of a task comprised of reaching a distant
target using a whip with only a single movement. The simulation results show
that DMPs may be used to compose control, thus enabling the manipulation of a
whip's dynamics.

\subsubsection{Human-Robot Interaction}
\paragraph{Active compliance.} 
Compliance is an important safety feature in HRI. Active compliance is online
adaptable, its dynamic response is determined by the sampling and response rate
of the controller. Conversely, passive compliance is inherent to the system.
\cite{basa2015learning} couple DMPs with RL to enhance the trajectory generation
process for point-to-point movements while ensuring that passive compliance is
maintained by the controller. This is done by learning a regular grid of goal
positions inside the workspace of the robot arm. Then, each learned nonlinear
function is associated with a grid node and can later on be mixed to generate
trajectories for unlearned targets inside the grid. This approach provides
trajectories for the gear-side of compliant joints, while maintaining the
inherent elasticity of the joints. An evaluation is conducted via simulation of
a 2-DoF robot limb with passive compliant joint drives.

Admittance and impedance control frameworks integrated with stabilized bimanual
controllers can achieve versatile behaviors in robot manipulation tasks. For
example, \cite{gao2018projected} introduce a force-admittance control method
targeted towards compliant DMPs. The approach ensures motion synchronization and
target force regulation under external perturbations in three steps. First, the
controller uses a projected constraint force obtained through the analysis of
load distribution in bimanual tasks via a grasp mapping technique. Second, the
controller's actuation force is fed into an admittance control framework. Third,
the virtual target pose is provided to an impedance controller modeled as a
mass-spring-damper system. Experiments demonstrate the compliance of the control
framework on DMP generated trajectories using a humanoid robot.

Admittance control is a way to allow a robot to compliantly follow a reference
trajectory. For instance, \cite{bian2020extended} present an approach that
enables a robot to simultaneously learn both trajectories and stiffness profiles
from humans via kinesthetic teaching. Specifically, the framework develops
DMP-stiffness primitives that simultaneously generalize both trajectories and
stiffness profiles in a unified manner. EMG signals, extracted from a human's
upper limbs, are used to model the muscle co-contraction levels and obtain
target stiffness profiles. Variation of virtual stiffness in the admittance
controller is adjusted by monitoring vibrations of the end-effector velocities.
Experimental tests are conducted on a rigid plate transporting task and a water
pumping task using a robot arm.

When the environment changes, learning a new robot trajectory alone may not be
enough to successfully complete the task. However, incorporating stiffness
adaptation can play an important role for contact-related tasks.
\cite{dong2021dmp} leverage this observation by proposing an adaptive stiffness
methodology for DMPs based on the position error as follows. First, the DMP
model of the stiffness trajectory is established via LfD. Second, the feedback
term of the position trajectory error is added to the DMP model to allow for
online modulation. To do this, impedance control is used to adjust the contact
force between the end effector and the object. This allows the robot to adjust
its stiffness accordingly to cope with different changes. Experimental results
using a robot manipulator show the effectiveness of the method on pick-and-place
tasks.

\cite{dou2022robot} create a torque-controlled skill learning framework that
unifies trajectory learning and active compliance control within a single
system. The framework includes kinesthetic teaching, task learning, a skill
library, and task generalization. Specifically, DMPs encode motion trajectories
and stiffness profiles, enabling compliance through a variable impedance
controller. A variable stiffness interface maps HRI forces to stiffness
adjustments, allowing robots to generalize compliant behaviors to new tasks.
Experiments conducted on a 6-DoF collaborative robot performing button-pressing
tasks demonstrate accurate trajectory reproduction, safe compliance, and
successful generalization across varying button positions and required forces.

Parallel robots are ideal for tasks that require precision and performance due
to their robust behavior. Yet, two fundamental issues need to be resolved to
ensure safe operation: (i) the force exerted on the human must be controlled and
limited; (ii) type II singularities (i.e, configurations where the robot loses
controllable DoF) should be avoided. \cite{escarabajal2023combined} provide a
solution using DMPs to simultaneously tackle both of these problems. To allow
for force control, an admittance controller is defined within the DMP structure.
Then, a type II singularity evasion layer is added to the system. This allows
both the admittance controller and the evader to exploit the behavior of the DMP
and its properties related to invariance and temporal coupling. The system is
deployed and tested on a parallel robot for knee rehabilitation.

Using LfD to directly extract compliant DMPs can improve the performance of
robots, but human motion is typically optimal for human biomechanics, not for
robot dynamics. \cite{hong2023human} solve this problem via a two-level
framework that extracts dynamic features such as inertia, damping, and
stiffness for either parameter tuning (LfD with DMPs) or skill transfer (RL
with DMPs). In the first case (LfD with DMPs), the forcing term of the DMP is
learned using LWR. In the second case (RL with DMPs), the forcing term is
substituted with the action of an RL agent, where proximal policy optimization
is used to train an actor-critic policy. A position controller is implemented
in a simulated HRI scenario to extract dynamic features from virtual reality
(VR) demonstrations and regenerate robot trajectories while maintaining
human-like motions.

Achieving a desired interaction force with a robot requires adaptation to
unpredictable human behavior. To realize compliant interaction forces,
\cite{xing2023dynamic} develop a framework that integrates DMPs with iterative
learning via the following steps. First, DMPs are employed to parameterize the
demonstration trajectories of the user. Second, a recursive least squares
estimator is designed to update the trajectory parameters and minimize the
interaction force error. Concretely, the desired reference trajectory is
iteratively obtained by resolving the DMPs. By parametrizing the trajectories
based on the DMP phase variable, the assumption made by traditional methods that
the iteration period should be invariant is removed. Experiments using a 2-DoF
robot with multiple human subjects on both 1D and 2D tasks demonstrate small
force errors and robustness against time-related uncertainties. 

Compliant motion control strategies are essential for providing flexibility and
safety in HRI scenarios. For example, \cite{zhang2024innovative} propose a
human-robot skill transfer algorithm based on DMPs as follows. To begin, human
demonstrations are encoded using multiple DMPs constrained by captured force and
position data. Next, the framework addresses safety through a force-based
impedance control strategy that enables active compliance during trajectory
reproduction. Lastly, the demonstration and reproduction trajectories are
temporally aligned using DTW. An evaluation on a simulated robot platform and a
real-world 7-DoF arm performing curve drawing and table-wiping tasks show that
the DMP model can accurately reproduce and generalize acquired skills while
maintaining compliance in both obstacle-free and obstacle-present environments. 

Compliant physical HRI is critical for rehabilitation robots. Recognizing this
importance, \cite{zhou2024dynamic} design a sitting/lying lower-limb
rehabilitation robot for patients with lower-extremity motor dysfunction. In
particular, the system includes a high-level planner consisting of a trajectory
generator based on DMPs, acceleration layer modulation generator, and velocity
layer modulation generator. Additionally, a linear active disturbance rejection
controller is utilized as a low-level position controller to ensure that each
joint can accurately and robustly track the desired trajectory under internal
and external disturbances. Simulation and experimental results show that the
strategy not only provides compliant physical HRI within the constrained joint
space, but also ensures accurate and robust trajectory tracking.

\paragraph{Active learning.} 
Robots are being deployed in increasingly complex environments where they have
to fulfill a range of different tasks. Therefore, in skill learning settings
where a robot actively selects among tasks that have been examined during
learning, it can be helpful to transfer knowledge of easy tasks to more
difficult tasks. To this end, \cite{fabisch2014active} introduce an active
context selection methodology. The framework models the learning process as a
stationary multi-armed bandit problem with custom intrinsic reward heuristics.
Although the approach does not make assumptions about the underlying policy
representation, DMPs are used to represent skills. Empirical results show that
active context selection improves skill learning on a ball-throwing problem
using a simulated robot arm. 

A fundamental problem in HRI research is deciding when to teach a robot a new
skill or when to rely on it to generalize its actions. \cite{maeda2017active}
approach this issue via the development of an active learning framework that
allows a robot to incrementally increase its skills by reasoning when to request
demonstrations. The methodology utilizes GPs to provide a measure of the
confidence in the extrapolated training set, which can then trigger a new
demonstration request. The assumption is that contexts that are close to each
other will have trajectories with similar profiles. DMPs are learned on the GP
output to guarantee that the desired position can be exactly achieved. The
combination of DMPs and GPs also provides a way to contextualize the
demonstration, providing the capability to address different demonstrations.
The system is tested in an industrial environment where a user has to program a
collaborative robot manipulator to reach different objects in its workspace.

\paragraph{Cooperative control.} 
Developing cooperative robotic technologies in spaces such as small-scale
manufacturing, assistance of the physically disabled, and surgery remains a
challenging problem. To address this dilemma, \cite{guerin2014adjutant} propose
a collaborative robotic system designed to be instructable for a range of tasks.
The framework leverages user interaction modalities necessary to execute a task,
and it requires minimal retraining when adapting to a new task. Specifically,
the system organizes robot capabilities and collaborative behaviors by relating
each capability to a specific user interface. DMPs are used to create reusable
tool motion representations for modeling each capability. An evaluation is
conducted on several real-world tasks including collaborative/automated drilling
and continuous sanding. 

Enhancing the generalizability of collaborative robotic systems is an open
problem in dynamic environments. Nevertheless, environmental properties along
with the human collaborator may be utilized by DMP frameworks to improve motor
skills and enhance generalization ability. For example,
\cite{cui2016environment} propose environment-adaptive IPs to learn complex
interactive skills. To do this, the IP methodology is extended via a set of
environmental features. This allows for maintaining a distribution over DMP
parameters conditioned on information about the environment. Experiments using a
dual-arm robot manipulator on the task of covering objects with plastic bags
demonstrates adaptation to novel environments. \cite{cui2019environment} extend
the framework to automatically derive suitable environmental parameters.

Cooperation between humans and robots is a challenge in non-structured
environments. \cite{petrivc2016cooperative} address this difficulty through a
control framework for human-robot cooperative tasks that combines an impedance
control approach with a computational model of Fitts's law. Fitts's law states
that the time it takes to move to a target is related to the distance to and the
size of the target, with larger and closer targets being easier and faster to
reach. Within the framework, a robot adapts to human motion via sensory feedback
by taking into account Fitts's law, where user-specific behavior is estimated
using recursive least squares updates. Trajectories are represented by DMPs and
adaptation relies on an IL controller. An evaluation employing a haptic robot
for an arm-reaching task shows adaptation to human motor control while
maintaining the desired accuracy of the movement.

Adapting different robot manipulation systems to new tasks and target platforms
can allow for solving unforeseen challenges and tasks. For instance,
\cite{gutzeit2018besman} propose a learning platform for automated robot skill
learning. The framework consists of five components including preprocessing of
human demonstrations, segmenting demonstrations into basic building blocks, IL
using DMPs, refinement through RL, and generalization to similar tasks. The core
components of the platform are empirically validated in a 10-participant study
according to automation level and minimum execution time. Experimental results
show that the learning platform allows for the transferring of skills from
humans to robots within a reasonable amount of time. 

\cite{nemec2018human} propose a framework to facilitate adaptive physical HRI
for cooperative tasks. The method combines trajectory adaptation with dynamic
stiffness control based on speed-scaled DMPs (\cite{nemec2013velocity}). By
analyzing task repetitions, a robot learns to be stiff in directions requiring
precision while remaining compliant along the motion direction. This allows
humans to modify both spatial paths and execution speeds through physical
interaction. Compliance is adjusted in the path operational space defined by
Frenet-Serret frames. Additionally, the system extends \cite{nemec2018adaptive}
by adding passivity-based control and improved speed adaptation mechanisms for
safe interaction. The framework is experimentally validated through (i)
demonstration of two-phase learning where spatial and velocity components are
taught separately, (ii) demonstration of trajectory coaching to enable rapid
local path modifications, and (iii) bimanual manipulation where dual robot arms
cooperate with a human to transport and insert a plate onto a vertical rod. 

Modeling the movement of target objects and responding with an accurate
predefined DMP requires real-time adaptation. Yet, many methods require a
computationally expensive forward simulation of the DMP at every time step which
makes it undesirable for integration in real-time control systems. To this end,
\cite{anand2021real} present a system for real-time adjustments and apply it to
the industrial handling of moving parts and human-robot collaborative tasks. Two
methods are proposed to achieve real-time temporal scaling: (i) using a control
law to vary the temporal scaling term of the standard exponential canonical
system, and (ii) leveraging a polynomial-based canonical system with a suitable
control law for temporal scaling. The framework is evaluated using a robot arm
where the desired tracking of a moving target is varied in real-time. 

Cooperative manipulation of dual-arm robots in domains such as medical and home
service requires careful coordination. \cite{lu2022dmps} tackle this issue by
enabling human-like cooperation for dual-arm robots via a DMP-based skills
learning framework. The proposed methodology includes three functionalities: (i)
DMP-based skill learning generalization given relative distance constraints,
(ii) trajectory replanning for joint distance restrictions, and (iii) a
redundant solution for two 7-DoF robots. The DMPs use a coupled acceleration
term that is computed based on a constant joint distance and constrained dynamic
relative distances between the robot and approaching objects. The idea takes
inspiration from BLFs to achieve transient performance control. An evaluation
using simulated environments shows effectiveness in producing human-like motions
with a dual-arm robot.

Tasks that are difficult with a single robotic arm, such as lifting heavy or
bulky objects, can be executed using dual-arm cooperation. Nevertheless,
cooperative manipulation requires compliant synchronized motions to handle
kinematic uncertainties. To make progress on this problem,
\cite{wang2024cooperative} develop a hierarchical control system to achieve
stable and adaptive manipulation. The DMP formulation introduces a hyperbolic
tangent as a delayed goal function to suppress large initial accelerations and
ensure fast convergence. A modular framework couples the trajectory model with
object detections for grasp-point localization and an impedance control strategy
to regulate contact forces. Both simulated and physical experiments using a
6-DoF manipulator demonstrate transportation of soft and rigid objects, with
stable handling maintained through impedance adjustment. 

\paragraph{Imitation learning.} 
Integrating robots into unstructured human environments requires them to be able
to easily learn new skills. To achieve this goal, \cite{kormushev2011imitation}
create a user-friendly framework that transfers robot task skills as follows.
First, a positional profile is obtained from demonstrations using kinesthetic
teaching. Next, a force profile is recorded from additional demonstrations via a
haptic device. This is done by having an operator input the desired forces that
the robot should exert on external objects during the task execution. Then, the
positional and force profiles are represented as DMPs and used to reproduce the
task satisfying both profiles. Finally, an active control strategy based on
task-space control with variable stiffness is developed to reproduce the skill.
The method is evaluated on two tasks: ironing clothes and opening doors.

The modeling and identification of movement prototypes is a key challenge in IL.
In particular, maintaining a minimal representative model capable of motion
generation remains a difficult problem. \cite{chang2013motion} present a
solution to this issue via a simple and efficient affinity propagation
clustering algorithm that identifies representative motions for several motion
groups. Prior knowledge of the number of clusters is not necessary, yet can be
used to fine-tune the results. Both HMMs and DMPs are used to model each motion
group. Using videos and motion capture data of human demonstrations,
experimental results show that the weight parameters of the DMP model can
distinguish between either coarsely distinct or finely distinct motion groups.  

Robot motion planning plays a key role in the interaction between users and
assistive technologies. Nonetheless, reproducing a user's personal motion style
is very challenging. \cite{lauretti2017learning} address this problem by
building an LfD system with the following properties: (i) easily trained by
humans that lack the technical skills to program a robot to perform daily living
tasks, (ii) can replicate personal motion styles, and (iii) can adapt to
environmental changes. During demonstrations, DMP trajectory parameters are
recorded and stored in a database. In operation, depending on the type of task
and target position, the system selects the proper DMP parameters from the
database and builds the trajectory to carry out the movement. The generalization
of the method is tested across three activities of daily living (eating,
pouring, drinking) by evaluating the success rate of the task execution.

In HRI, LfD has traditionally been applied to learn tasks from a single
demonstration modality, yet this can restrict the scalability of learning and
executing a series of tasks in real-world environments. \cite{wu2018multi}
mitigate this problem by developing a multimodal robot apprenticeship system
that learns multiple tasks from demonstration through natural interaction
modalities. The system utilizes a DMP+ model integrated in a dialogue system
with speech and ontology. Different from the original DMP+ formulation, a linear
decay system is used in place of the exponential one. This allows the DMP+
system to achieve arbitrary on-the-fly updates by making it easier to modify the
primitives (e.g., joining and inserting a new primitive into an existing task).
A gluing demonstration using a robot arm highlights how the system learns
multiple tasks.

\cite{vollmer2018user} investigate a user-centered setup that is intuitively
usable by non-experts and easily operated outside of a laboratory. To do this,
the study concentrates on robot learning of complex movement skills with a human
teacher using DMPs. Concretely, participants are asked to teach a semi-humanoid
robot a game of skill. The only feedback participants can provide is a discrete
rating (very good, good, average, not so good, not good at all) after each of
the robot's movement executions. The learning performance of the robot when
applying the user feedback is compared to a version of the learning where an
objectively determined cost function is applied. The findings suggest that
DMP-based optimization can be used for learning complex movement skills, thus
making the tedious definition of a cost function obsolete. 

Force-based sensing can enable a robot to identify physical constraints and act
accordingly. For instance, \cite{eiband2019learning} introduce a teaching
strategy and learning framework that allows for the generation of adaptive robot
behaviors in dynamic environments using only the sense of touch. The approach
represents skills hierarchically, using haptic exploration behaviors to feel the
environment and relative manipulation skills based on previously explored
events. DMPs are utilized to learn stable motions based on motion data generated
from a GMM. Experimental results demonstrate the utility of the proposed
exploration scheme by obtaining generalization to unseen object locations for a
pick-and-place manipulation task.

Parallel robots can provide physical therapists with an efficient and safe way
to perform ankle rehabilitation exercises. For example, \cite{abu2020passive}
present a framework that integrates LfD and ILC techniques to adaptively
generate personalized rehabilitation trajectories as follows. First, a therapist
designs an exercise routine with the patient by teaching a robot through LfD.
The learned trajectories are encoded using DMPs. Then, ILC iteratively adapts
these trajectories by incorporating force feedback to reduce patient discomfort
and ensure safe motion. Using these adapted trajectories as a baseline, the
framework gradually restores the originally designed exercise as the patient's
condition improves. The method is tested through both simulations and real-world
experiments, showing successful adaptation to individual patient needs. 

\cite{wang2020framework} present an HRI framework, based on DMPs, that takes
into account both positional and contact force profiles for skill transfer.
Different from the other methods that only involve motion information, the
system combines two sets of DMPs, which are built to model the motion trajectory
and force variation of a robot manipulator, respectively. This is done by a
hybrid force/motion control approach that ensures accurate tracking and
reproduction of the desired positional/force motor skills. Additionally, instead
of employing force sensors, a momentum-based force observer is applied to
estimate the contact force and thus simplify the control system. Using a
dual-arm robot, an evaluation is conducted to verify the effectiveness of the
learning framework on real-world scenarios such as cleaning a table.

Timber construction is a labor intensive assembly operation with opportunities
for automation. For instance, \cite{kramberger2022robotic} develop an LfD
framework for the collaborative assembly of timber truss structures connected by
novel lap-joint connections that interlock elements through assembly motions. To
do this, CDMPs are enhanced to enable a precise orientation transformation of
the entire encoded motion according to the pose specified by the global assembly
designer. Additionally, a force compliance controller is coupled directly to the
acceleration part of the CDMP to facilitate adaptation to environmental changes
during the assembly execution, therefore minimizing the interaction-force
torques and maximizing the success rate of the assembly. The effectiveness of
the system is shown on a timber structure assembly executed by two collaborative
robots.

In the case of robot-assisted rehabilitation of patients who have limited
mobility or injuries, unexpected forces may occur making traditional force
controllers incompatible. \cite{escarabajal2023imitation} mitigate this problem
by creating a framework for generating compliant motion trajectories for passive
rehabilitation exercises. The approach uses the prior trajectories of the
patient's healthy limb as a reference for generating safe motions. To obtain
safe back-and-forth motions, the patient's limb forces are first encoded using
GMR. Then, the motions are encoded via reversible DMPs
(\cite{sidiropoulos2021reversible}) based on the patient's input. An
experimental evaluation shows the success of self-paced patient exercises for
lower-limb rehabilitation using a 4-DoF parallel robot.

Capturing rich motion information from human demonstrations is important in
order to effectively transfer human-robot skills and ensure safety. For example,
\cite{li2023human} propose a technique that transmits skills from humans to
robots by utilizing LfD. The approach consists of four elements. First, a depth
sensor is used to recognize the demonstrated motion, hence creating a trajectory
path. Second, the path is modeled as a DMP. Third, GMR is employed to capture
the human motion via encoding of the trajectory. Finally, a recurrent neural
network accurately tracks the positioning along the learning path. Using a
mobile robot, experimental results demonstrate the method's effectiveness in
following human trajectories.

Flexible robots offer advantages in maneuverability and safety over rigid
robots, particularly in safety-critical applications or unstructured
environments. Yet, these advantages come with drawbacks such as unpredictable
movements and unusual behavior. Targeting these liabilities,
\cite{seleem2023imitation} create an imitation-based motion planning framework
with dynamic impedance control, enabling the manipulation of a robot in
non-static environments. The dynamic impedance controller is derived from the
Lagrangian formulation and Taylor expansion series, while the motion planning
method uses DMPs. The effectiveness of the approach is evaluated on experiments
with a two-section soft continuum robot where demonstrations are recorded via a
motion capture system. The results of the simulation highlight the ability of
the robot to return to the desired path, even after disturbances.

Long-horizon robot manipulation tasks involve intricate trajectories that
require efficient integration of high-level reasoning with motion planning. To
manage this complexity, \cite{liu2024enhancing} embed a large language model
within a hierarchical manipulation framework to facilitate motion execution.
The large language model decomposes complex tasks into subtasks using
environmental data from an object detection system, while a teleoperation
interface allows users to record manual demonstrations. The trajectories are
encoded and stored in a DMP library, from which the large language model
selectively executes motions. Using a 10-DoF robot, the system is able to
perform both one-shot and zero-shot manipulation tasks such as opening ovens and
cabinets, as well as multi-step operations including warming food.

\paragraph{Interaction learning.} 
In HRI, hand gestures offer an easy way to instruct and control machines (e.g.,
robots). For instance, \cite{strachan2004dynamic} present an approach to
gestural interaction via DMPs. This is done by modeling a gesture as a
second-order dynamic system followed by a learned nonlinear transformation. The
following body areas are employed as the gesture end points: left shoulder,
right shoulder, back pocket, and back of the head. DMPs allow for modeling each
gesture trajectory and they are better able to account for the variability
inherent in individual hand gestures. An evaluation using a handheld device with
a 3-DoF linear accelerometer shows that it is possible to learn models that
simulate and classify hand gestures even from single examples.

Action recognition (i.e., the understanding of what a human or another robot is
doing) is essential for robot interaction. \cite{parlaktuna2012closed} build
upon preceding work that employs DMPs for online action recognition
(\cite{akgun2010action}) by adapting DMPs to both recognize and generate robot
actions as follows. First, action states are recorded from multiple human
demonstrations. Then, a modified DMP formulation is used to acquire a new
behavior via the mapping from the position and velocity of the robot's end
effector to a function approximator. One function approximator is learned for
each action irrespective of the goal position. Finally, the weights are learned
via LWPR. The closed-loop primitive system is evaluated using an interactive
game that can be played between a human and humanoid robot.

Safe HRI scenarios require that a change in the robot's trajectory is made as
soon as an obstacle is detected. To handle this situation,
\cite{kulvicius2013interaction} develop a method that allows online alteration
of robot trajectories. In detail, a tightly coupled dual-robot system is
constructed to learn an adaptive sensor-driven interaction based on DMPs. The
behavior and cooperation of the robots is based purely on low-level sensory
information without any advanced planning. Therefore, the robots' roles do not
have to be explicitly defined beforehand, which leads to an alternating
leader/follower architecture. The trajectory planning and online modification
are done in the task space, taking into account collision avoidance only for the
robot's end effector. The system is tested in an HRI experiment involving a
robot handing off a tray of bottles to a human.

Motivated by \cite{strachan2004dynamic}, \cite{liu2014visual} devise a dynamic
hand gesture recognition approach for HRI applications. To characterize the
spatio-temporal variance in gestures, detected gestures are represented by DMPs.
Furthermore, an adaptive DMP learning method is proposed to deal with the
diversity and noise in different gestures as follows. First, a learned weight
vector shapes the attractor landscape of the DMP and modulates the variance. The
weight vector compactly encodes the spatio-temporal information of the original
gesture and naturally serves as a feature vector. Then, a support vector machine
(SVM) is constructed to classify new gestures. An experimental evaluation is
conducted on a recognition task using nine classes of human gestures. Related
work on the topic of hand gesture recognition using DMPs is presented by
\cite{wang2015study}.

Learning a wide range of complex hierarchically-organized tasks poses
significant challenges. To this end, \cite{duminy2016strategic} present a
cumulative learning system that makes hierarchical active decisions on how to
learn using previous experiences to drive learning. The socially-guided
framework with intrinsic motivation leverages interactive learning alongside
autonomous goal-babbling, where goals are prioritized according to ease of
reach. Experiments conducted using a humanoid robot show that the system learns
several motor skills modeled using DMPs. The proposed algorithm first focuses on
easy tasks before moving onto more difficult tasks. Lastly, the framework
consistently selects the best strategy with respect to the desired outcome,
learning to actively request social guidance from the teacher.  

Within the field of HRI, policy search requires either good heuristics or
initializations, particularly with complex robot learning tasks. Human
demonstrations may be used as a starting point for the policy search process.
Alternatively, \cite{schroecker2016directing} introduce an approach for
utilizing via points provided by a human to interactively guide DMP-based policy
search. By doing so, the method restricts the trajectory search space, thereby
significantly reducing the number of training samples. Using only a minimal
number of demonstrations, experiments on a simulated robot arm show efficient
learning of an object-insertion task. Furthermore, experiments on a letter
reproduction task demonstrate that the methodology can accurately reproduce
letter sequences. 
  
Not only is post-stroke rehabilitation demanding for patients, but it is also a
costly dilemma for healthcare systems. Based on this observation,
\cite{nielsen2017individualised} present an approach for the training of upper
extremities after a stroke by using a robot arm and DMPs as follows. First,
prerecorded trajectories are utilized to construct DMPs that act as basis
exercises via interaction learning (\cite{kulvicius2013interaction}). Force
feedback is used for online modification of the trajectories. The basic set of
primitives include translation of the arm, rotation of the wrist, and curved
movement. Then, the learned DMPs can be modified into individualised and
adaptive rehabilitation exercises that fit within the patient's physical
capabilities. Simulation results highlight the generation of novel trajectories.

DMPs can provide a way to assist kinesthetic teaching, a domain where full
cooperation and automation is challenging to achieve. For example,
\cite{caccavale2019kinesthetic} introduce a natural teaching framework to
address the difficulties related to the execution of structured tasks. To do
this, DMPs are leveraged to enable the teaching of tasks structured as behavior
trees. Concretely, during kinesthetic teaching, the approach annotates one-shot
demonstrations into segments. Concurrently, an attentional system relates the
generated segments to the task structure thus exploiting attentional
regulations. Using a robot arm, the kinesthetic framework is evaluated on a
human-robot co-working scenario and demonstrates that the robot is effective at
learning and executing structured tasks. 

Integrating collaborative robots into a laboratory requires safe and precise
manipulation in proximity to human operators. As an example,
\cite{baumkircher2022collaborative} create a framework that establishes the
feasibility of employing kinesthetic teaching to learn bacterial colony picking
tasks with sub-millimeter accuracy. This is done by using DMPs to generalize
tasks shown by a human operator. The system incorporates a custom end effector
that combines an RGB camera, 2D laser profiler, force sensor, and a sterilizable
needle for accurate colony localization. Experiments on 56 colonies demonstrate
an overall identification error rate of almost 11\%, which is comparable to
manual deposition. 

Mobility assistance and object manipulation are critical challenges when
developing robotic systems to assist the elderly in living independently. To
this end, \cite{ding2022intelligent} present a wheeled mobile manipulator that
uses cooperative DMPs to learn a user's movement pattern where the user may be
elderly, disabled, or able-bodied. Variable admittance control is adopted to
detect the walking intent of the user while cooperative DMPs model and adapt the
manipulator according to the user's walking pattern. To support object
manipulation, the demonstrator starts by collaborating with the user to execute
the task. Subsequently, the manipulator takes the place of the demonstrator to
assist the user. To learn and reproduce the demonstrator's experience, the
approach utilizes a GMM and GMR, respectively. Experiments using a four-wheeled
omnidirectional mobile manipulator show that a user could save up to 54\% of
their power and 65\% of their energy.

Instead of relying on visual cues, contact forces may also be used to interpret
user intention during HRI. For instance, \cite{lai2022user} propose a framework
that extends IPs (\cite{amor2014interaction}) and infers user intent using only
interaction forces in a human-robot dyad. Concretely, the parameters for robotic
motion are inferred without the robot endpoint position, conditioning only upon
partial observations of the interaction forces. To do this, DMPs are utilized to
encode robot trajectories while dense force data is integrated into the IP's
parameter distribution to generate motions consistent with the user's
intentions. Real-world experiments on a 7-DoF arm demonstrate that the framework
successfully infers user intent during reaching and obstacle-avoidance tasks.

Prior DMP research has focused on point-to-point skill learning where learned
skills are generalized to the same tool or manipulator. In contrast,
\cite{lu2022novel} develop a DMP-based framework for the use of multiple tools.
Specifically, two types of skills are learned: an object-effective skill and a
state-switching skill. The object-effective skill moves an object along the
intended trajectory under the influence of the tool. The idea is to acquire a
basic movement that can be adapted based on tools with different shapes and
necessary force ranges. Conversely, the state-switching skill considers the
contact points of the tool and the object in order to reduce the possibilities
of conflicts. Thus, a learned state-switching skill can be generalized to tools
with new shapes or contact regions. Experimental results using a haptic device
demonstrate the integration of the proposed skills.

For semi-automated robots working in remote environments, ensuring the
teleoperation performance of beginner operators is important, especially for
intricate tasks requiring refined motions. In light of this importance,
\cite{luo2023vision} create a vision-based robot learning method using a virtual
fixture, a perceptual overlay that can help the operator perform safe and
precise movements. The approach leverages the motions of a tutor to train a
novice teleoperator as follows. First, the tutor's manipulation skills are
learned via DMPs. Then, the DMP-generated trajectories are used to train the
novice. Lastly, a virtual fixture generates the force selection based on
position errors, providing guidance and allowing the trainee to finish the task
at hand.

Wearable sensing devices can be employed to achieve direct control of robots
during HRI. For example, \cite{odesanmi2023skill} develop a skill learning
framework that utilizes online human motion data acquired using wearable sensors
as an interactive interface for providing the anticipated motion to a robot in a
user-friendly way. More specifically, LfD is performed to teach the robot
fragile/non-fragile surface interaction and machining tasks. To realize
efficient and accurate HRI, DMPs are used to reproduce the human demonstrations
on the robot. An experimental evaluation is conducted via remote teleoperation
using wearable sensors to control a 6-DoF collaborative robot manipulator. The
results show that the trajectories learned by the proposed method are better
than those obtained by ProMPs. 

Existing work has treated teleoperation and LfD as independent paradigms,
leaving their synergistic potential in addressing shared task requirements
underexplored. Based on this observation, \cite{shi2025exploring} focus on task
precision as a unifying objective. Concretely, a mutual relationship between LfD
primitive arrangements and teleoperation scaling mechanisms is established
through their shared precision optimization objectives. To do this, an adaptive
teleoperation scaling framework is developed by integrating operator
physiological states using EMG signals and task-specific precision requirements.
Additionally, improved DMP trajectory fitting accuracy is realized by
integrating a scaling ratio. Two experiments involving a ring transfer task and
letter writing are conducted to test the effectiveness of the methodology. 

\paragraph{Object handover.} 
\cite{prada2012dynamic,prada2013dynamic} present an analysis of a modified DMP
model for an object handover HRI task. The model enables explicit control of the
transition between the feed-forward and feedback parts of the system. It also
accounts for moving goals. The method is benchmarked using a set of motion
capture sequences of two humans exchanging an object in various configurations
in an industrial scenario. For each configuration, the performance of the system
is repeatedly evaluated by training the desired motion with one movement
sequence and testing the system responses against the rest of the samples in the
same configuration using the human motion as the ground-truth data. The control
law and DMP-based controller implementation is further explained in proceeding
work \cite{prada2014implementation}.

The perception of HRI is an essential component of developing service robots as
the acceptance of such robots into everyday life will be determined by the
experience of the users. To this end, \cite{koene2014experimental} present a
DMP-based prototype system designed for fluent human-robot object handover
interaction. The work focuses on the handover of a torch (flashlight), hammer,
and hacksaw. Using a robot arm mounted with an anthropomorphic hand, an
evaluation is conducted where both the human and the robot are tracked within a
magnetic motion tracking system. A series of robotic manipulations are judged
during the handover according to user satisfaction, comfortability, ease, and
safety. The results show that user satisfaction was primarily determined by the
temporal aspects of the object handover interaction. 

Human motions are not completely repeatable. Therefore, HRI trajectories need to
allow for modulation via external sensory feedback. For instance,
\cite{gams2014couplingb} show how to couple DMPs on both the velocity and
acceleration levels for HRI tasks. In particular, ILC is utilized to achieve a
desired force contact behavior. An experimental evaluation demonstrates the
coupling of two robots and an operator to execute the task of closing a box.
Nonetheless, the admittance force control used in the proposed methodology has
limitations, the most serious being that the environment stiffness can directly
affect the force feedback gain. In follow-up work, \cite{nemec2014human} perform
force adaptation using an impedance control law, which decouples and linearizes
the robot dynamics at the torque level.

For seamless human-robot handover, a robot must predict where and when the
handover will take place and reach that point with its end effector sufficiently
fast to receive the object in a human-like manner. \cite{widmann2018human}
perform such a prediction by using prior knowledge on how humans move during a
handover. The approach utilizes DMPs as a representation of human trajectories
learned from demonstration with an extended Kalman filter. Specifically, the
parameters of the nonlinear dynamical system describing the handover place and
time are estimated online for handovers similar to the demonstration. The
estimated parameters are then used to design an adaptive controller to control
the robot in human-robot handover tasks. Simulations are performed to validate
the method on 1D trajectories, predict 2D motion, and evaluate robustness of the
predictor to modeling errors.

During collaborative object transfer, a physical coupling is established between
human and robot partners through which interaction forces emerge.  Disagreements
between intentions with respect to the target position and the desired time
duration can result in higher interaction forces. \cite{sidiropoulos2019human}
consider the problem of object transfer where the robot is aware of the pattern
of motion, yet it is agnostic to the target position and how fast the movement
should be executed. Concretely, a model reference with a DMP-based control input
that uses the desired pattern of motion along with estimates of the target
position and time scaling is proposed. An EKF-based observer estimates the
target position and time scaling with fading memory and parameter projection
such that the bounds originating from the physical interpretation are respected.
Estimates are based on the interaction force and used in the control signal
supplied to the robot. The methodology is extended in
\cite{sidiropoulos2021human} to include not only position, but also target
orientation.

Context-aware task execution is critical for enabling assistive service robots
to adapt to human-oriented tasks such as object handover. To this end,
\cite{abdelrahman2020context} propose an approach that builds on a hierarchical
apprenticeship learning approach using DMPs and a contextual REPS algorithm.
The method involves collecting human task demonstrations to encode motion
trajectories with DMPs, defining context variables like posture and obstacle
presence, and learning policies through a linear-Gaussian hierarchical
structure. By leveraging simulations to learn contextual handover skills before
transferring the policies, the approach allows a robot to adapt to human
postural contexts. The effectiveness of the context-aware behavior is validated
through a user study, where participants preferred the adaptive handover skill
over non-adaptive alternatives due to its perceived naturalness and suitability.

Although a robot can easily acquire motion skills via LfD, it becomes much more
difficult to learn compliant skills. \cite{zeng2021learning} handle this issue
using a two-stage approach. In the first stage, a human teacher demonstrates to
the robot how to perform a task whereby only motion trajectories are recorded
without the involvement of force sensing. To generate human-like motion, DMPs
are used to encode the kinematics data. In the second stage, inspired by
neuroscience findings in human motor learning, a biomimetic controller is
employed to obtain the desired robot compliant behaviors by simultaneously
adapting the impedance profiles and feed-forward torques in an online fashion.
The methodology is verified in three ways: simulations, a handover task with a
robot manipulator, and a human–robot collaboration sawing task.

Object handovers in collaborative tasks remain challenging due to disparate
positions and timing. To address this problem, \cite{wu2022adaptive} present a
DMP-based framework that learns handover motions from human demonstrations and
adapts online to human feedback. The key contributions include the following:
(i) uncertainty-aware learning with GPR, (ii) a state-dependent weighting
function for shape and goal attraction terms, (iii) orientation-based spatial
scaling, and (iv) online parameter adaptation. In addition, an interaction model
combining physical dynamics and probabilistic learning to study human-robot
handover coordination is introduced. Experiments conducted on a 7-DoF robot show
improved success rates and fluency compared to standard DMP approaches,
particularly for dynamic and passive handover scenarios, though performance
slightly decreased for normal handovers.

\cite{cai2023probabilistic} develop a framework to model human movements in
manufacturing scenarios and predict future motion. Specifically, the focus is on
modeling and predicting human-hand motion during object transfer, under the
assumption that the movements are target directed. A probabilistic DMP method
(\cite{meier2016probabilistic}) is implemented to recognize goal-directed
movements in the reaching phase and make online predictions. The probabilistic
representation allows for capturing uncertainty, and jointly estimates the
posterior distribution of the states and DMP weights via online observations. To
avoid frame-dependency, rotation and magnitude scaling is used, which allows for
the generalization of learned motion patterns. Experimental results show
satisfactory performance on hand-motion prediction in the presence of motion
variations, and generalization to tasks beyond those demonstrated. 

Object handovers require the coordination of joint movements from both humans
and robots. Perturbations that interrupt handovers can complicate this process.
\cite{iori2023dmp} alleviate this problem by developing an online trajectory
generation method based on DMPs. The approach allows a robot manipulator to
adapt to human motion through minor disturbances to the partner's trajectory. To
demonstrate the capability of the controller with different parameter settings
and against a non-reactive implementation, a qualitative analysis is performed
via a questionnaire provided to participants as part of a randomized trial. The
results show that subjects favored interactions with the proposed system.
Moreover, the method was shown to significantly increase the subjective
perception of the interaction with no statistically significant deterioration in
task performance. In follow-up work, \cite{perovic2023adaptive} enable humans to
optimize toward preferred behaviors, which results in more natural and intuitive
object handovers across two different scenarios.

DNNs can be trained to predict future poses and used to improve the dynamics of
cooperative tasks, which is crucial to ensuring safety in HRI scenarios. For
instance, \cite{mavsar2024simulation} develop a recurrent neural architecture
that is capable of transforming variable-length input motion RGBD videos into a
set of parameters that describe a robot trajectory. It allows for predicting
object handover trajectories after only receiving a few frames. To exploit the
network's predicted robot receiver trajectories, a motion generation system
based on third-order DMPs and quintic polynomials is designed. This enables
smooth switching, up to second-order derivatives, between the trajectories. The
accuracy of the giver and receiver trajectories is evaluated through
robot-to-robot handover tasks.

An important aspect of HRI is human-aware motion planning, which involves
determining the sequence of movements a robot must execute to accomplish a given
task. \cite{franceschi2025human} explore this area of research via an online
personalized motion planner based on DMPs. Although DMPs directly allow for
online trajectory modification, the framework decouples the motion planning and
execution step. This ensures that the executed trajectory is as collision-free
as possible, and that the online adaptation is limited to scaling the execution
along the predefined path. In addition, a velocity scaling law is developed to
make the interaction more comfortable for the human partner. The performance of
the system is measured by capturing stress-related metrics
(electroencephalography and electrodermal activity) on an object handover task
with a robot arm.

\paragraph{Virtual reality.}
\cite{lentini2020robot} develop a framework to quickly teach robots in an
intuitive way without programming. The methodology utilizes immersive
teleoperation via VR to facilitate the interaction between the operator and
robot. To do this, a tele-impedance approach is used to control the robot pose
and Cartesian stiffness at the end effector. A depth camera captures a point
cloud of the scene, which is then filtered to obtain the region of interest.
Next, the position, orientation, end-effector state, and impedance profile are
recorded. Finally, the hand closure and stiffness profile are obtained using
DMPs. Insertion and multiple object sorting tasks highlight the generalization
of the learned trajectories, the impedance profile, and the hand closure
signal.

Assessing the feasibility of a new robotic component, adding more collaborative
robots into an existing infrastructure, etc., is a significant challenge in
real-world deployments. To address this issue, \cite{tram2023intuitive} develop
a framework that enables natural HRI through a virtual representation of the
robot workspace. The system allows humans to interact with multiple
collaborative robots in realistic scenarios for specific use cases (e.g., LfD).
The framework consists of the following components: a VR head-mounted display,
rendering software, and a backend for robot simulation and planning. System
evaluation is performed by using DMPs to encode virtual demonstrations of
pick-and-place tasks, which are then executed on a physical robot arm.

\subsubsection{Motion Modeling}
\paragraph{Control policy generation.}
In bimanual manipulation tasks, the trajectories of both robot arms must be
temporally synchronized while satisfying certain spatial constraints. This
requires a high level of coordination between the two arms performing the task.
To solve this problem, \cite{thota2016learning} learn the dynamics of the task
from demonstration data using DMPs. Concretely, the dynamics of only one arm are
modeled via a set of DMPs. The learned set of DMPs is then utilized as the
desired trajectory for one arm, while a transformation is applied to obtain the
ideal trajectory for the second arm. Control laws are designed to provide
trajectory tracking and synchronization of the multi-arm system using
contraction analysis. Two experiments involving a bimanual manipulation task are
performed using the learned DMPs and the proposed control laws.

A robot with large DoF can better comply with and move in complex environments.
However, possessing many DoF is only an advantage if the system is capable of
coordinating them to achieve its desired goals. To this end,
\cite{travers2016shape,travers2018shape} fuse planning and control with a
coherent middle layer that adds robustness to and decreases the complexity of
coordinating many DoF during robot locomotion. Shape functions are introduced as
the element that helps define the middle layer of the articulated system.
Admittance control is specified in terms of shape parameters, which are used to
define shape functions. By mapping joint torques into equivalent forces on
shape, the shape-based compliant controllers autonomously adapt the robot's
shape to improve compliance during locomotion. These derived controllers can be
interpreted as special cases of DMPs. Experimental comparisons against CPG-based
and torque-modified CPG controllers are presented on snake-like and hexapod
robot platforms.

\cite{kim2022learning} apply PDMPs to cooperative manipulation tasks using
mobile robots. Instead of encoding multiple demonstrations for entire complex
tasks, the approach segments the overall task into sub-tasks, where each
sub-task is represented by a PDMP. A phase decision process is employed to
sequence the sub-tasks, which enables the PDMPs to be reconfigured online. The
framework also integrates GPR to generalize execution time and regional goals
for each phase. This effectively reduces the number of required demonstrations
for complex tasks while retaining scalable behavior. An experimental evaluation
demonstrates a hang-dry task where two mobile manipulators must maintain a
stretched cloth as they navigate obstacles.

To enhance learning efficiency and control performance, \cite{li2024efficient}
integrate RL and DMPs. Specifically, the approach combines a DMP-based policy
into an actor-critic framework using a deep deterministic policy gradient
algorithm. This is done by deriving the corresponding update formulas to learn
the networks that properly decide the parameters of the DMPs. In addition, an
inverse controller is developed to adaptively learn the translation from
observed states into various robot control signals via DMPs, eliminating the
requirement for prior human knowledge on controller design. The methodology is
evaluated on five robot arm control benchmark tasks highlighting the smoothness
of actions for complex control applications.

\paragraph{Human-motion reproduction.}
Many daily activities are composed of rhythmic patterns of behavior (e.g.,
walking, swimming, writing, etc.). Based on this insight,
\cite{lantz2004rhythmic} generate and model hand gestures with a mobile device
using DMPs. Receptive field weighted regression, an incremental function
approximation method that assumes the functions are linear sums of nonlinear
kernel functions, is used to determine the DMP parameters. The algorithm learns
the number of kernels, the kernel parameters, and the linear regression
parameters at the same time. Incremental learning of new gestures is done by
learning the linear regression parameters independently for each kernel
function. An evaluation is conducted by performing two sets of ten different
gestures using a handheld device equipped with an accelerometer.  

Gait analysis is the systematic study of human walking, which can be used make
informed diagnoses and plan optimal treatments. Temporal EMG profiles are
typically recorded from patients during walking for gait analysis. However, they
can change significantly with speed and make it difficult to distinguish among
gait patterns. \cite{xu2006internal} resolve this problem by developing a method
to model and classify gait patterns based on an internal model. The model
consists of two sets of differential equations (i.e, DMPs) with two single
hidden layer feed-forward networks in between them. The DMPs are used to
describe either discrete or rhythmic movements. The neural network allows the
internal model to capture nonlinear behavior in a gait pattern. Case studies on
speed dependent gait patterns are conducted to verify the effectiveness of the
approach.

When repeatedly executing reaching movements, humans develop stereotypical
strategies for grasping objects in the presence of obstacles.  Based on this
observation, \cite{stulp2009compactb} analyze human reaching trajectories
collected with a magnetic motion tracker to identify when obstacles influence
these motions and what strategies humans use to avoid them. Through
dimensionality reduction and clustering, four principal trajectories are
identified: a default trajectory and three obstacle avoidance strategies (going
over, left, or right of obstacles). These compact motion models are implemented
on a robot manipulator using DMPs to demonstrate that the robot can both
reproduce the principal trajectories and generate novel motions through
interpolation. The results show that the compact models capture natural human
motion patterns as the robot's executed trajectories closely match observed
human movements.

Due to the dissimilar dynamic and kinematic properties of human and robotic
mechanisms, there is no direct transfer of movements from one to the other.
Thus, the resulting movement of the robot will likely not accomplish the same
task. For instance, recorded joint movement of humans when squatting directly
copied to a humanoid robot will result in the robot tipping over. To this end,
\cite{gams2011constraining} show how DMPs can be used in a modified prioritized
task control method to implement the reflexive movement of a robot. In this
scenario, higher-priority movement acts as reflexive movement and only takes
over when the desired movement approaches a given threshold. A zero-moment-point
criterion is maintained by constraining the recorded movement when mapping it to
the robotic mechanism. The proposed algorithm is experimentally evaluated on
squatting movements using a custom-developed leg robot. 

A natural way to learn human movement is to teach constituent elements of basic
movements in isolation and then chain them together into complex movements.
Based on this insight, \cite{meier2012movement} develop a sequential movement
segmentation and recognition framework via DMPs. Concretely, an EM algorithm is
devised to estimate the open parameters of a DMP from a library using an
observed piece of the trajectory as input. If no matching primitive in the
library can be found, then a new primitive is created. This process allows for a
sequential segmentation of an observed movement into known and new primitives,
which can then be utilized for bootstrapping autonomous learning robots through
IL. Data collected from human movements and synthetic examples are employed to
demonstrate the methodology. 

To improve the generalization of DMPs in task space, \cite{chen2015efficient}
embed DMPs in deep autoencoders. In detail, deep autoencoders are first used to
learn a nonlinear dimensionality reduction of the data. Then, a representation
of movement in a latent feature space can be found in which a DMP optimally
generalizes. The proposed architecture allows for training the system as a whole
unit. Sparsity is added to the feature layer neurons to further improve the
model for multiple movements. This allows for the generation of new movements
that are not in the training data by switching on/off or interpolating one
hidden neuron. Experiments are conducted using 50-dimensional human-movement
data. \cite{chen2016dynamic} extends this work by embedding DMPs in a
time-dependent variational autoencoder's latent space, which improves
high-dimensional movement generalization. 

The HMM is a ubiquitous tool for modeling time series data. Nonetheless, the
working principles of HMMs and DMPs are very different. A DMP takes position,
velocity, and acceleration of a trajectory into account. Conversely, an HMM uses
only trajectory position values. To better understand the capabilities of HMMs
and DMPs, \cite{pehlivan2015dynamic} provide a systematic comparison between the
two methods using human-movement data. In addition to performance measurements,
the number of adaptable parameters that are used in each method and the
processing time are compared. The methodologies are tested against each other on
a recognition task involving human generated letter movements. The results show
that HMMs outperform DMPs, with a possible noise robustness advantage for DMPs.
Concretely, DMPs have a 78\% success rate compared to an 83\% success rate of
the HMMs on the non-modified, noiseless human movement data.

GPs can provide a measure of uncertainty in robot movements.
\cite{fanger2016gaussian} exploit this fact by developing a cooperation
framework for DMPs. To do this, a combination of GPs with DMPs is employed.
Concretely, the approach learns guaranteed converging point-to-point movements
from human examples, while exploiting the uncertainty information encoded in a
GP in terms of predicted variance. Logarithmic transformation of the GP input
data is performed to guarantee stability. This provides a measure of confidence
at each phase of a robot's motion. Furthermore, a dynamic leader-follower scheme
based on cooperative DMPs (\cite{umlauft2014dynamic}) is developed. Simulations
and robot experiments show the ability to learn cooperative tasks, including
role assignment from demonstrations. 

Variable impedance actuators store a significant amount of energy that can be
used for link speedup. However, modelling the way human motions exploit the
ability to store and release potential energy, via human muscle or tendons in
viscoelastic joints, is not possible with data-driven techniques. To address
this challenge, \cite{haddadin2016optimal} propose a DMP-based optimal control
framework for producing explosive motions based on prior work that studies
maximizing link velocity for elastic joints. The framework consists of computing
a sample set of optimal trajectories, encoding these trajectories into DMPs, and
then leveraging metric-based weight interpolation to generalize motions to
unforeseen scenarios. Experimental results show that the framework performs well
in both simulation and on an anthropomorphic robot, achieving near-optimal
motions in real-time.

In many cases an accurate kinematics model of a robot is not available (e.g.,
when the robot specifications are not released by the manufacturer). To confront
this dilemma, \cite{hiratsuka2016trajectory} develop an LfD system by learning a
kinematic mapping between a human model and a robot. The method first requires a
motion trajectory obtained from human demonstrations using a depth sensor, which
is then projected onto a corresponding human skeleton model. Then, the
kinematics mapping between the robot and human model is learned by employing
local Procrustes analysis, a data-driven technique based on manifold mapping,
which enables the transfer of the demonstrated trajectory from the human model
to the robot. Lastly, the transferred trajectory is modeled using DMPs, allowing
for real-time reproduction. Simulation results using a 4-DoF robot highlight the
imitation of various skills (e.g., writing/drawing) demonstrated by a human.

When training an LfD system, humans will naturally provide different
trajectories over multiple demonstrations for the same task. This can cause the
demonstrated movements to be more precise in some phases and less precise in
others. \cite{umlauft2017bayesian} address this issue via a Bayesian approach,
specifically GPR, that models uncertainties from the training data and then
infers them in regions with sparse information. This allows for precisely
controlling the DMP trajectory behavior in regions far away from the training
data while the appropriate Gaussian kernel choice provides smoothness of the
inferred uncertainties. A simulated and real-world evaluation using a 2-DoF
robot manipulator shows an increase in performance in terms of precision when
compared to a GMM.

As a trajectory encoding method, DMPs are widely used in motion synthesis and
generation. Nonetheless, they can also be applied to human action recognition.
For instance, \cite{zhang2017robust} use inertial measurement unit (IMU) data to
extract dynamic features from trajectories within a DMP framework. These
spatial-temporal invariant features are then used to solve recognition tasks
using k-nearest neighbors classification. However, the features can experience
drift in the time dimension, which reduces classification accuracy. To mitigate
this drift, a fast algorithm that combines DTW and k-nearest neighbors is
proposed. The method decreases the time complexity by reducing the search space
and length of the data. A series of comparative experiments on human recognition
and handwritten letter datasets highlights the performance of the algorithm.

The efficient ordering of motor skills can improve the efficacy of learning and
transferring such skills. For example, \cite{cho2018relationship} present a
framework to learn motor skills based on motion complexity as follows. First,
several human demonstrations are clustered and each motor skill is assigned a
task. Next, the spatial and temporal entropy of the motion trajectories is
utilized to compute motion complexity. To learn the ordered motion skills, the
reaction forces and moments are represented as HMMs while the control signals
are described as DMPs. The experimental findings show that motion complexity is
proportional to shape complexity. Furthermore, the evaluation demonstrates
use-cases that determine when to choose between the complex-to-simple and
simple-to-complex ordering criteria.

Effective motion planning of high-DoF deformable objects requires the management
of high-dimensional configuration spaces while satisfying dynamics constraints.
To this end, \cite{pan2018realtime} propose a two-stage algorithm to learn
various locomotion skills as follows. First, a multitask controller is
parametrized using DMPs, encoding the deformable object's movements. Then, a
neural network controller is trained to select a DMP for navigation through the
environment where avoiding obstacles is required. A finite element method
together with model reduction and contact invariant optimization are leveraged
to efficiently optimize the DMP's parameters. Finally, the neural network
controller's parameters are optimized using deep Q-learning. An evaluation
demonstrates real-time navigation of a deformable body using a trained DMP
controller on tasks including swimming and walking. 

\cite{wang2020robot} create an LfD framework that involves a teaching, learning,
and reproduction phase. The system consists of an adaptive admittance controller
that takes into account unknown human dynamics. The task model in the controller
is formulated via GMR to extract human motion characteristics. In the teaching
phase, an operator demonstrates how to perform the task and the robot's motion
is recorded. During the learning phase, an adaptive admittance controller is
employed to allow the operator to smoothly guide the robot during the
demonstration. DMPs are used to model the motion. In the reproduction phase, an
RBFNN controller is developed to achieve accurate motion reproduction. A set of
robot experiments are conducted to validate the controller, admittance model
estimation, and motion generalization. 

Motion retargeting can be used to efficiently adapt sign language motions to
robots. Nevertheless, to cope with dual-arm gestures, a motion retargeting
scheme that can address differences in body structure, dual-arm coordination,
motion similarity, and feasibility constraints is needed.
\cite{liang2021dynamic} address these problems through DMPs and graph
optimization as follows. First, parameterization of the dual-arm state using
wrist pose and elbow position is performed to handle variations in body
structure. Next, DMPs are learned in a leader-follower manner to allow relative
movements between elbows and wrists to be retained for new trajectories. Lastly,
an optimization procedure is employed to search and track deformed reference
trajectories for collision avoidance. The method is tested using a 26-DoF
dual-arm robot with multi-fingered hands on retargeted sign language motions. 

Incremental learning-based DMPs can continuously learn skills from new
demonstrations, however they cannot generate stylistic/reactive skills.
\cite{lu2021incremental} mitigate this issue through an incremental skill
learning framework based on DMPs, fuzzy logic, and the idea of a broad learning
system. This is accomplished via a system that acquires nonlinear reactive skill
terms with fuzzy sets and rules about shapes, locations of the obstacle, human
operational strategies, and muscle stiffness to create suitable reactions for
varying environmental conditions. The skills can then be generalized by changing
not only the start, end, and scaling parameters of a robot arm, but also
strategies, moving directions, and control impedances. To demonstrate the
method, an obstacle avoidance experiment with barriers is conducted using a
modified joystick controlled by EMG signals via the operator's limb.

\paragraph{Trajectory generation.}
To be successful, robot learning cannot be limited to direct imitation of
movements obtained during training. Instead, it must enable the generation of
motions in situations that a robot has never encountered before. Inspired by
motor-tape theories in which example movement trajectories are stored directly
in memory, \cite{ude2010task} propose a methodology for the generalization of
example trajectories to scenarios not observed during training using DMPs. This
is done by associating every example trajectory with parameters that describe
the characteristics of the task (e.g., its goal), which then serve as query
points into a trajectory database. A series of simulations (reaching) and
physical robot experiments (grasping, ball throwing, drumming) demonstrate the
generalization of the approach. While this method takes into account external
perceptual feedback to generalize example movements to different situations, its
computational cost is prohibitive for use in a real-time feedback loop.
Follow-up work by \cite{forte2011real,forte2012line} address this concern
through an approach that is efficient enough to be applied in such a loop. In
addition, \cite{kramberger2016generalization} propose a method for task-specific
generalization of orientation trajectories encoded CDMPs. \cite{zhou2017task}
further improve the generalization ability through LWR, a task-oriented loss
function, and modification of the model switching algorithm. 

\cite{ning2011accurate} modify the DMP framework to enable precise boundary
condition handling at both the start and goal positions while maintaining smooth
transitions. The approach introduces a second-order system combined with
Gaussian kernels for path modulation and a specialized boundary function
generator using polynomials and linear segments to guarantee boundary condition
satisfaction. In \cite{ning2012novel}, the framework is validated on several
robot manipulator simulations to demonstrate the advantages over spline-based
methods including: (i) natural handling of sharp corners, (ii) built-in
filtering through the second-order dynamics, (iii) generation of complex
trajectories without manual transition zones. Although the method achieves high
positional and velocity accuracy at goal locations, it sacrifices the
time-independence of DMPs.

\cite{nemec2012action} propose techniques for sequencing DMPs by focusing on
continuous transitions (up to second-order derivatives) between consecutive
primitives. This allows for smoother transitions between DMPs by solving the
issue of achieving continuous acceleration, which is critical in tasks involving
rapid or precise movements. To do this, two methodologies are introduced. The
first method employs third-order DMPs, with an appropriate initialization for
system variables, to allow for continuity in position, velocity, and
acceleration. The second method performs an online adjustment of the weights of
the Gaussian kernel functions in standard second-order DMPs to maintain
continuous acceleration during transitions. Experiments are conducted both in
simulation and using a robot arm on a range of tasks, including pouring, table
wiping, and carrying a glass of liquid. 

Morphed nonlinear phase oscillators such as rhythmic DMPs can be utilized for
modeling periodic movements such as quadruped locomotion. For example,
\cite{ajallooeian2013general} introduce an approach based on using phase-based
scaling functions to morph the limit cycle of an existing phase oscillator.
This methodology comprises a general family of nonlinear phase oscillators that
is a superset of proportional-derivative control and rhythmic DMP methods. The
morphed oscillators can be applied whenever it is necessary to encode a periodic
motion or pattern as a limit cycle of a dynamical system. Applications of this
methodology include modeling feline locomotion, imitating periodic tasks such as
drum-playing, neurorobotic applications (e.g., swimming, walking), and
model-free tracking for rehabilitation robotics.

The need to acquire numerous demonstrations in order to generalize example
trajectories to new situations is a major hindrance in LfD systems.
\cite{denivsa2013discovering,denivsa2013new,denivsa2013synthesizing,
denivsa2015synthesis,denivsa2017cooperative} solve this dilemma via an approach
that uses motion graphs, binary trees, and hierarchical search to generate
movements that were not directly demonstrated by the teacher. Specifically, a
database is utilized to find new connections between nodes and therefore novel
movements. If the path at the desired level does not exist, then a top-down
hierarchical database search is performed to find optimal partial paths. DMPs
are employed to combine the paths into smooth and continuous trajectories from
the given start-state and end-state vectors according to the desired level of
granularity. Simulated and physical experiments using a robot arm highlight the
ability to discover unique sets of movements for arbitrary configurations.

Estimating the parameters of a dynamical system is problematic when the provided
demonstrations are relatively sparse. This can lead to unexpected and
undesirable behaviors in the unexplored parts of the state space.
\cite{krug2013representing} address this complication for DMPs by formulating
parameter estimation as a nonlinear optimization problem instead of the commonly
used linear approximation. The approach lets DMPs learn separate dynamical
systems corresponding to multiple demonstrations. Not only does this reduce the
number of parameters necessary to achieve a good fit to the provided
demonstrations, but it also allows for better capturing a motion's actual
underlying dynamics. An assessment is conducted by parametrizing the proposed
model from demonstrations of robot grasping movements. In \cite{krug2015model},
the online optimization approach is extended to an MPC scheme.

Assembly operations may require a level of online trajectory adaptation due to
positioning inaccuracies and tight tolerances between objects. To solve this
problem, \cite{nemec2013transfer} develop an LfD approach that iteratively
adapts a learned trajectory to improve task performance. This is done via a
learning procedure that modifies demonstrated trajectories, encoded using DMPs,
such that the resulting forces and torques match the demonstrated ones. Adapting
the desired forces is accomplished using either an admittance/impedance control
law, where the robot motion is modified to reduce the force/torque error. To
ensure stability, the force adaptation is slowed down whenever the force/torque
error becomes large. The method is evaluated on learning peg-in-hole tasks for
workpieces with different geometries.

Due to differences between kinematics and dynamics, the direct transfer of
human-motion trajectories to humanoid robots does not result in dynamically
stable movements. \cite{vuga2013motion} undertake this problem by developing a
system that converts human movements, captured by an RGBD camera, into
dynamically stable humanoid movements. The proposed balance controller is based
on an approximate model of the robot dynamics, which is sufficient to stabilize
the robot during online imitation. Nonetheless, the resulting movements are not
guaranteed to be optimal since the dynamics are not exact. Therefore, the
initially acquired movement is subsequently improved via model-free RL where the
transferred motion is encoded using DMPs. An evaluation of the framework
highlights the capability to transfer human-walking patterns to a small-size
humanoid robot.

Elastic robots have the ability to mechanically store and release potential
energy, which makes them especially interesting for optimal control.  Yet,
solving any kind of optimal control problem for such highly nonlinear dynamics
is only feasible numerically (i.e., offline). To address this limitation,
\cite{weitschat2013dynamic} create a framework for executing near-optimal
motions for elastic arms in real-time as follows. First, a set of prototypical
optimal control problems is defined to represent a reasonable set of motions
that the arm can execute. Next, the optimal control problem is solved for some
of these prototypes in a roughly covered task space. Then, the resulting optimal
trajectories are encoded using DMPs. Lastly, a distance and cost function-based
metric forms the basis to generalize the learned parameterizations to a new
unsolved optimal control problem. Optimal throwing movements are experimentally
verified on an anthropomorphic hand-arm system using variable stiffness
actuation. 

Since MPs are typically small entities, a richness of motion can result from the
combination of them. \cite{lemme2014self} take advantage of this idea to
bootstrap new MPs by concatenating frequently used MPs to more complex ones.  In
detail, this consists of the following steps: (i) use an existing MP library to
decompose a complex trajectory and generate new training data for MP learning;
(ii) organize the new training data into training datasets for creating new or
refining existing MPs; (iii) consolidate the library by deciding which MP needs
to be deleted. The library holds additional information for each MP including
how frequently an MP was used for decomposition and how long it has been part of
the library. The bootstrapping cycle is demonstrated with DMPs via a robot LfD
scenario. Proceeding work focuses on the problem of segmenting demonstrated
trajectories into a library of MPs (e.g.,
\cite{lioutikov2015probabilistic,lioutikov2017learning})

In many applications (e.g., grasping an object, shooting a ball, etc.), distinct
goals require trajectories of different styles. \cite{zhao2014generating}
resolve this issue by determining how to reproduce a trajectory with a suitable
style. Concretely, a style-adaptive trajectory generation approach based on DMPs
is proposed where the style of the reproduced trajectories changes smoothly as
the new goal changes. To do this, the approach first adopts a point distribution
model to obtain the principal trajectories for various styles. Next, the model
of each principal trajectory is learned independently using DMPs. Lastly, the
parameters of the trajectory model are smoothly adapted according to the new
goal using an adaptive goal-to-style mechanism. The method is tested on an
adaptive shooting task and quantitatively compared against the original DMPs. 

Motion parameterization of DMP-controlled robotic movements leads to a
dimensionality problem where convergence of motion learning requires a
prohibitive number of experiments or simulations. \cite{colome2015friction}
propose three solutions to address this issue of dimensionality: (i)
prioritizing only the most significant directions during the exploration of the
parameter space, (ii) fixing a set of Gaussians to approximate a demonstration
and then utilizing a second set of Gaussians to optimize the trajectory, and
(iii) identifying the least significant DoF affecting task performance to update
via a joint coordination matrix. Experimental results show the first two
strategies result in fewer Gaussian computations and better performance.
Furthermore, the results demonstrate a significantly lower computational cost by
reducing the dimensionality of the exploration space. 

Various statistical methods have been developed to generalize DMPs to new robot
workspace configurations. However, these approaches can only be successful if
there is enough training data. To this end, \cite{forte2015exploration} propose
a technique for statistical generalization using a library of DMPs. The goal of
the generalization is to compute a trajectory that solves the desired task at a
given query point. Two different approaches to statistical generalization are
implemented: DMPs and GPR. For each query point, the generalization function
computes the corresponding DMP. This provides a low-dimensional parametrization
of the solution space, which can be exploited to accelerate the acquisition of
new example trajectories. Experimental results highlight the method's utility on
a feed-forward grasping task using a 7-DoF robot arm. 

LfD typically assumes full knowledge of the task parameters, which is
unrealistic in many real-world scenarios. \cite{alizadeh2016learning} address
this challenge by developing an approach that handles partially observable task
parameters during reproduction. The method first encodes demonstrated
trajectories using DMPs and pairs them with corresponding task parameters to
build a database. GPR then models the relationship between the DMP and the
parameters to generalize trajectory generation to new situations using only the
observable parameters. In experiments with a robot manipulator, a demonstrated
task involves passing through four coordinate frames (starting frame, reference
frame, randomly placed via-point frame, and target frame), where the via-point
location becomes unobservable during reproduction. Compared to using the
full-parameter set, where some parameters are unobservable during reproduction,
the results show that better generalization occurs when considering only the
observable parameters. 

Accurately modelling the robust, adaptive, and versatile motor control of
biological systems requires novel computations. To this end,
\cite{dewolf2016spiking} propose a spiking neuron model of a motor control
system. In particular, the model consists of spiking neurons in connection with
the anatomical areas of the premotor, primary motor, and cerebellar cortices.
The premotor cortex generates and models trajectories using DMPs. Additionally,
the model incorporates neural computations to adaptively control a nonlinear
three-link robot arm. Experiments show that the model is adaptive to either
changes in arm dynamics or kinematic structure, suggesting that the approach is
a possible model for changing body size or unexpected dynamic perturbations. 

Extending the velocity adaptation framework proposed by
\cite{nemec2013velocity}, \cite{nemec2016bimanual} develop a hybrid control
architecture for bimanual human-robot cooperation that combines speed-scaled
DMPs with a decoupled bimanual control scheme. The main innovation is a
stiffness adaptation method that varies based on both motion variance across
demonstrations and current execution speed. By defining a path-aligned
coordinate frame, different stiffness settings are enabled parallel and
perpendicular to the direction of motion. This improves precision while
maintaining compliance for human interaction. The technique is validated on a
dual robot manipulator system performing cooperative object transport tasks,
where it is able to adapt to changes in position goals and obstacles while
maintaining coordination between arms and end effector force constraints.

As state predictions become more accurate over time a robot's behavior needs to
be updated. However, state updates cannot be easily incorporated into most
trajectory generation solutions. To address this problem,
\cite{samant2016adaptive} modify DMPs by replacing the traditional exponential
canonical system with a piecewise linear system. This enables a more uniform and
faster learning of the DMP parameters. In addition, an adaptive learning
technique based on Lyapunov stability for updating the kernel weights, centers,
and widths is presented. The approach is validated on a robot manipulator for a
ball-hitting task, where an increase in accuracy compared to standard DMP
learning methods (including LWR and GMR) is demonstrated. The results show that
improved discrimination between learned trajectories is realized while fewer
kernels are required to achieve comparable accuracy. 

\cite{tan2016applying} develop an offline algorithm that enables a robot to
learn the internal dynamics of a demonstrated trajectory. Specifically, an
adaptive control technique is employed as the basis of an RL method for IL.
During the learning stage, the robot utilizes the proposed method to learn the
parameters of a dynamical system. DMPs are chosen as the basis for describing
the internal dynamics of the motions. This allows for adaptively generating
motions in similar, but slightly different, situations. In the generation stage,
the robot uses the learned parameters and a predefined controller to drive its
configuration states and move along the desired trajectory. Simulations and
physical experiments are conducted to validate that a robot can learn the
dynamics (i.e., model parameters) and generate similar trajectories.

Many tasks can be performed by robots with one or more discrete movements (i.e.,
point-to-point motions). Based on this observation, \cite{chen2017robot} propose
a methodology that allows a robot to learn point-to-point motions from human
demonstrations. The framework is based on discrete DMPs in joint space, where
each joint state is represented by one DMP. To learn from multiple
demonstrations of a specific task, DMPs are combined with a GMM using the
approach developed by \cite{yin2014learning}. An EM algorithm is employed to
learn the model. The nonlinear part of the DMP is learned via GMR. This permits
more features of the same skill to be extracted, which generates improved
trajectories. A motion capture sensor is used to extract the training
demonstrations. The performance of the system is evaluated through a virtual
robot platform.

Although DMPs have been developed for rhythmic movements, a predictive model
that allows a human to physically interact with a robot by arbitrarily modifying
its trajectory is an unsolved problem. In a series of works,
\cite{dimeas2018towards,dimeas2019progressive,kastritsi2018phri,kastritsi2018progressive}
enable this capability by adding a dynamical system that can speed up or slow
down a DMP, which allows the trajectory to be synchronized with the current
demonstration. Concretely, a variable stiffness controller is designed that
considers the tracking error in combination with the interaction forces to
modify the target stiffness of the robot. Variable stiffness is applied via an
energy tank method that preserves the control system's passivity with respect to
the tracking error, while all states remain bounded. Furthermore, the tracking
error between the DMP and the robot's pose is used as an indicator of agreement
between consecutive demonstrations. In follow-up work,
\cite{papageorgiou2020passive} address prior deficiencies in the methodology by
allowing the DMP to evolve in a virtual and synchronized manner with the human
teacher. Additionally, \cite{papageorgiou2020kinesthetic} prove the stability
and passivity of the approach under external interaction forces.

DMPs can only be used to handle a single demonstration, yet multiple
demonstrations are often necessary since it can be difficult to obtain the
optimal motion from a one-time teaching. To address this shortcoming,
\cite{yang2018biologically} integrate the features of multiple demonstrations by
combining DMPs with GMMs. Specifically, a fuzzy GMM is employed to fuse the
features of multiple demonstrations into the nonlinear term of a DMP. This is
done to improve the learning efficiency and increase the nonlinearity fitting
performance when compared to a conventional GMM. Moreover, a regression
algorithm for the fuzzy GMM is developed to retrieve the nonlinear terms based
on the geometric significance of GMR. The motion modeling scheme is evaluated
using a robot arm where a neural network-based controller tracks the generated
motions and compensates for unknown robot dynamics. 

Accurate target reaching is especially important in robot interaction tasks to
ensure safety in uncertain and dynamic environments. Taking these factors into
account, \cite{vlachos2020control} track the accuracy of a DMP encoded
trajectory while being compliant to contact forces arising from possible
collisions. The approach relies on a combination of a low feedback gain
controller with a virtual reference trajectory produced by a variant of the
learned DMP. This results in variable stiffness without the need for variable
control gains. In the case of constant and sinusoidal disturbance inputs, the
method is theoretically proven to achieve tracking error convergence without
requiring any prior training or iterative execution. Experimental results using
a robot arm show higher accuracy for target reaching and tracking when compared
to previous work on compliant DMPs.

Traditional position control strategies for robotic manipulators are designed to
achieve high positioning accuracy. Yet, in the presence of contact motion,
position control approaches may result in large contact forces and therefore
unsafe operation. With this problem in mind, \cite{liu2020learning} employ
improved CDMPs (\cite{liu2019modified}) by encoding the trajectory for both the
free motion and contact stage of the robotic task. This allows the output
trajectory to have a lower speed during the initial phase of contact, which
weakens the impact and leads to a stable interaction. By utilizing the force and
position feedback, the DMP model can not only reduce the trajectory error, but
it can also make the resulting trajectory smoother. The feasibility of the
system is shown in both simulation and through physical robot arm experiments on
a peg-in-hole task.

Combining perception feedback with learning-based open-loop trajectory
generation for a robot's end-effector control is an attractive solution for many
robotic manufacturing tasks. For instance,
\cite{rotithor2020combining,rotithor2022stitching} develop a common state-space
representation to analyze the switched system stability of image-based visual
servoing and DMP acceleration controllers. To do this, an image-based visual
servoing acceleration control law is designed and the corresponding closed-loop
dynamics are proven to be uniformly ultimately bounded if the initial state is
sufficiently close to the goal state. Furthermore, the switching between the
locally stable image-based visual servoing controller and globally stable DMP is
analyzed using multiple Lyapunov functions and the switched system dynamics are
proven to converge asymptotically. The system is tested on a pose regulation
task using an eye-in-hand configuration with 7-DoF arm.

In safety-critical scenarios, the velocity of actuators must be accounted for
during trajectory formulation. Therefore, imposing velocity constraints can be
useful, especially for online trajectory generation. For example,
\cite{dahlin2021temporal} create a framework using DMPs for adaptive trajectory
generation under velocity constraints. Specifically, a repulsive potential term
is utilized to keep velocities within predefined limits while preserving the
desired path shape. Lyapunov-like arguments are used to prove both performance
and stability of the proposed method. Simulations and physical experiments using
a robot manipulator verify the velocity constraints, and that the temporal
coupling allows for the modification of the trajectory's temporal evolution
while maintaining path shape invariance. 

Using low-dimensional latent state space dynamics models is a powerful tool for
vision-based control, yet the integration of simplified
proportional-integral-derivative controllers is an open problem.
\cite{jaques2021newtonianvae} introduce a latent dynamics learning framework
that is designed to induce proportional controllability from visual
demonstrations. The framework simplifies vision-based behavioral cloning,
providing interpretable  goal identification. In particular, goal discovery is
applied to a DMP-based behavior cloning of switching controllers for
demonstration. Experimental results show that the proposed technique allows for
robust control, even in the presence of noise. The model also provides
interpretable goal identification on simulated point mass and 2D/3D robot
reaching tasks. Similarly, physical experiments on a 7-DoF robot show enhanced
interpretability without any labeled or proprioception data. 

\cite{jiang2021multiple} realize autonomous trajectory planning via adapting
multiple demonstrations of different tasks. This is accomplished by designing an
LfD system that integrates DMPs with a Takagi-Sugeno fuzzy model as follows.
First, traditional DMPs are trained from a single human demonstration and used
to generate trajectories by changing the target position. Then, the DMPs are
combined with the fuzzy model to solve the problem of learning from multiple
demonstrations. Finally, a new trajectory is generated from the set of
demonstrations using the DMP and fuzzy model combination based on the target
position. When the goal position is altered, the scheme is able to regenerate
trajectories, even in the presence of obstacles. Simulation results highlight
the generation of a novel trajectory using many demonstrations under various
settings.

Installation/modification of robot-supported quality inspection is a tedious
process that requires continuous human engagement.
\cite{lonvcarevic2021specifying} address this challenge with an approach for
specifying visual inspection trajectories based on CAD models of the workpieces
to be inspected. An expert is involved only to select the desired points on the
inspection path. From the selected path, a temporal parametrization is computed
to ensure smoothness of the resulting robot trajectory. To optimize the speed of
the robot along the path, a learning technique is developed based on ILC and the
PoWER algorithm. The method relies on the ability to easily modulate the robot's
speed using CDMPs. Experimental results using a robot arm show that the system
achieves up to 53\% cycle time reduction from a manually specified motion
without degrading image quality.

Designing motions that are smooth and interpretable is a key problem in
robotics. Although DMPs are versatile, modular, and widely used, the best
attractor locations are often far in distance from the generated trajectory,
making them difficult to initialize or evaluate. To address this dilemma,
\cite{rouse2021visualization} introduce a stable heteroclinic adjusted framework
that preserves stability and learnability. The proposed methodology replaces the
attractor points of DMPs with the saddle points of stable heteroclinic channels.
This allows the saddle points to be visualized in the task space via the overlay
of the actual generated trajectories. Experimental results show that the
framework can follow a trajectory plotted in the task space while incurring a
comparable computational cost.

DMPs have many advantages such as conditional convergence, robustness to
disturbances, time independence, and more. Yet, DMPs cannot drive a system away
from its initial state if the start and goal positions are the same.
\cite{wang2021learning} mitigate this problem via a DNN. More exactly, instead
of using a specific formula to describe the forcing term in the DMP, a DNN is
utilized to fit the target nonlinear function with the demonstrated trajectory
information. The input layer to the DNN is a one-dimensional DMP, while the
output layer is composed of the fitted forcing function. Mean squared error is
used as the loss function. Simulation results show that the goal position can be
accurately reached and generalization ability achieved.

To overcome the fact that DMPs can only learn from one demonstration,
\cite{zhang2021robot} combine DMPs with a GMM and GMR for trajectory generation
and tracking. To do this, a GMM and GMR are first used to fit the demonstration
data. Then, DMPs are utilized to model the fitted motion and force trajectory
data. Finally, a neural network controller is employed to track the desired
trajectory generated by the DMPs. More specifically, to meet the real-time
requirements of robot control, a triangular expansion method is used to generate
enhancement nodes within an RBFNN. This allows the motion controller to reuse
learned motions and complete the task at hand without relearning the weight
parameters. The effectiveness of the proposed system is demonstrated on an
ultrasound scanning task using a robot arm.

A relevant DMP use case is handwriting encoding. However, certain types of
graphemes (letters and letter combinations) are nearly impossible to model via
vanilla DMPs. \cite{liendo2022dmp} highlight which cases of graphemes may or may
not be modeled using DMPs. More specifically, movements that correspond to
closed graphemes (i.e., graphemes that start and end at exactly the same point)
are less likely to be captured. Furthermore, examples that have very similar
initial and final positions also present issues for scaled DMP formulations. A
solution to avoid these cases is proposed by reformulating the DMP scaling and
rotation terms. This makes it possible to model closed paths where the distance
between the initial and goal positions tends towards zero. Simulation results
show increased generalizability on historically difficult to model graphemes.

Although DMPs are a flexible trajectory learning scheme widely utilized for
robot motion generation, most existing DMP-based methods focus on simple
go-to-goal tasks. Motivated to handle tasks beyond point-to-point planning,
\cite{wang2022temporal} propose a temporal logic guided optimization for complex
manipulation tasks. In particular, weighted truncated linear temporal logic is
incorporated into a policy improvement with black-box optimization
(PI$^{\text{BB}}$) algorithm. This not only enables the encoding of tasks that
involve a sequence of logically organized action plans with user preferences,
but it also provides a convenient and efficient means to design the cost
function. PI$^{\text{BB}}$ is then adapted to identify the optimal shape
parameters of a DMP. Simulations and physical robot manipulation experiments are
performed to test the effectiveness of the method.

Associations between sensor data and DMP parameters have been improved via deep
learning. Nonetheless, DMP parameters are difficult to learn for complicated
motions, which requires a large number of basis functions to reconstruct.
\cite{anarossi2023deep} tackle this issue by designing a deep segmented DMP
model. The proposed DNN architecture generates segmented variable-length motions
by utilizing a combination of a DMP multiple parameter predicting network, a
double stage decoder network, and a number of segments predictor. The method is
evaluated on both synthetic and real data, and benchmarked on high
generalization capability, task achievement, and data efficiency.

Skill transfer is an open challenge in LfD systems, especially how to best
represent skill features. To address this problem,
\cite{boas2023dmps,coelho2024dmps}, present progress towards transferring the
skills of human-like motions to robots via LfD. Designed for common industrial
tasks, the approach is based on DMPs for joint-state control of a collaborative
robotic arm. Kinematic differences between the human and robot chain are handle
by fine-tuning the DMPs using a covariance matrix adaptation evolutionary
strategy optimizer. Specifically, the weighted sum of the error between the
point-to-point trajectories of the demonstration and those generated by the DMPs
are minimized. The experimental LfD setup consists of lifting a box to twelve
different positions. 

The execution of periodic manipulation skills while maintaining geometric
constraints remains an open problem in robotics. \cite{abu2024learning} make
progress on this challenge by creating a Riemannian periodic DMP framework that
enables robots to encode general periodic manipulation skills with inherent
geometric constraints. In particular, a state-to-action collaborative model
dynamically adjusts end-effector stiffness based on the task state. To do this,
the framework introduces three variants based on quaternion, rotation, and
symmetric positive definite manifolds. Simulated and real-world experiments on
two 7-DoF robots performing drilling and sawing tasks demonstrate that the
system can successfully execute periodic changes in orientation and end-effector
stiffness.

When employing LfD, adapting learned motor skills to meet additional constraints
that arise during a task can be challenging. \cite{liu2023demonstration} handle
this issue via a noise exploration method that uses PI$^{\text{BB}}$ with an
adaptive covariance matrix as follows. First, the teaching trajectory is
captured with an inertial motion capture system. Next, a more reasonable
trajectory is synthesized as the input to a DMP using a GMM and GMR. Finally,
PI$^{\text{BB}}$ with an adaptive covariance matrix is used to carry out noise
exploration and ensure that a robot arm can pass the designated points during
path planning. The accuracy and robustness of the generalized LfD trajectories
are evaluated on handwriting and obstacle avoidance tasks.

Generating a trajectory where the outcome of the task is not ensured by a simple
repetition of the motion is a difficult problem. \cite{lonvcarevic2023encoder}
address this complexity through an approach that retains the correspondence of
the executed dynamic task. Concretely, two different deep encoder-decoder
networks are developed and combined with ProDMP (\cite{li2023prodmp}) trajectory
encoding to actively predict the behavior of a task. The first architecture
consists of a dual parameter-based decoder that takes the posterior distribution
as input and outputs the mean value of the ProDMP weights and goal. Conversely,
the second Monte Carlo-based architecture samples actions directly in the latent
space instead of the space of the ProDMP weights and goal. The approach is
tested using a 7-DoF robot that manipulates a box by dragging it on a table
using only a rod as an end effector. In follow-up work,
\cite{lonvcarevic2024effects} evaluate the performance of RL using ProDMPs on
the box manipulation scenario.

The generalization of skill learning using DMPs is still an unsolved problem for
dynamic tasks such as reaching for a moving target or obstacle avoidance.
\cite{si2023composite} provide a solution to this issue by developing a
composite DMP skill learning framework that considers both position and
orientation constraints. To do this, skills associated with position and
orientation are modeled by composite DMPs simultaneously via coupling with a
temporal parameter. The nonlinear forcing terms in the DMP model are
approximated by employing an RBFNN. This allows composite DMPs to reach moving
goals by generalizing with respect to spatial and temporal scaling. Moreover,
the framework is guaranteed to converge to moving targets even when perturbed by
obstacles. The approach is validated through simulation and robot experiments.

The successful execution a task can require the use of different motion patterns
that take into account not only the initial and target positions, but also the
features relating to the overall layout of the scene. To increase the
applicability of DMPs to a wider range of tasks, \cite{sidiropoulos2023from} add
a deep residual CNN into the DMP framework. In detail, the CNN is first trained
on a set of RGB images that correspond to demonstrated trajectories. Then, the
method infers the DMP parameters (i.e., the weights that determine the motion
pattern and the initial and target positions) for a planar trajectory given the
knowledge of the task directly from raw RGB images. Lastly, the predicted
trajectory is scaled by the dimensions of the image and projected from the 2D
image plane to the 3D task plane. The approach is tested on two planar tasks
involving cluttered scenes using a robot arm.

A robot motion skill learned from demonstration only has practical value when
the acquired skill can be extended to novel situations. Since it is impractical
to demonstrate every possible robot motion, trajectory generalization is
essential. In light of this observation, \cite{teng2023fuzzy} integrate
Takagi-Sugeno fuzzy inference with a DMP model to enable robot motion skill
learning and transfer. Human demonstrations are encoded using Gustafson-Kessel
fuzzy clustering to extract geometric features and generate if-then rules for
reactively regenerating demonstrated movements. Experiments conducted with a
7-DoF arm performing plant pruning and pick-and-place tasks demonstrate the
system's ability to replicate acquired skills to new trajectories.

The arm is the primary tool for a collaborative robot, thus a human-like level
of motion is essential for many tasks. To this end, \cite{zhang2023human}
explore the use of DMPs to model the motion of a human arm to achieve
generalization capability on diverse tasks using a robot manipulator. DMPs are
utilized to extract parameters from demonstration trajectories involving the
position and orientation of the human hand and elbow elevation angle.
Nevertheless, the absence of the elbow elevation angle to the goal prevents the
DMPs from planning the elbow's motion. To fix this problem, elbow elevation
angles are recorded using a motion capture system and a DNN is trained to
predict the elbow elevation angle. The method is evaluated via a simulated robot
arm by performing a pouring motion on seven different targets.

In manufacturing processes (gluing, painting, polishing, welding, etc.)
trajectory planning methods often struggle to adapt to changing requirements,
which leads to repeated robot programming. To accommodate these needs,
\cite{shen2024research} develop a trajectory learning and modification system
based on DMPs. Specifically, a force-controlled dynamic coupling term is
proposed to dynamically sense the range of a target for trajectory modification
based on the coupling force and trajectory velocity. Additionally, a virtual
coupling force based on an RBFNN is employed to provide precise and quantifiable
control of the coupling force's trajectory modification influence within the
target range. The feasibility of the system is demonstrated on a manufacturing
case study involving trajectory planning for vehicle body polishing.

Energy-efficient motion planning is a central challenge in manufacturing
robotics. \cite{xu2025generalizing} address this problem via a framework that
combines RL, dynamic modeling, and DMPs to automate manufacturing with minimal
energy consumption. In particular, the proposed method integrates a deep
Q-learning kinematic skill composition module with a dynamic simulation-driven
trajectory selector that minimizes energy consumption across varying robot
types and loads. Additionally, normalized and parameter optimized DMPs ensure
consistent obstacle avoidance without extensive retuning. Simulation and
real-world experiments using 6- and 7-DoF robots demonstrate that the system
reduces trajectory deviation and energy use, while maintaining adaptability and
stability in the presence of unexpected obstacles.

\paragraph{Trajectory modification.}
During learning, the interaction between a student and teacher usually involves
natural communication (e.g., gestures or speech). Based on this observation,
\cite{petrivc2014onlineb} develop an approach where a human can interactively
change a robot's motion to achieve the desired outcome via hand gestures. The
coaching gestures are specified by pointing towards the part of the movement
that needs to be changed. For example, a pointing gesture defines the direction
and magnitude of change. The framework consists of a two-layered system for
imitation movement (\cite{gams2009line}). To extract the phase and frequency
from an arbitrary periodic signal, DMPs are combined with an adaptive frequency
oscillator (\cite{petrivc2011line}). The method is implemented in simulation and
on a humanoid robot where gestures are recorded by a depth sensor and body
tracker.

During reaching or pouring tasks, the target object (e.g., cup) can move or the
amount of liquid to pour may change. To handle these types of unpredictable
scenarios, \cite{hangl2015reactive} propose a methodology for the manipulation
of dynamic systems. The framework is designed to smoothly react to such changes
as follows. First, the robot learns a set of execution trajectories for solving
various tasks in distinct situations using DMPs. Then, upon encountering a new
scene, the robot adapts its trajectory to create a novel motion generated by a
weighted linear combination of previously learned trajectories. The weights are
computed via a task-dependent metric, which is learned in the state space of the
robot. Finally, RL is employed to further optimize the metric. Experiments
performed on a robot arm show the system's success on pouring and reaching tasks
under a wide range of perturbations. 

In some situations, an operator may want to correct the last part of a DMP
trajectory while leaving the first part intact. Such a correction should result
in a new DMP where the initial part of the trajectory is the original
trajectory, and the final part of the trajectory resembles the correction.
\cite{karlsson2017autonomous} enable this partial trajectory update via a method
that allows an operator to lead the manipulator backwards, approximately along
the part of the trajectory that should be adjusted, followed by the desired
trajectory. The proposed framework determines what parts of the original and
corrective trajectories to retain, and then merges the trajectories without
introducing any discontinuities. The system is evaluated on a peg-in-hole task
using a robot manipulator.

To ensure the safety and adaptiveness of robots, it is important to have systems
that are robust to noisy environments. \cite{hu2018evolution} deal with this
problem by introducing covariance matrix adaptation evolution strategies that
reduce the influence of noise using a variable impedance controller. To do this,
two hierarchies are proposed for the control of a robot. The first hierarchy
leverages a high-level neural-dynamic network optimization for redundancy
resolution. The second hierarchy utilizes low-level fusing with DMPs for
learning trajectories in joint space. Experiments on a 4-DoF robotic arm with a
three-finger hand show that the methodology learns the variable impedance
controller without a model of the dynamic equations nor the robotic system.

Robot safety is ensured by limiting velocity, force, and power. However, this
results in large cycle times and is thus inefficient in industrial applications
where no amortization of the robotic system can be expected.
\cite{weitschat2018safe} mitigate this problem while still complying with safety
standards for collaborative robots. The approach is based on the projection of
human-arm motion into the robot's path to estimate a potential collision.
Specifically, the time needed by the robot to reach the goal position under
human-in-the-loop constraints is minimized. To do this, the segmented path is
optimized by solving a nonlinear programming problem. Then, DMPs are used to
provide flexible motion from the resulting optimized path. Experimental results
performed on an 8-DoF robot arm show that the method is suitable for
human-in-the-loop as well as spatial constraints.

To foster trust and cooperation during human-robot collaboration, it is desired
for a robot to intuitively recognize the subject's motion, adapting as necessary
to the movements. Achieving this goal requires real-time execution of motion
recognition and generation. For this purpose, \cite{kordia2021movement} present
an approach that identifies partially observed trajectories, represented as
DMPs, among a library of previously-observed motions. Fundamentally, the method
can be divided into two steps: motion recognition of an observed trajectory and
inferring its target. After target inference, the library trajectory is
modulated to match the observed trajectory. Experimental results using both a
simulated and physical dual-arm robot illustrate that the methodology
efficiently adapts to rotations, modulations, and scaling of trajectories. 

A common approach to creating complex robot behaviors is to compose skills such
that the resulting behavior satisfies a high-level task. Nonetheless, when a
task cannot be carried out with a given set of skills, it is hard to know how to
modify the skills to make the task possible. \cite{pacheck2023physically} solve
this problem by encoding robot skills in linear temporal logic formulas that
capture both safety constraints and goals for reactive tasks. Concretely, given
a task, the goal is to automatically find a strategy that enables the robot to
complete it. Otherwise, the task is repaired by finding and physically
implementing skills that enable task success. Skills are treated as trajectories
generated by DMPs, where the first step of skill repair is finding a new skill
that has trajectories as close as possible to the original while satisfying the
new constraints. Simulated and physical demonstrations conducted on a
manipulator and mobile robot show the generality of the approach.

\subsubsection{Motion Planning}
\paragraph{Driver assistance/behavior.}
Advanced motion planning algorithms are crucial for the development of
intelligent vehicle systems. To improve vehicle motion planning, a method for
segmenting unlabeled trajectory data into a library of DMPs is presented by
\cite{wang2018learning}. The library is used to infer true positive segment
points from the possible initial cutting points. The inferred segmentations are
then employed to optimize the DMPs in the library. To do this, probabilistic
inference based on an iterative EM algorithm is applied to segment the collected
trajectories and learn a set of DMPs. Mutual dependencies are utilized between
the segmentation and representation of the DMPs, and the initial segmentation.
This allows for solving both the segmentation of unlabeled trajectory data and
the DMP generation. The model is trained and validated via driving data
collected from a vehicle platform.

Decomposing complex driving tasks into a series of DMPs can improve the
efficiency of a vehicle motion planning system. For instance,
\cite{wang2019regeneration} propose an approach that separates basic shape
parameters and fine-tuning shape parameters from the same type of demonstration
trajectories in a DMP library \cite{wang2018learning}. This simplifies and
constrains the shape adjustment during regeneration. In follow-up work, a
modified DMP method suitable for vehicle motion planning is presented by
\cite{wang2019motion}. The technique provides a way to describe and utilize
driving behavior at the trajectory level. Later on, the problem of coordinated
motion planning in heterogeneous autonomous vehicle environments is addressed by
\cite{guan2023coordinated}. Specifically, a coordinated motion planning method
is developed via the generation, extension, and selection of driving
behavior-based DMPs. 

\paragraph{Obstacle avoidance.}
Compliant robot adaptation to obstacles is crucial in human environments. This
requires that a preplanned robot movement be adapted to an obstacle that may be
moving or appear unexpectedly. In this area, \cite{park2008movement} were the
first to demonstrate the combination of DMPs with potential fields for online
obstacle avoidance. Compared to potential functions for static obstacle
avoidance, the approach shows using a dynamic potential field that depends on
the relative velocity between the robot's end effector and an obstacle results
in smoother avoidance movements. This is implemented by adding the gradient of
the potential field to the acceleration term of the differential equation in the
DMP. The method is evaluated via simulations along with an anthropomorphic robot
arm.

\cite{tan2011potential} develop a DMP system that can generate new trajectories,
with similar dynamics to the demonstrated trajectories, and avoid obstacles in
the environment. To do this, potential fields are incorporated into the original
DMP algorithm. The key idea is to move the goal state to a virtual goal position
by adding an impedance factor when the current state is in the impedance area
around an obstacle. The force generated by the potential field is decomposed
into tangent and centrifugal directions. Although the virtual goal state changes
the trajectory around the obstacle, it still keeps the dynamics of the generated
trajectory similar to the demonstration since it is calculated by the DMP.
Simulations are conducted to verify the effectiveness of the framework. In
proceeding work, \cite{tan2011conceptual} integrate the DMP potential field
method into an LfD cognitive architecture.

\cite{gams2013modulation} address the broader challenge of force-based motion
adaptation for physical interaction tasks. The approach modulates DMPs by
incorporating force feedback at both the velocity and acceleration levels
through coupling terms learned using ILC. Specifically, the framework enables
force profile adaptation for single robots interacting with the environment and
coordination between multiple robots through virtual forces, as demonstrated in
tasks like bimanual object manipulation where dual robot manipulators must avoid
obstacles. A key contribution is proving the stability of the coupled DMP system
with the force modulation technique. Rather than modifying the original DMP, the
method learns feed-forward coupling terms that can simultaneously handle both
contact force adaptation and multi-robot coordination.

The extensible nature of DMPs has led to the inclusion of additional
perturbation terms to achieve motion objectives beyond simply repeating the
encoded behavior, including obstacle avoidance. For example,
\cite{rai2014learning} propose a methodology for learning coupling terms to
enable reactive obstacle avoidance. The key innovation is a set of features that
capture human-like avoidance behaviors, consisting of components that respond to
obstacle distance, heading angle, and a spherical repulsion field.  These
features are linearly combined with weights learned through Bayesian linear
regression with automatic relevance determination. The approach is validated
using both simulated experiments and human demonstrations collected via motion
capture. Experimental results show that the learned coupling terms can
successfully reproduce human-like avoidance behaviors while maintaining
convergence guarantees to the goal.

Trajectories generated via LfD may find motion planning solutions to a task in a
short time. Nevertheless, prior constrained LfD methods are limited in scope and
poorly scalable. For this purpose and to improve adaptability,
\cite{kim2015adaptability} propose a framework that combines LfD with sequential
quadratic programming. The approach separates the addition of constraints (e.g.,
obstacles) from the initial trajectory creation process by first using DMPs to
generate motion trajectories. It then applies sequential quadratic programming
to satisfy hard constraints, achieving safer motion planning in new
environments. Simulation results on a 3-link robot arm show the effectiveness of
the methodology on a target reaching task with obstacles.

One approach to moving heavy/bulky objects is to employ cooperative aerial
transportation. Yet, there are a number planning and control issues associated
with multiple aerial manipulators that must be solved. \cite{lee2016planning}
mitigate these problems by designing an augmented adaptive sliding mode
controller to handle uncertainties and grasping errors as follows. First, to
transport an object to a desired position, the necessary path of each aerial
manipulator is obtained using Bezier curves with a modified RRT$^\ast$
algorithm. Then, to avoid obstacles, DMPs are utilized to modify the generated
trajectories in the $x$-$y$ direction. In proceeding work, \cite{kim2017motion}
speed up the proposed methodology for increased performance. Using custom-made
aerial manipulators, experiments are performed where an elongated object is
moved by two hexacopters while avoiding obstacles.

Many path planning algorithms are not only constrained to a model of the
environment, but they also require large amounts of computation and struggle
with real-time performance. \cite{jiang2016mobile,mei2017mobile} address these
limitations for mobile robots through a DMP library framework. The proposed path
planning system is split into two phases. In the first phase, LfD is employed to
operate a mobile robot while simultaneously recording the robot's position,
velocity, and acceleration. The recorded robot trajectories are used to learn
the DMP weights, which are then stored in the library. In the second phase, the
DMP library is utilized for online path planning decision making. Specifically,
the optimal primitive is chosen from the DMP library based on the search
criteria (e.g., minimum distance between the robot and target). Simulation
results show the generalization and adaptability of the methodology.

The flexibility of the DMP framework is highlighted by its adaptability to a
variety of forcing terms. This is key when integrating obstacle avoidance
capabilities into DMPs, a topic that has been a significant focus of research.
For example, \cite{chi2017learning} use a mobile manipulator (wheelchair-mounted
robot arm) with DMPs to perform tasks in human-centered environments. The goal
of the approach is to develop a general learning framework for service robots to
support elderly and disabled people. To realize this, dynamic potential fields
are integrated with DMPs to perform static obstacle avoidance
\cite{chi2019learning}. Experimental demonstrations involving the placement of a
cup on a table show that the system can accomplish the learned tasks and
generate similar motion trajectories even in the presence of obstacles.

Adapting a robotic assembly system to a new task requires manual setup, which
represents a major bottleneck for the automation of small batch sizes.
\cite{ossenkopf2017reinforcement} tackle this problem by designing a learning
process that considers mechanical constraints of a serial manipulator. In
particular, DMPs are chosen as a policy representation with REPS as the learning
method. Additionally, two enhancements for the REPS algorithm are developed that
enable a robot to: (i) detect collisions during the learning process, (ii) react
to collisions, (ii) learn from collisions, and (iv) avoid planning movements
outside the robot's maximum joint angles. Simulation results on a reaching task
show that exceeding a robot's maximum joint angles can be significantly reduced
while detecting and avoiding obstacles.

Dynamically changing environments require reactive motion plans. Based on this
observation, \cite{rai2017learning} add reactivity to previously learned
nominally skilled behaviors. To do this, a reactive modification term is learned
for movement plans represented by nonlinear differential equations.
Specifically, DMPs are utilized to represent the skills and a neural network is
employed to predict the coupling term given sensory inputs. This provides a
compact representation of the coupling term while maintaining generalizability
across varying task parameters. Evaluations are conducted using an
anthropomorphic robot on the task of obstacle avoidance. In follow-up work,
\cite{sutanto2018learning} use phase-modulated neural networks to improve the
learning performance and test it on a different task (scraping). 

Two common approaches for robot obstacle avoidance are to treat obstacles as
either points or volumes. A DMP modification to support volumetric obstacle
avoidance is proposed by \cite{ginesi2019dynamic} as follows. First, a
trajectory without obstacles is learned. Next, obstacles are modeled via
superquadric potential functions, which allows for the description of objects
with generic shapes. Lastly, the obstacle avoidance information is included
within the DMPs. The framework can be applied offline (i.e., when the position
and shape of the obstacles are known a priori) and online (i.e., when the
obstacles are retrieved by visual information). Experiments are administered
using both simulation and an industrial robot manipulator. In follow-up work,
\cite{ginesi2021dynamic} extend the methodology to include the velocity of the
system in the definition of the potential for smoother obstacle avoidance
behavior.

\cite{lauretti2019hybrid} introduce a hybrid joint/Cartesian DMP formulation to
achieve both obstacle avoidance and anthropomorphic motion using redundant
manipulators in human-interactive tasks. The main idea is to combine both joint
space and Cartesian-space DMPs through a multi-priority coupling equation that
ensures accurate end-effector tracking, while maintaining human-like joint
configurations. Compared to traditional Cartesian DMP methods with inverse
kinematics, the hybrid approach achieves similar end effector accuracy while
producing more human-like motions as measured by joint limits and convex hull
metrics. User studies indicate that human subjects felt more comfortable and
perceived the robot as more natural when moving with the hybrid approach. The
method is demonstrated using a robot manipulator on reaching and pouring tasks,
achieving a 100\% success rate in avoiding obstacles.

Inspired by human studies of obstacle avoidance and route selection,
\cite{pairet2019learning} develop a DMP-learning-based hierarchical framework
that generates reactive, but bounded, obstacle avoidance behaviors. The approach
consists of layered perception-decision-action analysis. DMPs are used to encode
desired policies and define obstacle avoidance behaviors as coupling terms. At
the action level, the coupling terms are reformulated to provide deadzone-free
behaviors and guide the obstacle avoidance reactivity to satisfy task-dependent
constraints. In the perception-decision level, actions are regulated according
to a system-obstacle low-dimensional geometric descriptor and exploration of the
parameter space. The methodology is evaluated in the presence of unplanned
obstacles via simulated environments and a robot arm.

\cite{sharma2019dmp} develop a DMP framework that tracks a desired trajectory
with automatic goal adaptation and obstacle avoidance for nonholonomic mobile
robots. This is done by first deriving weight update laws using a gradient
descent approach to train two RBFNNs that learn the forcing functions and shape
the trajectory associated with the DMP model. Then, the dynamics of the steering
angle are modified using fuzzy logic rules to consider both sides of asymmetric
obstacles. The strategy is also extended to handle multiple static and dynamic
obstacles. Convergence of the robot goal position is proven using Lyapunov
stability analysis and experiments are conducted using both simulations and a
ground-based robot. 

Different potential field functions have been adopted for DMPs to improve
obstacle avoidance performance, yet the profiles of the potentials are rarely
incorporated into RL frameworks. To this end, \cite{li2021reinforcement} develop
an RL-based method to learn not only the profiles of potentials, but also the
motion shape parameters. This is done by initializing a DMP with a trajectory
and predefined potential. Next, the DMP and potential function parameters are
iteratively updated via PI$^2$ to realize the desired behavior while avoiding
obstacles. Concurrently, the cost for each trajectory is computed with a
task-specific cost function. During execution, exploration is ensured by adding
noise to the DMP and potential function parameters. The system is validated on
avoiding objects with different shapes/dimensions using simulations along with a
7-DoF redundant manipulator.

Enabling robots to share skills, control laws, and outcomes through collective
learning remains a big problem. Towards tackling this challenge,
\cite{lu2021dynamic} propose a cloud robotic skill learning framework that
allows robots to apply acquired skills to new tasks in the presence of both
point and non-point obstacles. For the case of point obstacles, an acceleration
term is added to the DMP and its parameters are estimated through iterative
identification to isolate a pure skill uninfluenced by obstacles. With non-point
obstacles, a space transformation method maps trajectories from obstacle-free to
obstacle-condensed spaces, providing adaptation to complex shapes and
multi-obstacle environments. Simulations show that the framework can reconstruct
generalized trajectories that preserve obstacle-avoidance behavior while
maintaining convergence and smoothness. 

Shared control can make it easier to remotely operate a robot, hence decreasing
its dependence on the operator. Yet, addressing how to blend operator inputs
with a robot's own policy without defining an explicit arbitration function is a
complex problem. \cite{maeda2022blending} tackle this challenge by performing
policy blending for shared control during assisted teleoperation. The method
adopts DMPs to implicitly blend input with the robot's policy by treating user
commands as disturbances that the primitive naturally recovers from.
Furthermore, the approach blends human and robot policies without requiring
their weightings. User studies with a teleoperated mobile manipulator
demonstrate that the methodology substantially decreases human intervention on
the task of avoiding dynamic obstacles. 

Mobile collaborative robots can provide benefits such as higher redundancy and a
wider working range. Nonetheless, resolving challenges in coordination,
compliance, collision avoidance, and physical accessibility (e.g., operating in
narrow spaces or reaching distant task targets) through base-arm collaboration
remains an open problem. \cite{tu2022whole} approach this dilemma via a
framework that addresses two key issues in whole-body coordination: (i) the
mobile base and arm exhibit different dynamic characteristics, and (ii) motion
must safely satisfy physical constraints. The proposed controller optimizes
movement capability, reduces structural vibration at the end effector, and
ensures safe obstacle avoidance. Experiments with a 3-DoF mobile base and a
6-DoF position-controlled manipulator validate the system on a wiping task while
avoiding obstacles.  

\cite{zhai2022motion} utilize DMPs to create a dynamic obstacle-avoidance
algorithm for task space motion planning of a robot manipulator. The approach
develops an improved steering angle scheme by incorporating the distance between
the manipulator's end effector and the obstacle. This not only allows for the
ability to avoid obstacles, but it also solves problems such as trajectory
jitter. Furthermore, a dynamic approximation method is implemented to improve
the similarity between the planned and desired trajectory, thus reducing the
loss of free space. Lyapunov stability analysis is used to prove convergence of
the system to the target state. The methodology is analyzed and validated by
performing a series of numerical simulations and experiments with a robot arm.

When DMPs are employed for a robotic manipulation task, the common solution is
to encode the end-effector pose using a separate but phase-coupled DMP for
position and orientation. To address the algorithmic inefficiencies in this
approach, \cite{liendo2023improved} design an improved dual quaternion-based DMP
formulation that encodes both position and orientation in a unified $SE(3)$
space while incorporating obstacle-avoidance dynamics. The formulation extends
existing work with a steering term that modulates motion in response to nearby
obstacles, ensuring smooth convergence to the goal pose. Furthermore, the
approach is framed within the awareness and intelligence layers of human-robot
collaboration systems of systems. Simulation results on synthetic minimum-jerk
trajectories demonstrate stability and accurate trajectory regeneration. 

For non-point obstacles, the locally flat isopotential of a potential field
increases the risk of getting stuck in a local minimum. To address this problem,
\cite{liu2023superquadrics} utilize DMP coupling terms to modify the planned
trajectory of a robot. Specifically, the approach combines superquadrics and a
steering angle technique to construct new coupling terms for obstacle avoidance
without relying on potential fields. Moreover, the original center point
steering angle formula is modified to improve the ability to avoid obstacles on
a wider range of angles. Not only does the method guarantee convergence of the
dynamic system, but it is also applicable to most types of superquadrics,
including those with flat boundaries. A set of experiments conducted in
simulation highlight the avoidance of volumetric obstacles. In
\cite{liu2025novel}, the framework is extended to account for additional
(quaternion and Minkowski sum) coupling terms.

\cite{shaw2023constrained} present a version of a ``constrained DMP," which is
should not be confused with \cite{duan2018constrained}. This approach focuses on
constraint satisfaction in the robot workspace. To ensure the safe learning of
motor skills, a nonlinear optimization method is introduced. This formulation
perturbs the DMP forcing weights in order to incorporate a zeroing barrier
function that guarantees that the constraints are satisfied. A zeroing barrier
function is a way to determine whether or not a dynamical system is
forward-invariant in a safety set. The challenge is to find a suitable function
that can be used to certify the satisfaction of all constraints desired by the
user. The effectiveness of the DMPs with operational constraints is demonstrated
with various experiments, including obstacle avoidance and the drawing of
shapes.

There is no guarantee that trajectories produced by DMPs will not exceed a
robot's kinematic constraints (i.e., the constraints relating to the robot's
position, velocity, and acceleration). To address this problem,
\cite{sidiropoulos2023novel} enforce kinematic constraints on DMPs while
avoiding obstacles. Concretely, an optimization framework is developed to
achieve generalization of a learned trajectory to a new fixed target and time
duration under position, velocity, and acceleration kinematic constraints. These
constraints include via points and obstacles for static scenes (offline), and an
MPC extension to handle constraints in real-time with a varying target and/or
time duration (online). The applicability of the system is demonstrated in
experimental scenarios that involve object handover and placing an object inside
a bin with dynamic obstacles.

An obstacle avoidance solution that satisfies robot kinematic constraints is
also put forth by \cite{niu2023generalized}. Different from previous work on
this topic, a composite DMP algorithm is developed as follows. First, a GMM and
GMR are used to process multiple demonstration trajectories for learning DMPs.
Then, to avoid exceeding the position, velocity, and acceleration constraints
set by the robot or task, the optimization approach proposed by
\cite{sidiropoulos2023novel} is utilized. This results in generalized
trajectories that are close to the learned trajectories and accomplish obstacle
avoidance. Simulated experiments are conducted to verify that the resulting
trajectories meet kinematic constraints while staying clear of obstacles in the
workspace.

Deep RL algorithms with experience replay have been employed to solve sequential
learning problems, yet real-world robotic applications face serious challenges
such as smooth movements. To this end, \cite{yuan2023hierarchical} develop a
hierarchical RL control framework using MoMPs that enables smooth trajectories
in the presence of obstacles. The method consists of a lower-level controller
learning hierarchy and an upper-level policy learning hierarchy. In the
lower-level controller learning hierarchy, modified MoMPs are used to generate
smooth trajectories. In the upper-level policy learning hierarchy, a local
proximal policy optimization algorithm is designed to learn a mapping function
from a range of situations to the corresponding metaparameters in order to adapt
to new tasks. The system is evaluated using a robot arm on the task of
maintaining smooth motions while avoiding collisions.

To address limited generalization performance and low adaptability in robot arm
trajectory planning algorithms, \cite{jia2024trajectory} create a DMP-based
framework that integrates DTW, GMMs, and GMR to preprocess multiple
demonstrations in order to obtain an ideal reference trajectory. This is done by
adding an artificial potential field coupling term to enable a robot arm to
deviate smoothly from obstacles and return to the motion trajectory. Simulation
results show that the methodology can accurately reproduce and generalize
demonstrated trajectories, maintaining consistent motion trends even when start
and goal positions are close or inverted. The method is also verified on a 6-DoF
collaborative robot in an obstacle-present setting.

\cite{lu2024dynamic} design a shared-control teleoperation framework that
integrates IL and bilateral control to achieve system stability via DMPs. This
is done by first creating a DMPs-based observer to capture human operational
skills through offline demonstrations. The learning results are then used to
predict action intention using teleoperation. Specifically, the DMPs-based
observer is employed to estimate human intentions on the leader side, while a
high-gain observer predicts the state of the robot in real time. The convergence
of the DMPs-based observer under time delays and teleoperation stability are
proved by building two Lyapunov functions. The effectiveness of the system is
validated on obstacle avoidance experiments.

To avoid obstacles and reduce the overall control effort,
\cite{theofanidis2024safe} present a framework where a robot learns motor skills
through human demonstrations with safety guarantees as follows. First, a
DMP-based approach is utilized to enable encoding of distributions from multiple
demonstrations. Then, a combined control Lyapunov function (CLF) and control
barrier function (CBF) controller executes the planned trajectories by tracking
the mean trajectory while ensuring that the robot's end effector remains within
a safe zone. The methodology extends to both position and orientation control
through a null-space projection controller. The probabilistic DMP formulation is
demonstrated via simulations and pick-and-place experiments with a robot
manipulator using known obstacle positions.

Mobile robots face challenges when executing accurate real-time trajectory
planning, especially in non-static settings. \cite{li2025model} address this
shortcoming by combining DMPs with MPC. The DMPs generate smooth reference
trajectories while MPC is utilized to predict obstacles and adjust the path in
real time. More specifically, the MPC scheme employs an operator splitting
quadratic program for path adjustment, modeling obstacles as both soft and hard
constraints. A super twisting terminal sliding mode control strategy is used for
tracking to improve accuracy in nonlinear environments and ensure smooth
convergence. The effectiveness of the method is demonstrated on a two-wheeled
differential robotic platform by showing the ability to avoid and adapt to both
single and double obstacles. 

Collaborative multi-arm robot manipulation can enable fine control, but as the
adoption of these systems increases so does the need for safety-aware motion
planning, particularly in dynamic environments. In light of this necessity,
\cite{singh2025collaborative} create a hierarchical framework that integrates RL
with DMPs to enable collaborative, adaptive, and collision-free task completion.
At the higher level, a deep Q-network learns to generate trajectories from a
skill library, while at the lower level an optimized and normalized DMP employs
heuristic collaborative phase control and dynamic potential fields for
trajectory execution. Real-world demonstrations on a 7-DoF manipulator show
that the approach generates constraint-aware trajectories, avoids obstacles, and
can accomplish block stacking, table cleaning, and water pouring tasks.

\subsubsection{Robotic Prosthetics}
\paragraph{Exoskeletons.}
Sit-to-stand movements pose a major challenge for individuals with lower-limb
weakness. To mitigate this problem, \cite{kamali2016trajectory} present a
control framework for knee exoskeletons that combines DMP-based trajectory
generation with impedance control. The approach predicts intended sit-to-stand
motions from demonstrated trajectories using initial knee and ankle angles
measured at seat-off. The formulation provides a flexible motion representation,
while impedance control assists by tracking the predicted trajectory and
allowing speed adjustments. Via a series elastic actuator, the system adapts
movement timing based on human-exoskeleton interaction forces. Experimental
results demonstrate a reduction in average knee power requirements by
approximately 30\% across varying seat heights and foot positions, with
assistance levels remaining statistically consistent across conditions.

Exoskeletons are designed to enclose human limbs and augment their existing
capabilities by providing additional power to the joints. Control methods are
the interface between the human and exoskeleton, and they must be able to
predict user intentions and apply them to the mechanism at the correct time to
achieve human-robot cooperation. For example, \cite{peternel2016adaptive}
develop an approach to adaptively learn elbow exoskeleton joint torque behavior.
The system uses EMG-based muscle activity feedback to include the human in the
robot control loop. In particular, a control method dynamically adapts the shape
of the robot joint torque trajectories in accordance with human behavior. DMPs
encode the torque-control trajectories and couple them with adaptive oscillators
to control their phase and frequency. A set of experiments demonstrates how the
exoskeleton adapts to the user and offloads muscle effort via the proposed
methodology. 

\cite{huang2016hierarchical,huang2018hierarchical} put forth a hierarchical
interactive learning control strategy to cover the shortages of both
sensor-based and model-based controllers in lower exoskeletons. To do this, two
learning hierarchies are combined: a high-level motion learning hierarchy and a
low-level controller learning hierarchy. In the high-level motion learning
hierarchy, trajectories are modeled using DMPs. Then, to describe varying
walking patterns, LWR is utilized to update the DMPs incrementally through
human-exoskeleton interactions. In the low-level controller learning hierarchy,
RL is employed to perform online updates of the model-based controller
parameters. This allows for adaptation to varying dynamics in order to reduce
interaction forces between the human and exoskeleton. The strategy is tested
with varying walking speeds both in simulation and via an exoskeleton system. 

Human-powered lower exoskeletons are increasingly being used for locomotion
assistance and strength augmentation. In these applications, it is crucial to
achieve robust control, which requires the proactive modeling of human movement
trajectories. \cite{huang2016learning,huang2019learning} tackle this challenge
through the introduction of coupled cooperative primitives for lower exoskeleton
control, where DMPs are employed to model normal walking motion from clinical
gait datasets. To apply this method of control, impedance models first capture
the interaction between the pilot and exoskeleton. Then, RL is utilized to learn
the model parameters online. The coupled cooperative primitives scheme is
verified on both a 1-DoF exoskeleton platform and a human-powered augmentation
lower exoskeleton system.

Desired exoskeleton trajectories are predefined from a healthy subject or
extrapolated from clinical gait analysis datasets. However, an exoskeleton
should have the ability to adapt to the motions of the user and allow
corresponding changes for different walking situations. To this end,
\cite{chen2017step} present a step length adaptation method with dynamic gait
models to model gait flexibility. This is done by modeling the system as a
special hybrid human-exoskeleton agent. DMPs are utilized to model the
exoskeleton gait by combining the relationship between the center of mass of the
agent and the gait length. To learn a dynamic gait online, PI$^2$ is employed to
update the model parameters using the energy cost of the user and the stability
of the agent as the cost term. Experiments are carried out in both simulation
and on a real-time exoskeleton system.

The goal of a robotic exoskeleton is to learn a human's experiences and skills
such that it can actively cooperate with the human. To this end,
\cite{deng2018learning} develop a high-level model that can learn human
demonstrations and a low-level admittance control scheme that enables back
drivability. Concretely, the underlying features and constraints of the
interaction are statistically analyzed and learned using GMMs based on logged
HRI data. Then, during reproduction the robot predicts the interaction force
similar to those in the demonstrations via GMR. This provides the human with
external assistance to ensure that the task can be successfully completed. An
experimental evaluation is carried out using a dual-arm exoskeleton on tasks
involving the motion of drawing lines.

In upper-limb exoskeletons, the reference joint position is typically obtained
by means of Cartesian motion planners and inverse kinematics via the inverse
Jacobian. Yet, if used to operate non-redundant exoskeletons, then this approach
does not ensure that the anthropomorphic criteria is satisfied within the entire
human-robot workspace. To address this concern, \cite{lauretti2018learning}
develop a motion planner based on DMPs that guarantees successful task execution
by preserving the configurations of an assisted human arm. Recorded trajectories
are replicated by means of a weighted sum of optimally spaced Gaussian kernels.
DMP parameters are then extracted from demonstrated movements with LWR and used
to train a neural network to predict the joint target positions. The system is
validated on the following daily living activities: drinking, pouring, and
lifting a light sphere. 

Most exoskeletons use a predefined or preplanned gait movement. These systems
lack flexibility and adaptability, which limits their use in outdoor
environments. \cite{ma2018gait} solve this problem by developing a gait planning
method with DMPs that enables a lower-limb exoskeleton to walk up fixed-size
stairs smoothly. In particular, the approach can perform online adjustment of
the exoskeleton gait trajectory by detecting the position of the stair edge at
each step. Simulations and experiments with a lower-limb exoskeleton are
conducted to verify the methodology and validate its control effect.
Nonetheless, the stair parameters can only be tuned by hand and not by
perception algorithms, which limits the applicability of the system in
activities of daily living.

Rehabilitation exoskeletons that rely on predefined average gait patterns often
fail to adapt to individual patient needs and environments.
\cite{hwang2019intuitive} propose a solution to this problem through an approach
that combines end-point reference generation via inverse kinematics with DMPs
for adaptive motion learning. The method uses Gaussian basis functions and LWR
to learn trajectories, while an amplitude re-scaling gain flexibly adjusts the
stride length and a controller ensures tracking. Validation is performed using
simulations and experiments on an exoskeleton, showing successful stride-length
modulation and natural trajectory tracking. \cite{hwang2021gait} extend the
methodology to address patient-specific conditions when crutches are used for
support. 

To assist the act of human walking while wearing a lower-limb exoskeleton,
\cite{yuan2019dmp} propose a DMP trajectory-learning strategy based on RL. This
is done via a two-level planning scheme as follows. In the first level, an
inverted pendulum approximation considers the locomotion parameters to guarantee
that the zero-moment point within the ankle joint of the support leg is in phase
with the single support. In the second level, joint trajectories are modeled and
learned by DMPs. Additionally, RL is used to learn the trajectories and
eliminate the effects of uncertainties and disturbances in the joint space.
Experiments conducted with human subjects show that the trajectories of the
exoskeleton can be constantly adjusted and that the resulting target
trajectories are smooth.

For lower-limb exoskeletons, creating stable human-like gaits for walking on
slopes is a critical problem. \cite{zou2019adaptive} address this issue by
constructing an adaptive gait planning approach. Concretely, a human-exoskeleton
system is first modeled as a 2D linear inverted pendulum with an external force.
Then, a dynamic gait generator based on an extension of conventional capture
point theory and DMPs is developed. This allows the system to generate
human-like gait trajectories and adapt to slopes with different gradients. The
framework is evaluated via simulations and an exoskeleton system designed for
providing walking assistance to paraplegic patients. In follow-up work,
\cite{zou2020slope} extend the method by designing a slope gradient estimator
that takes advantage of local sensor data from the exoskeleton system.

Existing massage robots are often bulky, expensive, and limited in adaptability.
To alleviate these issues, \cite{li2020enhanced} propose an enhanced robot
massage system that combines force sensing and DMPs to generate multiple
trajectory patterns. The approach integrates several key components including:
(i) DTW to preprocess multiple human demonstrations, (ii) DMPs with GMMs to
model and evaluate trajectories, (iii) GMR to synthesize smooth generalized
trajectories with minimal position errors, and (iv) a hybrid position/force
controller to ensure HRI safety. Using a robot manipulator, experimental results
highlight the accurate reproduction of massage paths, adaptation to body shape
variations, and attention to safety when performing massage tasks with a
consistent force.  

\cite{li2020skill} consider both motion modeling and dynamic controller
performance when developing a two-level skill learning-based control strategy
for HRI scenarios. The hierarchical control strategy is structured as follows.
A high-level learning strategy learns human motor skills from multiple
demonstrations using DMPs and GMMs, which guarantees convergence to a goal
position A low-level control strategy is realized by combining an admittance
controller for transferring interaction forces to reference trajectories in the
task space. Specifically, an integral BLF RBFNN controller tracks the
trajectories in joint space to ensure compliant motions of the exoskeleton
during cooperation. The approach is evaluated using a dual-arm exoskeleton robot
on manipulation tasks.

\cite{qiu2020exoskeleton} present an online learning algorithm for active
assistance control of lower-limb exoskeletons. The objective is to estimate
human walking intention and ensure that the exoskeleton does not impede the
user's motion. To do this, DMPs are utilized to learn the periodical human joint
trajectory during stable walking. The trained model can then be deployed to
predict a smooth and personalized future joint trajectory. Smooth joint
trajectories are crucial for human joint torque estimation. Otherwise, joint
trajectory noise may be significantly amplified by numerical differentiation.
Human walking experiments demonstrate that the method can not only predict a
smooth trajectory and joint torque profile in real time, but it can also
compensate for the phase delay caused by sensor signal filtering.
\cite{qiu2021exoskeleton} broaden this work by addressing the instability of
variable-speed walking with frequency adaptive DMPs. 

Although lower-limb exoskeletons are utilized for both walking assistance and
rehabilitation related applications, dealing with stairs is still a challenge.
To solve this issue, \cite{xu2020stair} create a bio-inspired gait trajectory
generator based on stair size information and DMPs as follows. First, a
perception component consisting of two time-of-flight sensors is built. This
allows the system to capture the stair size and edge position. Then, the
recorded data is fed to a stair gait trajectory generator. Specifically, the
stair edge position and size parameters are used as the obstacle and goal impact
factors to generate appropriate trajectories corresponding to different
environmental factors. The implementation is run on a lower-limb exoskeleton and
the collected power and torque data are analyzed during a stair-climbing task.

Dynamic control of robotic exoskeletons is crucial to ensure safe and
synergistic assistive actions. Although progress has been made in exoskeleton
controllers for rhythmic tasks, control methods for assisting discrete movements
remain limited by their task-specificity. Based on this observation,
\cite{lanotte2021adaptive} formulate a single adaptive DMP-based impedance
control framework for the assistance of discrete actions. The method defines the
initial MP as a dimensionless generic trajectory that represents the general
type of motion for the human joint to be assisted. This generic MP is then
modulated in real-time, in both amplitude and duration, to represent a wide
range of movements in a user-independent fashion. The approach is validated on
two distinct experiments involving lifting tasks.

Attaining flexible limb control with minimal constraints on a human gait is a
challenging problem. Nonetheless, achieving robust limb control may be done by
leveraging the user's reference trajectory to inform lower-limb control in terms
of the touchdown position, step length, and time. For example,
\cite{luo2022trajectory} present a gait trajectory generation method based on
DMPs and a path-based controller. The DMPs provide flexible online adjustment to
the reference trajectory, while the controller guides the limb's swing motion
using a lookahead distance to search for the target joint angle. Experiments
highlight reduced gait trajectory deviation and improved leg motion performance.
Furthermore, the results show that the approach does not limit the gait pattern
variance of the subject.

To assist patients with lower-limb disabilities, \cite{zhang2022motion} develop
a trajectory learning scheme for a 6-DoF exoskeleton where the hip and knee of
each artificial leg provides two electric-powered DoFs for flexion/extension.
This is done via a five-point segmented planning strategy for creating gait
trajectories. Concretely, the gait zero moment point stability margin is used as
a parameter to ensure the stability of the human-exoskeleton system. Based on
the segmented gait trajectory planning scheme, multiple-DMP sequences are used
to model the generation trajectories. In addition, to eliminate the effect of
uncertainties in joint space, RL is employed to learn the trajectories.
Experimental results demonstrate the removal of interferences and uncertainties,
thus ensuring smooth and efficient movement.

Continuous gait phase estimation can facilitate real-time synchronization of
wearable robots (e.g., exoskeletons) with the user's movements, providing
seamless operation across varying walking speeds. For instance,
\cite{eken2023continuous,eken2024continuous} implement a gait phase estimation
methodology for common locomotion activities (e.g., level-ground walking, stair
and ramp negotiation). In particular, the approach investigates the use of
adaptive DMP-based gait phase estimation to generate phase-dependent torque
profiles using a powered unilateral hip exoskeleton. Tests are conducted with
six able-bodied participants along with pathological gaits using a dataset from
six stroke survivors. Experimental results demonstrate that the approach can
effectively generate phase-locked torque profiles for able-bodied participants.

Robot grasping skills are typically manually designed with little consideration
for dynamic modification based on real-time tactile feedback. In light of this
observation, \cite{lu2023visual} develop a wearable hand exoskeleton capable of
capturing visual-tactile and motion data to model human grasping skills. The
exoskeleton is worn by a human demonstrator, allowing the operator to experience
real-time force feedback through their fingers. The collected exoskeleton data
is used to build a human grasping skill model and robot grasping strategy
through a combination of DMPs to mimic human-grasp actions with varying
velocities, a linear array network, and tactile-based motion planning. The
framework is validated on grasping tasks using a robot arm and a self-designed
gripper.

Robotic mirror therapy is an approach for hemiplegic patients where the motion
of the healthy limb is transmitted to the impaired limb. Specifically, a
wearable robot assists the impaired limb to mimic the healthy limb's action and
thus stimulate the active participation of the injured muscles.
\cite{xu2023dmp,xu2023learning} formulate this physical interaction as an
impedance model coupled with a DMP model. To do this, the subject's muscle
strength is evaluated via skin surface EMG signals and transferred to the
robot's joint stiffness. Then, to adapt the human-robot coupled DMP parameters
to handle varying subject performance, a PI$^2$ with covariance matrix
adaptation RL algorithm is designed to optimize the robot's movement trajectory
and stiffness profile where the training safety and rehabilitation improvement
are both guaranteed. The methodology is validated using a lower extremity
rehabilitation robot.  

Adaptive gait control is a significant challenge when using a lower-limb
exoskeleton for walking assistance. \cite{yu2024modified} address this problem
using DMPs to achieve gait adjustments with varying assistance levels.
Specifically, an adaptive frequency oscillator is utilized such that a periodic
trajectory can be obtained as the input trajectory for further refinement.
Through HRI, the DMPs generate trajectories adaptive to the user's intention and
walking status. Compared to traditional DMP methods, the approach combines
time-dependent interaction forces with threshold forces directly into the model,
which reduces the algorithmic complexity and improves the trajectory tracking
ability of the exoskeleton. An experimental evaluation is conducted on five
subjects wearing a lower-limb exoskeleton on level-ground walking.

Adaptive compliance control is crucial for robots to handle a spectrum of
rehabilitation needs and enhance training safety. To this end,
\cite{zhou2024trajectory} create a trajectory deformation-based multimodal
strategy for a wearable lower-limb rehabilitation robot that includes a
high-level trajectory planner and a low-level position controller. DMPs are
integrated into the trajectory planner to realize compliant HRI, and make
learning and adjusting rehabilitation task trajectories more convenient. A
multimodal adaptive regulator based on the amplitude modulation and deformation
factors of the algorithm, enables matching the varying motor abilities of a user
by smoothly switching between different control modes. Four healthy participants
and two stroke survivors are employed to conduct robot-assisted walking
experiments using the proposed system.

To assist people with impaired mobility, \cite{hao2025hierarchical} develop a
hierarchical, task-oriented whole-body locomotion framework via adaptive DMPs.
The framework integrates gait planning, phase estimation, and a task-switching
mechanism that provides smooth transitions between straight and turning gaits.
The proposed method generalizes joint trajectories using DMPs, while updating
parameters online to enhance adaptability and trajectory convergence. Phase
estimation from joint angle measurements ensures synchronized gait transitions
and switching occurs only when both feet are supported. An experimental
evaluation conducted with a 10-DoF exoskeleton demonstrates that the system can
enable wearers to perform both straight walking and turning maneuvers while
pushing a cart. 

Advancements in lower-limb exoskeleton control have mainly focused on enhancing
walking capabilities across diverse terrains (e.g., stairs, ramps, etc.), yet
achieving seamless transitions between these terrains remains a challenge.
\cite{huang2025lower} address this problem via a hierarchical interactive
learning control framework. The system consists of high-level learning and
low-level control layers. The high-level learning layer employs DMPs to
piecewise learn desired joint torque curves. The low-level control layer then
uses the learned DMP to output torque based on the gait phase and locomotion
pattern, while PI$^2$ is utilized to dynamically adjust the DMP control
parameters in real time to minimize human-exoskeleton interaction forces.
Experimental results show that the framework enables smooth transitions among
different terrains, reduces reliance on accurate dynamic models, and decreases
human oxygen consumption.

\paragraph{Powered prosthetics.}
Most knee prostheses are energetically passive devices with limited ability to
reproduce healthy biological joint behavior. Conversely, powered prostheses
utilize battery-powered servo motors to actively adjust joint movement, thus
achieving the ability to overcome obstacles. For instance,
\cite{zhang2021dynamical} create a trajectory-generating powered prosthetic to
help amputees cross over obstacles as follows. First, DMPs are used to plan an
obstacle-free trajectory from the initial to the target state where the forcing
term is computed using GMR. Second, a modified obstacle-avoiding algorithm is
added as a coupling term to the second-order system of DMPs to generate the
planned trajectory. This involves incorporating the distance factor and dynamic
obstacle avoidance. The method is tested in simulation by generating desired
trajectories for crossing over obstacles.

In a powered prosthetic, key design features include the ability to decode the
user's intent to start, stop, or change locomotion. Based on this knowledge,
\cite{eken2023locomotion} develop a locomotion mode recognition algorithm using
adaptive DMP models (\cite{lanotte2021adaptive}) as locomotion templates. The
goal of the method is to accurately and timely classify the user's activity to
avoid mismatches between the action of the device and the actual locomotion in
terms of modality and temporal onset. Concretely, the algorithm takes as input
accelerometer and gyroscope signals of a thigh-mounted IMU and foot-contact
information to perform discrete classifications using a set of SVMs during the
swing phase before the end of the current step. The system is tested in
simulation and with human amputee participants.

Amputees are unable to directly use perceptual information to control the
movements and gait patterns of their prosthesis. To aid amputees in walking and
smoothly crossing over obstacles, \cite{hong2023vision} propose a locomotion
coordination control method for powered lower-limb prostheses. This includes a
vision-based system to detect obstacles, enabling the prosthesis to make
obstacle avoidance decisions, and fuzzy-based DMPs to help amputees maneuver
around obstacles more smoothly. The coordination control is partitioned into
obstacle detection, trajectory shaping, and joint control. The type-2 fuzzy
models are trained according to nonlinear forcing terms, which allows them to
handle high-dimensional training data with little computational complexity
overhead. Experiments show that the approach enables a powered prosthetic to
adaptively switch between level walking and obstacle avoidance gaits.

\subsection{ProMPs}
\begin{table*}
\caption{Applications of ProMPs organized by topic area.}
\label{tab:promp_applications}
\renewcommand{\arraystretch}{1.5}
\centering
\scalebox{1}{
  \begin{tabular}{M{0.2\textwidth-2\tabcolsep - 1.25\arrayrulewidth} M{0.8\textwidth-2\tabcolsep - 1.25\arrayrulewidth}}
    \toprule
    \textbf{Area} & \textbf{References} \\\midrule
    Contact Manipulation & 
    Assembly: \cite{ewerton2015modeling,carvalho2022adapting}.
    \newline
    Deformable Objects: \cite{motokura2020plucking}. 
    \newline
    Hitting: \cite{gomez2016using}. 
    \newline
    Insertion: \cite{zang2023peg}.
    \newline
    Pick-and-Place: \cite{conkey2019active,mghames2020interactive,jankowski2021probabilistic,schale2021incremental}.
    \newline
    Polishing/Wiping: \cite{wang2023promp,unger2024prosip}. 
    \\\cline{1-2}
    Human-Robot Interaction & 
    Active Compliance: \cite{guan2021improvement}.
    \newline
    Imitation Learning: \cite{maeda2017probabilistic,knaust2021guided,hu2023autonomous,yao2023improved,liao2024bmp}.
    \newline
    Movement Coordination: \cite{maeda2014learning,ewerton2015motor,dermy2017prediction,faria2017me,maeda2017phase,maeda2018probabilistic,oguz2017progressive,koert2018online,parent2020variable,luo2021dual,oikonomou2022reproduction,qian2022environment}.
    \\\cline{1-2}
    Motion Modeling & 
    Human-Motion Reproduction: \cite{xue2021using}.
    \newline
    Trajectory Generation:
    \cite{lockel2020probabilistic,davoodi2021safe,sanni2022deep,vorndamme2022integrated}. 
    \\\cline{1-2}
    Motion Planning & 
    Multitask Learning: \cite{yue2024probabilistic}.
    \newline
    Obstacle Avoidance: \cite{koert2016demonstration,colome2017free,osa2017guiding,wilbers2017context,shyam2019improving,low2020identification,colome2021topological,low2021prompt,davoodi2021probabilistic,davoodi2022safe,davoodi2022rule,fu2022online}.
    \\\cline{1-2}
    Robotic Prosthetics & 
    Exoskeletons: \cite{jamvsek2021predictive,wang2023probabilistic}.
    \\
    \bottomrule
  \end{tabular}}
\end{table*}

\subsubsection{Contact Manipulation}
\paragraph{Assembly.}
ProMPs utilize probability theory to seamlessly combine primitives, specify via
points, and correlate joints by conditioning. They have been used to model
spatio-temporal variability and learn multiple interaction patterns from
unlabeled demonstrations. For example, \cite{ewerton2015modeling} propose a
mixture of IPs that allow robots to learn interactions from multiple unlabeled
demonstrations and deal with nonlinear correlations between the interacting
partners. Specifically, a multimodal prior probability distribution is obtained
over parameterized demonstration trajectories of a robot and human working in
collaboration. During execution, the algorithm selects the mixture component
with the highest probability given the observation of the human partner. Then,
this component is conditioned to infer the appropriate robot reaction. Using a
7-DoF robot arm, the approach is shown to adapt to human movements from an
arbitrary number of demonstrations executed at different speeds.

For object manipulation tasks, the accuracy of ProMPs is often insufficient,
especially when learned from external observations in Cartesian space and then
executed with limited controller gains. To address this issue,
\cite{carvalho2022adapting} combine ProMPs and residual RL. This permits both
corrections in position and orientation during task execution. In detail, the
method learns a residual on top of a nominal ProMP trajectory using a soft
actor-critic. This allows for the incorporation of variability in the
demonstrations as a decision variable and reduces the search space for residual
RL. Utilizing a 7-DoF robot arm, the approach is applied to a block insertion
task resulting in the robot successfully learning to complete the insertion,
which was not possible with basic ProMPs.

\paragraph{Deformable objects.}
Hand plucking objects (e.g., fruits, leaves, etc.) is tedious and monotonous
process for humans and thus there is a strong incentive for automation.
\cite{motokura2020plucking} tackle this agricultural problem by
utilizing ProMPs to develop a harvesting robot capable of plucking tea leaves.
To properly pluck the leaves, the robot must reproduce a complicated human hand
motion that requires pulling while rotating. This can vary depending on the
condition of the tree such as the maturity of the leaves, thickness of the
petioles, and thickness and length of the branches. By probabilistically
combining learned motions at a ratio determined by the branch stiffness, the
appropriate motion is generated for each situation. The method is experimentally
confirmed to reproduce the desired plucking motion by inputting the mean of the
probabilistic distribution of the generated motion to a robot manipulator.

\paragraph{Hitting.}
Robot table tennis serves as an important testbed for learning new approaches
to trajectory generation due to its strong requirements in motor abilities.
\cite{gomez2016using} present two methods for adjusting ProMPs in the context
of robot table tennis. The first technique adapts a probability distribution of
hitting movements learned in joint space to have a desired end-effector
position, velocity, and orientation. The second approach finds the initial time
and duration of the ProMP in order to intercept a moving object (e.g., table
tennis ball). These methods rely on simple operations from probability theory,
hence offering a more principled approach to solving some of the challenges of
robot table tennis compared to previous work on this topic. The methods also
have the potential to generalize to other motor tasks. Table tennis experiments
are conducted using a 7-DoF robot arm and a camera system to track the position
of the ball. 

\paragraph{Insertion.}
Real-world demonstration trajectories are often sparse and imperfect, which
makes it challenging to comprehensively learn directly from action samples.
\cite{zang2023peg} address this problem through a streamlined IL approach for
assembly tasks as follows. First, the demonstration trajectories are aligned
using DTW, which allows for segmenting them into several time stages. The
trajectories are then extracted via ProMPs and utilized as global strategy
action samples. In particular, the current state of the assembly task is the via
point of the ProMP model to obtain the generated trajectories, while the time
point of the via point is calculated according to the probability model of the
different time stages. The action of the current state is treated according to
the target position of the next time state. Finally, a neural network is trained
to obtain the global assembly strategy by BC. The method is applied to a
peg-in-hole task in simulated and physical experiments using a robot manipulator.

\paragraph{Pick-and-place.}
Although ProMPs are typically initialized from human demonstrations and they
achieve task generalization through probabilistic operations, there is no way to
know how many demonstrations should be provided or what constitutes a good
demonstration for obtaining generalization. To achieve better generalization,
\cite{conkey2019active} propose an active learning approach that learns a
library of ProMPs capable of task generalization over a given space.
Specifically, uncertainty sampling techniques are used to generate a task
instance for a teacher to provide as a demonstration to the robot. If possible,
the demonstration is merged into an existing ProMP. Otherwise, if the
demonstration is too dissimilar from existing demonstrations then a new ProMP is
created. Grasping experiments show that the method achieves better task
generalization using fewer demonstrations when compared to a random sampling
over the space.

Picking from a dense cluster of fruit is a challenge in terms of trajectory
planning. For instance, conventional planning approaches may not find
collision-free movements for a robot to reach and pick a ripe fruit.
\cite{mghames2020interactive} tackle this problem by creating an interactive
ProMP framework that features pushing actions in complex clusters. Different
degrees of nonlinearity in the system behavior are generated as follows. First,
the planner learns two ProMPs from demonstrations. Then, it conditions the
resulting ProMPs to pass through selective neighbors (i.e., movable obstacles).
Lastly, it finds the pushing direction of each neighbor and augments them with
an updated pose. The approach is experimentally tested on different unripe
strawberry configurations in a simulated polytunnel via a robot arm mounted on
a mobile base. 

ProMPs allow for trajectory policy representations acquired from LfD. However,
to reproduce a learned skill with a real robot, a feedback controller that can
deal with perturbations and react to dynamic changes in the environment is
required. To this end, \cite{jankowski2021probabilistic} develop a
probabilistic control technique based on trajectory distributions. Concretely,
the approach merges the probabilistic modeling of the demonstrated movements
and the feedback control action for reproducing the demonstrated behavior. The
proposed probabilistic control encodes a compromise between a robust trajectory
tracking controller and an imitation controller that adapts its gains to the
demonstrated variations of the movement. Although tested using ProMPs, the
approach can extend to other forms of trajectory distributions that have a
linear relation between the corresponding latent and trajectory space. The
system is evaluated by teaching a robotic arm to drop a ball into a moving box
using only a few demonstrations.

ProMPs are a promising way to teach a robot new skills, yet most algorithms to
learn ProMPs from human demonstrations operate in batch mode, which is not
ideal. \cite{schale2021incremental} address this concern by proposing an
incremental learning algorithm. The approach allows humans to teach robots by
incorporating demonstrations sequentially as they arrive. This enables a more
intuitive learning process and it facilitates a faster establishment of
successful cooperation between a human and robot compared to training in batch
mode. Additionally, a built-in forgetting factor allows for corrective
demonstrations resulting from the human's learning curve or changes in task
constraints. Compared to batch ProMP algorithms, successful human-robot
cooperation is demonstrated on a pick-and-place task. 

\paragraph{Polishing/wiping.}
Focusing on the less studied problem of learning force-relevant skills,
\cite{wang2023promp} develop an arc-length ProMP method to improve the
generalization of polishing skills to different robotic polishing tools. To do
this, the technique learns the mapping between a contact force and polishing
trajectory, and the temporal scaling and force scaling factors to allow better
robustness of force planning for speed scaling while polishing. Specifically, a
force scaling factor for different polishing discs is proposed according to the
contact force model. Additionally, polishing contact position learning provides
a basis for trajectory generalization. Arc-length ProMPs are demonstrated using
LfD on robotic disc polishing surfaces, speed-scaling experiments, and
polishing tool experiments.

LfD tasks that involve interactions on free-form 3D surfaces present novel
challenges in modeling and execution, especially when geometric variations
exist between demonstrations and robot execution. \cite{unger2024prosip}
address this problem by developing a ProMP-based framework that systematically
incorporates the surface path and local surface features into the learning
procedure. By considering the surface path and the local surface features, the
framework is independent of time. Furthermore, by describing the tool motion
via a projection of the tool center point onto the surface, the system is also
independent from the robotic platform. Simulations and an experimental setup
with a 9-DoF robot on bathroom-sink edge-cleaning tasks using a sponge tool
highlight the generalization capability of the method to various object
geometries.

\subsubsection{Human-Robot Interaction}
\paragraph{Active compliance.}
Compliant and natural HRI requires adaptive skill generalization and human-like
stiffness modulation. To this end, \cite{guan2021improvement} create an LfD
framework for stiffness-adaptive skill generalization that enhances
collaboration between humans and robots. Concretely, the methodology employs
ProMPs to jointly encode motion and stiffness, capturing their intrinsic
coupling within a unified model. Skin surface EMG signals from an armband
estimate human arm endpoint stiffness, while a hand and finger tracking device
provides the motion data. An experimental evaluation on object handover,
object-matching, and pick-and-place tasks demonstrate that a robot arm can infer
appropriate motion and stiffness using only partial human observations.

\paragraph{Imitation learning.}
Recognizing human actions and generating the respective movement from an
assistive robot is consolidated by \cite{maeda2017probabilistic}, where an
interaction learning method is devised. The approach uses IL to construct a
mixture model of HRI primitives called interaction ProMPs, which allows a robot
to infer its trajectory from human observations. Specifically, the
probabilistic model provides a prior that can be used for both recognizing
human intent and for creating the corresponding commands for a robot
collaborator. Furthermore, the technique is scalable in relation to the number
of tasks and it can learn nonlinear correlations between the trajectories that
describe the HRI scenario. The scalability and ability to learn these
correlations between interaction trajectories are evaluated on human-robot
handover and assembly tasks via a robot arm.

Enabling non-expert users to teach tasks to robots is important for
everyday-life applications such as assisted living and industrial production
processes. To realize this potential, \cite{knaust2021guided} propose a skill
learning framework that provides a graphical user interface for teaching robots
via ProMPs. In particular, the system combines IL with sequential behavior
trees, allowing an operator to record single and bimanual ProMPs, and
automatically integrate them into executable skill trees. A user study with ten
participants employing two robot arms shows that the ProMP-based teaching
achieves higher task success rates and is perceived as more intuitive and
useful than point-to-point motion teaching, though users found it slightly more
complex. 

Ultrasound scanning is a common way to view the anatomical structures of soft
tissues. However, ultrasound scanning is physically repetitive and straining to
clinicians. \cite{hu2023autonomous} alleviate the repetitiveness of the task by
learning a scanning navigation strategy. In the demonstration phase,
interaction ProMPs (\cite{maeda2017probabilistic}) are used to learn the robot
trajectories. A DNN is also utilized to recognize the desired ultrasound image
shown during the demonstrations and a confidence map is created. In the
reproduction phase, the ultrasound image features are extracted by the DNN. The
image features and confidence map are then used to evaluate the ultrasound image
quality and generate the robot movement. The approach is evaluated via an
autonomous ultrasound scanning experiment for breast seroma using a robot arm.

Although existing IL methods can be effective, they often struggle to generalize
across a variety of robotic tasks. \cite{yao2023improved} introduce a solution
to this problem with an IL framework that extrapolates to unseen object
locations. This is done by utilizing task-parametrized GMMs and ProMPs to
achieve extrapolation, multi-modal trajectory modeling, and joint-position
orientation learning. The method incorporates multiple local reference frames
for object-based adaptation, supports via-point conditioning, and extends to
mixture-of-Gaussians weight modeling for handling varying movement modes. An
additional paired-object parameterization improves performance on tasks that
involve aiming. Evaluations are conducted on pick-and-place, pouring, puck
shooting, and sweeping tasks using a robotic gripper.

\cite{liao2024bmp} present B-splines for motion representation within the ProMP
framework. B-splines are a well-known concept in motion planning. They are able
to generate complex, smooth trajectories with only a few control points while
satisfying boundary conditions (i.e., passing through a specified position with
a desired velocity). Nevertheless, prior usages of B-splines ignores
higher-order statistics in the trajectory distribution. This limits their usage
in IL and RL where modeling the trajectory distribution is essential. The
integration of B-splines with ProMPs enables the use of probabilistic techniques
for learning and ensures that boundary conditions are satisfied. An experimental
evaluation shows that B-splines are generally more expressive than ProDMPs and
ProMPs. Moreover, the results highlight that B-splines provide the ability to
specify arbitrary boundary conditions.

\paragraph{Movement coordination.}
Semi-autonomous robots that assist humans must be able to adapt and learn new
skills on-demand, without the need for an expert programmer. To provide this
capability, \cite{maeda2014learning} propose a framework that utilizes ProMPs
to allow robots to collaborate with human coworkers. The approach is based on
probabilistically modeling interactions via a distribution of observed
trajectories as follows. First, a prior model of the interaction is created
using a collection of trajectories in a lower dimensional weight space. Then,
the model is used to recognize the intention of the observed human and to
generate the MP of the action given the same observations. Finally, the MP of
the human can be employed to control a robot assistant. Experiments include
interactions with an anthropomorphic robot manipulator and humans assembling a
box to demonstrate the method's capabilities on collaborative tasks. 

When learning skills from multiple demonstrations, major challenges include the
occurrence of occlusions, lack of sensor coverage, and variability in the
execution speed of the demonstrations. \cite{ewerton2015motor} present an
approach to handle these issues through an EM algorithm that learns ProMPs.
Specifically, multiple demonstrations are learned not only over trajectories,
but also over speed profiles (or the phase of the movement). The phase of the
movement is a function of time that can be related with a movement. By changing
the phase of a movement, it is possible to change its execution speed. This
allows for learning a distribution over phase profiles, which is crucial to
learn motor skills performed at varying speeds and to allow a robot to react to
human actions executed at different speeds. The method is evaluated on object
handover tasks to a robot at various speeds.  

Inferring the intention of a human partner during collaboration, via predicting
a future trajectory, is crucial for designing anticipatory behaviors in HRI
scenarios (e.g., cooperative assembly, co-manipulation, transportation, etc.).
On this topic, \cite{dermy2017prediction} propose a ProMP-based approach for
predicting the intention of a user physically interacting with a robot. To do
this, the robot first recognizes which ProMP best matches the early
observations of a movement initiated by the human partner. Then, it estimates
the future trajectory given the early observations and the prior distribution.
Finally, the corresponding trajectory can be used by the robot to autonomously
finish the movement without relying on the human. The approach is tested in
simulation and on a humanoid robot with tasks involving reaching trajectories
and collaborative object sorting.

Social interactions between humans and robots requires that humans be able to
interpret a robot's intentions such that the actions of the robot do not
confuse or surprise the humans. Based on this observation, \cite{faria2017me}
evaluate the impact of different motion types in a collaboration task between a
robot and multiple people. Concretely, the effect of different motion types in
both the fluency of the collaboration and in the efficiency of the intention
transmission by the robot is investigated using interaction ProMPs
(\cite{maeda2017probabilistic}). A probabilistic model of the intended motion
correlating both the DoF of the robot and additional DoF foreign to the robot
(e.g., of a human) is first built. Then, at run time one type of movement is
selected and the trajectory is generated from the corresponding ProMP. The
system is evaluated via a user study where a robot arm interacts with and
serves water to three humans. 

Utilizing interaction ProMPs (\cite{maeda2017probabilistic}),
\cite{maeda2017phase,maeda2018probabilistic} develop a single probabilistic
framework that allows a robot to estimate the phase of human movements online
and match the outcome of the estimation to handle different tasks and their
respective robot motions. The interaction ProMPs are conditioned on the
observations of a human, which enables a robot to be controlled based on the
posterior distribution over robot trajectories. Not only does the proposed
phase estimation method allow a robot to react faster since the trajectory
inference can be done with partial and occluded observations, but it also
eliminates the need for time alignment. Evaluation of the framework is
performed using a 7-DoF robot arm equipped with a 5-finger hand on multitask
collaborative assembly.

In close HRI scenarios, trajectory generation not only has to consider the
robot's capabilities (e.g., joint limitations), but also variations in human
motions. To this end, \cite{oguz2017progressive} formulate a stochastic
optimization model to generate robot motions that (i) take into account human
motion prediction, (ii) minimize the trajectory cost of the robot, and (iii)
avoid collisions without obstructing the human partner's task execution. This
is accomplished by recording human motions for a desired set of tasks, and then
utilizing ProMPs for offline training. At interaction time, the learned ProMPs
generate predictive trajectory distributions for the human's motion given the
observed data. Based on this prediction, a final posture is optimized for the
robot to avoid collisions with the human. The method is compared to a
state-of-the-art motion planner on avoidance performance in online HRI
scenarios. 

Cooperative robot assistants will need to adapt to user needs along with an
abundance of tasks, which renders pre-programming of all possible scenarios
practically impossible. To address this problem, \cite{koert2018online} present
an online learning approach that employs an incremental mixture model of
interaction ProMPs (\cite{maeda2017probabilistic}). At a high level,
interaction ProMPs are used as a representation of cooperative skills by
capturing correlations between movements as well as the inherent variance. In
practice, the model chooses a corresponding robot response to a human movement
where the human intention is extracted from previously demonstrated movements.
Compared to batch MP methods for HRI, the method builds a library of skills
online in an open-ended manner and updates existing skills using new
demonstrations. The system is evaluated on both a straightforward benchmark
task and in an assistive human-robot collaboration scenario with a 7-DoF robot
arm.

Assistive robots are expected to correctly perform a given task and be
compliant with people. Therefore, in HRI applications Cartesian impedance
control is often preferred over joint control since the desired interaction
feedback can be more naturally described and the robot's force readily
adjusted. Based on this observation, \cite{parent2020variable} use ProMPs to
develop a method for controlling a robot arm in task space with variable
stiffness. This allows for continuously adapting the force exerted in each
phase of motion according to the application requirements. Furthermore,
dimensionality reduction is utilized to avoid obstacles via the redundant DoF
of the task while constantly adapting the stiffness of the relevant DoF. The
results, tested on an assisted water pouring task using an anthropomorphic
robot, show a suitable trade-off between compliance and precision. 

Many existing applications of ProMPs for human-robot collaboration can be
categorized as follows: (i) they estimate the posterior distribution of the
weights and therefore lack an estimation of the phase parameter, or (ii) they
depend on the prior distribution of the phase parameter. These methods may lead
to a misinterpretation of the basis matrix when the divergence between the
prior distribution and the posterior distribution of the phase parameter
becomes large, which can result in a divergence of the estimation of the
weights. \cite{luo2021dual} tackle this issue via a dual-filtering method for
ProMPs. Originally devised for system identification
\cite{nelson1976simultaneous}, the dual-filtering approach aims to
simultaneously estimate both the weights and the phase parameter online.
Experiments are conducted to compare the prediction performance and phase
parameter estimation against prior methods.

ProMPs have been applied to the area of soft robotics for the purpose of
qualitatively reproducing human demonstrations. For example,
\cite{oikonomou2022reproduction} propose a control methodology to support
elderly people in independently completing bathing tasks. This is performed by
replicating human demonstrations via a ProMP library. Specifically, a base of
parameterized movements is constructed and dynamically combined to emulate
complicated trajectories. Inverse kinematics are approximated via model
learning, and a replanning process manages the asynchronous activation of
ProMPs. The outcome is a mapping at the primitive level between the task space
and the actuation space, which allows for planning in the task space and
transferring the skill to the actuation space. The approach is validated on a
bio-inspired soft-robotic arm to demonstrate its efficacy in simplifying the
control of robots with complex dynamics.

\cite{qian2022environment} develop an LfD framework that accommodates the
adaptation of robot behaviors to parameter variance for HRI tasks. To do this,
interaction ProMPs (\cite{maeda2014learning}) are extended to learn the
regression between the environmental parameters and the weight vectors, thus
allowing latent environmental constraints on human-robot joint trajectories to
be learned. In particular, the focus is on handling the temporal variance of
demonstrations through an algorithm that provides more accurate phase estimation
rather than temporal normalization. As a result, the restriction of a Gaussian
or uniform distribution on the phase variables is mitigated. Experimental
results on the joint human-robot tasks of push-button assembly and object
covering show increased robustness to ambiguity in partner activity as well as
environmental changes.

\subsubsection{Motion Modeling}
\paragraph{Human-motion reproduction.}
Traditionally, human motion analysis is performed using manual feature
definitions. Not only do these approaches require data post-processing or
specific domain knowledge to achieve meaningful feature extraction, but they are
also prone to noise and lack reusability for similar types of analysis.
\cite{xue2021using} address this problem by using ProMPs to learn features
directly from the data. Building on this foundation, the method focuses on the
effects of various noninvasive brain stimulation techniques on human motor
coordination, especially arm motions. In particular, by integrating ProMPs with
KL divergence the motion differences under stimulated and non-stimulated
conditions are quantified. Experimental results show that a more robust and
comprehensive method for analyzing the impact of these brain stimulation
techniques on human motion is achieved. 

\paragraph{Trajectory generation.}
ProMPs have been applied to learning human behavior in the context of automotive
racing. For instance, recognizing the challenges associated with the variability
and complexity of human driving styles, \cite{lockel2020probabilistic} propose a
probabilistic modeling of human behavior to better understand and imitate human
drivers. To do this, ProMPs are first used to describe a global target
trajectory distribution that expresses human variability. Then, clothoids are
constructed to represent a path from the current vehicle position to a future
target position in a compact way. Finally, a neural network maps the path
information and basic vehicle states to on-board control actions, thus imitating
the expert driving style. Experiments conducted in a simulated racing
environment highlight the advantages of the approach compared to IL algorithms.

Although ProMPs have been mainly used for determining robot trajectories, they
can also be utilized to model the trajectories of humans. For example,
\cite{davoodi2021safe} propose the use of ProMPs to predict the probabilistic
motion of humans in a shared environment. MPC and CBFs are combined to guide a
robot along a predefined trajectory while ensuring it always maintains a desired
distance from a human worker's motion distribution defined by a ProMP. This
approach ensures the safety of human workers among robots and that the system
state never leaves more than a desired distance from the distribution mean,
while handling nonlinearities in the system dynamics and minimizing control
effort. A case study demonstrates that the method runs in real-time and that it
can provide the designer with the ability to emphasize specific safety
objectives.

Robot programming for autonomous breast palpating is a complex problem. Even
though LfD reduces the time and cost, many methods cannot model the manipulation
path/trajectory as an explicit function of the visual sensory information. To
address this dilemma, \cite{sanni2022deep} propose an LfD approach using ProMPs
that directly maps visual sensory information into the learned trajectory.
Typically, the weights of a ProMP are learned from demonstrations where they can
be later adapted according to different trajectory reproduction needs. This is
effective for simple robotic tasks, but it is too complex for breast palpation
where the task trajectory and the geometry of the breast are related. Therefore,
a deep learning model (e.g., CNNs) is used to capture the correlation between
the visual sensory information and ProMP weights. A series of robot breast
palpation experiments show the effectiveness of the approach on complex path
planning tasks.

Deploying learned ProMP trajectories in scenarios that involve cooperative
task-space constraints is a challenge. Moreover, difficulties arise when
modeling bimanual coordinated movements directly with ProMPs in the joint or
task spaces. This is due to the need to ensure correct alignment of trajectories
between training and execution, as well as stable coordination, in the presence
of external disturbances and obstacles. To handle these issues,
\cite{vorndamme2022integrated} propose an integrated solution that actively
adapts and improves robot motion by exploiting the cooperative task variability
underlying human demonstrations. ProMPs are used to represent trajectories of
object positions. This allows for constructing distributions that are
conditioned on arbitrary time steps and points inside a confidence interval of
the demonstrations. It also only relies on a small amount of data for training.
Evaluations involving adaptive deflection, obstacle avoidance, and the
generalization and transfer of grasping strategies are conducted using a
bimanual robot. 

\subsubsection{Motion Planning}
\paragraph{Multitask learning.}
ProMPs provide generalization inside the area covered by a demonstration, but
they can also generate unpredictable motions when the required action
significantly deviates from the demonstration region.
\cite{yue2024probabilistic} address this problem via a task-parameterized
multitask learning framework that allows learning of operational skills for
robots. This is done by first employing ProMPs for single-task learning and
establishing a joint Gaussian distribution connecting weights and trajectory
features. Then, the trajectory features are modulated to the target trajectory
features using conditional probability. Finally, multiple tasks are iteratively
processed with each task performing extrapolation learning where a trajectory
feature is modulated to a target trajectory feature for a single task, and then
all trajectory features are modulated to their corresponding target features.
The system is validated through simulation and a predefined task with a 7-DoF
robot. 

\paragraph{Obstacle avoidance.}
Robots not only need the ability to generalize motions learned from human
demonstrations, but they must also adapt these motions to different workspaces
not present in the original demonstrations. Working towards this goal,
\cite{koert2016demonstration} take on the challenges of collaborative robots
adapting to diverse tasks and environments. Instead of traditional
pre-programming or kinesthetic teaching, an offline optimization algorithm for
motion planning that optimizes trajectory distributions using ProMPs is
introduced. Collision-free trajectories for obstacles not present during the
human demonstrations are then sampled from the optimized distribution over the
trajectories. Lastly, the optimized distribution is represented as a
probabilistic model such that the learned robot motions can be generalized to
satisfy different via points. The approach is validated using a 7-DoF
lightweight arm that grasps and places a ball into different boxes while
avoiding novel obstacles. 

A few demonstrations can be used to obtain an initial ProMP encoded motion,
however generalization to context variables requires a much larger number of
samples. In light of this observation, \cite{colome2017free} develop a
methodology for defining interest points, sampling them, and efficiently
building a model without the need of a large number of demonstrations. The
proposed contextual representation not only allows for easy adaptation to
changing situations through context variables by reparametrizing motion with
them, but it can also initialize ProMPs with synthetic data by setting an
initial position, a final position, and a number of trajectory interest points.
Furthermore, an obstacle avoidance method based on quadratic optimization is
implemented. Both simulations and physical robots are utilized to test the
effectiveness of the system in avoiding obstacles. In proceeding work,
contextual representations are topologically extended by
\cite{colome2021topological} to generate natural, low-acceleration trajectories.

Although optimization-based methods can explicitly optimize a trajectory via
leveraging prior knowledge of the system, these approaches often require
hand-coded cost functions thus limiting them to simple tasks.
\cite{osa2017guiding} address this issue by merging optimization-based motion
planning with LfD through ProMPs. Concretely, a distribution of the
trajectories demonstrated by human experts is used to define the cost function
of the optimization process. The cost function consists of several objectives
such as smoothness, obstacle avoidance, and similarity to the demonstrated
distribution. The weight of each cost term can be tuned to allow the
trajectories to match the demonstrations. In addition, the cost function is
dependent on the learned distribution. Simulated and physical robot experiments
demonstrate that the approach avoids obstacles and encodes the demonstrated
behavior in the resulting optimized trajectory.

Human-like robot skills can be generalized to different task variations. This
usually requires adapting to new start/goal positions and environmental
changes. To this end, \cite{wilbers2017context} investigate how to modify
ProMPs to allow for context changes that are not included in the training data.
Specifically, an optimization technique is presented that maximizes the
expected return while concurrently minimizing the KL divergence to the
demonstrations. At the same time, the method learns how to linearly combine the
updated primitive with the demonstrations such that only relevant parts of the
primitive are changed. This allows for the adaptation of the primitive to an
unseen situation such that it is still applicable to the context in which it
was already generalized to. The approach is evaluated on obstacle avoidance and
broken joint scenarios in simulation and on a 7-DoF robot arm.

Local trajectory optimization techniques are a powerful tool for robot motion
planning. Yet, these methods can become stuck in local optima and fail to find
a valid (i.e., collision free) trajectory. Additionally, they often require
fine-tuning of a cost function to obtain desirable motions.
\cite{shyam2019improving} address the challenge of combining local trajectory
optimization with LfD to replicate human-like trajectory planning. This is done
by utilizing ProMPs to learn trajectory distributions from kinesthetic
demonstrations. The distributions are then combined with optimization-based
obstacle avoidance, which yields trajectories that are natural-looking and
reliable. Compared to planners such as STOMP \cite{kalakrishnan2011stomp} and
CHOMP \cite{zucker2013chomp}, the ProMP-based planner optimizes in the
parameter space hence reducing computational time. The approach is demonstrated
on obstacle avoidance tasks with a set of performance metrics that include
planning time, trajectory success rate, and overall trajectory cost. 

Motion planning for autonomous robots is a daunting problem not only due to the
complexity of planning under kinematic/dynamic constraints and the uncertainty
of observations, but also because of the effect of unobservable environmental
variables. \cite{low2020identification} handle this issue for mobile robots by
segmenting continuous human-executed trajectories into unit motions represented
by ProMPs. The segmented trajectories are then merged into a motion library and
later retrieved for execution on a robot. In follow-up work,
\cite{low2021prompt} create a framework that combines ProMPs with stochastic
optimization for trajectory planning. The system consists of two motion
planners: feasibility-based trajectory sampling and stochastic gradient-based
trajectory optimization. The versatility of the approach is illustrated by
showing the ability to handle significantly skewed ProMPs (e.g, as induced by a
steering failure in a ground-based robot).

Control methods for executing desired motions typically suffer from a number of
drawbacks such as the reliance on linear control and sensitivity to initial
parameters. \cite{davoodi2021probabilistic} tackle this problem through the use
of feedback linearization, quadratic programming, and multiple CBFs to guide a
robot along a trajectory within the distribution defined via a ProMP.
Concretely, a controller is designed such that the system output tracks a
trajectory within the distribution generated by a ProMP. The approach provides
an easy way to define trajectories for CLFs and barriers for CBFs using the
mean, standard deviation, or other moments of the distribution. Furthermore,
the method guarantees that the system state never leaves more than a desired
distance from the distribution mean. A stability analysis of the proposed
control laws and demonstrations on a 2-link and 6-link robot are conducted. In
follow-up work, \cite{davoodi2022safe} automate the design of CLFs and CBFs
from the distribution given by a ProMP, and \cite{davoodi2022rule} introduce a
rule-base control technique by utilizing a finite-state machine.

\cite{fu2022online} develop an interaction ProMP (\cite{maeda2017probabilistic})
framework that combines online with offline static obstacle avoidance. Online
obstacle avoidance is performed by an algorithm that fuses two ProMP
trajectories via human-robot collaboration. Offline obstacle avoidance is based
on PI$^2$ with a covariance matrix adaptation algorithm. Concretely, a path is
first planned by observing the human trajectory. If an obstacle appears in the
executed path, then the product of a Gaussian distribution is used to fuse the
two ProMP trajectories and achieve fast online static obstacle avoidance.
Lastly, the offline obstacle avoidance algorithm is employed to update the
parameters of the ProMP such that an optimal obstacle avoidance trajectory can
be planned after the human trajectory is observed again. The effectiveness of
the method is tested on 2D trajectory obstacle avoidance and using a robot arm.

\subsubsection{Robotic Prosthetics}
\paragraph{Exoskeletons.}
There are many manual tasks where an exoskeleton could significantly reduce the
amount of physical effort required by a human (e.g., object manipulation on
assembly lines or logistic centers, etc.). In these repetitive tasks, prediction
can be achieved by using a probabilistic trajectory representation. For
example, \cite{jamvsek2021predictive} develop a predictive exoskeleton control
method based on ProMPs combined with a flow controller for arm-motion
augmentation. Concretely, ProMPs are used to generate real-time predictions of
user movements, which are then employed in a velocity-field-based exoskeleton
control approach. An evaluation on a haptic robot highlights the accurate
prediction of user movement intentions and a decrease in the overall physical
effort exerted by the participants to achieve a given task.

Interactions with exoskeleton systems require adaptive motion models that align
with human intent in order to generate physically-natural trajectories. To this
end, \cite{wang2023probabilistic} introduce a framework that permits human-like
trajectories for lower-limb exoskeletons. The methodology combines ProMPs with
PI$^{\text{BB}}$ as follows. First, a motion model is learned offline via
ProMPs, which can then generate reference trajectories for online use by an
exoskeleton controller. Second, PI$^{\text{BB}}$ is adopted to learn and update
the model for online trajectory generation. This provides the system the
ability to adapt to and eliminate the effects of uncertainties. Simulation
results highlight fast convergence, while real-world experiments using a 4-DoF
exoskeleton show adaptation to new users within 3-4 gait cycles. 

\subsection{KMPs}
\begin{table*}
\caption{Applications of KMPs organized by topic area.}
\label{tab:kmp_applications}
\renewcommand{\arraystretch}{1.5}
\centering
\scalebox{1}{
  \begin{tabular}{M{0.2\textwidth-2\tabcolsep - 1.25\arrayrulewidth} M{0.8\textwidth-2\tabcolsep - 1.25\arrayrulewidth}}
    \toprule
    \textbf{Area} & \textbf{References} \\\midrule
    Contact Manipulation & 
    Dexterous Manipulation: \cite{katyara2021leveraging,katyara2021vision}. 
    \newline
    Pick-and-Place: \cite{sun2024fruit}.
    \\\cline{1-2}
    Human-Robot Interaction & 
    Imitation Learning: \cite{silverio2019uncertainty,dall2024imitation}.
    \newline
    Object Handover: \cite{yan2022probabilistic}.
    \\\cline{1-2}
    Motion Planning & 
    Autonomous Vehicles: \cite{deng2021autonomous}. 
    \newline
    Obstacle Avoidance: \cite{liu2024policy,xiao2024kmp}.
    \\
    \bottomrule
  \end{tabular}
}
\end{table*}

\subsubsection{Contact Manipulation}
\paragraph{Dexterous manipulation.}
Postural synergies are patterns of coordinated joint movements that a body's
nervous system employs to simplify control, as opposed to independently
controlling each muscle. The concept has also been extended to robot
manipulation (e.g., robotic hands) by reducing the complexity of numerous DoF.
For example, \cite{katyara2021leveraging} devise a kernelized synergy approach
that enables behaviors to be adapted dynamically for various manipulation tasks
without recalculating the basic movement patterns for each new object. The
method is demonstrated via simulated and real-world robotic systems by assessing
performance using a force closure quality index. \cite{katyara2021vision} extend
the work by integrating real-time visual feedback into the kernelized synergies
framework. This enables a robotic system to recognize and adapt to changes in
object orientation and position dynamically. The approach employs a perception
pipeline that utilizes random sample consensus for object segmentation and SVMs
for recognition. Using a robotic hand, the technique is demonstrated on diverse
manipulation tasks. Specifically, a robot performs tasks that require real-time
adjustments based on visual analysis such as sorting objects of different shapes
and sizes in unstructured environments. 

\paragraph{Pick-and-place.}
The robotic harvesting of delicate fruits, such as grapes, is exceptionally
challenging due to the risk of damaging the crop during manipulation or
placement. \cite{sun2024fruit} make progress on this problem with a manual skill
imitation learning framework for safe and compliant fruit collection as follows.
First, human demonstrations of placing grapes are captured using a motion
capture system. Next, the trajectories are preprocessed, aligned, and modeled
via GMMs and GMR to create a generalized reference trajectory. Finally, KMPs
are used to further imitate the reference trajectory by employing a genetic
algorithm that optimizes the hyperparameters. Synthetic experiments show
improved similarity to the reference trajectory, while real-world experiments
with a 7-DoF arm demonstrate compliant harvesting of fresh grapes.

\subsubsection{Human-Robot Interaction}
\paragraph{Imitation learning.}
A prominent feature of probabilistic approaches to IL is that they not only
extract a mean trajectory from task demonstrations, but they also provide a
variance estimation. Yet, the meaning of this variance can change across
different techniques to indicate either variability or uncertainty. To tackle
this issue, \cite{silverio2019uncertainty} utilize KMPs for IL by predicting
variability, correlations, and uncertainty using a single model. Concretely, the
following insights are obtained: (i) KMPs are shown to predict full covariance
matrices and uncertainty; (ii) a linear quadratic regulator formulation is
created that yields control gains that are a function of both covariance and
uncertainty; (iii) a fusion of controllers is developed, which allows for
demonstrating one complex task as separate subtasks whose activation depends on
an individual uncertainty level. The approach is validated via a toy example
with synthetic data and a robot-assisted painting task.

Ultrasound imaging is commonly used to diagnose vascular diseases, yet the
imaging process can be inconsistent due to its high dependence on the operator's
skill. \cite{dall2024imitation} address this problem via an IL method based on
KMPs that standardizes the contact force profile during ultrasound exams using
sonographer demonstrations. To do this, a demonstration acquisition setup is
developed to allow the recording of ultrasound images, interaction forces, and
torques, together with position and orientation of the ultrasound probe without
drastically altering the probe's ergonomics. Then, the output force of the KMPs
is conditioned on the relative scanning position to allow adaptation to
patient-specific vascular anatomy. Realistic anatomical models and healthy
volunteers are used to evaluate the ability of KMPs to replicate both the forces
and the quality of images obtained by a human operator.

\paragraph{Object handover/transfer.}
Object handover is a basic capability for robots interacting with humans on
collaborative tasks. To ensure that humans can perform handover operations
naturally, \cite{yan2022probabilistic} apply KMPs to adapt learned receiving
skills and fulfill safety constraints based on online predicted results.
Concretely, GMR is utilized to model and predict hand trajectories and then a
Gaussian model is applied to model handover positions online. KMPs are then used
to learn the receiving skills. The KMPs preserve the probabilistic properties
exhibited in multiple demonstrations, thus permitting the robot to learn motor
skills more accurately. Experiments demonstrate that the handover system can
efficiently solve human-to-robot dual-arm handover operations for large box-type
objects.

\subsubsection{Motion Planning}
\paragraph{Autonomous vehicles.}
In autonomous vehicle motion planning, modeling human driver behavior is one way
to encode driving skills. Yet, this is extremely challenging due to the
difficulty of balancing the variability of human behavior in various driving
environments. \cite{deng2021autonomous} address this issue via a motion planning
approach based on KMPs. Specifically, a GMM and GMR are first employed to
adaptively learn human driving behavior in a stochastic manner. Then, the KL
divergence is utilized to minimize information loss between the imitating
reference trajectories and the new via points. This not only allows for
adaptation to new tasks, but it also enables the generation of motion
trajectories for a variety of driving situations. Using an urban driving
simulator, experimental results show the capability of generating robust
autonomous vehicle trajectories.

\paragraph{Obstacle avoidance.}
Dynamic environments require robots to acquire skills that adapt to changing
conditions such as obstacle collisions. To this end, \cite{liu2024policy} create
an algorithm that integrates demonstration learning with trajectory
optimization. Human demonstrations are modeled using multivariate GPR and
obstacle avoidance is formulated as a relative entropy policy search problem by
minimizing the KL divergence between the optimized and demonstrated trajectory
distributions while maximizing obstacle avoidance rewards. The resulting optimal
trajectory distribution is encoded using KMPs. Simulations and real-world
experiments with a 7-DoF arm demonstrate smooth obstacle avoidance that
preserves the characteristics of the original motion.

IL techniques can adapt a robot trajectory to an obstacle, yet the solution may
not always be ideal. In addition, can be difficult for a non-expert user to teach
a robot to avoid obstacles via demonstrations. To address this problem,
\cite{xiao2024kmp} develop an approach that combines human supervision with KMPs
as follows. First a reference database of KMPs is created by demonstrations
using a GMM and GMR. Next, KMPs are extracted from the database to modulate the
trajectory of a robot end effector in real-time to avoid obstacles based on
human-interaction feedback. This allows a user to test different obstacle
avoidance trajectories on the current task until a satisfactory solution is
found. Obstacle avoidance experiments using a robot arm show that the method can
adapt the trajectories of the end effector according to the user's demands. 

\subsection{CNMPs}
\begin{table*}
\renewcommand{\arraystretch}{1.5}
\caption{Applications of CNMPs organized by topic area.}
\label{tab:cnmp_applications}
\centering
\scalebox{1}{
  \begin{tabular}{M{0.2\textwidth-2\tabcolsep - 1.25\arrayrulewidth} M{0.8\textwidth-2\tabcolsep - 1.25\arrayrulewidth}}
    \toprule
    \textbf{Area} & \textbf{References} \\\midrule
    Contact Manipulation & 
    Deformable Objects: \cite{akbulut2022bimanual}.
    \\\cline{1-2}
    Motion Planning & 
    Autonomous Vehicles: \cite{nakahara2023image}. 
    \\
    \bottomrule
  \end{tabular}
}
\end{table*}

\subsubsection{Contact Manipulation}
\paragraph{Deformable objects.}
Although LfD provides a means to endow robots with dexterous skills, compounding
errors in long and complex tasks reduces its applicability. Thus, it is critical
to ensure that an MP ends in a state that allows the successful execution of the
subsequent primitive. \cite{akbulut2022bimanual} address this problem by
learning an explicit correction policy when the expected transition state
between primitives is not achieved. To do this, a correction policy is learned
via BC and CNMPs. The learned correction policy is then able to create diverse
movement trajectories in a context dependent manner. Compared to learning the
complete task as a single action, the advantage of the proposed system is
evaluated using a simulated table-top setup where an object is pushed through a
corridor in two steps. Then, the applicability of the method to bimanual rope
manipulation is demonstrated by teaching a humanoid robot the skill of making
knots. 

\subsubsection{Motion Planning}
\paragraph{Autonomous vehicles.}
Significant computing resources are required to handle large environments and
complex obstacle layouts in outdoor robot navigation scenarios, which can make
real-time path planning difficult. To tackle this issue,
\cite{nakahara2023image} develop a path-planning method based on CNMPs.
Specifically, the proposed framework can allow a robot to learn how to navigate
through a static environment using only image data and location coordinates. To
do this, the robot first requires an image and a start coordinate to find the
path to the target point. Then, a CNMP dynamically generates the optimal motion
to the target point. The CNMP-based navigation method allows for reduced
computational complexity, improved real-time performance, and robustness
against environmental changes. Experimental results show that the methodology
can be utilized for the autonomous navigation of an outdoor wheelchair robot.

\subsection{FMPs}

In developing FMPs, \cite{kulak2020fourier} demonstrated several experiments
that involve polishing, wiping, and 8-shape drawing tasks via a robot arm.
However, as a relatively new MP framework, FMPs have not yet seen widespread
adoption. 


\section{Discussion}
\label{sec:discussion}
\subsection{Open Questions in Movement Primitives for Robotics}
From this survey it is clear that MPs have been used to tackle important
problems in robotics over the last two decades. However, these solutions are
often bespoke. We broadly categorize these issues as follows:
\begin{itemize}
  \item \textbf{Environments:} How can MPs be utilized to handle dynamic
  environments, adapt to new tasks in real-time, and be robust to physical
  disturbances? Continuously learning from human demonstrations, incorporating
  perception, and integrating other sensory information into MPs for various
  types of robots will be important in addressing these issues.
  \item \textbf{Safety:} How can MPs be used to generate safe, natural,
  expressive, and efficient movements for various domains with unique safety
  challenges? This will involve understanding the specific constraints and
  requirements of different application domains and tailoring MPs to address
  these challenges.
  \item \textbf{Scalability:} How can MP frameworks be efficiently scaled to
  handle more complex tasks and high-dimensional systems, such as multi-robot
  cooperation or humanoid robots performing dexterous manipulation? Real-time
  applications require more efficient algorithms. For multi-robot systems,
  distributed and parallel computing paradigms may be applicable.
  \item \textbf{Integration:} How can MPs be more easily integrated with other
  motion planning systems and effectively combined with different learning
  techniques to improve overall performance and generalization? For MP tasks
  where online adaptation is required, the problem is bounded by computational
  capability and the complexity of the models and methods. Robot-specific hybrid
  methods may facilitate tighter integration of control with MPs.
  \item \textbf{Compositionality:} How can multiple MPs be combined or sequenced
  to generate complex, coordinated behaviors? Developing methods for composing
  and blending primitives may lead to more versatile robotic systems.
  \item \textbf{Robustness:} How can MPs be made more robust to uncertainties,
  noise, or partial failures in the system? Fault-tolerant MPs could enhance the
  reliability and safety of robotic applications.
  \item \textbf{Transferability:} How can MPs learned on one robot or task be
  effectively transferred to another robot or task? Creating effective transfer
  learning methods could reduce the amount of training data required and speed
  up the learning process.
  \item \textbf{Interpretability:} In a similar vein to safety, how can MPs be
  made more interpretable and intuitive for human users, particularly in
  collaborative tasks where humans and robots need to work together seamlessly?
  There is a need for visualization tools that provide insights into the
  underlying structure and functioning of MPs in the context of the tasks the
  human is tackling.
  \item \textbf{Benchmarking:} How can the performance of different MP
  frameworks be effectively compared and evaluated? Establishing standardized
  benchmarks and evaluation metrics could facilitate the development and
  adoption of new MP approaches.
\end{itemize}

\subsection{Practical Challenges in Applying Movement Primitives}
\subsubsection{Data Collection}
\hfill\vspace{1.5mm}\\
Collecting accurate and sufficient data for training MPs can be difficult and
time-consuming. Depending on the method, MPs require a nontrivial amount of
data, which requires significant human effort to capture demonstrations that
are representative of the desired behaviors. Furthermore, it can be challenging
to collect clean data depending on the tasks and methods used to express and
record the demonstration data. For example, some MPs necessitate extensive
preprocessing while others may need to find a way to translate joystick
commands into representative trajectories or steering wheel commands. Likewise,
other MPs can use kinesthetic teaching, which appears to be more intuitive, but
also requires either access to a physical robot or a suitable VR interface.

\subsubsection{Parameter Tuning}
\hfill\vspace{1.5mm}\\
Choosing the appropriate parameters for MPs is crucial for ensuring successful
learning and adaptation of the demonstrated motion. With the exception of FMPs
to some degree, every framework described in this survey requires the
implementer to tune multiple parameters. The tuning process is generally
challenging due to the high dimensionality and nonlinear relationships between
these parameters and their effects on MP performance. It requires extensive
testing and validation to achieve optimal performance, as slight modifications
can significantly impact how well the MP captures the desired movement at the
cost of performance. Moreover, the best settings are often dependent on the
specific task and the environment, making the tuning process heavily
context-dependent.

\subsubsection{Model Generalization and Complexity}
\hfill\vspace{1.5mm}\\
MP frameworks can be computationally expensive and require a large amount of
data to be trained effectively. This poses a problem for real-time applications,
such as robot control or HRI. In addition, MPs are sensitive to the choice of
parameters and they expect proper initialization. MPs may not generalize well to
new tasks or environments, and therefore demand retraining or fine-tuning.

\subsubsection{Stochastic Environments}
\hfill\vspace{1.5mm}\\
MP methods can be insensitive to disturbances and uncertainty. For instance,
some frameworks may not be able to handle nonlinear and non-stationary movements
effectively, which can pose a challenge in real-world environments where humans
are present. Since many MP applications are driven by the need for human
demonstration or collaboration, it is crucial for these frameworks to operate
safely despite having a limited model of objects within the environment.

\subsubsection{Domain Specific Challenges}
\hfill\vspace{1.5mm}\\
MPs may need to be adapted to the specific dynamics of different types of
robots, such as legged robots, flying robots, and underwater robots.  Moreover,
they can require further adaptation to the specific constraints and requirements
of different application domains (e.g., rehabilitation robotics, manufacturing
robots, search and rescue robots, etc.). These challenges arise because
different domains often involve diverse methods of demonstration and sensing,
which can impact MP performance and necessitate additional efforts to preprocess
data and fine-tune MP parameters.

\subsubsection{Human Interfaces}
\hfill\vspace{1.5mm}\\
MPs can be used to make robots more human-like, which can make it easier for
humans to interact with and understand the robot's movements. Nonetheless, it
can be a challenge to make sure that the robot's movements are consistent with
human expectations. Many MP frameworks possess properties to ``focus'' the MP
based on the human's immediate needs of a task; however a unified interface for
these capabilities targeted to the end-user teacher is missing.

\subsubsection{Technical Challenges of Industrial Adoption} 
\hfill\vspace{1.5mm}\\
In industrial settings, the use of MPs can present several technical challenges
that must be addressed before widespread support can be realized. One of the
most significant challenges is the real-time implementation of MPs, which can be
especially challenging for complex and high-dimensional frameworks.  Real-time
implementation requires efficient algorithms and hardware with sufficient
processing power to handle the computations necessary for MP execution.

Another significant issue for MPs in industry is explainability. Although the
application of various MP frameworks are intuitive in practice, the theory may
not be understandable to the end user. Translating the state of a primitive into
terms that can be reliably interpreted is important as it makes it easier to
diagnose issues that arise during execution. This is a major concern for
applications that require high reliability and safety, such as the aerospace
industry or medical domain. Developing methods for interpreting and explaining
the behavior of MPs can help address this challenge and increase user
confidence.

Repairability is another critical challenge for the acceptance of MPs in
industrial applications. In complex systems, the ability to diagnose and repair
issues is crucial to minimize downtime and ensure reliable operation. MP systems
that offer online tuning capabilities can address this challenge by allowing
operators to adjust parameters and update the model during runtime. This
enables operators to adapt to changing environments and maintain optimal
performance.

\section{Conclusion}
\label{sec:conclusion}
For over two decades, MPs have been established as one of the most popular and
widely used approaches for generating repeatable movements in robotics. MPs
learn from demonstration data and can be utilized across a vast range of tasks
and environments. Moreover, adjustable parameters allow for modifying MPs in a
deterministic fashion and facilitates anthropomorphic behavior via human
demonstrations. The representation of control and motion planning using MPs has
expedited their deployment to many time-varying problems. Over the years, the
original MP formulation has been extended to overcome specific limitations and
applied to a variety of applications. This has resulted in a very large number
of publications.

In this survey, we presented an overview of MP frameworks along with a variety
of MP applications to highlight their capabilities and relevant uses. Our goal
was to compile and categorize the vast MP literature. To do this, we performed
an exhaustive review guided by unbiased criteria. Another aim of this survey
was to provide helpful guidelines that aid the practitioner in selecting the
right MP framework for a given application. MPs are still a highly relevant and
active area of research. To this end, we provided a comprehensive discussion
that will help understand what has been done in the field, included a set of
open questions to be addressed by the research community, and listed the
current challenges in deploying MPs.

\section*{Acknowledgments}
Nolan B. Gutierrez was supported by U.S. Department of Defense SMART
Scholarship.

\bibliographystyle{SageH}
\bibliography{movement_primitives_in_robotics_a_comprehensive_survey}

\end{document}